\documentclass[twoside,11pt,hyphens]{article}

\usepackage{subcaption}
\usepackage[preprint]{jmlr2e}
\usepackage{mathtools}
\usepackage{booktabs}       

\usepackage{bookmark}
\usepackage{verbatim}
\usepackage{lastpage}
\newcommand\given[1][]{\:#1\vert\:}
\newcommand\KLsep[1][]{\:#1\Vert\:}

\DeclarePairedDelimiter{\abs}{\lvert}{\rvert}
\DeclarePairedDelimiter{\norm}{\lVert}{\rVert}
\DeclareMathOperator*{\argmin}{arg\,min}
\DeclareMathOperator*{\st}{s.t.}
\newcommand{\ones}{\mathbf{1}}
\newcommand{\bb}{\mathbf{b}}
\newcommand{\bB}{\mathbf{B}}
\newcommand{\bH}{\mathbf{H}}

\newcommand{\bP}{\mathbf{P}}
\newcommand{\br}{\mathbf{r}}
\newcommand{\boldf}{\mathbf{f}}
\newcommand{\cA}{\mathcal{A}}
\newcommand{\cD}{\mathcal{D}}
\newcommand{\cE}{\mathcal{E}}

\newcommand{\cX}{\mathcal{X}}

\newcommand{\tlambda}{\tilde{\lambda}}
\newcommand{\trho}{\tilde{\rho}}
\newtheorem{assumption}{Assumption}
\usepackage{xcolor}

\newcommand{\EE}[1]{\mathbb{E}\left[#1\right]}

\newcommand{\defined}{\triangleq}
\newcommand{\mname}{$\mathsf{FairScoreTransformer}$}

\usepackage{tikz-cd}

\jmlrheading{22}{2021}{1--\pageref{LastPage}}{10/20; Revised 5/21}{10/21}{20-1143}{Dennis Wei, Karthikeyan Natesan Ramamurthy, and Flavio P. Calmon}

\ShortHeadings{Optimized Score Transformation for Consistent Fair Classification}{Wei, Natesan Ramamurthy, and Calmon}
\firstpageno{1}

\begin{document}

\title{Optimized Score Transformation for\\Consistent Fair Classification}

\author{\name Dennis Wei \email dwei@us.ibm.com \\
       \name Karthikeyan Natesan Ramamurthy \email knatesa@us.ibm.com \\
       \addr IBM Research\\
       1101 Kitchawan Road\\
       Yorktown Heights, NY 10598, USA
       \AND
       \name Flavio P. Calmon \email flavio@seas.harvard.edu \\
       \addr John A. Paulson School of Engineering and Applied Sciences\\
       Harvard University\\
       150 Western Ave\\
       Allston, MA 02134, USA}

\editor{Maya Gupta}

\maketitle

\begin{abstract}%
This paper considers fair probabilistic binary classification where the outputs of primary interest are predicted probabilities, commonly referred to as scores. We formulate the problem of transforming scores to satisfy fairness constraints that are linear in conditional means of scores while minimizing a cross-entropy objective. The formulation can be applied directly 
to post-process classifier outputs and we also explore a pre-processing extension, thus allowing maximum freedom in selecting a classification algorithm. 
We derive a closed-form expression for the optimal transformed scores and a convex optimization problem for the transformation parameters. In the population limit, the transformed score function is the fairness-constrained minimizer of cross-entropy with respect to the true conditional probability of the outcome. 
In the finite sample setting, we propose a method called \mname{} to approach this solution using a combination of standard probabilistic classifiers and ADMM. We provide several consistency and finite-sample guarantees for \mname, relating to the transformation parameters and transformed score function that it obtains. Comprehensive experiments comparing to 10 existing methods show that \mname{} has advantages for score-based metrics such as Brier score and AUC while remaining competitive for binary label-based metrics such as accuracy.
\end{abstract}

\begin{keywords}
  algorithmic fairness, machine learning fairness, probabilistic classification, post-processing
\end{keywords}

\section{Introduction}
\label{sec:intro}

Recent years have seen a surge of interest in the problem of 
\emph{fair classification}, which is concerned with disparities in classification output or performance 
when conditioned on protected attributes such as race, gender, or ethnicity. 
Many measures of fairness have been introduced \citep{pedreschi2012study,dwork2012fairness,kamiran2013,hardt2016,zafar2017fairness,chouldechova2017,kleinberg2017,kilbertus2017,kusner2017,zafar2017preference,nabi2018,kearns2018preventing,heidari2018,chiappa2019} and fairness-enhancing interventions have been proposed to mitigate these disparities
\citep{friedler2019comparative}. Roughly categorized, these interventions either (i) change data used to train 
a classifier (pre-processing) \citep{kamiranC2012,hajian2013,zemel2013,feldman2015,calmon2017}, (ii) change a classifier's output 
(post-processing) \citep{kamiran2012,fish2016,hardt2016,pleiss2017,woodworth2017}, or (iii) directly change 
a classification model to ensure fairness (in-processing) \citep{calders2010,kamishima2012,zafar2017fairness,zafar2017constraints,dwork2018,agarwal2018,krasanakis2018,donini2018,celis2019}.

This paper differs from many of the above works in placing more emphasis on probabilistic classification, in which the outputs of interest are predicted probabilities of belonging to one of the classes as opposed to binary predictions. The predicted probabilities are often referred to as \emph{scores}. They are desirable because they indicate confidences in predictions (when well-calibrated) and provide more information for decision-making. For example, in a loan approval scenario, a score of $0.7$ may indicate that a loan applicant is predicted to have a $70\%$ chance of repaying the loan on time, given their credit history features.

Our objective is to produce probabilistic scores satisfying fairness criteria. These scores can be useful in a number of decision-making scenarios. In health risk assessment for example, the scores represent risks of developing a condition or requiring medical intervention (e.g.,~stroke, \citealp{lip2010refining}, ICU admission, \citealp{zhao2020prediction}) and are the final output of interest. In other applications, the scores are an intermediate output that is passed to a subsequent decision-making stage. However, this subsequent stage may not be fully known or defined, may take additional inputs, and/or may be performed by a different party. An important example is where the decision-maker is a human (e.g.,~a hiring manager) who, in addition to considering a score (e.g.,~predicted probability of succeeding in a new job), may have to weigh other information (e.g.,~reports from human interviewers), and whose fairness and other decision-making properties cannot be well-controlled. In this case, it may be desirable to enforce fairness in the scores given to the human decision-maker (perhaps in addition to measures that encourage the decision-maker to be more fair). Even in the straightforward case where the scores are thresholded to produce a binary decision, exact knowledge of protected attributes may be lacking to use existing post-processing methods for fairness \citep{kamiran2012,hardt2016,pleiss2017,yang2020fairness}. Moreover, our experimental results in Section~\ref{sec:expt} suggest that thresholding fairer scores can be competitive with fairness methods that directly target binary outputs.

We make several contributions to the subject of fair probabilistic classification. In Section~\ref{sec:prob}, we propose an optimization formulation for transforming scores to satisfy fairness constraints while minimizing a cross-entropy objective. The formulation accommodates any fairness criteria that can be expressed as linear inequalities involving conditional means of scores, including variants of statistical parity (SP) \citep{pedreschi2012study} and equalized odds (EO) \citep{hardt2016,zafar2017fairness}. 

In Section~\ref{sec:opt}, we study solutions to the optimization problem of fair score transformation that we have formulated. Given an input score function $r(x)$, we derive a closed-form expression for the optimal transformed scores $r'(x)$ and a convex dual optimization problem for the Lagrange multipliers that parametrize the transformation. In the population limit, the optimal input score function (i.e.,~the unconstrained optimum) is the conditional distribution $p_{Y\given X}$ of the outcome $Y$ given features $X$. In this case, the transformed scores minimize cross-entropy with respect to $p_{Y\given X}$ while satisfying the fairness constraints. 

In Section~\ref{sec:proc}, we consider the finite sample setting and propose a method called \mname{} (FST) to approximate the optimal solution found in Section~\ref{sec:opt}. FST takes a practical ``plug-in'' approach, using standard probabilistic classifiers (e.g.,~logistic regression) to approximate $p_{Y\given X}$ and estimating other probabilities as needed. In particular, if protected attributes are not known at test time, FST can instead use estimates of them based on the available features. We find that the dual problem is well-suited to the alternating direction method of multipliers (ADMM) and describe an ADMM algorithm to solve it. The closed-form expression for the transformed scores and the low dimension of the dual problem (a small multiple of the number of protected groups) make FST computationally lightweight. 

FST lends itself naturally to post-processing, with scores as input and fairer scores as output. To increase flexibility, we also explore a pre-processing extension of FST in which the output scores are used to re-weight the training data. The re-weighted data can then be published as an output in its own right, allowing others to train fairer models using standard algorithms that do not explicitly account for fairness. We envision therefore that FST will be particularly beneficial in situations that make post- and pre-processing attractive, as also articulated by e.g.,~\citet{hajian2013,calmon2017,agarwal2018,madras2018,salimi2019capuchin}: a) when it is not possible or desirable to modify an existing classifier (only post-processing is possible); b) when freedom is desired to select the most suitable classifier for an application, whether it maximizes performance or has some other desired property such as interpretability (post- and pre-processing apply); and c) when standard training algorithms are used without the additional complexity of fairness constraints or regularizers (post- and pre-processing again). In-processing meta-algorithms \citep{agarwal2018,celis2019} can also support situation b) but not a) or c), while standard in-processing does not support any of a)--c). As discussed in Section~\ref{sec:relWork} and summarized in Table~\ref{tab:capabilities}, FST is considerably more flexible than existing post- and pre-processing methods in handling more cases.

The conference version \citep{wei2020optimized} of this work focused on formulating and solving the optimization problem (Sections~\ref{sec:prob} and \ref{sec:opt}) and translating the solution into a practical procedure (Section~\ref{sec:proc}). This has left a gap however between the solution in the ideal population setting ($r(x) = p_{Y\given X}(1\given x)$) and the approximate result of the FST procedure. In this extended version, we address this gap by providing consistency and finite-sample guarantees. In Section~\ref{sec:consistency}, under suitable assumptions on the convergence of the estimated score function $\hat{r}(x)$ and other estimated probabilities, we prove that:
\begin{enumerate}
    \item Optimal solutions to the empirical version of the dual problem solved by FST become asymptotically optimal for the population version of the dual problem. For finite sample sizes, the optimality gap is bounded with high probability (Theorem~\ref{thm:dualConsistency}).
    \item The plug-in solution for the transformed scores asymptotically satisfies the population fairness constraints (i.e.,~fairness consistency). For finite sample sizes, the degree of infeasibility is bounded with high probability (Theorem~\ref{thm:primalFeas}).
    \item The plug-in solution asymptotically minimizes cross-entropy with respect to $p_{Y\given X}$ subject to the fairness constraints (Theorem~\ref{thm:primalOpt}).
\end{enumerate}
We have accordingly refined the presentation in Sections~\ref{sec:opt} and \ref{sec:proc}, for example clearly distinguishing between the empirical and population dual problems and explicitly defining the plug-in primal solution. Of note, we have clarified that the characterization of the optimal solution in Section~\ref{sec:opt} applies to any input score function $\hat{r}(x)$, not just $r(x) = p_{Y\given X}(1\given x)$.

We have conducted comprehensive experiments, reported in Section~\ref{sec:expt} and Appendix~\ref{sec:exptAdd}, comparing FST to 10 existing methods, a number that compares favorably to recent meta-studies \citep{friedler2019comparative}. On score-based metrics such as Brier score and AUC, FST achieves better fairness-utility trade-offs and hence is indeed advantageous when scores are of interest. At the same time, it remains competitive on binary label-based metrics such as accuracy.

In summary, \mname{} enables fairness-enhancing post-processing that 
\begin{itemize}
    \item is principled, optimal in the population limit, and comes with consistency and finite-sample guarantees (Sections \ref{sec:prob}, \ref{sec:opt}, and \ref{sec:consistency}),
    \item is computationally lightweight (Section \ref{sec:proc}),
    \item performs favorably compared to the state-of-the-art and can handle lack of protected attributes at test time (Section \ref{sec:expt} and Appendix~\ref{sec:exptAdd}).
\end{itemize}

The organization of the paper is recapitulated below: Section~\ref{sec:prob} formulates the optimization problem of transforming scores to satisfy fairness constraints. Section~\ref{sec:opt} specifies the optimal solution to the problem in terms of a closed-form transformation and a dual optimization problem. Section~\ref{sec:proc} describes the \mname{} procedure that approximates the optimal solution given a finite sample. Section~\ref{sec:consistency} provides theoretical 
results for the \mname{} solution. Section~\ref{sec:expt} discusses empirical evaluation of \mname{} and comparisons to existing methods. Section~\ref{sec:concl} concludes the paper.

\subsection{Related Work}
\label{sec:relWork}

Existing post-processing methods for fairness 
include those from \citet{kamiran2012,fish2016,hardt2016,pleiss2017,jiang2019wasserstein,chzhen2019leveraging,yang2020fairness}; limitations of post-processing are studied by \citet{woodworth2017}. While these methods take predicted scores as input, most \citep{kamiran2012,fish2016,hardt2016,chzhen2019leveraging} are designed to 
produce only binary output and not scores. The method of \citet{pleiss2017} 
maintains calibrated probability estimates, which is 
a requirement that we do not enforce herein. Furthermore, \citet{kamiran2012,fish2016,hardt2016,pleiss2017,yang2020fairness} all assume exact knowledge of the protected attribute.
\citet{kamiran2012,fish2016,jiang2019wasserstein} address only SP (\citet{kamiran2012} as originally proposed), 
\citet{hardt2016,pleiss2017} address disparities in error rates, and \citet{chzhen2019leveraging} address only equal opportunity. Our approach does not have these limitations. It produces scores as well as binary outputs, can handle estimated protected attributes, and accommodates a wider range of fairness criteria.

Pre-processing methods range from reweighing, resampling, and relabeling training data \citep{kamiranC2012}, to performing probability transformations on features \citep{feldman2015}, to modifying both labels and features through optimization \citep{calmon2017} or labels and protected attributes using classification rules \citep{hajian2013}. The above methods 
only address SP or the related notion of disparate impact \citep{feldman2015}. Learning representations that are invariant to protected attributes \citep{zemel2013,louizos2016,edwards2016,xie2017,xu2018} can also be seen as pre-processing, 
and recent adversarial approaches \citep{beutel2017,zhang2018,madras2018} permit control of EO as well as SP. Representation learning however does not preserve the original data domain and its semantics, while adversarial algorithms 
can produce unstable results and be computationally challenging.

Several works by \citet{agarwal2018,celis2019,menon2018,corbett-davies2017,jiang2019,yang2020fairness} have technical similarities to the approach herein but 
focus on binary outputs, with $0$-$1$ risk \citep{celis2019,agarwal2018} or cost-sensitive risk \citep{menon2018,corbett-davies2017,yang2020fairness} as the objective function, and/or lead to in-processing algorithms \citep{celis2019,agarwal2018,cotter2019optimization}. \citet{celis2019} come closest in also solving a fairness-constrained classification problem via the dual problem. However, \citet{celis2019} along with \citet{agarwal2018} propose in-processing algorithms that solve multiple instances of a subproblem whereas we solve only one instance. \citet{celis2019} also address a larger class of fairness measures that are linear-fractional in the classifier output. \citet{cotter2019optimization} propose incorporating rate constraints when training predictive models in order to meet target fairness, churn, or other performance requirements. These constraints are cast in terms of indicator functions and are inherently non-convex and non-differentiable, motivating an oracle-based in-processing optimization algorithm. Unlike \citet{cotter2019optimization}, the optimized transformation introduced here circumvents non-differentiability issues by formulating fairness constraints in terms of scores (as opposed to a sum of indicator functions). The resulting optimization is convex and solvable using standard methods. 

Similar to us, \citet{menon2018,corbett-davies2017,yang2020fairness} also characterize optimal fair classifiers in the population limit in which probability distributions are known; however, \citet{menon2018,corbett-davies2017} do not propose algorithms for computing the Lagrange multipliers or thresholds that parametrize 
the solution. The recent work of \citet{yang2020fairness} provides such a characterization in a very general multi-class setting with overlapping protected groups. They propose two algorithms inspired by the Bayes-optimal fair classifier. The first is an in-processing approach that generalizes the algorithm of \citet{agarwal2018}. The second is similar to ours in also taking a plug-in post-processing approach and optimizing Lagrange multipliers. In their case, the Lagrange multipliers determine thresholds to apply to the ``plugged-in'' probabilistic classifier. However, both algorithms of \citet{yang2020fairness} return a \emph{randomized} classifier \citep[similar to][]{agarwal2018}, i.e., a probability distribution over a set of classifiers, and they also assume knowledge of the protected attributes.

\section{Problem Formulation}
\label{sec:prob}

We represent one or more protected attributes such as gender and race by a random variable $A$ and an outcome variable by $Y$. We make the common assumption that $Y \in \{0,1\}$ is binary-valued. 
It is assumed that $A$ takes a finite number of values in a set $\cA$, corresponding to protected groups.  Let $X$ denote features (drawn from domain $\cX$) used to predict $Y$ in a supervised classification setting.  We consider two scenarios in which $X$ either includes or does not include $A$, like in other works in fair classification
\citep[e.g.,][]{kamiranC2012,agarwal2018,donini2018}.  While it is recognized that 
the former scenario can achieve better trade-offs between utility and fairness, the latter is needed in applications where disparate treatment laws and regulations forbid the explicit use of $A$.  To develop our approach in this section and Section~\ref{sec:opt}, we work in the population limit and make use of probability distributions involving $A$, $X$, $Y$.  Section~\ref{sec:proc} discusses how these distributions are approximated using a training sample. In general, we use capital letters (e.g.,~$A$, $X$) to refer to random variables, and lowercase letters ($a$, $x$) to their realizations.

As stated earlier, we focus more heavily on probabilistic classification in which the output of interest is the predicted probability of being in the positive class $Y=1$ rather than a binary prediction.  The optimal probabilistic classifier is the conditional probability $r(x) \equiv p_{Y\given X}(1\given x)$, which we refer to as the 
\emph{population score} because it is only known in the population limit.  Bayes-optimal binary classifiers can be derived from $r(x)$ by thresholding, specifically at level $c \in [0,1]$ if $c$ and $1-c$ are the relative costs of false positive and false negative errors.  Score functions will thus play the central role in our development.

We propose a mathematical formulation and method called \mname~(FST) that leads directly to a post-processing solution. The goal is to transform $r(x)$ into a \emph{transformed score} $r'(x)$ that satisfies fairness conditions while minimizing the loss in optimality compared to $r(x)$. The transformed score $r'(x)$ is taken 
as the classification output and can be thresholded to provide a binary prediction. We elaborate on the utility and fairness measures considered in Sections~\ref{sec:prob:utility} and \ref{sec:prob:fairness}.

We also consider a pre-processing extension of FST in which $r'(x)$ is used to transform the training data and train a new classifier, which provides the final output. For this case, we additionally define a \emph{transformed outcome} variable $Y' \in \{0,1\}$ and let $r'(x) = p_{Y'\given X}(1\given x)$ be the conditional probability associated with it.  
The overall procedure consists of two steps, performed in general by two different parties: 1) The \emph{data owner} transforms the outcome variable from $Y$ to $Y'$; 2) The \emph{modeler} trains a classifier with $Y'$ as target variable and $X$ as input, without regard for fairness.  The transformed score $r'(x)$ plays two roles in this procedure.  The first is to specify the probabilistic mapping from $X$ to $Y'$ in step 1). As discussed in Section~\ref{sec:proc:pre}, we realize this mapping by re-weighting the training data. The second role stems from the main challenge faced by pre-processing methods, namely that the predominant fairness metrics depend on the output of the classifier trained in step 2) but this classifier is not under direct control of the pre-processing in step 1).  
In recognition of this challenge, we make the following assumption, also discussed by \citet{madras2018,salimi2019capuchin}:
\begin{assumption}[pre-processing]
\label{ass:pre-process}
The classifier trained by the modeler approximates the transformed score $r'(x)$ if it is a probabilistic classifier or a thresholded version of $r'(x)$ if it is a binary classifier.
\end{assumption}
This assumption is satisfied for modelers who are ``doing their job'' in learning to predict $Y'$ from $X$ since the optimal classifier in this case is $r'(x)$ or a function thereof.  Given the assumption, we will use $r'(x)$ as a surrogate for the actual classifier output.  The assumption is not satisfied if the modeler is not competent or, worse, malicious in trying to discriminate against certain protected groups.

We note that this pre-processing extension is not specific to FST and could be applied to other methods that produce a fair output score similar to $r'(x)$, for example in-processing methods that work with probabilistic classifiers.

\subsection{Utility Measure}
\label{sec:prob:utility}

We propose to measure the loss in optimality, i.e.,~utility, between the transformed score $r'(x)$ and population score $r(x)$ using the following cross-entropy:
\begin{equation}\label{eqn:utility}
\mathbb{E}\bigl[-\log p_{Y'\given X}(Y\given X)\bigr] = \mathbb{E} \bigl[ -r(X) \log r'(X) - (1-r(X)) \log(1 - r'(X)) \bigr],
\end{equation}
where the right-hand side results from expanding the expectation over $Y$ conditioned on $X$, and $p_{Y'\given X}$ is used only as notational shorthand in the post-processing case since $Y'$ is not generated. For simplicity, we shall also use the following notation for cross-entropy:
\begin{equation}\label{eqn:crossentropy}
H_b(p,q )\defined -p\log q-(1-p)\log(1-q).
\end{equation}
The utility measure in \eqref{eqn:utility} is equivalent to $\EE{H_b\left(r(X),r'(X)\right)}.$

One way to arrive at \eqref{eqn:utility} is to assume that $r'(x)$, which is the classifier output in the post-processing case and a surrogate thereof in the pre-processing extension, is evaluated against the observed outcomes $y_1,\dots,y_n$ in a training set using the cross-entropy a.k.a.~log loss.  This yields the empirical version of the left-hand side of \eqref{eqn:utility},
\[
-\frac{1}{n} \sum_{i=1}^n \log p_{Y'\given X}(y_i\given x_i).
\]
The use of log loss is well-motivated by the desire for $r'(x)$ to be close to the true conditional probability $r(x)$. 

An equivalent way to motivate \eqref{eqn:utility} in the pre-processing context is to measure the utility lost in transformation by the Kullback-Leibler (KL) divergence between the original and transformed joint 
distributions
\begin{equation}
    \label{eqn:utilityKL}
    D_{\mathrm{KL}}\bigl(p_{X,Y} \KLsep p_{X,Y'}\bigr) = \mathbb{E}_{p_{X,Y}} \left[ \log\frac{p_{X,Y}}{p_{X,Y'}} \right]= \mathbb{E}_{p_{X,Y}}[\log p_{Y\given X}] - \mathbb{E}_{p_{X,Y}}[\log p_{Y'\given X}].    
\end{equation}
On the right-hand side, the first term depends on the data distribution but not $r'(x)$ and the second term is exactly \eqref{eqn:utility}.

Starting from a different premise, \citet{jiang2019} proposed a similar mathematical formulation in which the arguments of the KL divergence are reversed from those in \eqref{eqn:utilityKL}, i.e.,~the given distribution is the second argument while the distribution to be determined is the first.  
The form of the solution of \citet{jiang2019} is therefore different from the one presented herein.  The order of arguments in \eqref{eqn:utilityKL} is justified by the connection to log loss in classification discussed above. The order in \eqref{eqn:utilityKL} also agrees with the common interpretation of the first argument as a given distribution and the second argument as an approximation or deviation from the given distribution.

\subsection{Fairness Measures}
\label{sec:prob:fairness}

We consider fairness criteria expressible as linear inequalities involving 
conditional means of scores,
\begin{equation}\label{eqn:fairnessLinIneq}
    \sum_{j=1}^J b_{lj} \mathbb{E}\bigl[ r'(X) \given \cE_{lj} \bigr] \leq c_l, \quad l = 1,\dots,L,
\end{equation}
where $\{b_{lj}\}$ and $\{c_l\}$ are real-valued coefficients 
and the conditioning events $\cE_{lj}$ are defined in terms of   $(A, X, Y)$ but do not depend on $r'$. Special cases of \eqref{eqn:fairnessLinIneq} correspond to the well-studied notions of statistical parity (SP) and equalized odds (EO). More precisely, we focus on the following variant of SP:
\begin{equation}\label{eqn:MSP}
    -\epsilon \leq \mathbb{E}[r'(X)\given A=a] - \mathbb{E}[r'(X)] \leq \epsilon \qquad \forall a \in \cA,
\end{equation}
which we refer to as \emph{mean score parity} (MSP) following \citet{coston2019}. Condition \eqref{eqn:MSP} corresponds to approximate mean independence of random variable $R' = r'(X)$ with respect to $A$.
Similar notions can also be put in the form of \eqref{eqn:fairnessLinIneq}, for example bounds on the ratio 
\[
1-\epsilon \leq \frac{\mathbb{E}[r'(X)\given A=a]}{\mathbb{E}[r'(X)]} \leq 1+\epsilon,
\]
referred to as \emph{disparate impact} by \citet{feldman2015}, as well as \emph{conditional statistical parity} \citep{kamiran2013,corbett-davies2017}.

For EO, we add the condition $Y=y$ to the conditioning events in \eqref{eqn:MSP}, resulting in \begin{equation}\label{eqn:MEO}
    -\epsilon \leq \mathbb{E}[r'(X)\given A=a, Y=y] - \mathbb{E}[r'(X) \given Y=y] \leq \epsilon \qquad \forall a \in \cA, \; y \in \{0,1\}.
\end{equation}
For $y = 0$ (respectively $y = 1$), $\mathbb{E}[r'(X) \given Y=y]$ is the false (true) positive rate (FPR, TPR) generalized for a probabilistic classifier, and $\mathbb{E}[r'(X) \given A=a, Y=y]$ is the corresponding group-specific rate.  Following \citet{pleiss2017}, we refer to \eqref{eqn:MEO} for $y=0$ or $y=1$ alone as approximate equality in generalized FPRs or TPRs, and to \eqref{eqn:MEO} for $y=0$ and $y=1$ together as generalized EO (GEO). The correspondences between \eqref{eqn:MSP}, \eqref{eqn:MEO} and \eqref{eqn:fairnessLinIneq} are detailed in Appendix~\ref{sec:proof:dualspecialize}.

The fairness measures \eqref{eqn:fairnessLinIneq} in our formulation are defined in terms of probabilistic scores.  Parallel notions defined for binary predictions, i.e.,~by replacing $r'(X)$ with a thresholded version $\ones(r'(X) > t)$, are more common in the literature.  For example, the counterpart to \eqref{eqn:MEO} is (non-generalized) EO while the counterpart to \eqref{eqn:MSP} is called \emph{thresholded score parity} by \citet{coston2019}.  While our formulation does not optimize for these binary prediction measures, we nevertheless use them for evaluation in Section~\ref{sec:expt}.

The form of \eqref{eqn:fairnessLinIneq} is inspired by but is less general than the linear conditional moment 
constraints of \citet{agarwal2018}, which replace $r'(X)$ in \eqref{eqn:fairnessLinIneq} by 
an arbitrary bounded function $g_j(A, X, Y, r'(X))$. We have restricted ourselves to \eqref{eqn:fairnessLinIneq} so that a closed-form optimal solution can be derived 
in Section~\ref{sec:opt}.  We note however that in both of 
the examples of \citet{agarwal2018} and many 
fairness measures, $g_j(A, X, Y, r'(X)) = r'(X)$ and the additional generality is not required.

\subsection{Optimization Problem}
\label{sec:prob:opt}

The transformed score $r'(x)$ is obtained by minimizing the cross-entropy in \eqref{eqn:utility} (equivalently maximizing its negative) subject to fairness constraints \eqref{eqn:fairnessLinIneq}:
\begin{equation}\label{eqn:primalGeneral}
        \max_{r'} \;\; -\EE{ H_b\left(r(X),r'(X)\right) } \quad
        \st \quad \sum_{j=1}^J b_{lj} \mathbb{E}\bigl[ r'(X) \given \cE_{lj} \bigr] \leq c_l, \quad l = 1,\dots,L.
\end{equation}
Section~\ref{sec:opt} characterizes the optimal solution to this  problem. 

\subsection{Sufficiency of Pre-Processing Scores}

In the pre-processing extension of FST, the proposed optimization \eqref{eqn:primalGeneral} transforms only scores and uses them to generate a weighted data set, as described further in Section~\ref{sec:proc:pre}. Can a better trade-off between utility and fairness  be achieved by also pre-processing features $X$, i.e., mapping each pair $(X,r(X))$ into a new $(X',r'(X))$? Note that pre-processing both scores/labels and input features is suggested by \citet{hajian2013,feldman2015,calmon2017}. When utility and fairness are measured according to the objective and constraints in \eqref{eqn:primalGeneral}, the answer is negative: a transformed feature $X'$ would not impact the constraints in \eqref{eqn:primalGeneral}, since they only depend on the marginals of $r'(X)$ conditioned events $\mathcal{E}_{l,j}$ given in terms of $A$ and $Y$. Moreover, a transformed feature would also not change the objective value, which only depends on $r(X)$ and $r'(X)$. In other words, a transformed score/label pair would satisfy the Markov relation:

\begin{center}
   \begin{tikzcd}
 (A,Y)  \arrow[r] & X \arrow[r] \arrow[rd] & \left\{r(X), r'(X)\right\} \\
 & & X'
\end{tikzcd} 
\end{center}
\noindent The quantities in formulation \eqref{eqn:primalGeneral} only depend on the upper branch of the above graph  and, hence, are invariant to the mapping from $X$ to $X'$. Thus, for the metrics considered here, pre-processing the scores is sufficient.

\section{Characterization of Optimal Fairness-Constrained Score}
\label{sec:opt}

In this section, we consider a slight generalization of problem \eqref{eqn:primalGeneral} in which $r(X)$ is replaced by an arbitrary score function $\hat{r}(X)$:
\begin{equation}\label{eqn:primal_rhat}
    \max_{r'} \;\; -\EE{ H_b\left(\hat{r}(X),r'(X)\right) } \quad
    \st \quad \sum_{j=1}^J b_{lj} \mathbb{E}\bigl[ r'(X) \given \cE_{lj} \bigr] \leq c_l, \quad l = 1,\dots,L.
\end{equation}
In later sections, $\hat{r}(X)$ will be an estimate of $r(X)$, thus justifying the hat notation.

We derive a closed-form expression for the optimal solution to problem \eqref{eqn:primal_rhat} using the method of Lagrange multipliers.  We then state the dual optimization problem that determines the Lagrange multipliers.  These results are specialized to the cases of MSP \eqref{eqn:MSP} and GEO \eqref{eqn:MEO}.

Define Lagrange multipliers $\lambda_l \geq 0$, $l = 1,\dots,L$ for the constraints in \eqref{eqn:primal_rhat}, and let $\lambda \defined (\lambda_1,\dots,\lambda_L)$.  Then the Lagrangian function is given by 
\begin{equation}\label{eqn:Lagrangian1}
    L(r', \lambda) = -\mathbb{E} \bigl[ H_b\left(\hat{r}(X),r'(X)\right) \bigr] - \sum_{l=1}^L \sum_{j=1}^J \lambda_l b_{lj} \mathbb{E}\bigl[ r'(X) \given \cE_{lj} \bigr] + \sum_{l=1}^L c_l \lambda_l.
\end{equation}
The dual optimization problem corresponding to \eqref{eqn:primalGeneral} is 
\[
\min_{\lambda \geq 0} \max_{r'} L(r', \lambda).
\]

Note that $L(r',\lambda)$ is a strictly concave function of $r'$ and the fairness constraints in \eqref{eqn:primal_rhat} are affine functions of $r'$. Consequently, as long as the constraints in \eqref{eqn:primal_rhat} are feasible, the optimal transformed score $r^*$ can be found by maximizing $L(r',\lambda)$ with respect to $r'$, resulting in an optimal solution $r^*$ that is a function of $\lambda$, and then minimizing $L(r^*,\lambda)$ with respect to~$\lambda$ \citep[Section 5.5.5]{boyd2004}. Substituting the optimal $\lambda^*$ into the solution for $r^*$ found in the first step then yields the optimal transformed score. Note that this procedure would not necessarily be correct if a linear objective function were considered \citep[e.g., 0-1 loss in][]{celis2019} due to lack of strict concavity. The next proposition states the general form of the solution to the inner maximization of $L(r',\lambda)$ above. Its proof is in Appendix~\ref{sec:proof:rstar}.

\begin{proposition}
\label{prop:rstar}
Let $L(r',\lambda)$ be as given in \eqref{eqn:Lagrangian1}. Then for fixed $\lambda$, $r^*(\lambda) = \arg \max_{r'} L(r',\lambda)$ is given by
\begin{equation}\label{eqn:r*}
    r^*\bigl(\mu(x); \hat{r}(x)\bigr) = \begin{cases}
    \dfrac{1 + \mu(x) - \sqrt{(1 + \mu(x))^2 - 4 \hat{r}(x) \mu(x)}}{2\mu(x)}, & \mu(x) \neq 0\\
    \hat{r}(x), & \mu(x) = 0,
    \end{cases}
\end{equation}
where
\begin{equation}\label{eqn:mu}
    \mu(x) \defined \sum_{l=1}^L \sum_{j=1}^J \lambda_l b_{lj} \frac{\Pr(\cE_{lj} \given X=x)}{\Pr(\cE_{lj})}.
\end{equation}
\end{proposition}

\begin{figure}[t]
  \centering
  \begin{subfigure}[b]{0.48\columnwidth}
  \includegraphics[width=\columnwidth]{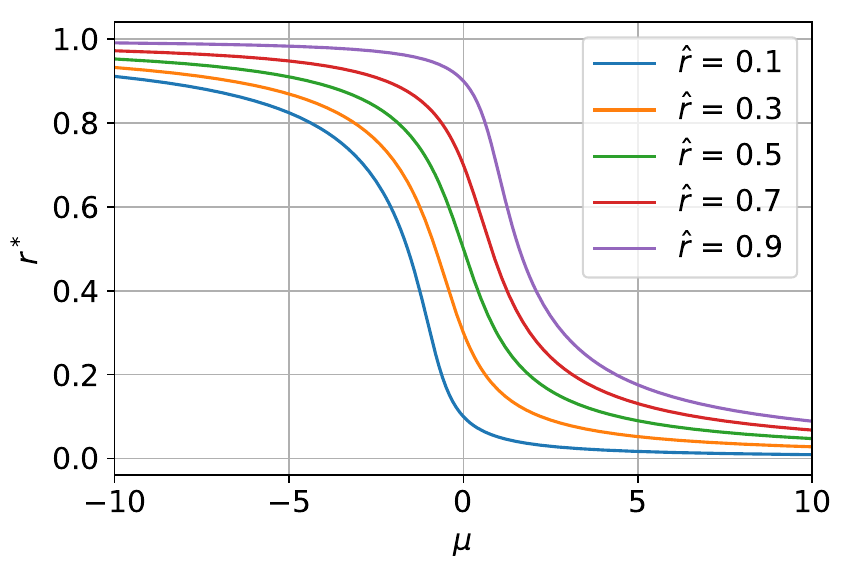}
  \caption{$r^*$ vs.~$\mu$ for fixed values of $\hat{r}$.}
  \label{fig:rStar_mu}
  \end{subfigure}
  \begin{subfigure}[b]{0.48\columnwidth}
  \includegraphics[width=\columnwidth]{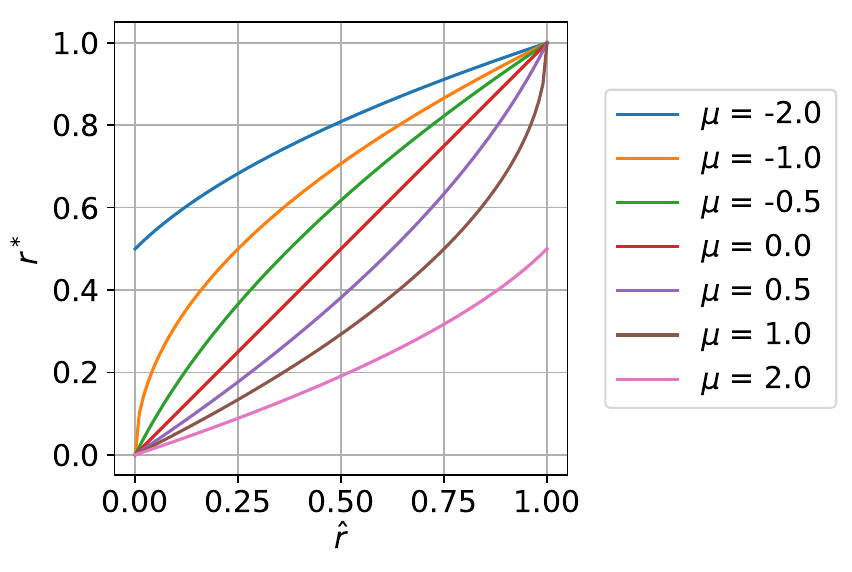}
  \caption{$r^*$ vs.~$\hat{r}$ for fixed values of $\mu$.}
  \label{fig:rStar_r}
  \end{subfigure}
  \caption{Optimal transformed score $r^*(\mu; \hat{r})$ (Equation~\ref{eqn:r*}) as a function of $\mu$ and $\hat{r}$.}
  \label{fig:rStar}
\end{figure}

We can interpret the optimal primal solution \eqref{eqn:r*} as a prescription for \emph{score transformation} controlled by $\mu(x)$, which is in turn a linear function of $\lambda$. When $\mu(x) = 0$, the score is unchanged from the input $\hat{r}(x)$, and as $\mu(x)$ increases or decreases away from zero, the score $r^*(\mu(x); \hat{r}(x))$ decreases or increases smoothly from $\hat{r}(x)$, as seen in Figure~\ref{fig:rStar_mu}. Figure~\ref{fig:rStar_r} shows that the transformed score $r^*$ has a rank-preserving property stated in Lemma~\ref{clm:monotonic}.

\begin{lemma}\label{clm:monotonic}
The transformed score $r^*(\mu; \hat{r})$ is monotonically increasing in $\hat{r}$ for fixed $\mu$, i.e.,~if $r_1 < r_2$ then $r^*(\mu; r_1) < r^*(\mu; r_2)$. 
\end{lemma}
\begin{proof}
This is confirmed analytically by a positive partial derivative:
\[
\frac{\partial r^*(\mu; \hat{r})}{\partial \hat{r}} = \frac{1}{\sqrt{(1+\mu)^2 - 4\hat{r}\mu}} > 0.
\]
\end{proof}

It is shown in Appendix~\ref{sec:proof:rstar} that the result of substituting the optimal primal solution \eqref{eqn:r*} into the first two terms of the Lagrangian \eqref{eqn:Lagrangian1} is the expectation of the function
\begin{equation}
\label{eqn:g}
g\bigl(\mu(x); \hat{r}(x)\bigr) \defined -H_b\Bigl(\hat{r}(x),r^*\left(\mu(x);\hat{r}(x) \right) \Bigr)  - \mu(x) r^*\bigl(\mu(x); \hat{r}(x)\bigr).
\end{equation}
The dual problem is therefore 
\begin{equation}\label{eqn:dualGeneral}
    \begin{split}
        \min_{\lambda} \quad &\mathbb{E}\left[ g\bigl(\mu(X); \hat{r}(X)\bigr) \right] + \sum_{l=1}^L c_l \lambda_l \\
        \st \quad &\mu(X) = \sum_{l=1}^L \sum_{j=1}^J \lambda_l b_{lj} \frac{\Pr(\cE_{lj} \given X)}{\Pr(\cE_{lj})}, \qquad \lambda \geq 0.
    \end{split}
\end{equation}
The solution to the above minimization provides the values of $\lambda^*$ for the optimal transformed score \eqref{eqn:r*}. Like all Lagrangian duals, \eqref{eqn:dualGeneral} is a convex optimization (although it is no longer apparent from Equation~\ref{eqn:dualGeneral} that this is the case).  Furthermore, \eqref{eqn:dualGeneral} is typically low-dimensional in cases where the number of dual variables $L$ is a small multiple of the number of protected groups $\abs{\cA}$.

We now specialize and simplify \eqref{eqn:dualGeneral} to MSP \eqref{eqn:MSP} and GEO \eqref{eqn:MEO} fairness constraints. The following proposition follows from the correspondences between \eqref{eqn:MSP}, \eqref{eqn:MEO} and \eqref{eqn:fairnessLinIneq} and is proved in Appendix~\ref{sec:proof:dualspecialize}.

\begin{proposition}
\label{prop:dualspecialize}
Under the MSP constraint \eqref{eqn:MSP}, the dual optimization \eqref{eqn:dualGeneral} reduces to
\begin{equation}\label{eqn:dualMSP_rhat}
    \begin{split}
        \min_{\lambda} \quad &\mathbb{E}\left[ g\bigl(\mu(X); \hat{r}(X)\bigr) \right] + \epsilon \norm{\lambda}_1\\
        \st \quad &\mu(X) = \sum_{a\in\cA} \lambda_a \left(\frac{p_{A\given X}(a\given X)}{p_A(a)} - 1\right).
    \end{split}
\end{equation}
For the GEO constraint \eqref{eqn:MEO}, \eqref{eqn:dualGeneral} reduces to
\begin{equation}\label{eqn:dualMEO_rhat}
    \begin{split}
        \min_{\lambda} \quad &\mathbb{E}\left[ g\bigl(\mu(X); \hat{r}(X)\bigr) \right] + \epsilon \norm{\lambda}_1, \\
        \st \quad &\mu(X) = \sum_{y\in\{0,1\}} \frac{p_{Y\given X}(y\given X)}{p_{Y}(y)} \sum_{a\in\cA} \lambda_{a,y} \left( \frac{p_{A\given X,Y}(a\given X,y)}{p_{A\given Y}(a\given y)} - 1\right).
    \end{split}
\end{equation}
In the case $\hat{r}(X) = r(X)$, we refer to \eqref{eqn:dualMSP_rhat}, \eqref{eqn:dualMEO_rhat} as the \emph{population dual} problem.
\end{proposition}
In \eqref{eqn:dualMSP_rhat}, \eqref{eqn:dualMEO_rhat}, there is no longer a non-negativity constraint on $\lambda$ but instead an $\ell_1$ norm, and the problem dimension is only $|\cA|$ 
in \eqref{eqn:dualMSP_rhat} and $2|\cA|$ in \eqref{eqn:dualMEO_rhat}.
Moreover, both dual formulations are well-suited for decomposition using the alternating direction method of multipliers (ADMM), as discussed further in Section~\ref{sec:proc:ADMM}.

In the case where the features $X$ include the protected attribute $A$, we have $p_{A\given X}(a\given X) = p_{A\given X,Y}(a\given X,y) = \ones(a = A)$, where $A$ is the component of $X$ that is given. The constraints in \eqref{eqn:dualMSP_rhat} and \eqref{eqn:dualMEO_rhat} then simplify to
\begin{equation}\label{eqn:muMSPAtrue}
\mu(X) = \frac{\lambda_A}{p_A(A)} - \sum_{a\in\cA} \lambda_a,
\end{equation}
\begin{equation}\label{eqn:muMEOAtrue}
\mu(X) = \sum_{y\in\{0,1\}} \frac{p_{Y\given X}(y\given X)}{p_{Y}(y)} \left( \frac{\lambda_{A,y}}{p_{A\given Y}(A\given y)} - \sum_{a\in\cA} \lambda_{a,y} \right)
\end{equation}
respectively. Interestingly, the only difference between the cases of including or excluding $A$ is that in the latter, the constraints in \eqref{eqn:dualMSP_rhat}, \eqref{eqn:dualMEO_rhat} indicate that $A$ should be inferred from the available features $X$ and possibly $Y$, whereas in the former, $A$ can be used directly.

\subsection{Comparison to Optimal Fair Binary Classifiers}
\label{sec:opt:compare}

As discussed in Section~\ref{sec:relWork}, optimal fair classifiers have been characterized by \citet{menon2018,corbett-davies2017,yang2020fairness} in the case of binary outputs and cost-sensitive risk. While the score transformation discussed herein is optimized for different, score-based measures of utility and fairness, it is still of interest to compare the result of thresholding the transformed score to these optimal fair binary-output classifiers. 

We focus on \citet{menon2018}, who give the most concrete expressions for optimal classifiers compared to \citet{corbett-davies2017,yang2020fairness}. In accordance with \citet{menon2018}, we consider the population limit, e.g.,~$\hat{r}(X) = r(X)$, use a binary protected attribute, $\cA = \{0,1\}$, and consider the fairness measures statistical parity (SP) and equal opportunity (EOpp) with respect to $A$. To be closer to fairness constraints \eqref{eqn:MSP}, \eqref{eqn:MEO}, we consider the ``mean difference'' (MD) measure of \citet[Equation~6]{menon2018}. In the SP-MD case, \citet{menon2018} minimize the following cost-sensitive risk (Problem 3.2 therein):
\begin{equation}\label{eqn:menonCostSens}
(1-c) \Pr(\hat{Y}=0\given Y=1) + c \Pr(\hat{Y}=1\given Y=0) - \lambda \left(\Pr(\hat{Y}=1\given A=0) - \Pr(\hat{Y}=1\given A=1)\right),
\end{equation}
where the first and second terms are the FNR and FPR, weighted by $1-c$ and $c$, and the last two terms are the difference between positive prediction rates. For EOpp, the last two terms are additionally conditioned on $Y=1$. 

\begin{table}[t]
    \small
    \centering
    \begin{tabular}{lccc}
    \toprule
    fairness & $A$ & \multicolumn{2}{c}{$h(X)$}\\
    criterion & known & \citeauthor{menon2018} & from optimal fair score \\
    \midrule
    SP & no & $r(X) - \lambda(\bar{\eta}(X) - 1/2)$ & $r(X) - c(1-c) \left(\lambda_0 \theta_0(X) + \lambda_1 \theta_1(X) \right),$ \\
    & & & $\theta_a(X) = \frac{(1-\bar{\eta}(X))^{1-a} \bar{\eta}(X)^a}{p_A(a)} - 1, \;\; a = 0,1$ \\
    \midrule
    & yes & $r(X) + (-1)^A (1/2) \lambda$ & $r(X) + (-1)^A c(1-c) p_A(1-A) \tilde{\lambda},$\\
    & & & $\tilde{\lambda} = \frac{\lambda_1}{p_A(1)} - \frac{\lambda_0}{p_A(0)}$\\
    \midrule
    EOpp & no & $\left(1 - \frac{\lambda}{p_Y(1)} (\bar{\eta}(X) - 1/2) \right) r(X)$ & $\left(1 - \frac{c(1-c)}{p_{Y}(1)} \left(\lambda_{0} \theta_0(X) + \lambda_{1} \theta_1(X) \right) \right) r(X),$ \\
    & & & $\theta_a(X) = \left( \frac{(1-\bar{\eta}(X))^{1-a} \bar{\eta}(X)^a}{p_{A\given Y}(a\given 1)} - 1\right)$\\
    \midrule
    & yes & $\left(1 + (-1)^A \frac{1}{2 p_Y(1)} \lambda \right) r(X)$ & $\left(1 + (-1)^A \frac{c(1-c) p_{A\given Y}(1-A\given 1)}{p_{Y}(1)} \tilde{\lambda} \right) r(X),$ \\
    & & & $\tilde{\lambda} = \left(\frac{\lambda_{1}}{p_{A\given Y}(1\given 1)} - \frac{\lambda_{0}}{p_{A\given Y}(0\given 1)} \right)$\\
    \bottomrule
    \end{tabular}
    \caption{Comparison with optimal fair binary classifiers of \citet{menon2018}. All binary classifiers are of the form $\ones(h(X) > c)$ where $h(X)$ is given below.}
    \label{tab:menon}
\end{table}

Table~\ref{tab:menon} summarizes the expressions for fair binary classifiers, which are all of the form $\ones(h(X) > c)$ where $h(X)$ is given in the table. For the rightmost column, we assume that the optimal transformed score $r^*(\mu(X); r(X))$ is thresholded at the cost-sensitive threshold $c$ to obtain a binary prediction. Derivations of all expressions are given in Appendix~\ref{sec:proof:menon}. We use the notation $\bar{\eta}(X)$ of \citet{menon2018} for the conditional probability of $A$ given $X$, where $\bar{\eta}(X) = p_{A\given X}(1\given X)$ in the SP case and $\bar{\eta}(X) = p_{A\given X,Y}(1\given X,1)$ for EOpp. 

Overall, while the expressions from \citet{menon2018} and from thresholding the optimal transformed score are different, they do have notable similarities. The four cases in Table~\ref{tab:menon} are discussed further below, where ``$A$ known'' means that $A$ is included in $X$ or is perfectly predicted by $X$.
\begin{itemize}
    \item \textbf{SP, $A$ not known:} The two $h(X)$ expressions are similar in that an affine function of $\bar{\eta}(X)$ is added to the original score $r(X)$. In the case of \citet{menon2018}, the affine function is proportional to the trade-off parameter $\lambda$, whereas for the thresholded optimal fair score, the affine function is proportional to $c(1-c)$ and also depends on $p_A(a)$. The parameters $\lambda_0$, $\lambda_1$ are chosen to optimize the dual objective \eqref{eqn:dualMSP_rhat}.
    \item \textbf{SP, $A$ known:} In this case, the similarity between the two expressions becomes more apparent. On the right-hand side, the quantity $\tilde{\lambda} = \lambda_1 / p_A(1) - \lambda_0 / p_A(0)$ plays the role of $\lambda$ on the left side, and the main difference is the scaling by the factor $p_A(1-A)$ ($p_A(1)$ for $A = 0$ and $p_A(0)$ for $A = 1$), in addition to the factor $c(1-c)$. While these differences are likely due to optimizing for different criteria, the overall similarity between the two formulas is noteworthy. Indeed, if $p_A(0) = p_A(1)$, then the expressions coincide after defining $\lambda_0$, $\lambda_1$ appropriately.
    \item \textbf{EOpp, $A$ not known:} The two expressions again share similarities: now the modification to $r(X)$ is multiplicative, the multiplicative factor is affine in $\bar{\eta}(X)$, and $p_Y(1)$ appears in the denominator on both sides.
    \item \textbf{EOpp, $A$ known:} As in the SP, $A$ known case, the quantity $\tilde{\lambda} = \lambda_1 / p_{A\given Y}(1\given 1) - \lambda_0 / p_{A\given Y}(0\given 1)$ plays the role of $\lambda$ and the scale factors are $p_{A\given Y}(1\given 1)$ for $A = 0$ and $p_{A\given Y}(0\given 1)$ for $A = 1$. These are the analogues of the quantities in the SP, $A$ known case, now conditioned on $Y = 1$. Again if $p_{A\given Y}(0\given 1) = p_{A\given Y}(1\given 1)$, then the two expressions are equivalent.
\end{itemize}

\section{Proposed FairScoreTransformer Procedure}
\label{sec:proc}

We now consider the finite sample setting in which the probability distributions of $A,X,Y$ are not known and we have instead a training set $\mathcal{D}_n\defined \{(a_i, x_i, y_i), i=1,\dots,n\}$.  This section presents the proposed \mname{} (FST) procedure that approximates the optimal fairness-constrained score in Section~\ref{sec:opt}. We focus on the cases of MSP and GEO. The procedure consists of the following steps: 
\begin{enumerate}
    \item Estimate the population score and other probabilities required to define the dual problem \eqref{eqn:dualMSP_rhat} or \eqref{eqn:dualMEO_rhat}.
    \item Solve the dual problem to obtain dual variables $\hat{\lambda}$ (the ``fit'' step).
    \item Transform scores using \eqref{eqn:mu}  and \eqref{eqn:r*} (``transform'' step).
    \item For the pre-processing extension of FST, modify the training data.
    \item For binary-valued predictions, binarize scores.
\end{enumerate} 
The following subsections elaborate on steps 1--4. Step 5 is done simply by selecting a threshold $t \in [0,1]$ to maximize accuracy on the training set. 

\subsection{Estimation of Original Score and Other Probabilities}
\label{sec:proc:origScore}

In some post-processing applications, estimates $\hat{r}(x)$ of the population scores $r(x)$ 
may already be provided by an existing base classifier.  If no suitable base classifier exists, any probabilistic classification algorithm may be used to estimate $r(x)$.  We experiment with logistic regression and gradient boosting machines in Section~\ref{sec:expt}.  We naturally recommend selecting a model and any hyperparameter values to maximize performance in this regard, i.e.,~to yield accurate and calibrated probabilities.  This can be done through cross-validation on the training set using an appropriate metric such as Brier score \citep{hernandez-orallo2012}. 

In the case where $A$ is one of the features in $X$, the other probabilities required are $p_A(a)$ for MSP \eqref{eqn:muMSPAtrue} and $p_Y(y)$, $p_{A\given Y}(a\given y)$ for GEO \eqref{eqn:muMEOAtrue} ($p_{Y\given X}(y\given x)$ is already estimated by $\hat{r}(x)$ and $p_{A\given X}$, $p_{A\given X,Y}$ are delta functions).  Since $Y$ is binary and $\abs{\cA}$ is typically small, it suffices to use the empirical estimates of these probabilities.  If $A$ is not included in $X$, then it is also necessary to estimate it using $p_{A\given X}(a\given X)$ for MSP \eqref{eqn:dualMSP_rhat} and $p_{A\given X,Y}(a\given X,y)$ for GEO \eqref{eqn:dualMEO_rhat}.  Again, any probabilistic classification algorithm can be used, provided that it can handle more than two classes if $\abs{\cA} > 2$. 

We highlight that FST translates the effort of ensuring fair classification into training well-calibrated models for predicting $Y$ and, if necessary, $A$. This echoes the plug-in approach advocated by \citet{menon2018,chzhen2019leveraging}. 

\subsection{ADMM for Optimizing Dual Variables}
\label{sec:proc:ADMM}

In the finite sample case, we solve an empirical version of the dual problem in Proposition~\ref{prop:dualspecialize}. We write $\mu(x) = \lambda^T \boldf(x)$, where $\mathbf{f}:\mathcal{X}\to \mathbb{R}^L$ is defined by the expression for $\mu(x)$ in \eqref{eqn:dualMSP_rhat} or \eqref{eqn:dualMEO_rhat} (explicit definitions for $\mathbf{f}$ are given in Equations~\ref{eqn:fMSP}, \ref{eqn:fMEO} for the case where $A$ is known exactly), and $L$ is the dimension of $\lambda$. Let $\hat{r}(x)$ denote the estimate of $r(x)$ obtained in Section~\ref{sec:proc:origScore}, and $\hat{\boldf}(x)$ be an empirical version of $\boldf(x)$ in which all probabilities (e.g.,~$p_A(a)$ for MSP in Equations~\ref{eqn:dualMSP_rhat}, \ref{eqn:muMSPAtrue}) are replaced by their estimates, again as discussed in Section~\ref{sec:proc:origScore}. With these definitions, both optimizations in Proposition \ref{prop:dualspecialize} have the general form
\begin{equation}
      \label{eqn:dualSpecF}
        \min_{\lambda\in\mathbb{R}^L} \quad \frac{1}{n}\sum_{i=1}^n g\bigl(\mu(x_i); \hat{r}(x_i)\bigr)  + \epsilon \norm{\lambda}_1 \quad
        \st \quad \mu(x_i) = \lambda^T\hat{\mathbf{f}}(x_i), \quad i=1,\dots,n,
\end{equation}
where the expectation in the objective has also been approximated by the average over the training data set. 

Formulation \eqref{eqn:dualSpecF} is well-suited for ADMM because the objective function is separable between $\mu(x)$ and $\lambda$, which are linearly related through the constraint. We present one ADMM decomposition here and alternatives in Appendix~\ref{sec:proc:ADMMalt}. Under the first decomposition, application of the scaled ADMM algorithm \citep[Section 3.1.1]{boyd2011distributed} to \eqref{eqn:dualSpecF} yields the following three steps in each iteration $k = 0,1,\dots$:
\begin{subequations}
\begin{align}
\mu^{(k+1)}(x_i) &= \arg\min_{\mu} \; \frac{1}{n} g\bigl(\mu; \hat{r}(x_i)\bigr) + \frac{\rho}{2}\left(\mu- (\lambda^{(k)})^T\hat{\boldf}(x_i)+c^{(k)}(x_i) \right)^2 \quad \forall i=1,\dots,n \label{eq:iter-mu}\\
\lambda^{(k+1)} &= \arg\min_{\lambda}~ \epsilon\|\lambda \|_1+\frac{\rho}{2}\sum_{i=1}^n\left(\mu^{(k+1)}(x_i)- \lambda^T\hat{\boldf}(x_i) +c^{(k)}(x_i)\right)^2 \label{eq:iter-lambda}\\
c^{(k+1)}(x_i) &= c^{(k)}(x_i)+\mu^{(k+1)}(x_i)- \left(\lambda^{(k+1)}\right)^T\hat{\boldf}(x_i) \quad \forall i=1,\dots,n.\label{eq:iter-c}
\end{align}
\end{subequations}
Here $c^{(k)}(x_i)$ are Lagrange multipliers for the $n$ equality constraints in \eqref{eqn:dualSpecF}.

The first update \eqref{eq:iter-mu}  can be computed in parallel for each sample $x_i$ in the data set. Given an $x_i$, finding $\mu(x_i)$ is a single-parameter optimization where the objective possesses   closed-form expressions for its derivatives. For simplicity of notation, let $\hat{r}_i \defined \hat{r}(x_i)$, $b_i\defined (\lambda^{(k)})^T\hat{\boldf}(x_i)-c^{(k)}(x_i)$,   and $$\mathsf{obj}(\mu)\defined \frac{1}{n}g\bigl(\mu; \hat{r}_i\bigr) + \frac{{\rho}}{2}\left(\mu- b_i \right)^2.$$  The first two derivatives of $\mathsf{obj}(\mu)$ are 
\begin{align*}
\frac{\partial \mathsf{obj}(\mu)}{\partial \mu} = -\frac{r^*(\mu;\hat{r}_i)}{n}+{\rho}(\mu-b_i),\qquad
\frac{\partial^2 \mathsf{obj}(\mu)}{\partial \mu^2} = & 
\begin{cases}
\frac{1}{2n\mu^2}\left(1-\frac{1+\mu(1-2\hat{r}_i)}{\sqrt{(1+\mu)^2-4\hat{r}_i\mu}} \right)+{\rho},& \mu\neq 0, \\
\frac{r(1-r)}{n}+{\rho},&\mu=0,
\end{cases}
\end{align*}
using \eqref{eqn:df(tlambda)}, \eqref{eqn:d2f(tlambda)} in Appendix~\ref{sec:proc:ADMMalt} for the derivatives of $g(\mu; \hat{r}_i)$. It can be confirmed from the second derivative that $\mathsf{obj}(\mu)$ is convex (expected since the dual problem is convex) so that the first-order condition $\partial\mathsf{obj}(\mu) / \partial\mu = 0$ is necessary and sufficient for optimality. In Appendix~\ref{sec:proc:muCubic}, we show that this condition leads to a cubic equation with a closed-form solution.

The second update \eqref{eq:iter-lambda} reduces to an $\ell_1$-penalized quadratic minimization over (at most) $2|A|$ variables. Specifically,
\begin{align}
\lambda^{(k+1)} &= \arg\min_{\lambda}~ \epsilon\|\lambda \|_1+\lambda^T\mathbf{v} +\lambda^T\mathbf{F}\lambda, \label{eqn:step2}
\end{align}
where 
\begin{equation}
\nonumber
\mathbf{v} \defined -\rho\sum_{i=1}^n\hat{\boldf}(x_i)\left(\mu^{(k+1)}(x_i)+c^{(k)}(x_i)\right),\qquad \mathbf{F}\defined \frac{\rho}{2}\sum_{i=1}^n\hat{\boldf}(x_i)\hat{\boldf}(x_i)^T.
\end{equation}
The ADMM approach thus handles the non-smooth $\ell_1$ term in the objective \eqref{eqn:dualSpecF} by solving $\ell_1$-penalized quadratic subproblems \eqref{eqn:step2}, for which many solvers exist. Moreover, the values of $\mathbf{v}$ and $\mathbf{F}$ above can be pre-computed prior to solving \eqref{eqn:step2}. In fact, $\mathbf{F}$ can be computed once at the start of the iterations. The ensuing minimization only involves $|\cA|$ variables under the MSP constraint \eqref{eqn:MSP}, and $2|\cA|$ variables under the GEO constraint \eqref{eqn:MEO}.

From \eqref{eq:iter-mu}--\eqref{eq:iter-c}, it is seen that the computational complexity of each ADMM iteration scales linearly with 
$n$. We have fixed the ADMM penalty parameter $\rho = 1$ and have not attempted to tune it for faster convergence.

\subsection{Score Transformation}
\label{sec:proc:transform}

Let $\hat{\lambda}$ denote an optimal solution to the empirical dual problem \eqref{eqn:dualSpecF}. We propose using a plug-in solution for the transformed score $r'(x)$, obtained by substituting finite-sample estimates into formula \eqref{eqn:r*} for $r^*$, namely $r(x) = \hat{r}(x)$ and $\mu(x) = \hat{\lambda}^T \hat{\boldf}(x)$:
\begin{equation}\label{eqn:plug-in}
r'(x) = r^*\bigl(\hat{\lambda}^T \hat{\boldf}(x); \hat{r}(x)\bigr).
\end{equation}
Sections~\ref{sec:consistency:primalFeas} and \ref{sec:consistency:primalOpt} discuss the consistency properties of this plug-in solution.

\subsection{Additional Steps for Pre-Processing}
\label{sec:proc:pre}

In the pre-processing extension of FST, the transformed score $r'(x)$ is used to generate samples of a transformed outcome 
$Y'$.  Since $r'(x) = p_{Y'\given X}(1\given x)$ is a probabilistic mapping, we propose generating a \emph{weighted} data set $\cD' = \{(x_i, y'_i, w_i)\}$ with weights $w_i$ that reflect the conditional distribution $p_{Y'\given X}$.  Specifically, $\cD' = \cD'_0 \cup \cD'_1$ with $\cD'_0 = \{(x_i, 0, 1-r'(x_i)), i=1,\dots,n\}$ and $\cD'_1 = \{(x_i, 1, r'(x_i)), i=1,\dots,n\}$. With these weights, $Y'$ follows the conditional distribution given by $r'(x)$, and $\cD'$ is twice the size of the original data set. The data owner passes the transformed data set 
$\cD'$ to the modeler, who uses it to train a classifier for $Y'$ given $X$ 
without fairness constraints. Per Assumption~\ref{ass:pre-process}, the output of this new classifier is expected to approximate $r'(x)$.

\section{Consistency and Finite-Sample Guarantees for FairScoreTransformer}
\label{sec:consistency}

In this section, we present results guaranteeing the consistency of the FST procedure of Section~\ref{sec:proc}, again focusing on the cases of MSP and GEO. For two of the three theorems presented, finite-sample bounds are also provided. We consider in particular steps 2 and 3 of the procedure and make the following statements respectively:
\begin{enumerate}
    \item Optimal solutions to the empirical dual problem \eqref{eqn:dualSpecF} become asymptotically optimal for the population dual problem (Equations~\ref{eqn:dualMSP_rhat} or \ref{eqn:dualMEO_rhat} with $\hat{r}(X) = r(X)$) as the sample size $n \to \infty$ and the estimates $\hat{r}(x)$, $\hat{\boldf}(x)$ converge to their respective true quantities. For finite sample sizes, the optimality gap is bounded with high probability.
    \item The finite-sample plug-in solution \eqref{eqn:plug-in} for the transformed score $r'(x)$ becomes asymptotically feasible and optimal for the population primal problem \eqref{eqn:primalGeneral}, again as $n \to \infty$ and $\hat{r}(x)$, $\hat{\boldf}(x)$ converge. For finite sample sizes, the degree of infeasibility is bounded with high probability.
\end{enumerate}
Asymptotic feasibility in statement 2 may also be referred to as \emph{fairness consistency}, in that score functions that satisfy the fairness constraints on the training data also asymptotically satisfy them on the population. 

We first summarize the assumptions that are made before formally stating the results. This is followed by more detailed discussion of the assumptions, their basic implications, and outlines of the proofs. Proofs of lemmas are deferred to Appendix~\ref{sec:proofs}.

\subsection{Assumptions}
\label{sec:consistency:ass}

To simplify the proofs, we assume in this section that $A$ is available at test time, as stated below for easy reference: 
\begin{assumption}\label{ass:Atrue}
The protected attributes $A$ are known at test time.
\end{assumption}

We make the assumption that the probabilities $p_A(a)$ (MSP case) and $p_{A,Y}(a,y)$ (GEO case) together with their estimates are bounded away from zero.
\begin{assumption}\label{ass:pAY}
    For the MSP case, $p_A(a)$ and its estimate $\hat{p}_A(a)$ are bounded away from zero, i.e.,~$p_A(a) \geq \eta$ and $\hat{p}_A(a) \geq \eta$ for all $a \in \cA$ and some $\eta > 0$. For the GEO case, $p_{A,Y}(a,y) \geq \eta$ and $\hat{p}_{A,Y}(a,y) \geq \eta$ for all $a \in \cA$, $y \in \{0,1\}$, and some $\eta > 0$.
\end{assumption}

To ensure consistency of FST, we naturally assume that $\hat{r}(X)$ is a consistent estimator of the population score $r(X)$.
More specifically, we assume for theoretical purposes that $\hat{r}(X)$ is estimated from a data set of size $m$ that is independent of the data set of size $n$ used to approximate the expectation in \eqref{eqn:dualSpecF} (this might be obtained by splitting a larger data set into subsets of size $m$ and $n$.) The finite-sample bounds in the assumptions and theorems below are thus stated in terms of $m$. Different definitions of consistency suffice to prove different results. For the first definition, we view $\hat{r}(X)$ and $r(X)$ as random variables over $[0,1]$ induced by $X$ and define $D_{\mathrm{TV}}(R_1, R_2)$ to be the total variation distance between two such random variables,
\[
D_{\mathrm{TV}}(R_1, R_2) = \sup_{\mathcal{R}\subset[0,1]} \abs[]{\Pr(R_1 \in \mathcal{R}) - \Pr(R_2 \in \mathcal{R})}.
\]
\begin{assumption}\label{ass:rhatTV}
There exists a bound $E_{\mathrm{TV}}(m,\delta)$ as a function of $m$ and $\delta \in (0,1]$ such that 
\begin{enumerate}
    \item With probability at least $1-\delta$,
    \[
        \sum_{a\in\cA} p_A(a) D_{\mathrm{TV}}\bigl(\hat{r}(X) \given A=a, r(X) \given A=a\bigr) \leq E_{\mathrm{TV}}(m,\delta); 
    \]
    \item $E_{\mathrm{TV}}(m,\delta)$ is decreasing in $m$ for fixed $\delta$ (and decreases to zero as $m \to \infty$);
    \item $E_{\mathrm{TV}}(m,\delta)$ is increasing in $1/\delta$ for fixed $m$.
\end{enumerate}
\end{assumption}
The second definition 
involves convergence of $\hat{r}(X)$ to $r(X)$ in $L_1$ norm:
\begin{assumption}\label{ass:rhatL1}
There exists a bound $E_{L_1}(m,\delta)$ as a function of $m$ and $\delta \in (0,1]$ such that 
\begin{enumerate}
    \item With probability at least $1-\delta$,
    \[
        \EE{\abs{\hat{r}(X) - r(X)}} \leq E_{L_1}(m,\delta);
    \]
    \item $E_{L_1}(m,\delta)$ is decreasing in $m$ for fixed $\delta$ (and decreases to zero as $m \to \infty$);
    \item $E_{L_1}(m,\delta)$ is increasing in $1/\delta$ for fixed $m$.
\end{enumerate}
\end{assumption}
\noindent The third definition requires $\hat{r}(X)$ to converge to $r(X)$ in terms of the expectation of a Kullback-Leibler (KL) divergence. Define 
\begin{equation}\label{eqn:KL}
D_{\mathrm{KL}}(p\KLsep q) = p \log\left(\frac{p}{q}\right) + (1-p) \log\left(\frac{1-p}{1-q}\right)
\end{equation}
to be the KL divergence between Bernoulli random variables with parameters $p$ and $q$. The following assumption is stated only in terms of convergence in probability (as $m \to \infty$) as we do not make use of finite-sample bounds.
\begin{assumption}\label{ass:rhatKL}
The estimate $\hat{r}(X)$ converges to the population score $r(X)$ such that\\ $\EE{D_{\mathrm{KL}}(r(X) \KLsep \hat{r}(X))} \overset{p}{\to} 0$.
\end{assumption}
\noindent Note that the expectations in Assumptions~\ref{ass:rhatL1} and \ref{ass:rhatKL} are with respect to $X$. 

Lastly, we use the following assumption to show that it is sufficient to consider a bounded feasible set for the dual problem.
\begin{assumption}\label{ass:epsilon}
The fairness constraint parameter $\epsilon > 0$.
\end{assumption}
We also require a technical assumption to prove one of the lemmas, which we discuss in Section~\ref{sec:consistency:primalOpt}.

\subsection{Results}
\label{sec:consistency:results}
Property 2 stated at the beginning of Section~\ref{sec:consistency} is of primary importance as it pertains to the overall plug-in solution \eqref{eqn:plug-in} for the primal problem \eqref{eqn:primalGeneral}. The first theorem below addresses the degree to which the plug-in solution satisfies the population fairness constraints.

\begin{theorem}\label{thm:primalFeas}
In the MSP case, under Assumptions~\ref{ass:Atrue}, \ref{ass:pAY}, \ref{ass:epsilon}, with probability at least $1-\delta_1-\delta_2$ and $m > (2/\eta) \log(2L/\delta_1)$, the finite-sample plug-in solution $r'(x)$ in \eqref{eqn:plug-in} satisfies 
\begin{multline*}
    \abs*{\mathbb{E}[r'(X)\given A=a] - \mathbb{E}[r'(X)]}
    \leq \epsilon + \frac{\sqrt{2 \log(2L/\delta_1)}}{\sqrt{m p_A(a)} - \sqrt{2 \log(2L/\delta_1)}}\\
    {} + \left(\frac{1}{\eta} - 1\right) \left( \frac{4\log 2}{\epsilon} \left(\frac{1}{\eta} - 1\right) \sqrt{\frac{2\log(2L)}{n}} + \sqrt{\frac{2\log(2L/\delta_2)}{n}} + \frac{2}{\sqrt{n}} \right) \quad \forall \; a \in \cA,
\end{multline*}
where the terms after $\epsilon$ represent the excess with respect to the population fairness constraint \eqref{eqn:MSP}. In the GEO case, under Assumptions~\ref{ass:Atrue}, \ref{ass:pAY}, \ref{ass:rhatL1}, \ref{ass:epsilon}, with probability at least $1-\delta_1-\delta_2-\delta_3$ and $m > (2/\eta) \log(2(L+2)/\delta_1)$, the plug-in solution satisfies
\begin{align*}
    &\abs*{\mathbb{E}[r'(X)\given A=a, Y=y] - \mathbb{E}[r'(X) \given Y=y]}\\
    &\leq \epsilon + \frac{\sqrt{2 \log(2(L+2)/\delta_1)}}{\sqrt{m p_{A,Y}(a,y)} - \sqrt{2 \log(2(L+2)/\delta_1)}} + \frac{\sqrt{2 \log(2(L+2)/\delta_1)}}{\sqrt{m p_{Y}(y)} - \sqrt{2 \log(2(L+2)/\delta_1)}}\\
    &\quad {} + \left(\frac{1}{\eta} - 1\right) \left( \frac{4\log 2}{\epsilon} \left(\frac{1}{\eta} - 1\right) \sqrt{\frac{2\log(2L)}{n}} + \sqrt{\frac{2\log(2L/\delta_2)}{n}} + \frac{2}{\sqrt{n}} \right)\\
    &\quad {} + \left(\frac{1}{\eta} - 1\right) E_{L_1}(m, \delta_3) \quad \forall \; a \in \cA, \; y \in \{0,1\},
\end{align*}
where the terms after $\epsilon$ are the excess with respect to constraint \eqref{eqn:MEO}.
\end{theorem}

\noindent The next theorem asserts the asymptotic optimality of the plug-in primal solution.

\begin{theorem}\label{thm:primalOpt}
Under Assumptions~\ref{ass:Atrue}, \ref{ass:pAY}, \ref{ass:rhatTV}, \ref{ass:rhatKL}, \ref{ass:epsilon}, \ref{ass:strongConvex} and as $n \to \infty$, 
\[
-\EE{H_b\left(r(X), r^*\bigl(\hat{\lambda}^T \hat{\boldf}(A, \hat{r}(X)); \hat{r}(X)\bigr)\right)} + \EE{H_b\left(r(X), r^*\bigl(\lambda^{*T} \boldf(A, r(X)); r(X)\bigr)\right)} \overset{p}{\to} 0.
\]
The first term is the population primal objective evaluated at the plug-in solution while the second term is the optimal objective value.
\end{theorem}
Unlike Theorems~\ref{thm:primalFeas} and \ref{thm:dualConsistency}, Theorem~\ref{thm:primalOpt} does not provide finite-sample guarantees. We discuss reasons for not doing so in Section~\ref{sec:consistency:primalOpt}.

Property 1 (beginning of Section~\ref{sec:consistency}) pertains to the near optimality of empirical dual solutions. It is used to prove Theorem~\ref{thm:primalOpt} and may also be of independent interest. Let $J(\lambda)$ and $\hat{J}(\lambda)$ denote the objective functions in the population dual \eqref{eqn:dualMSP_rhat}, \eqref{eqn:dualMEO_rhat} and empirical dual \eqref{eqn:dualSpecF} respectively.
\begin{theorem}\label{thm:dualConsistency}
Let $\hat{\lambda} \in \argmin \hat{J}(\lambda)$ and $\lambda^* \in \argmin J(\lambda)$ be optimal solutions to the empirical dual problem \eqref{eqn:dualSpecF} and population dual problem \eqref{eqn:dualMSP_rhat}, \eqref{eqn:dualMEO_rhat} respectively. Under Assumptions~\ref{ass:Atrue}, \ref{ass:pAY}, \ref{ass:rhatTV}, \ref{ass:epsilon}, with probability at least $1 - \delta_1 - \delta_2 - \delta_3$ and $m > (2/\eta) \log(2(L+2)/\delta_3)$, we have
\begin{align*}
    &J(\hat{\lambda}) - J(\lambda^*)\\ 
    &\qquad \leq 2\log(2)  \left(1 + \frac{1}{\epsilon} \left(\frac{1}{\eta} - 1\right) \right) \left( 4 \sqrt{\frac{2 \log(2L)}{n}} + \sqrt{\frac{2 \log(2/\delta_1)}{n}} + E_{\mathrm{TV}}(m, \delta_2) \right) + \Delta,
\end{align*}
where in the MSP case, 
\[
    \Delta = \frac{2\log 2}{\epsilon} \sum_{a\in\cA} \frac{\sqrt{2 \log(2L/\delta_3)}}{\sqrt{m p_A(a)} - \sqrt{2 \log(2L/\delta_3)}},
\]
in the GEO case,
\begin{align*}
    \Delta = \frac{2\log 2}{\epsilon} &\left( \sum_{y\in\{0,1\}} \sum_{a\in\cA} \frac{\sqrt{2 \log(2(L+2)/\delta_3)}}{\sqrt{m p_{A,Y}(a,y)} - \sqrt{2 \log(2(L+2)/\delta_3)}} \right.\\
    &\left. \quad {} + \sum_{y\in\{0,1\}} \frac{\sqrt{2 \log(2(L+2)/\delta_3)}}{\sqrt{m p_{Y}(y)} - \sqrt{2 \log(2(L+2)/\delta_3)}} \right),
\end{align*}
and for an upper bound that covers both cases,
\[
    \Delta = \frac{2(L+2) \log 2}{\epsilon} \frac{\sqrt{2 \log(2(L+2)/\delta_3)}}{\sqrt{\eta m} - \sqrt{2 \log(2(L+2)/\delta_3)}}.
\]
\end{theorem}

We make the following remarks about the form of the bounds in Theorems~\ref{thm:primalFeas} and \ref{thm:dualConsistency}.
\begin{enumerate}
    \item The bounds are functions of two sample sizes: $n$, the number of data points that define the empirical dual \eqref{eqn:dualSpecF}, and $m$, the number of data points used to estimate $r(X)$ and $p_A(a)$ (in the MSP case) or $p_{A,Y}(a,y)$ (GEO case). The bounds have a familiar $1/\sqrt{n}$ dependence on $n$, and also on $m$ for the terms that correspond to estimation of $p_A(a)$ or $p_{A,Y}(a,y)$. The performance in estimating $r(X)$ is abstracted away by the error terms $E_{\mathrm{TV}}(m,\delta)$ and $E_{L_1}(m,\delta)$ defined in Assumptions~\ref{ass:rhatTV} and \ref{ass:rhatL1}.
    \item The dimension $L$ of the dual variable $\lambda$, already no more than $2\abs{\cA}$ to begin with, enters mostly in logarithmic form.
    \item The fairness tolerance $\epsilon$ and probability lower bound $\eta$ appear in the denominator (apart from the leading $\epsilon$ in Theorem~\ref{thm:primalFeas}). This agrees with the intuition that the problem becomes harder for stricter fairness constraints (smaller $\epsilon$) and smaller groups (smaller $\eta$). Some of the terms further specify the dependence on individual probabilities $p_A(a)$, $p_{A,Y}(a,y)$, $p_Y(y)$, which could be bounded by $\eta$ to simplify expressions.
\end{enumerate}

\subsection{Discussion and Basic Implications of Assumptions}
\label{sec:consistency:implications}

We now elaborate upon the assumptions stated in Section~\ref{sec:consistency:ass}.

\subsubsection{Assumption~\ref{ass:Atrue}} Under this assumption, $\mu(X) = \lambda^T \boldf(X)$ is given by \eqref{eqn:muMSPAtrue} (MSP) or \eqref{eqn:muMEOAtrue} (GEO). In this case, $\boldf$ depends on $X$ only through $A$ and $r(X)$ and we will often use the notation $\boldf(A, r(X))$ to make this clear. Below we give expressions for $\boldf$ for future reference. For the MSP case \eqref{eqn:muMSPAtrue}, $\boldf$ has $\abs{\cA}$ components and the $a$th component is given by 
\begin{equation}\label{eqn:fMSP}
f_a(X) = f_a(A) = \frac{\ones(A=a)}{p_A(a)} - 1.
\end{equation}
For the GEO case \eqref{eqn:muMEOAtrue}, $\boldf$ has $2\abs{\cA}$ components and the $(a,y)$ component is 
\begin{equation}\label{eqn:fMEO}
f_{a,y}(X) = f_{a,y}(A, r(X)) = \begin{cases}
\frac{1-r(X)}{p_Y(0)} \left(\frac{\ones(A=a)}{p_{A\given Y}(a\given 0)} - 1\right), & y = 0\\
\frac{r(X)}{p_Y(1)} \left(\frac{\ones(A=a)}{p_{A\given Y}(a\given 1)} - 1\right), & y = 1.
\end{cases}
\end{equation}
For the estimate $\hat{\boldf}$ of $\boldf$, $p_A$ in \eqref{eqn:fMSP} is replaced by its estimate $\hat{p}_A$, and $p_Y$, $p_{A\given Y}$ (equivalently $p_{A,Y}$) in \eqref{eqn:fMEO} are replaced by their estimates $\hat{p}_Y$, $\hat{p}_{A\given Y}$ ($\hat{p}_{A,Y}$).

The proofs can be extended to the case in which $A$ is not known by also assuming a consistent estimator of the conditional probability $p_{A\given X}$ in the MSP case or $p_{A\given X,Y}$ in the GEO case and accounting for the error of this estimator.

\subsubsection{Assumption~\ref{ass:pAY}} 
This assumption is reasonable in that if a protected group is to be considered, it should represent a constant fraction of the population (and have non-negligible probabilities of being in classes $0$ and $1$). The boundedness of the estimated probabilities can be ensured by truncating them, i.e.,~setting $\hat{p}_A(a) \leftarrow \max\{\hat{p}_A(a), \eta\}$. If the minimum probability $p_A(a)$ or $p_{A,Y}(a,y)$ is known or imposed, $\eta$ can be set equal to this minimum probability. Note also that we must have $\eta \leq 1/\abs{\cA}$ for MSP and $\eta \leq 1/\bigl(2\abs{\cA}\bigr)$ for GEO, as otherwise $p_A$, $p_{A,Y}$ would sum to more than $1$.

We further assume that the estimates $\hat{p}_A(a)$ and $\hat{p}_{A,Y}(a,y)$ are given by the corresponding empirical probabilities in a data set of size $m$. Each of these empirical probabilities is a binomial random variable with sample size parameter $m$ and scaled by $1/m$. Among many possible concentration inequalities,
we make use of the following bound on the relative error. It follows from a Chernoff bound, as shown in Appendix~\ref{sec:consistency:chernoffRel} for completeness.
\begin{lemma}\label{lem:chernoffRel}
With probability at least $1-\delta$, for any single $a \in \cA$ or $(a,y) \in \cA \times \{0,1\}$,
\begin{align*}
    \abs*{\frac{p_A(a)}{\hat{p}_A(a)} - 1} &\leq \frac{\sqrt{2 \log(2/\delta)}}{\sqrt{m p_A(a)} - \sqrt{2 \log(2/\delta)}}, \qquad &m p_A(a) > 2\log(2/\delta),\\
    \abs*{\frac{p_{A,Y}(a,y)}{\hat{p}_{A,Y}(a,y)} - 1} &\leq \frac{\sqrt{2 \log(2/\delta)}}{\sqrt{m p_{A,Y}(a,y)} - \sqrt{2 \log(2/\delta)}}, \qquad &m p_{A,Y}(a,y) > 2\log(2/\delta).
\end{align*}
\end{lemma}
Note also that under Assumption~\ref{ass:pAY}, truncating the estimated probabilities at $\eta$ can only decrease the error and hence does not affect the bounds above.

\subsubsection{Assumption~\ref{ass:rhatTV}--\ref{ass:rhatKL}}
In Assumptions~\ref{ass:rhatTV} and \ref{ass:rhatL1}, the properties of $E_{\mathrm{TV}}(m,\delta)$ and $E_{L_1}(m,\delta)$ imply that $D_{\mathrm{TV}}\bigl(\hat{r}(X) \given A=a, r(X) \given A=a\bigr)$ and $\EE{\abs{\hat{r}(X) - r(X)}}$ converge to zero in probability, similar to Assumption~\ref{ass:rhatKL}. This is true because for any deviation $E_{\mathrm{TV}}(m,\delta) > 0$ (similarly $E_{L_1}(m,\delta)$) and keeping $E_{\mathrm{TV}}(m,\delta)$ fixed, increasing $m$ requires increasing $1/\delta$ to compensate. Taking $m \to \infty$ thus drives $\delta$ (the probability of exceeding $E_{\mathrm{TV}}(m,\delta)$) to zero.

In Appendix~\ref{sec:consistency:KL_L1}, it is shown that Assumption~\ref{ass:rhatKL} implies the convergence in probability version of Assumption~\ref{ass:rhatL1}.
\begin{assumption}\label{ass:rhatL1convProb}
The estimate $\hat{r}(X)$ converges to the population score $r(X)$ in $L_1$ norm: $\EE{\abs{\hat{r}(X) - r(X)}} \overset{p}{\to} 0$.
\end{assumption}
\begin{lemma}\label{lem:KL_L1}
Assumption~\ref{ass:rhatKL} implies Assumption~\ref{ass:rhatL1convProb}.
\end{lemma}
We list Assumption~\ref{ass:rhatL1convProb} separately as the proof of Lemma~\ref{lem:primalConsistency2} below (for Theorem~\ref{thm:primalOpt}) requires only Assumption~\ref{ass:rhatL1convProb}, not Assumption~\ref{ass:rhatKL}.

\subsection{Proof Outlines}
\label{sec:consistency:outlines}

We begin with the proof of Theorem~\ref{thm:dualConsistency} as it contains elements that are reused in the proofs of Theorem~\ref{thm:primalFeas} and \ref{thm:primalOpt}.

\subsubsection{Asymptotic Dual Optimality (Theorem~\ref{thm:dualConsistency})}
\label{sec:consistency:dual}

\begin{proof}
We prove the theorem by deriving a uniform convergence bound on the absolute difference $\abs{\hat{J}(\lambda) - J(\lambda)}$. Then if $\varepsilon$ is such a bound (that holds with high probability),
we have 
\begin{equation}\label{eqn:JoptGap}
J(\hat{\lambda}) \leq \hat{J}(\hat{\lambda}) + \varepsilon \leq \hat{J}(\lambda^*) + \varepsilon \leq J(\lambda^*) + 2\varepsilon,
\end{equation}
where the second inequality is by definition of $\hat{\lambda}$.

Toward proving uniform convergence, we first establish that it suffices to solve the dual problem over a closed and bounded (and hence compact) feasible set. The same argument applies to both the population and empirical duals. Indeed, it always suffices to restrict to a sub-level set defined by the objective value of an initial solution. We take $\lambda = 0$ as the initial solution and consider $\{\lambda: J(\lambda) \leq J(0)\}$ and $\{\lambda: \hat{J}(\lambda) \leq \hat{J}(0)\}$. These sub-level sets are contained within an $\ell_1$ ball as proved in Appendix~\ref{sec:consistency:Lambda0}.
\begin{lemma}\label{lem:Lambda0}
Given Assumption~\ref{ass:epsilon}, define the $\ell_1$ ball
\begin{equation}
\nonumber
\Lambda_0 = \left\{\lambda: \norm{\lambda}_1 \leq \frac{\log 2}{\epsilon} \right\}.
\end{equation}
Then we have $\{\lambda: J(\lambda) \leq J(0)\} \subset \Lambda_0$ and $\{\lambda: \hat{J}(\lambda) \leq \hat{J}(0)\} \subset \Lambda_0$.
\end{lemma}
Henceforth we take $\Lambda_0$ to be the compact feasible set for the dual problem.

We then consider the supremum over $\Lambda_0$ of the absolute difference $\abs{\hat{J}(\lambda) - J(\lambda)}$ as the quantity of interest for uniform convergence. We use the triangle inequality and separate suprema to decompose this into three terms:
\begin{align}\label{eqn:JunifConv}
\sup_{\lambda\in\Lambda_0} \abs[\big]{\hat{J}(\lambda) - J(\lambda)} \leq& \sup_{\lambda\in\Lambda_0} \abs*{\frac{1}{n} \sum_{i=1}^n g\bigl(\lambda^T \hat{\boldf}(a_i, \hat{r}(x_i)); \hat{r}(x_i)\bigr) - \EE{g\bigl(\lambda^T \hat{\boldf}(A, \hat{r}(X)); \hat{r}(X)\bigr)}}\nonumber\\
&{} + \sup_{\lambda\in\Lambda_0} \abs*{\EE{g\bigl(\lambda^T \hat{\boldf}(A, \hat{r}(X)); \hat{r}(X)\bigr)} - \EE{g\bigl(\lambda^T \hat{\boldf}(A, r(X)); r(X)\bigr)}} \nonumber\\
&{} + \sup_{\lambda\in\Lambda_0} \abs*{\EE{g\bigl(\lambda^T \hat{\boldf}(A, r(X)); r(X)\bigr)} - \EE{g\bigl(\lambda^T \boldf(A, r(X)); r(X)\bigr)}}.
\end{align}
 The first right-hand side quantity in \eqref{eqn:JunifConv} is the difference between the empirical average and expectation of the same quantity. The second difference is due to having $\hat{r}(X)$ instead of $r(X)$, and the third is due to having $\hat{\boldf}$ instead of $\boldf$.

The following three lemmas, proven in Appendix~\ref{sec:consistency:dualProofs}, provide bounds on 
the three right-hand side terms in \eqref{eqn:JunifConv}. Combining them with probabilities $\delta_1$, $\delta_2$, $\delta_3$ and including the factor of $2$ from \eqref{eqn:JoptGap} completes the proof of the theorem.
\begin{lemma}\label{lem:dualConsistency1}
Under Assumptions~\ref{ass:Atrue}, \ref{ass:pAY}, \ref{ass:epsilon} and with probability at least $1 - \delta$,
\begin{align*}
    &\sup_{\lambda\in\Lambda_0} \abs*{\frac{1}{n} \sum_{i=1}^n g\bigl(\lambda^T \hat{\boldf}(a_i, \hat{r}(x_i)); \hat{r}(x_i)\bigr) - \EE{g\bigl(\lambda^T \hat{\boldf}(A, \hat{r}(X)); \hat{r}(X)\bigr)}}\\
    &\qquad \leq \left(1 + \frac{1}{\epsilon} \left(\frac{1}{\eta} - 1\right) \right) (\log 2) \left(4 \sqrt{\frac{2 \log(2L)}{n}} + \sqrt{\frac{2 \log(2/\delta)}{n}} \right).
\end{align*}
\end{lemma}
\begin{lemma}\label{lem:dualConsistency3}
Under Assumptions~\ref{ass:Atrue}, \ref{ass:pAY}, 
\ref{ass:rhatTV}, \ref{ass:epsilon} and with probability $1-\delta$,
\begin{align*}
\sup_{\lambda\in\Lambda_0} \abs*{\EE{g\bigl(\lambda^T \hat{\boldf}(A, \hat{r}(X)); \hat{r}(X)\bigr)} - \EE{g\bigl(\lambda^T \hat{\boldf}(A, r(X)); r(X)\bigr)}}\\
\leq \left(1 + \frac{1}{\epsilon} \left(\frac{1}{\eta} - 1\right) \right) (\log 2) E_{\mathrm{TV}}(m, \delta).
\end{align*}
\end{lemma}
\begin{lemma}\label{lem:dualConsistency2}
Under Assumptions~\ref{ass:Atrue}, \ref{ass:pAY}, and \ref{ass:epsilon} and with probability at least $1-\delta$ and $m > (2/\eta) \log(2(L+2)/\delta)$, in the MSP case,
\begin{align*}
&\sup_{\lambda\in\Lambda_0} \abs*{\EE{g\bigl(\lambda^T \hat{\boldf}(A, r(X)); r(X)\bigr)} - \EE{g\bigl(\lambda^T \boldf(A, r(X)); r(X)\bigr)}}\\ 
&\qquad \leq \frac{\log 2}{\epsilon} \sum_{a\in\cA} \frac{\sqrt{2 \log(2L/\delta)}}{\sqrt{m p_A(a)} - \sqrt{2 \log(2L/\delta)}},
\end{align*}
and in the GEO case,
\begin{align*}
    &\sup_{\lambda\in\Lambda_0} \abs*{\EE{g\bigl(\lambda^T \hat{\boldf}(A, r(X)); r(X)\bigr)} - \EE{g\bigl(\lambda^T \boldf(A, r(X)); r(X)\bigr)}}\\ 
    &\qquad \leq \frac{\log 2}{\epsilon} \left( \sum_{y\in\{0,1\}} \sum_{a\in\cA} \frac{\sqrt{2 \log(2(L+2)/\delta)}}{\sqrt{m p_{A,Y}(a,y)} - \sqrt{2 \log(2(L+2)/\delta)}} \right.\\
    &\qquad \qquad \qquad {} + \left. \sum_{y\in\{0,1\}} \frac{\sqrt{2 \log(2(L+2)/\delta)}}{\sqrt{m p_{Y}(y)} - \sqrt{2 \log(2(L+2)/\delta)}} \right),
\end{align*}
where $L = \dim(\lambda)$.
\end{lemma}
\end{proof}

\subsubsection{Asymptotic Primal Feasibility (Theorem~\ref{thm:primalFeas})}
\label{sec:consistency:primalFeas}

\begin{proof}
By retracing the derivation of dual problems \eqref{eqn:dualMSP_rhat}, \eqref{eqn:dualMEO_rhat} from the primal problem \eqref{eqn:primalGeneral}, it can be verified that the \emph{empirical} primal corresponding to the empirical dual \eqref{eqn:dualSpecF} is
\begin{equation}\label{eqn:primalEmp}
    \max_{r'} \quad -\frac{1}{n} \sum_{i=1}^n H_b\bigl(\hat{r}(x_i), r'(x_i)\bigr) \qquad
    \st \qquad \abs*{\frac{1}{n} \sum_{i=1}^n \hat{f}_l(a_i, \hat{r}(x_i)) r'(x_i)} \leq \epsilon \quad \forall \; l, 
\end{equation}
where $l = a$ ranges over $\cA$ in the MSP case and $l = (a,y)$ ranges over $\cA \times \{0,1\}$ in the GEO case. Since $\hat{\lambda}$ optimizes the empirical dual, it follows from the discussion in Section~\ref{sec:opt} that the plug-in solution \eqref{eqn:plug-in} satisfies the primal fairness constraints in \eqref{eqn:primalEmp}. The task is to bound the amount by which the plug-in solution violates the population MSP \eqref{eqn:MSP} or GEO \eqref{eqn:MEO} constraints.

Using the definitions of $\boldf(A, r(X))$ for the MSP \eqref{eqn:fMSP} and GEO \eqref{eqn:fMEO} cases, it can be seen that constraints \eqref{eqn:MSP} and \eqref{eqn:MEO} are equivalent to 
\begin{equation}\label{eqn:fairnessf}
\abs*{\EE{f_l(A, r(X)) r'(X)}} \leq \epsilon \quad \forall \; l,
\end{equation}
where $l$ ranges over the same values as in \eqref{eqn:primalEmp}. Therefore by the triangle inequality, the violation of constraint $l$ in \eqref{eqn:fairnessf} is bounded by the difference 
\[
\abs*{\frac{1}{n} \sum_{i=1}^n \hat{f}_l(a_i, \hat{r}(x_i)) r'(x_i) - \EE{f_l(A, r(X)) r'(X)}}.
\] We apply the triangle inequality again to separate this difference into three terms that are analyzed below:
\begin{align}
    &\abs*{\frac{1}{n} \sum_{i=1}^n \hat{f}_l(a_i, \hat{r}(x_i)) r'(x_i) - \EE{f_l(A, r(X)) r'(X)}}\nonumber\\ 
    &\qquad \leq \abs*{\frac{1}{n} \sum_{i=1}^n \hat{f}_l(a_i, \hat{r}(x_i)) r'(x_i) - \EE{\hat{f}_l(A, \hat{r}(X)) r'(X)}}\nonumber\\
    &\qquad \quad {} + \abs*{\EE{\hat{f}_l(A, \hat{r}(X)) r'(X)} - \EE{\hat{f}_l(A, r(X)) r'(X)}}\nonumber\\
    &\qquad \quad {} + \abs*{\EE{\hat{f}_l(A, r(X)) r'(X)} - \EE{f_l(A, r(X)) r'(X)}}.\label{eqn:fairConsistency}
\end{align}

For the first right-hand side term in \eqref{eqn:fairConsistency}, we substitute in the plug-in solution \eqref{eqn:plug-in} for $r'(X)$. To remove the dependence on $\hat{\lambda}$ (which is a function of the samples $i=1,\dots,n$ to which it is fit), we consider a uniform bound over $\Lambda_0$:
\begin{align*}
&\abs*{\frac{1}{n} \sum_{i=1}^n \hat{f}_l(a_i, \hat{r}(x_i)) r^*\bigl(\hat{\lambda}^T \hat{\boldf}(a_i, \hat{r}(x_i)); \hat{r}(x_i)\bigr) - \EE{\hat{f}_l(A, \hat{r}(X)) r^*\bigl(\hat{\lambda}^T \hat{\boldf}(A, \hat{r}(X)); \hat{r}(X)\bigr)}}\\
&\leq \sup_{\lambda\in\Lambda_0} \abs*{\frac{1}{n} \sum_{i=1}^n \hat{f}_l(a_i, \hat{r}(x_i)) r^*\bigl(\lambda^T \hat{\boldf}(a_i, \hat{r}(x_i)); \hat{r}(x_i)\bigr) - \EE{\hat{f}_l(A, \hat{r}(X)) r^*\bigl(\lambda^T \hat{\boldf}(A, \hat{r}(X)); \hat{r}(X)\bigr)}}.
\end{align*}
The following bound is derived in Appendix~\ref{sec:consistency:primalFeasProofs} using statistical learning theory tools similar to the proof of Lemma~\ref{lem:dualConsistency1}.
\begin{lemma}\label{lem:fairConsistency0}
Under Assumptions~\ref{ass:Atrue}, \ref{ass:pAY}, \ref{ass:epsilon}, with probability at least $1-\delta$,
\begin{align*}
&\sup_{\lambda\in\Lambda_0} \abs*{\frac{1}{n} \sum_{i=1}^n \hat{f}_l(a_i, \hat{r}(x_i)) r^*\bigl(\lambda^T \hat{\boldf}(a_i, \hat{r}(x_i)); \hat{r}(x_i)\bigr) - \EE{\hat{f}_l(A, \hat{r}(X)) r^*\bigl(\lambda^T \hat{\boldf}(A, \hat{r}(X)); \hat{r}(X)\bigr)}}\\
&\qquad \leq \left(\frac{1}{\eta} - 1\right) \left( \frac{4\log 2}{\epsilon} \left(\frac{1}{\eta} - 1\right) \sqrt{\frac{2\log(2L)}{n}} + \sqrt{\frac{2\log(2L/\delta)}{n}} + \frac{2}{\sqrt{n}} \right) \quad \forall \; l.
\end{align*}
\end{lemma}

In the case of MSP, the second term in \eqref{eqn:fairConsistency} is zero because $\boldf$ does not depend on its second argument $r(X)$. Using \eqref{eqn:fMSP}, the third term in \eqref{eqn:fairConsistency} reduces as follows: 
\begin{align}
    \abs*{\EE{\left(\frac{\ones(A=a)}{\hat{p}_A(a)} - \frac{\ones(A=a)}{p_A(a)}\right) r'(X)}}
    &= \abs*{\EE{\left(\frac{p_A(a)}{\hat{p}_A(a)} - 1\right) r'(X) \given A=a}}\nonumber\\
    &= \abs*{\frac{p_A(a)}{\hat{p}_A(a)} - 1} \abs*{\EE{r'(X) \given A=a}}\nonumber\\
    &\leq \abs*{\frac{p_A(a)}{\hat{p}_A(a)} - 1}\nonumber\\
    &\leq \frac{\sqrt{2 \log(2L/\delta)}}{\sqrt{m p_A(a)} - \sqrt{2 \log(2L/\delta)}},\label{eqn:fairConsistency2_0}
\end{align} 
where the first inequality is due to $\abs{r'(X)} \leq 1$, and the second inequality from Lemma~\ref{lem:chernoffRel} holds with probability at least $1-\delta/L$. By a union bound, \eqref{eqn:fairConsistency2_0} is true for all $a \in \cA$ with probability at least $1-\delta$ and $m$ large enough for the denominator to be positive.

In the GEO case, we prove in Appendix~\ref{sec:consistency:primalFeasProofs} that the second and third terms in \eqref{eqn:fairConsistency} are bounded as follows.
\begin{lemma}\label{lem:fairConsistency1}
In the GEO case, under Assumptions~\ref{ass:Atrue}, \ref{ass:pAY}, \ref{ass:rhatL1} and with probability at least $1-\delta$,
\[
\abs*{\EE{\hat{f}_{a,y}(A, \hat{r}(X)) r'(X)} - \EE{\hat{f}_{a,y}(A, r(X)) r'(X)}}
\leq \left(\frac{1}{\eta} - 1\right) E_{L_1}(m, \delta) \quad \forall \; (a,y).
\]
\end{lemma}
\begin{lemma}\label{lem:fairConsistency2}
In the GEO case, under Assumptions~\ref{ass:Atrue} and \ref{ass:pAY} and with probability at least $1-\delta$ and $m > (2/\eta) \log(2(L+2)/\delta)$,
\begin{align*}
&\abs*{\EE{\hat{f}_{a,y}(A, r(X)) r'(X)} - \EE{f_{a,y}(A, r(X)) r'(X)}}\\ 
&\leq \frac{\sqrt{2 \log(2(L+2)/\delta)}}{\sqrt{m p_{A,Y}(a,y)} - \sqrt{2 \log(2(L+2)/\delta)}} + \frac{\sqrt{2 \log(2(L+2)/\delta)}}{\sqrt{m p_{Y}(y)} - \sqrt{2 \log(2(L+2)/\delta)}} \quad \forall \; (a,y).
\end{align*}
\end{lemma}
\end{proof}

\subsubsection{Asymptotic Primal Optimality (Theorem~\ref{thm:primalOpt})}
\label{sec:consistency:primalOpt}

\begin{proof}
We use the triangle inequality to bound the difference by the absolute sum of three differences:
\begin{align}
    &\abs*{\EE{H_b\left(r(X), r^*\bigl(\hat{\lambda}^T \hat{\boldf}(A, \hat{r}(X)); \hat{r}(X)\bigr)\right)} - \EE{H_b\left(r(X), r^*\bigl(\lambda^{*T} \boldf(A, r(X)); r(X)\bigr)\right)} }\nonumber\\
    &\quad \leq \abs*{\EE{H_b\left(r(X), r^*\bigl(\hat{\lambda}^T \hat{\boldf}(A, \hat{r}(X)); \hat{r}(X)\bigr)\right)} - \EE{H_b\left(r(X), r^*\bigl(\hat{\lambda}^T \hat{\boldf}(A, \hat{r}(X)); r(X)\bigr)\right)}}\nonumber\\
    &\quad \quad {} + \abs*{\EE{H_b\left(r(X), r^*\bigl(\hat{\lambda}^T \hat{\boldf}(A, \hat{r}(X)); r(X)\bigr)\right)} - \EE{H_b\left(r(X), r^*\bigl(\hat{\lambda}^T \boldf(A, r(X)); r(X)\bigr)\right)}}\nonumber\\
    &\quad \quad {} + \abs*{\EE{H_b\left(r(X), r^*\bigl(\hat{\lambda}^T \boldf(A, r(X)); r(X)\bigr)\right)} - \EE{H_b\left(r(X), r^*\bigl(\lambda^{*T} \boldf(A, r(X)); r(X)\bigr)\right)} }\nonumber.
\end{align}
The first difference is due to having $\hat{r}(X)$ instead of $r(X)$ as the second argument to $r^*$, i.e.,~as the input score to the transformation. The second difference is due to having $\hat{\boldf}(A, \hat{r}(X))$ versus $\boldf(A, r(X))$, and the third to $\hat{\lambda}$ versus $\lambda^*$.

The following lemmas, proven in Appendix~\ref{sec:consistency:primalOptProofs}, ensure that the three differences above converge to zero. 
\begin{lemma}\label{lem:primalConsistency1}
Under Assumption~\ref{ass:rhatKL},
\[
\abs*{\EE{H_b\left(r(X), r^*\bigl(\hat{\lambda}^T \hat{\boldf}(A, \hat{r}(X)); \hat{r}(X)\bigr)\right)} - \EE{H_b\left(r(X), r^*\bigl(\hat{\lambda}^T \hat{\boldf}(A, \hat{r}(X)); r(X)\bigr)\right)}} \overset{p}{\to} 0.
\]
\end{lemma}
\begin{lemma}\label{lem:primalConsistency2}
Under Assumptions~\ref{ass:Atrue}, \ref{ass:pAY}, \ref{ass:epsilon}, and \ref{ass:rhatL1convProb} (implied by Assumption~\ref{ass:rhatKL}),
\[
\abs*{\EE{H_b\left(r(X), r^*\bigl(\hat{\lambda}^T \hat{\boldf}(A, \hat{r}(X)); r(X)\bigr)\right)} - \EE{H_b\left(r(X), r^*\bigl(\hat{\lambda}^T \boldf(A, r(X)); r(X)\bigr)\right)}} \overset{p}{\to} 0.
\]
\end{lemma}
\begin{lemma}\label{lem:primalConsistency6}
Under Assumptions~\ref{ass:Atrue}, \ref{ass:pAY}, \ref{ass:rhatTV}, \ref{ass:epsilon}, \ref{ass:strongConvex}, 
\[
\abs*{\EE{H_b\left(r(X), r^*\bigl(\hat{\lambda}^T \boldf(A, r(X)); r(X)\bigr)\right)} - \EE{H_b\left(r(X), r^*\bigl(\lambda^{*T} \boldf(A, r(X)); r(X)\bigr)\right)}} \overset{p}{\to} 0.
\]
\end{lemma}
\end{proof}

The proof of Lemma~\ref{lem:primalConsistency6} leverages the asymptotic dual optimality of $\hat{\lambda}$ as $n, m \to \infty$, implied by Theorem~\ref{thm:dualConsistency}. In addition, we use the following assumption, where we define $s(\mu; r) = -\partial r^*(\mu; r) / \partial\mu$.
\begin{assumption}\label{ass:strongConvex}
For any empirical dual solution $\hat{\lambda}$ and any $\bar{\lambda}$ on the line segment between $\hat{\lambda}$ and a population dual solution $\lambda^*$ (i.e.,~$\bar{\lambda} = \alpha \hat{\lambda} + (1-\alpha) \lambda^*$ for $\alpha \in [0,1]$), there exists $\tau > 0$ such that 
\[
\EE{s\bigl(\bar{\lambda}^T \boldf(A, r(X)); r(X)\bigr) \abs*{\bigl(\hat{\lambda} - \lambda^*\bigr)^T \boldf(A, r(X))}^2} \geq \tau \EE{\abs*{\bigl(\hat{\lambda} - \lambda^*\bigr)^T \boldf(A, r(X))}^2}.
\]
\end{assumption}
Assumption~\ref{ass:strongConvex} is a form of strong convexity assumption on the first term\linebreak[4] $\EE{g\bigl(\lambda^T \boldf(A, r(X)); r(X)\bigr)}$ in the population dual objective function, as will be seen in the proof of Lemma~\ref{lem:primalConsistency6}. The right-hand expectation in Assumption~\ref{ass:strongConvex} is an $L_2$ norm between $\hat{\mu}(A, r(X)) = \hat{\lambda}^T \boldf(A, r(X))$ and $\mu^*(A, r(X)) = \lambda^{*T} \boldf(A, r(X))$, while the left-hand expectation is an $L_2$ norm weighted by $s(\mu; r)$. It can be seen from Figure~\ref{fig:rStar_mu} and verified using the expression in \eqref{eqn:d2f(tlambda)} that $s(\mu; r) \geq 0$ everywhere and $s(\mu; r) > 0$ for $r \in (0,1)$. Hence, the assumption of a lower bound $\tau > 0$ is reasonable. However, whether Assumption~\ref{ass:strongConvex} is satisfied depends on the distribution of the induced random variable $r(X)$ in a way that does not seem straightforward to characterize. Since $s(\mu; r)$ can be zero for $r = 0$ or $r = 1$, one requirement may be that $r(X)$ not have all of its probability mass at $0$ and $1$. It might also be possible in future work to prove Lemma~\ref{lem:primalConsistency6} without Assumption~\ref{ass:strongConvex}.

Due in part to Assumption~\ref{ass:strongConvex}, in this work we do not pursue finite-sample guarantees or convergence rates to augment Theorem~\ref{thm:primalOpt}. Such bounds or rates would depend on the parameter $\tau$, which is not easy to interpret (and moreover may not be necessary). In addition, the proof of Lemma~\ref{lem:primalConsistency1} is fairly involved and obtaining a rate for it does not appear straightforward.

\section{Empirical Evaluation}
\label{sec:expt}

This section discusses experimental evaluation of the proposed FST methods for MSP and GEO constraints and both the direct post-processing solution as well as the pre-processing extension.

\subsection{Experimental Setup}
\label{sec:expt:setup}

We begin by describing the experimental setup, covering data sets, fairness methods, base classifiers, and metrics.

\subsubsection{Data Sets} Four data sets were used, 
the first three of which are standard in the fairness literature: 1) Adult Income, 2) ProPublica's COMPAS recidivism, 3) German credit risk, 4) Medical Expenditure Panel Survey (MEPS). Specifically, we used versions pre-processed by
an open-source library for algorithmic 
fairness \citep{aif360-oct-2018}. Each data set was randomly split $10$ times into training ($75\%$) and test ($25\%$) sets and all methods were subject to the same splits.

To facilitate comparison with other methods in Sections~\ref{sec:expt:Atest} and \ref{sec:expt:noAtest}, we used binary-valued protected attributes and consider gender and race for both adult and COMPAS, age for German, and race for MEPS. The resulting data set statistics are shown in Table~\ref{tab:datasets}. In Section~\ref{sec:expt:adult_both}, we also evaluate FST on the Adult Income data set with both gender and race as protected attributes (i.e.,~four protected groups corresponding to the combinations).

\begin{table}[t]
    \small
    \centering
    \begin{tabular}{lrrrr}
    \toprule
    & Adult & COMPAS & German & MEPS \\
    \midrule
    number of instances & $45222$ & $6167$ & $1000$ & $15830$ \\
    number of features & $13$ & $10$ & $20$ & $41$ \\
    \hspace{5mm} after one-hot encoding & $98$ & $401$ & $58$ & $138$\\
    percentage in positive class & $24.8$ & $54.5$ & $70.0$ & $17.2$ \\
    protected attribute 1 & gender & gender & age & race \\
    \hspace{5mm} percentage in majority group & $67.5$ & $81.0$ & $85.1$ & $64.3$ \\
    protected attribute 2 & race & race \\
    \hspace{5mm} percentage in majority group & $86.0$ & $65.9$ \\
    \bottomrule
    \end{tabular}
    \caption{Data set statistics}
    \label{tab:datasets}
\end{table}

\subsubsection{Methods Compared}
Since FST is intended for post- and pre-processing, comparisons to other post- and pre-processing methods are most natural as they accommodate situations a)--c) in Section~\ref{sec:intro}. For post-processing, we have chosen the method of \citet{hardt2016} (HPS) and the reject option method of \citet{kamiran2012}, both as implemented by \citet{aif360-oct-2018}, as well as the Wass-1 Post-Process $\hat{p}_S$ method (WPP) of \citet{jiang2019wasserstein}. For pre-processing, the massaging and reweighing methods of \citet{kamiranC2012} and the optimization method of \citet{calmon2017} (OPP) were chosen.  Among in-processing methods, meta-algorithms that work with essentially any base classifier can handle situation b). The reductions method of \citet{agarwal2018} (`red') was selected from this class. We also compared to in-processing methods specific to certain types of classifiers, which do not allow for any of a)--c): fairness constraints (FC) \citep{zafar2017constraints}, disparate mistreatment (DM) \citep{zafar2017fairness}, and fair empirical risk minimization (FERM) \citep{donini2018}. Lastly, availability of code was an important criterion.

The methods in the previous paragraph have various limitations, summarized by Table~\ref{tab:capabilities}, that affect the design of the experiments. 
First, the post-processing methods \citep[specifically the WPP variant for the last one]{hardt2016, kamiran2012, jiang2019wasserstein} require knowledge of the protected attribute $A$ at test time. Accordingly, the experiments presented in Section~\ref{sec:expt:Atest} include $A$ in the features $X$ to make it available to all methods; experiments without $A$ 
at test time \citep[excluding][]{hardt2016, kamiran2012, jiang2019wasserstein} are presented in Section~\ref{sec:expt:noAtest}.  We also encountered computational problems with the methods of 
\citet{calmon2017,zafar2017fairness} and thus perform separate comparisons with FST on reduced feature sets, reported in Appendix~\ref{sec:exptAdd}. 

Three versions of FST were evaluated: direct post-processing (FSTpost), the pre-processing extension (FSTpre), and a second post-processing version (FSTbatch) that assumes that test instances can be processed in a batch rather than one by one.  In this case, the fitting of the dual variables that parametrize FST (Section~\ref{sec:proc:ADMM})
can actually be done on test data since it does not depend on labels $y_i$ (and uses only predicted probabilities for $A$ if $A$ is unavailable at test time).

\subsubsection{Base Classifiers} We used $\ell_1$-regularized logistic regression (LR) and gradient boosted classification trees (GBM) from scikit-learn \citep{scikit-learn} as base classifiers. These are used in different ways depending on the method: 
Post-processing methods operate on the scores produced by the base classifier, pre-processing methods train the base classifier after modifying the training data, and the reductions method repeatedly calls the base classification algorithm with different instance-specific costs. For FSTpre, the same base classifier is used both to obtain weights $w_i$ as well as to fit the re-weighted data.
In Appendix~\ref{sec:exptAdd}, we used linear SVMs \citep[with the scaling of][to output probabilities]{platt1999} to compare with FERM \citep{donini2018}. We found it impractical to train nonlinear SVMs on the larger data sets for reductions and FERM since reductions needs to do so repeatedly and FERM uses a slower specialized algorithm.  For a similar reason, $5$-fold cross-validation to select parameters for LR (regularization parameter $C$ from $[10^{-4}, 10^4]$) and GBM (minimum number of samples per leaf from $\{5,10,15,20,30\}$) was done only once per training set. All other parameters were set to the scikit-learn defaults. The base classifier was then instantiated with the best parameter value for use by all methods.

\begin{table}[t]
    \small
    \centering
    \begin{tabular}{cccccccccc}
    \toprule
    method & pre & in & post & SP & EO & no $A$ at test time & scores & approx fair & any classifier\\
    \midrule
    massage & \checkmark & & & \checkmark & & \checkmark & \checkmark & & \checkmark\\
    reweigh & \checkmark & & & \checkmark & & \checkmark & \checkmark & & \checkmark\\
    OPP & \checkmark & & & \checkmark & & \checkmark & \checkmark & \checkmark & \checkmark\\
    HPS & & & \checkmark & & \checkmark & & & & \checkmark\\
    reject & & & \checkmark & \checkmark & $\star$ & & & \checkmark & \checkmark\\
    WPP & & & \checkmark & \checkmark & & & \checkmark & & \checkmark \\
    FC & & \checkmark & & \checkmark & & \checkmark & \checkmark & \checkmark &\\
    DM & & \checkmark & & & \checkmark & \checkmark & \checkmark & \checkmark &\\    
    FERM & & \checkmark & & & \checkmark & \checkmark & & \checkmark &\\    
    reductions & & \checkmark & & \checkmark & \checkmark & \checkmark & \checkmark & \checkmark & \checkmark\\
    \textbf{FST} & \checkmark & & \checkmark & \checkmark & \checkmark & \checkmark & \checkmark & \checkmark & \checkmark\\
    \bottomrule
    \end{tabular}
    \caption{Capabilities of methods in comparison. $\star$ refers to an extension implemented by \citet{aif360-oct-2018}.}
    \label{tab:capabilities}
\end{table}

\subsubsection{Metrics} Classification performance and fairness were evaluated using both score-based metrics (log loss, Brier score, and AUC for performance, differences in mean scores (MSP) and GEO for fairness) and binary label-based metrics (accuracy, differences in mean binary predictions (SP) and non-generalized EO). While FST optimizes log loss (recall from Section~\ref{sec:prob:utility}), we find that results for Brier score are highly similar and thus defer the log loss results to Appendix~\ref{sec:exptAdd:logloss}.
We account for the fact that the reductions method \citep{agarwal2018} returns a \emph{randomized} classifier, i.e.,~a probability distribution over a set of classifiers. For the binary label-based metrics, we used the methods provided with the code\footnote{\url{https://github.com/microsoft/fairlearn}} for reductions to compute the metrics. The score-based metrics were computed by evaluating the metric for each classifier in the distribution and then averaging, weighted by their probabilities.

\begin{figure}[t]
  \centering
  \begin{subfigure}[b]{0.32\columnwidth}
  \includegraphics[width=\columnwidth]{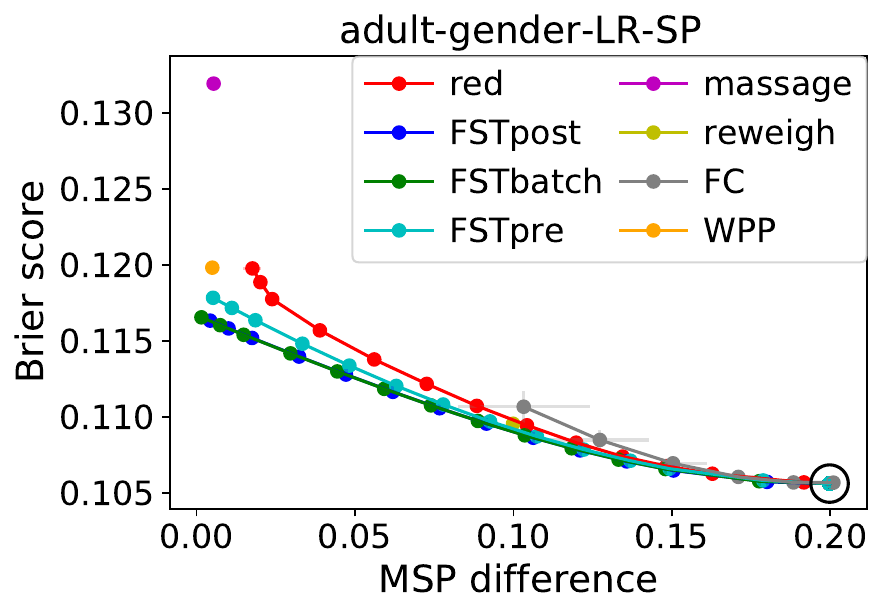}
  \label{fig:adult_1_LR_SP_acc_Brier}
  \end{subfigure}
  \begin{subfigure}[b]{0.32\columnwidth}
  \includegraphics[width=\columnwidth]{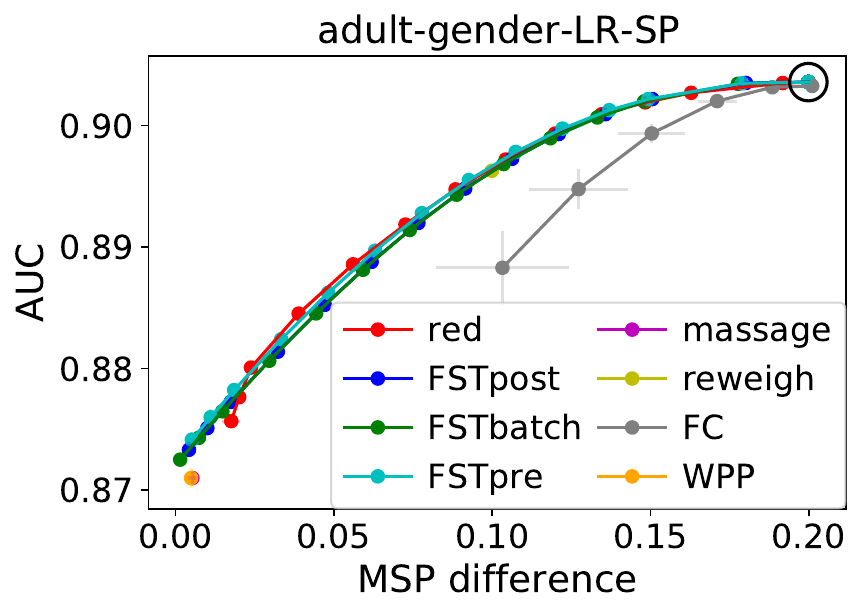}
  \label{fig:adult_1_LR_SP_acc_AUC}
  \end{subfigure}
  \begin{subfigure}[b]{0.32\columnwidth}
  \includegraphics[width=\columnwidth]{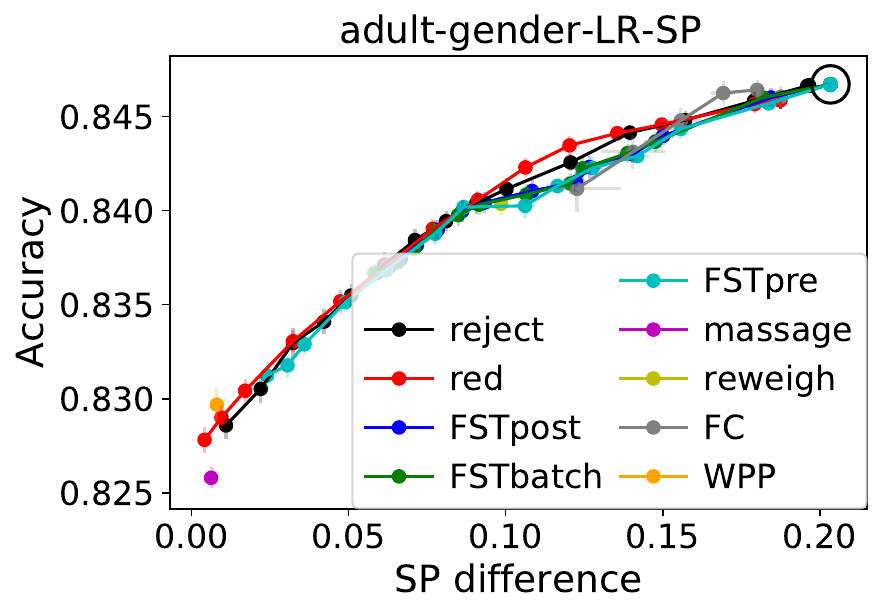}
  \label{fig:adult_1_LR_SP_acc_acc}
  \end{subfigure}
  \begin{subfigure}[b]{0.32\columnwidth}
  \includegraphics[width=\columnwidth]{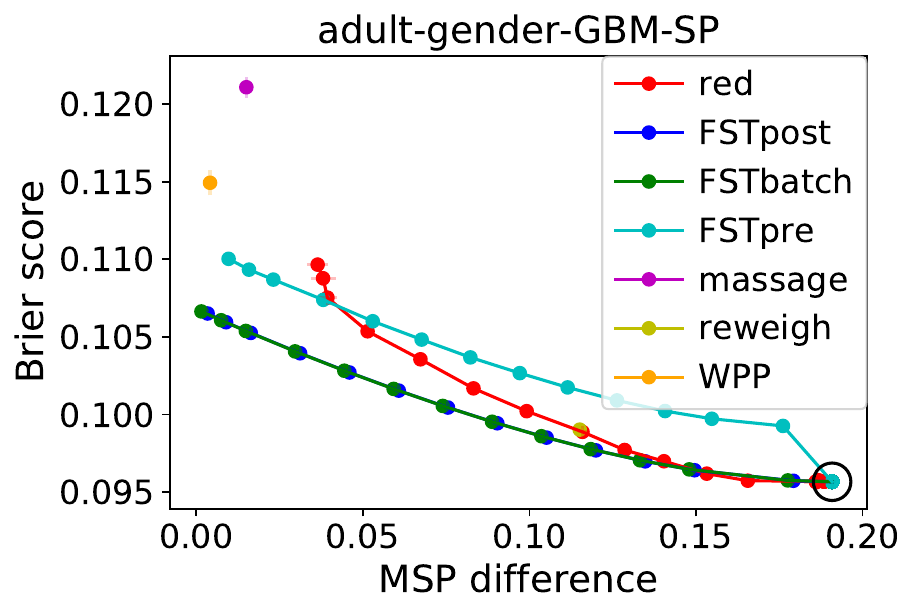}
  \label{fig:adult_1_GBM_SP_acc_Brier}
  \end{subfigure}
  \begin{subfigure}[b]{0.32\columnwidth}
  \includegraphics[width=\columnwidth]{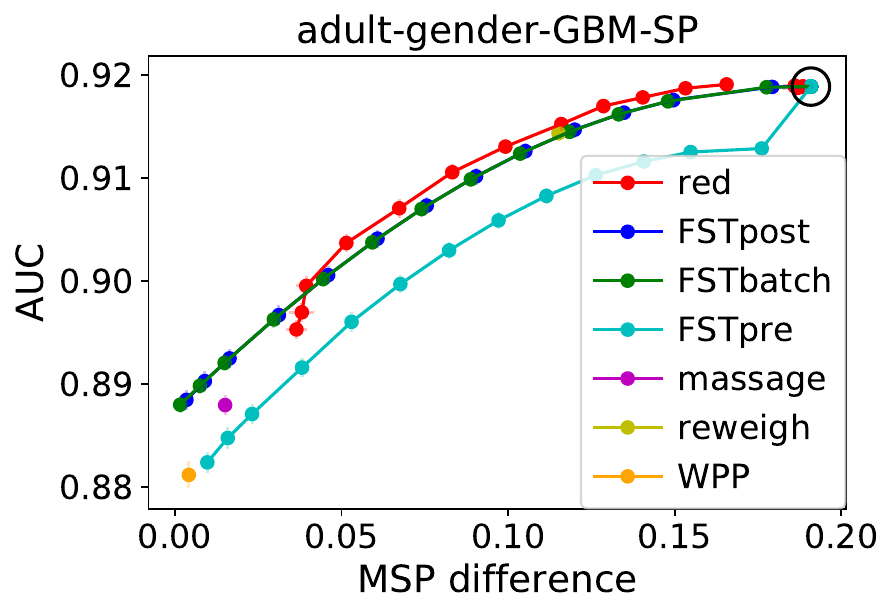}
  \label{fig:adult_1_GBM_SP_acc_AUC}
  \end{subfigure}
  \begin{subfigure}[b]{0.32\columnwidth}
  \includegraphics[width=\columnwidth]{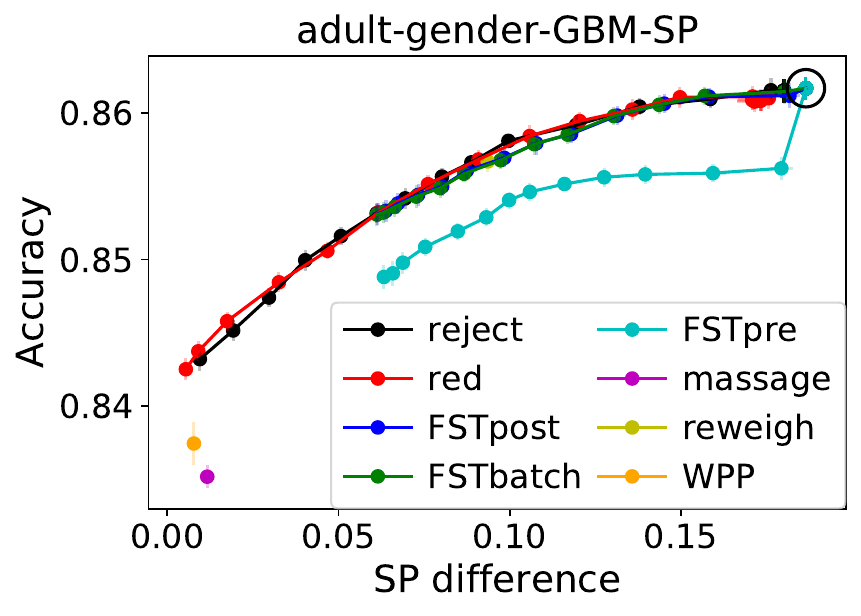}
  \label{fig:adult_1_GBM_SP_acc_acc}
  \end{subfigure}
  \begin{subfigure}[b]{0.32\columnwidth}
  \includegraphics[width=\columnwidth]{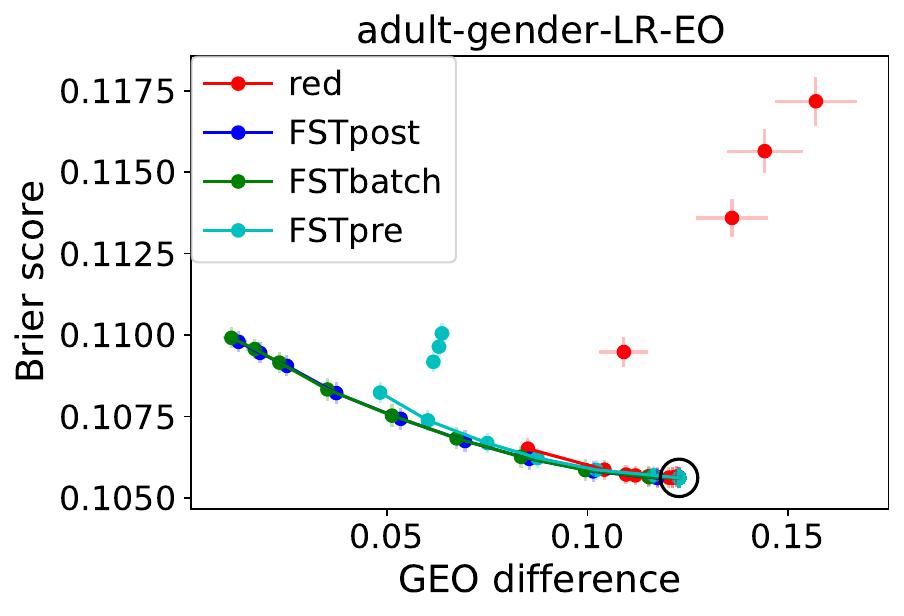}
  \label{fig:adult_1_LR_EO_acc_Brier}
  \end{subfigure}
  \begin{subfigure}[b]{0.32\columnwidth}
  \includegraphics[width=\columnwidth]{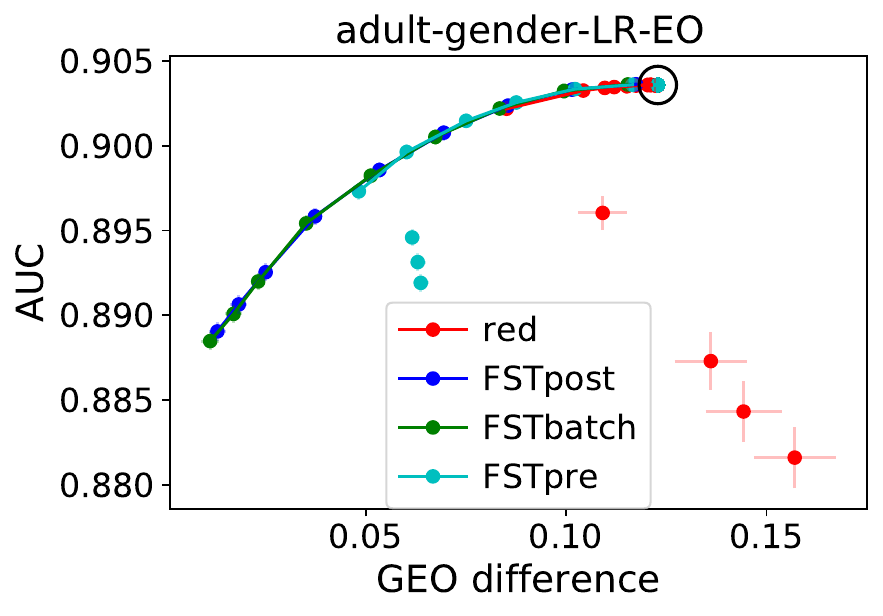}
  \label{fig:adult_1_LR_EO_acc_AUC}
  \end{subfigure}
  \begin{subfigure}[b]{0.32\columnwidth}
  \includegraphics[width=\columnwidth]{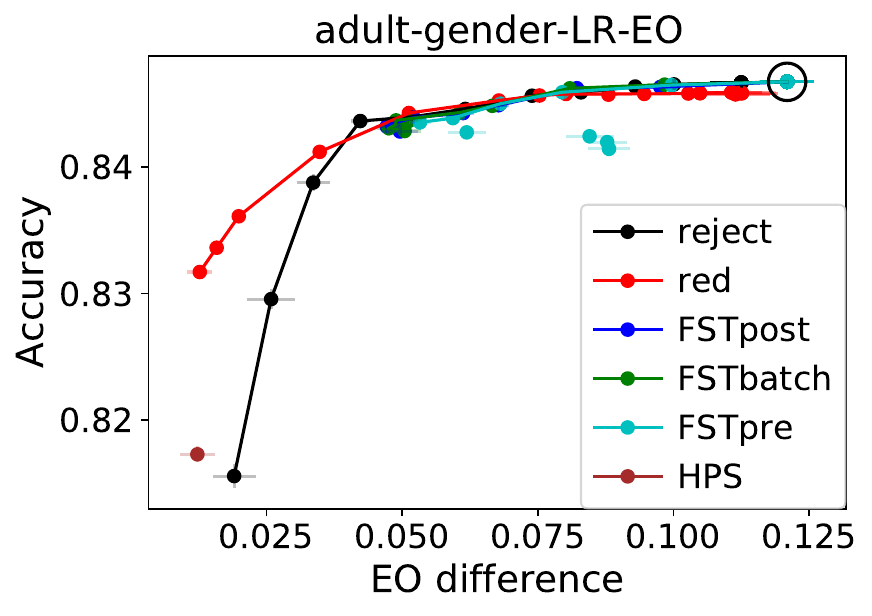}
  \label{fig:adult_1_LR_EO_acc_acc}
  \end{subfigure}
  \begin{subfigure}[b]{0.32\columnwidth}
  \includegraphics[width=\columnwidth]{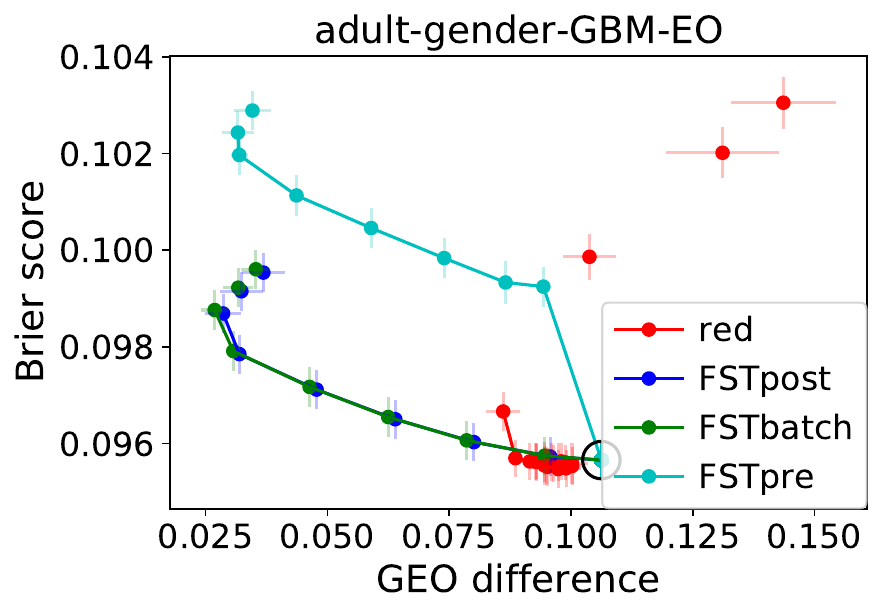}
  \label{fig:adult_1_GBM_EO_acc_Brier}
  \end{subfigure}
  \begin{subfigure}[b]{0.32\columnwidth}
  \includegraphics[width=\columnwidth]{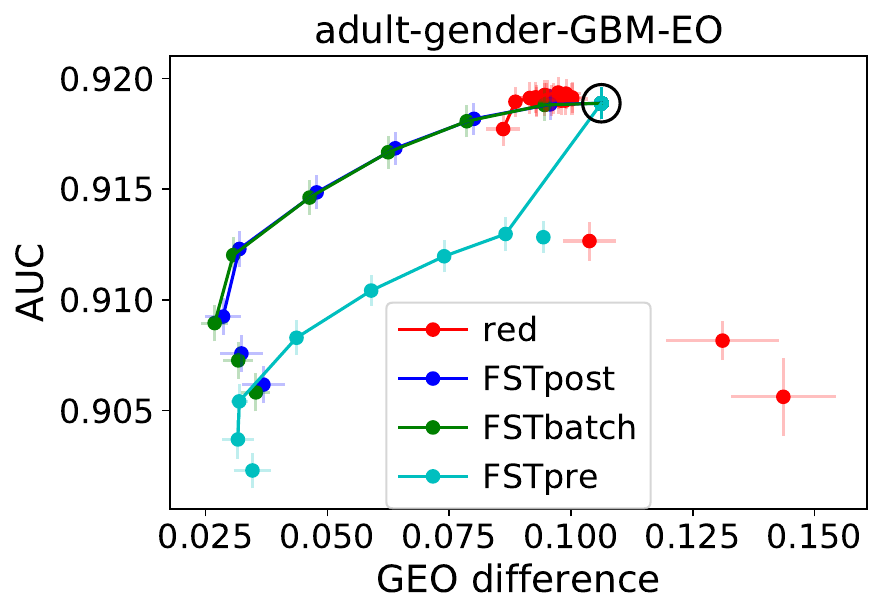}
  \label{fig:adult_1_GBM_EO_acc_AUC}
  \end{subfigure}
  \begin{subfigure}[b]{0.32\columnwidth}
  \includegraphics[width=\columnwidth]{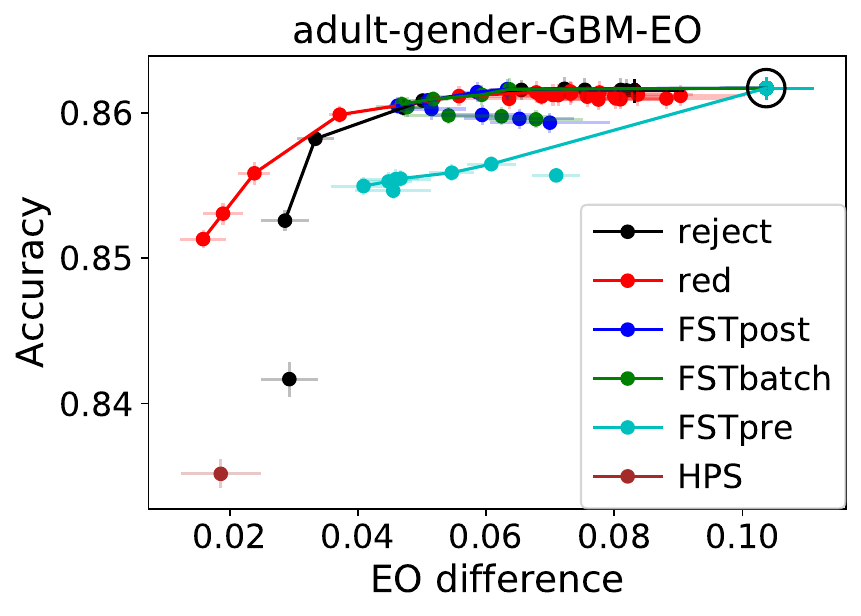}
  \label{fig:adult_1_GBM_EO_acc_acc}
  \end{subfigure}
  \caption{Trade-offs between fairness and classification performance on the Adult Income data set with gender as the protected attribute and the protected attribute included in the features. Pareto efficient points are connected by line segments to ease visualization. Horizontal and vertical bars represent standard errors in the means over $10$ train-test splits. The point achieved by a classifier without fairness constraints is marked by a black circle.}
  \label{fig:adult_1}
\end{figure}

\begin{figure}[t]
  \centering
  \begin{subfigure}[b]{0.32\columnwidth}
  \includegraphics[width=\columnwidth]{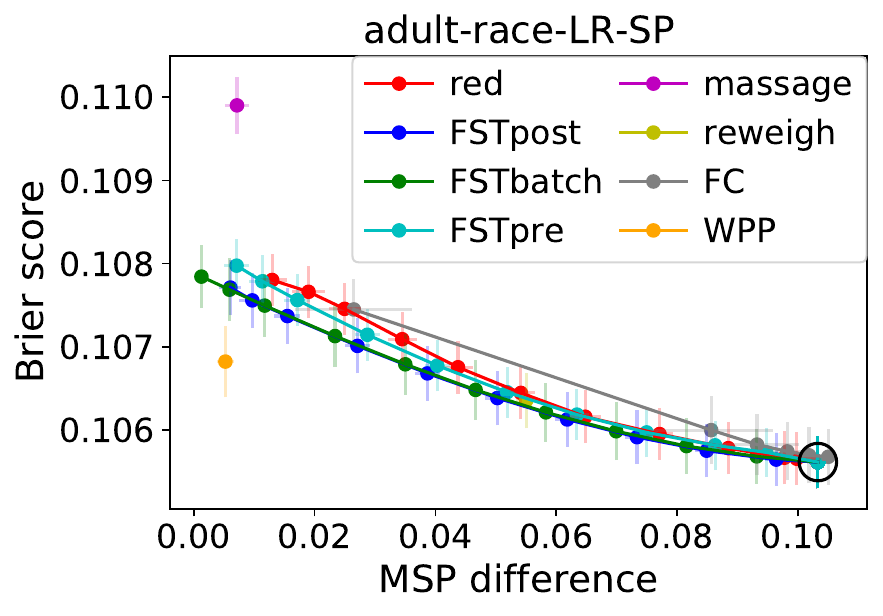}
  \label{fig:adult_2_LR_SP_acc_Brier}
  \end{subfigure}
  \begin{subfigure}[b]{0.32\columnwidth}
  \includegraphics[width=\columnwidth]{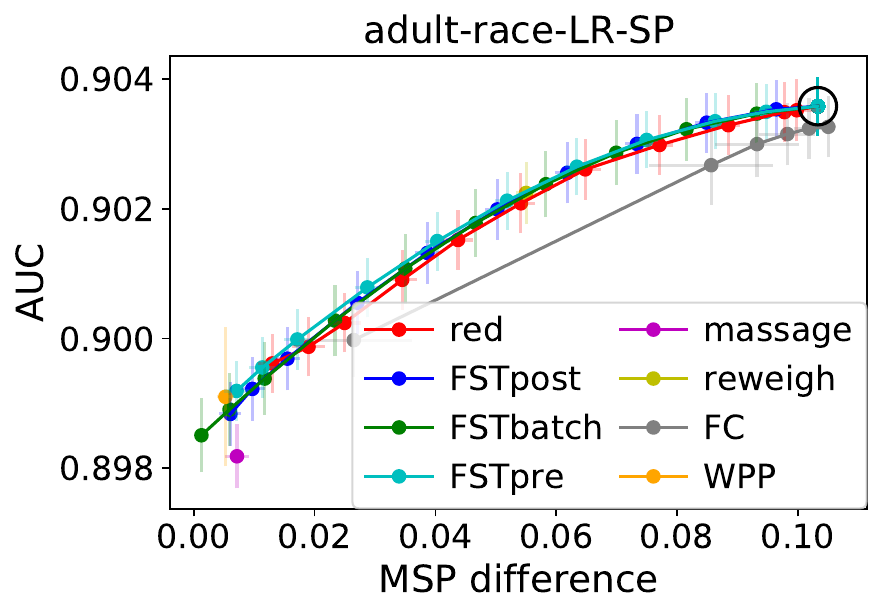}
  \label{fig:adult_2_LR_SP_acc_AUC}
  \end{subfigure}
  \begin{subfigure}[b]{0.32\columnwidth}
  \includegraphics[width=\columnwidth]{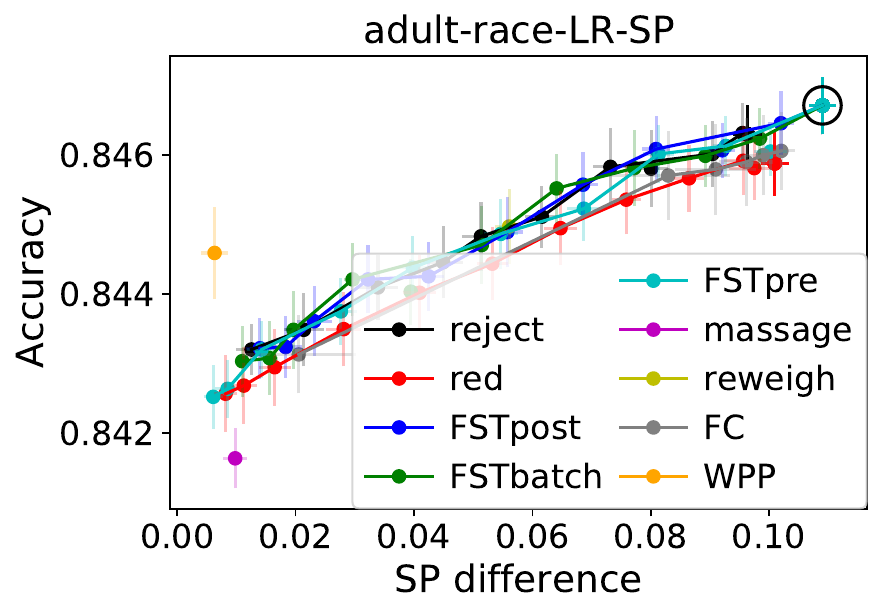}
  \label{fig:adult_2_LR_SP_acc_acc}
  \end{subfigure}
  \begin{subfigure}[b]{0.32\columnwidth}
  \includegraphics[width=\columnwidth]{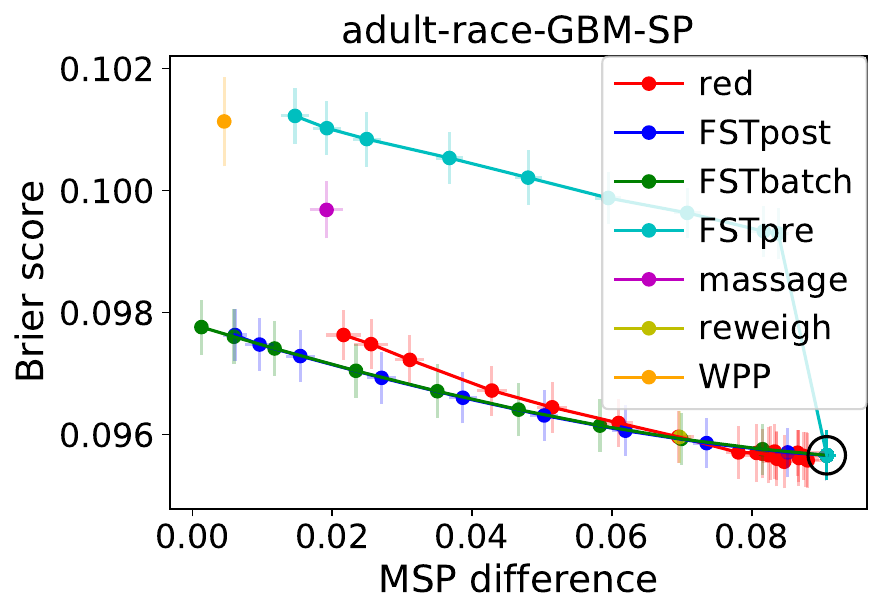}
  \label{fig:adult_2_GBM_SP_acc_Brier}
  \end{subfigure}
  \begin{subfigure}[b]{0.32\columnwidth}
  \includegraphics[width=\columnwidth]{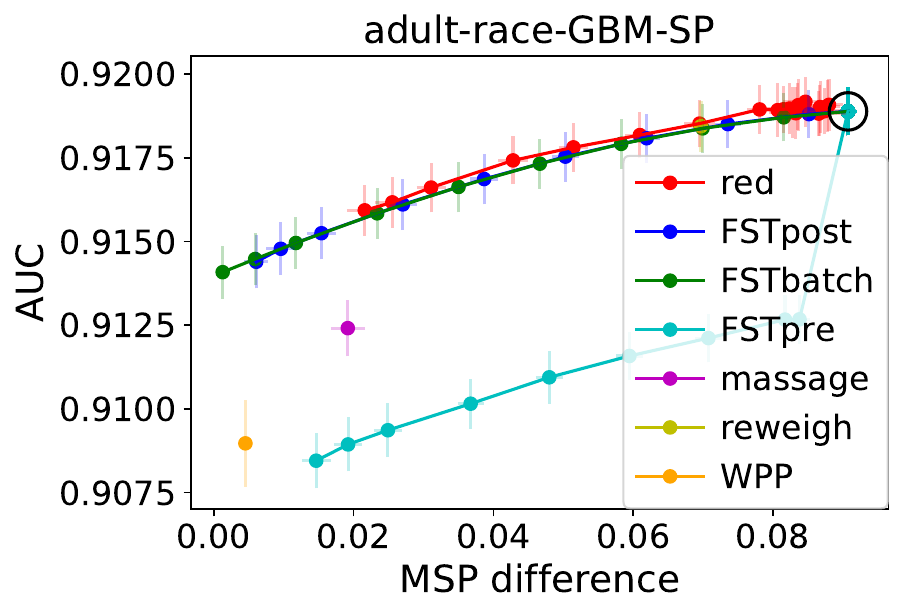}
  \label{fig:adult_2_GBM_SP_acc_AUC}
  \end{subfigure}
  \begin{subfigure}[b]{0.32\columnwidth}
  \includegraphics[width=\columnwidth]{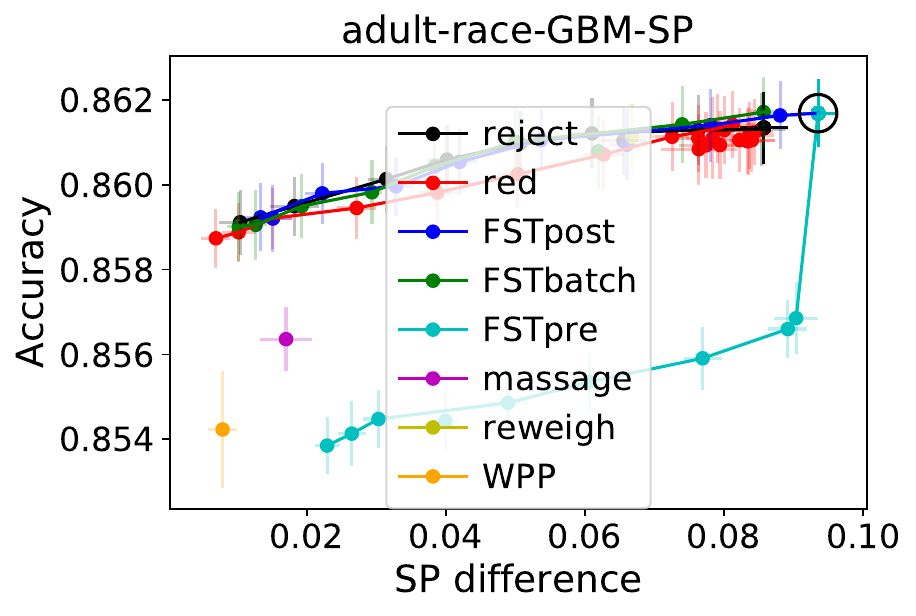}
  \label{fig:adult_2_GBM_SP_acc_acc}
  \end{subfigure}
  \begin{subfigure}[b]{0.32\columnwidth}
  \includegraphics[width=\columnwidth]{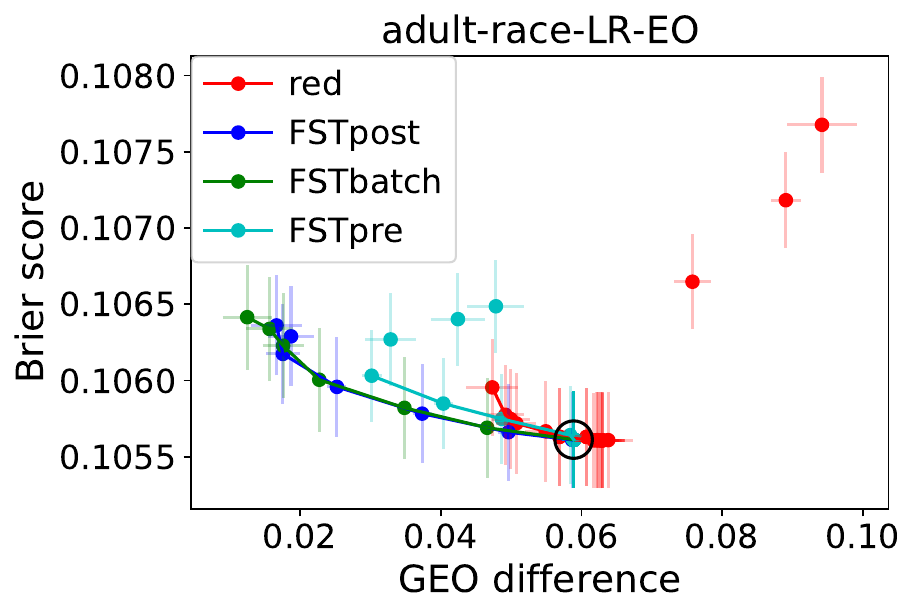}
  \label{fig:adult_2_LR_EO_acc_Brier}
  \end{subfigure}
  \begin{subfigure}[b]{0.32\columnwidth}
  \includegraphics[width=\columnwidth]{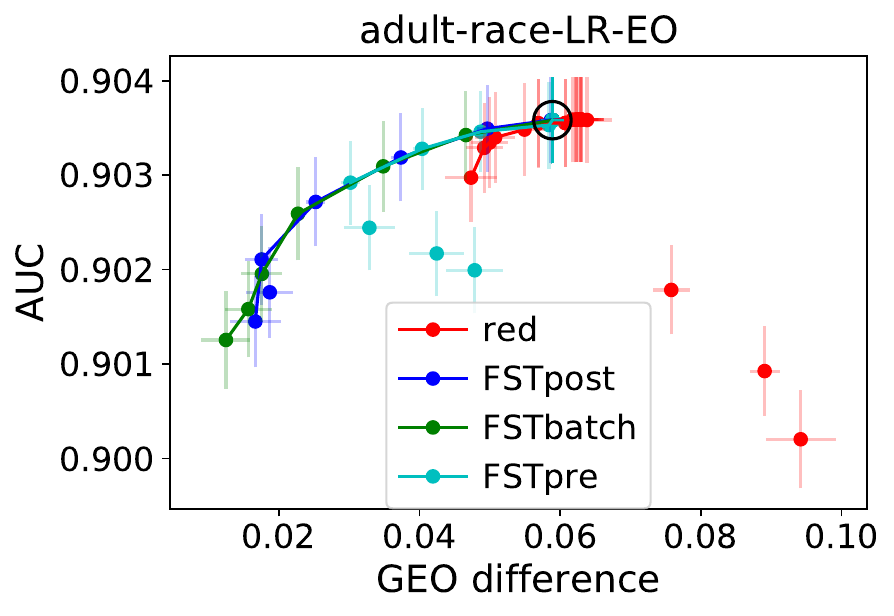}
  \label{fig:adult_2_LR_EO_acc_AUC}
  \end{subfigure}
  \begin{subfigure}[b]{0.32\columnwidth}
  \includegraphics[width=\columnwidth]{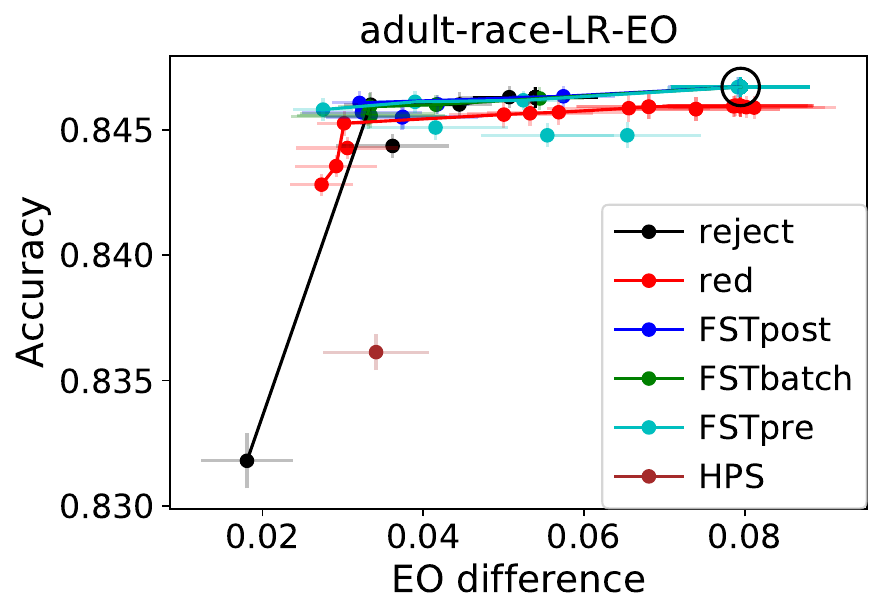}
  \label{fig:adult_2_LR_EO_acc_acc}
  \end{subfigure}
  \begin{subfigure}[b]{0.32\columnwidth}
  \includegraphics[width=\columnwidth]{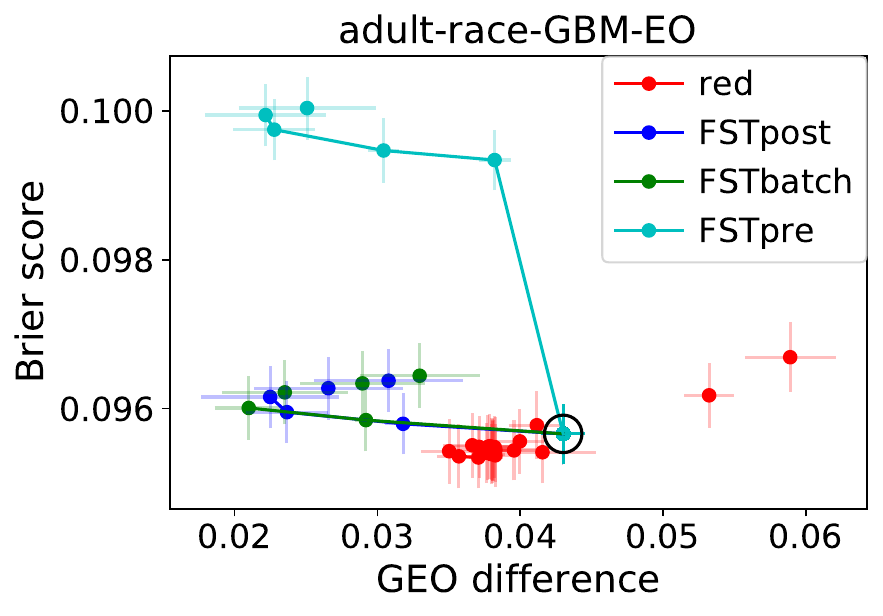}
  \label{fig:adult_2_GBM_EO_acc_Brier}
  \end{subfigure}
  \begin{subfigure}[b]{0.32\columnwidth}
  \includegraphics[width=\columnwidth]{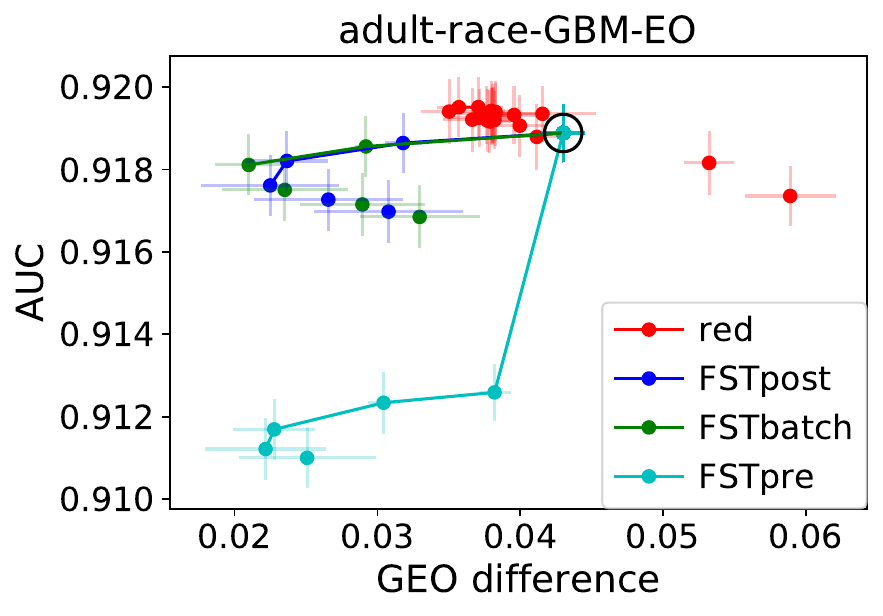}
  \label{fig:adult_2_GBM_EO_acc_AUC}
  \end{subfigure}
  \begin{subfigure}[b]{0.32\columnwidth}
  \includegraphics[width=\columnwidth]{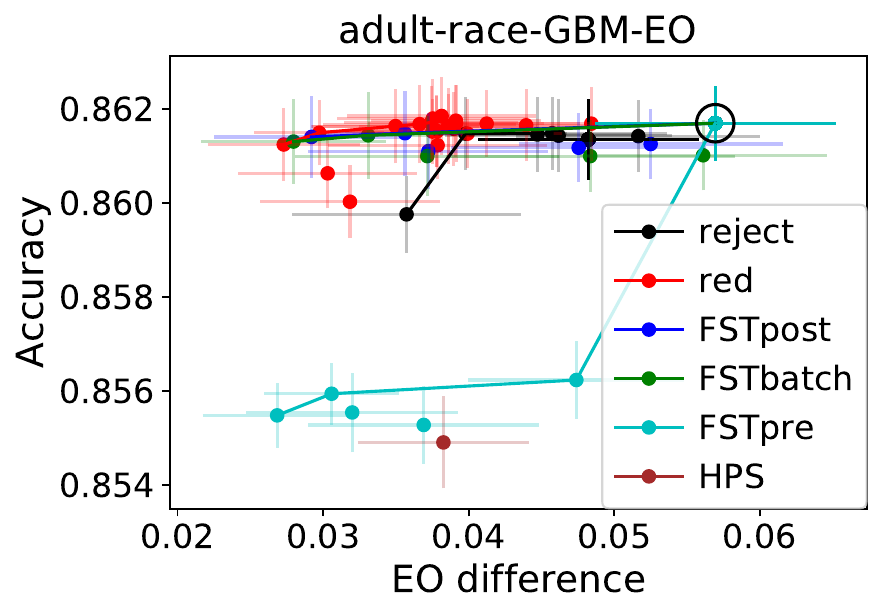}
  \label{fig:adult_2_GBM_EO_acc_acc}
  \end{subfigure}
  \caption{Trade-offs between fairness and classification performance on the Adult Income data set with race as the protected attribute and the protected attribute included in the features.}
  \label{fig:adult_2}
\end{figure}

\begin{figure}[t]
  \centering
  \begin{subfigure}[b]{0.32\columnwidth}
  \includegraphics[width=\columnwidth]{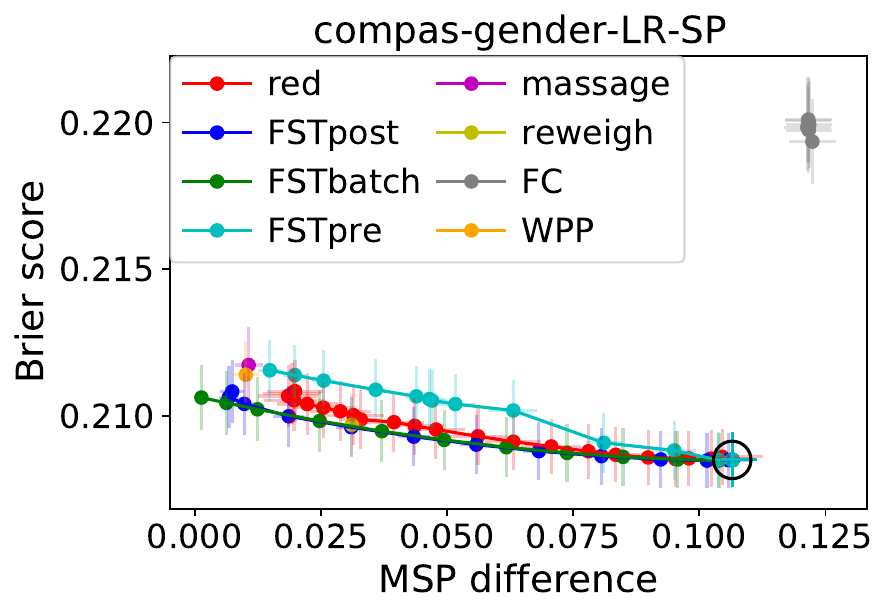}
  \label{fig:compas_1_LR_SP_acc_Brier}
  \end{subfigure}
  \begin{subfigure}[b]{0.32\columnwidth}
  \includegraphics[width=\columnwidth]{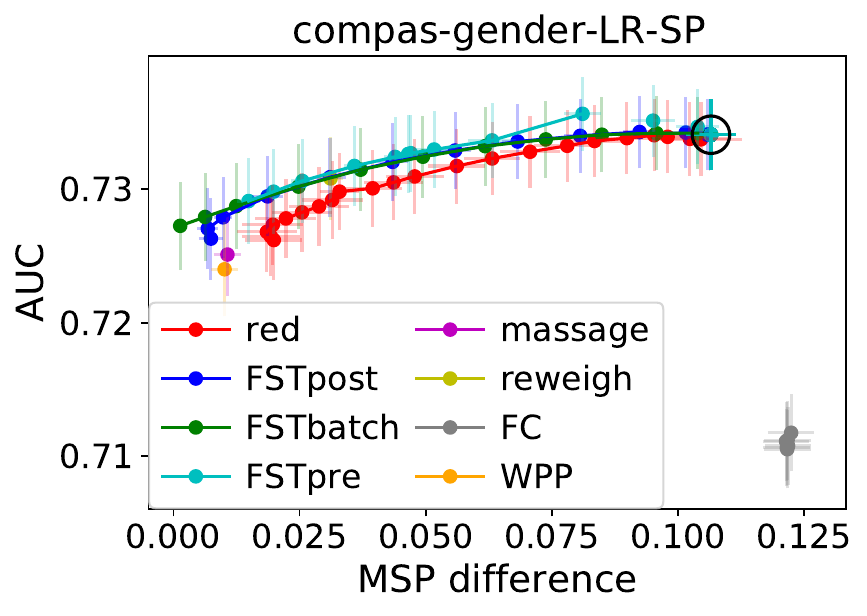}
  \label{fig:compas_1_LR_SP_acc_AUC}
  \end{subfigure}
  \begin{subfigure}[b]{0.32\columnwidth}
  \includegraphics[width=\columnwidth]{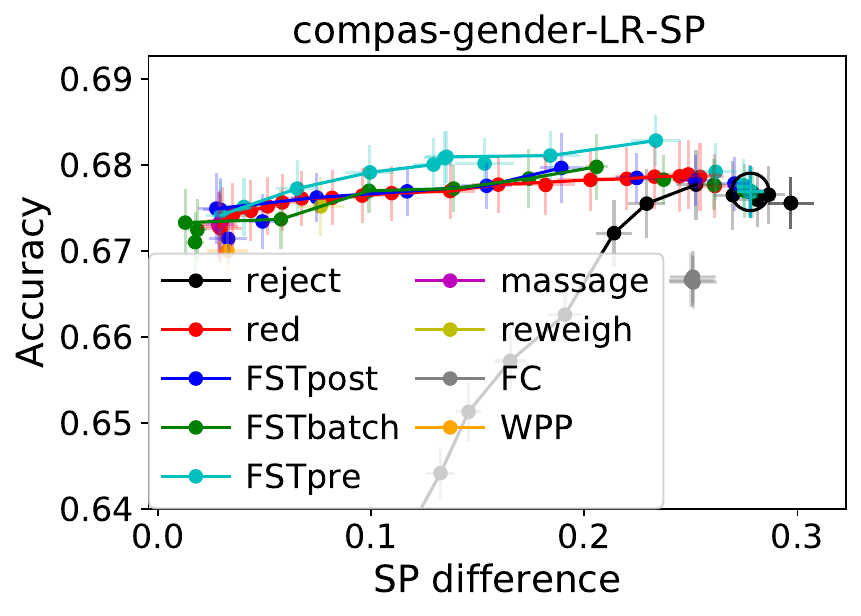}
  \label{fig:compas_1_LR_SP_acc_acc}
  \end{subfigure}
  \begin{subfigure}[b]{0.32\columnwidth}
  \includegraphics[width=\columnwidth]{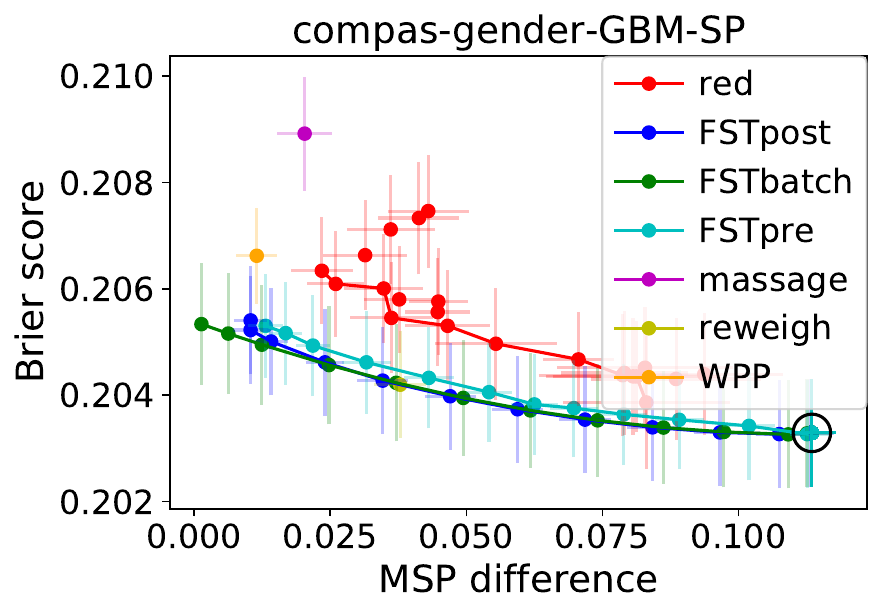}
  \label{fig:compas_1_GBM_SP_acc_Brier}
  \end{subfigure}
  \begin{subfigure}[b]{0.32\columnwidth}
  \includegraphics[width=\columnwidth]{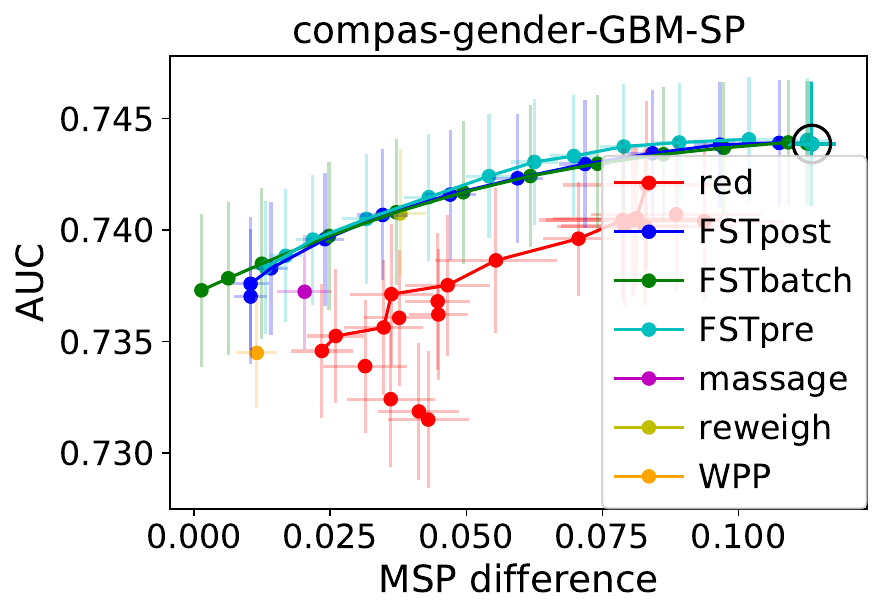}
  \label{fig:compas_1_GBM_SP_acc_AUC}
  \end{subfigure}
  \begin{subfigure}[b]{0.32\columnwidth}
  \includegraphics[width=\columnwidth]{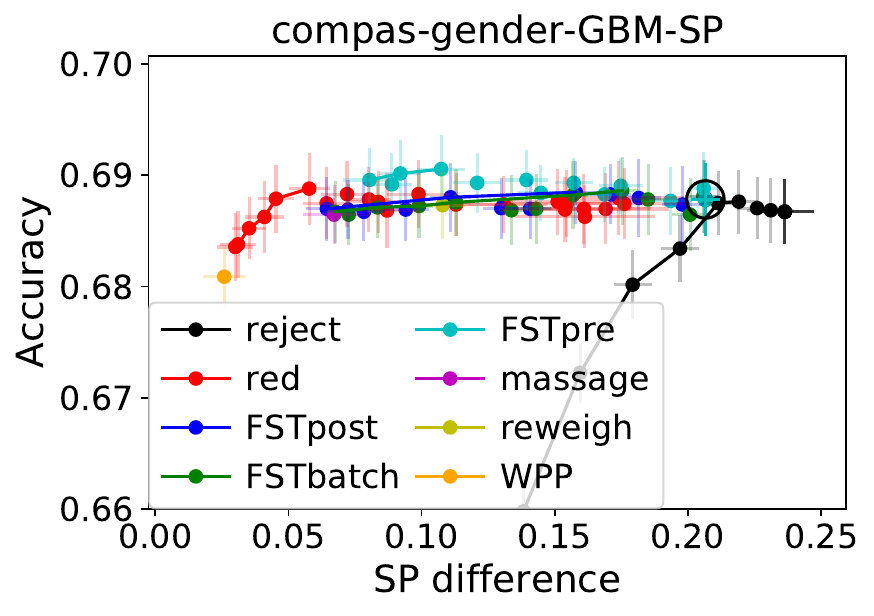}
  \label{fig:compas_1_GBM_SP_acc_acc}
  \end{subfigure}
  \begin{subfigure}[b]{0.32\columnwidth}
  \includegraphics[width=\columnwidth]{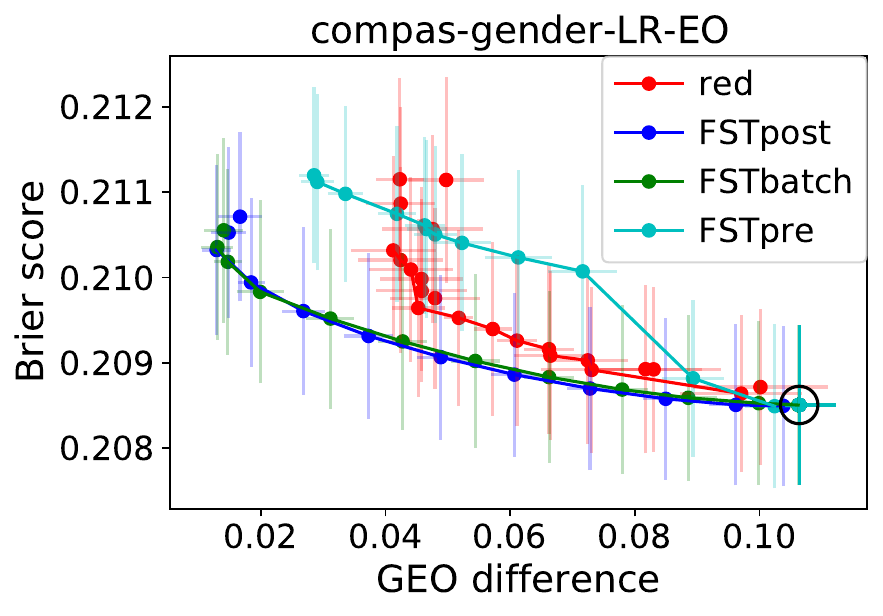}
  \label{fig:compas_1_LR_EO_acc_Brier}
  \end{subfigure}
  \begin{subfigure}[b]{0.32\columnwidth}
  \includegraphics[width=\columnwidth]{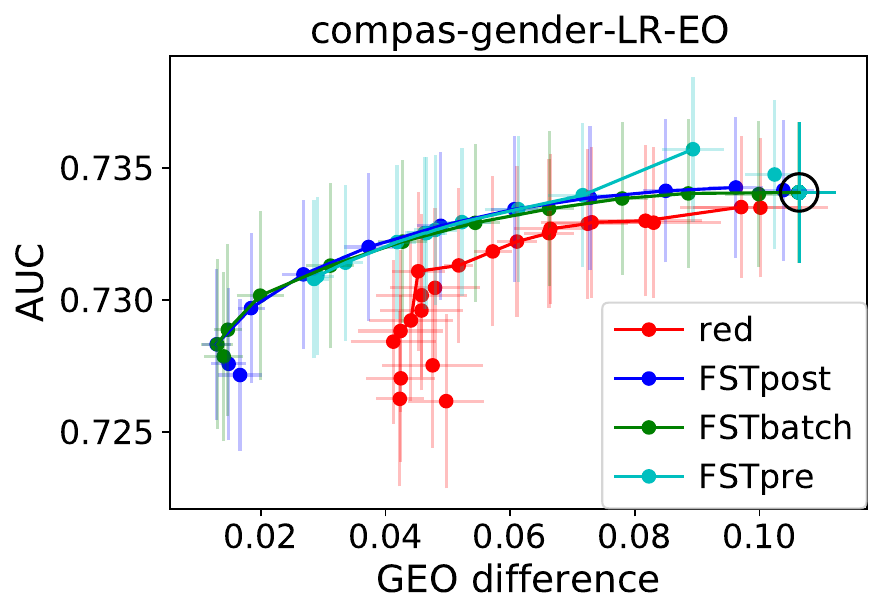}
  \label{fig:compas_1_LR_EO_acc_AUC}
  \end{subfigure}
  \begin{subfigure}[b]{0.32\columnwidth}
  \includegraphics[width=\columnwidth]{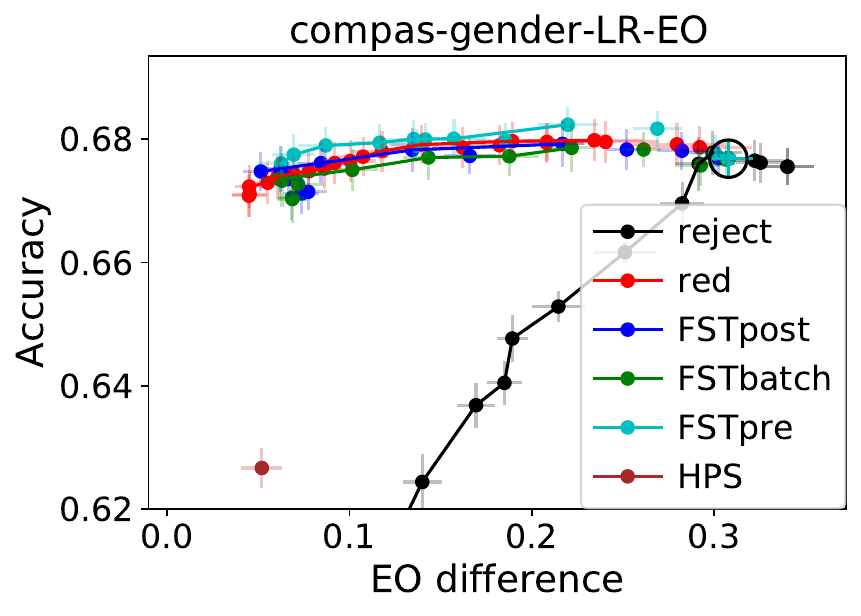}
  \label{fig:compas_1_LR_EO_acc_acc}
  \end{subfigure}
  \begin{subfigure}[b]{0.32\columnwidth}
  \includegraphics[width=\columnwidth]{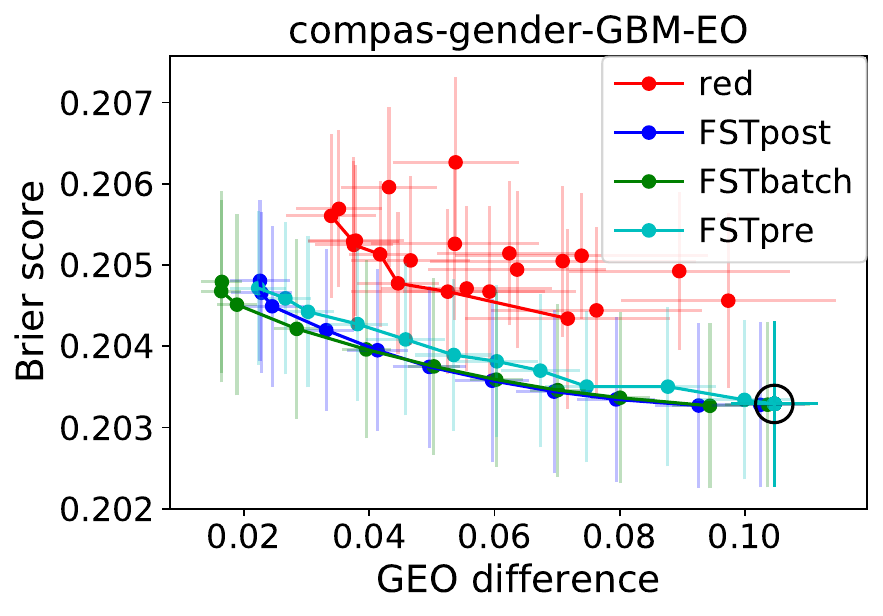}
  \label{fig:compas_1_GBM_EO_acc_Brier}
  \end{subfigure}
  \begin{subfigure}[b]{0.32\columnwidth}
  \includegraphics[width=\columnwidth]{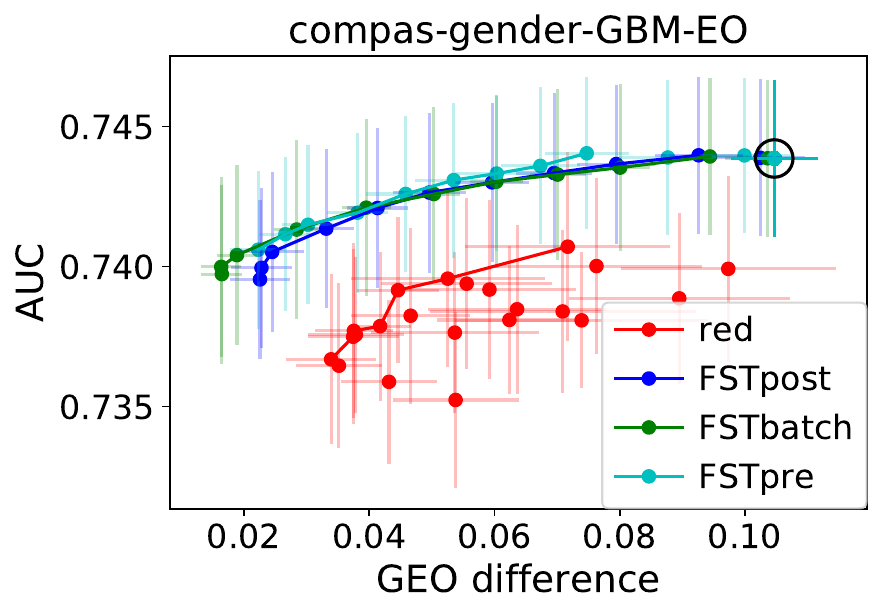}
  \label{fig:compas_1_GBM_EO_acc_AUC}
  \end{subfigure}
  \begin{subfigure}[b]{0.32\columnwidth}
  \includegraphics[width=\columnwidth]{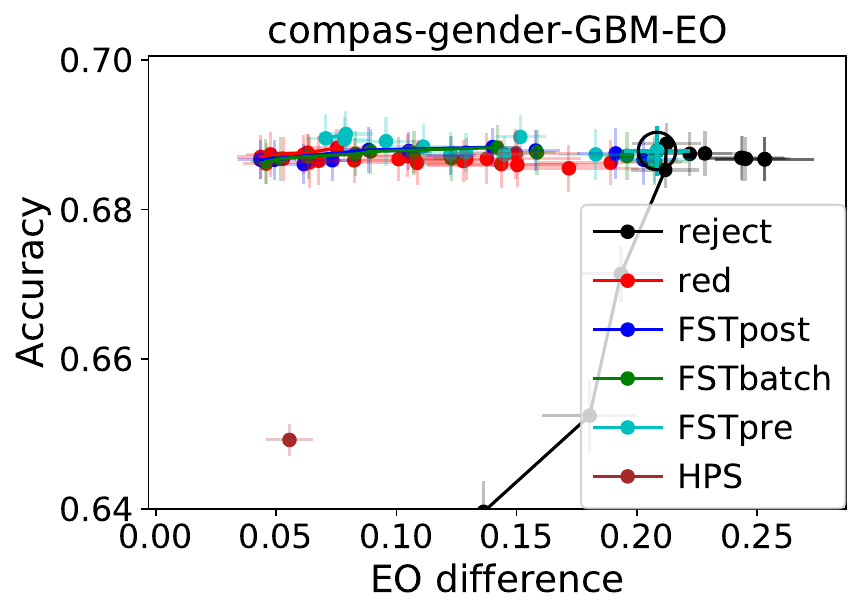}
  \label{fig:compas_1_GBM_EO_acc_acc}
  \end{subfigure}
  \caption{Trade-offs between fairness and classification performance on the COMPAS data set with gender as the protected attribute and the protected attribute included in the features.}
  \label{fig:compas_1}
\end{figure}

\begin{figure}[t]
  \centering
  \begin{subfigure}[b]{0.32\columnwidth}
  \includegraphics[width=\columnwidth]{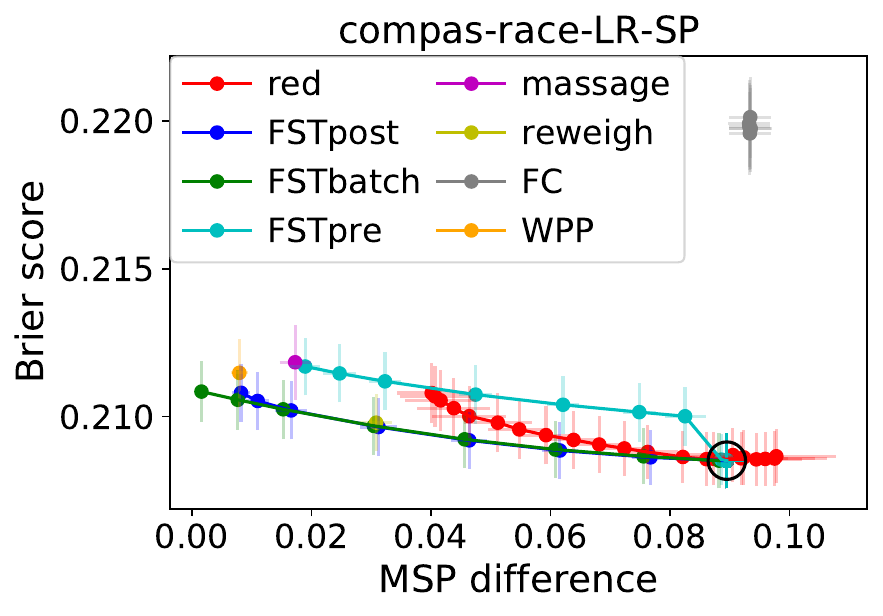}
  \label{fig:compas_2_LR_SP_acc_Brier}
  \end{subfigure}
  \begin{subfigure}[b]{0.32\columnwidth}
  \includegraphics[width=\columnwidth]{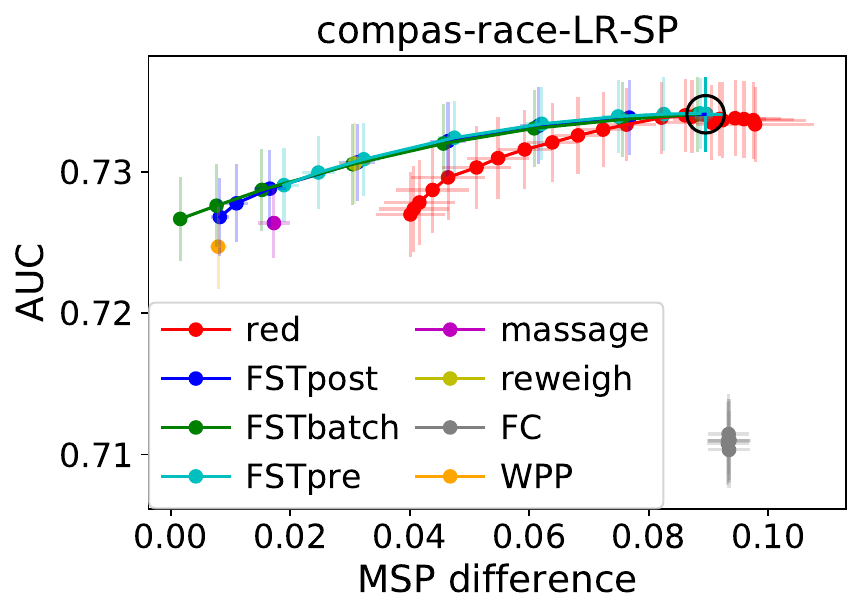}
  \label{fig:compas_2_LR_SP_acc_AUC}
  \end{subfigure}
  \begin{subfigure}[b]{0.32\columnwidth}
  \includegraphics[width=\columnwidth]{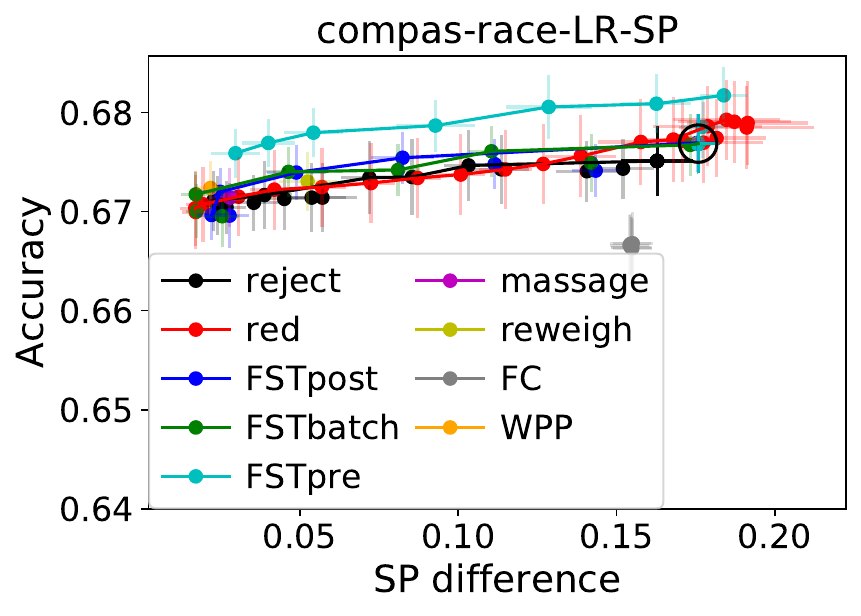}
  \label{fig:compas_2_LR_SP_acc_acc}
  \end{subfigure}
  \begin{subfigure}[b]{0.32\columnwidth}
  \includegraphics[width=\columnwidth]{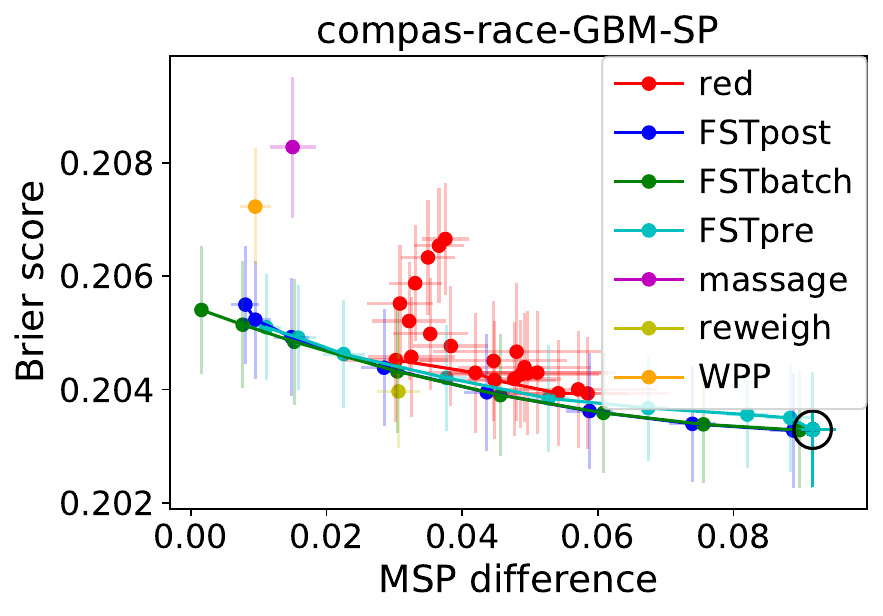}
  \label{fig:compas_2_GBM_SP_acc_Brier}
  \end{subfigure}
  \begin{subfigure}[b]{0.32\columnwidth}
  \includegraphics[width=\columnwidth]{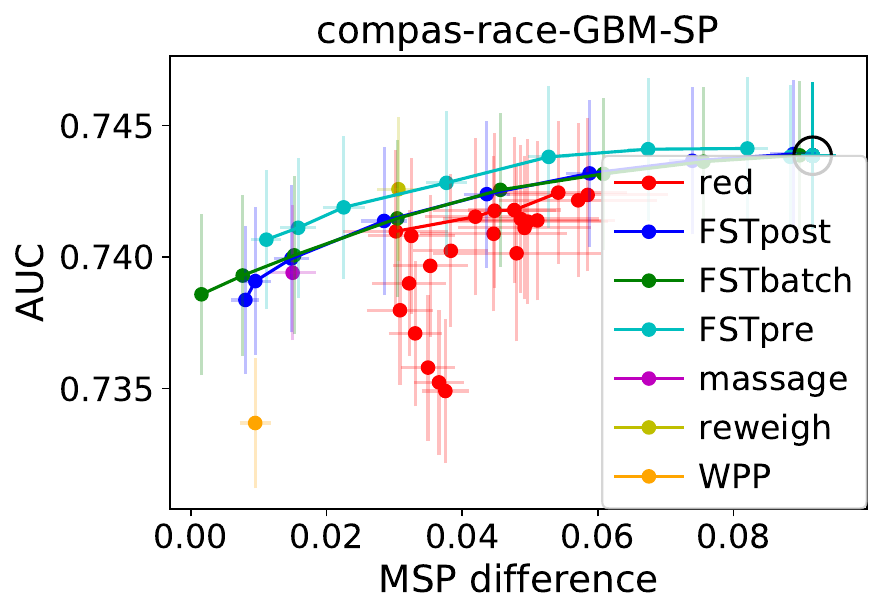}
  \label{fig:compas_2_GBM_SP_acc_AUC}
  \end{subfigure}
  \begin{subfigure}[b]{0.32\columnwidth}
  \includegraphics[width=\columnwidth]{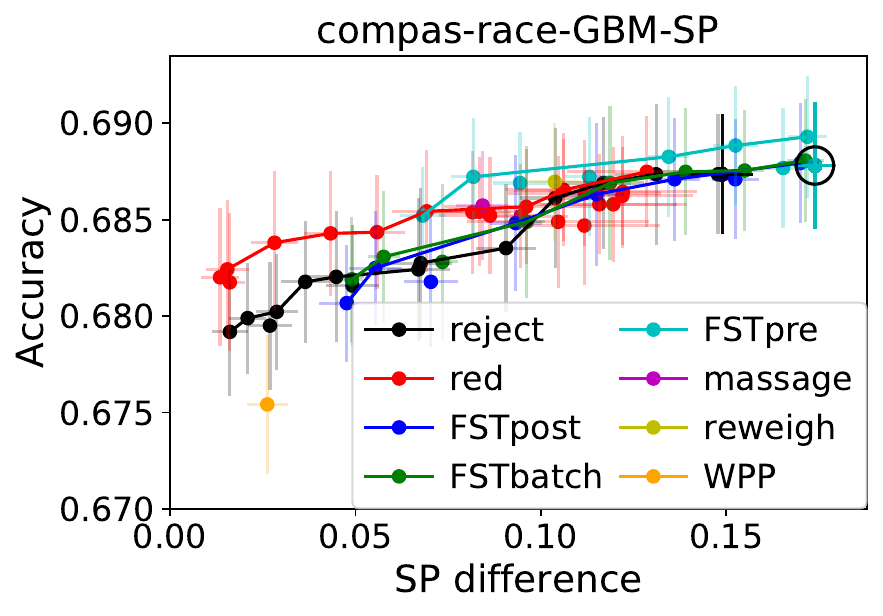}
  \label{fig:compas_2_GBM_SP_acc_acc}
  \end{subfigure}
  \begin{subfigure}[b]{0.32\columnwidth}
  \includegraphics[width=\columnwidth]{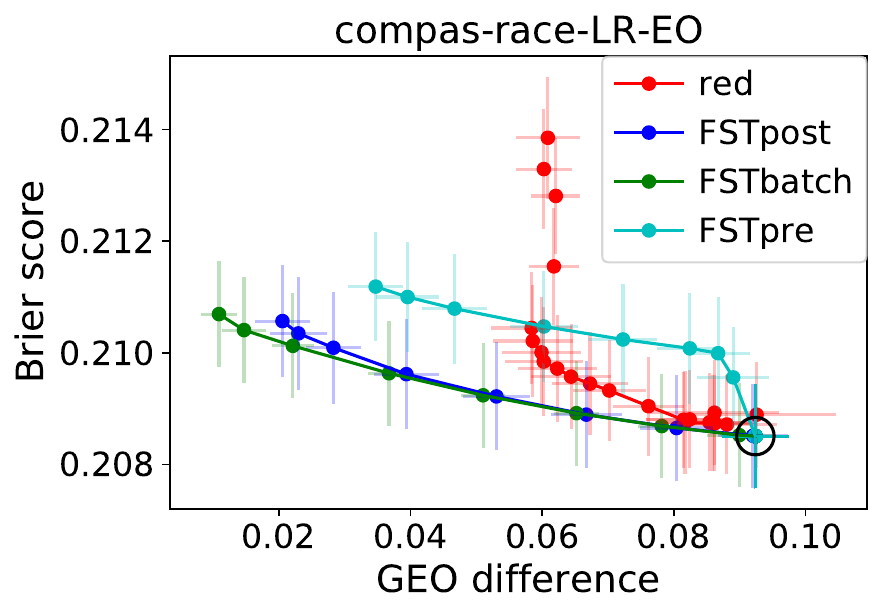}
  \label{fig:compas_2_LR_EO_acc_Brier}
  \end{subfigure}
  \begin{subfigure}[b]{0.32\columnwidth}
  \includegraphics[width=\columnwidth]{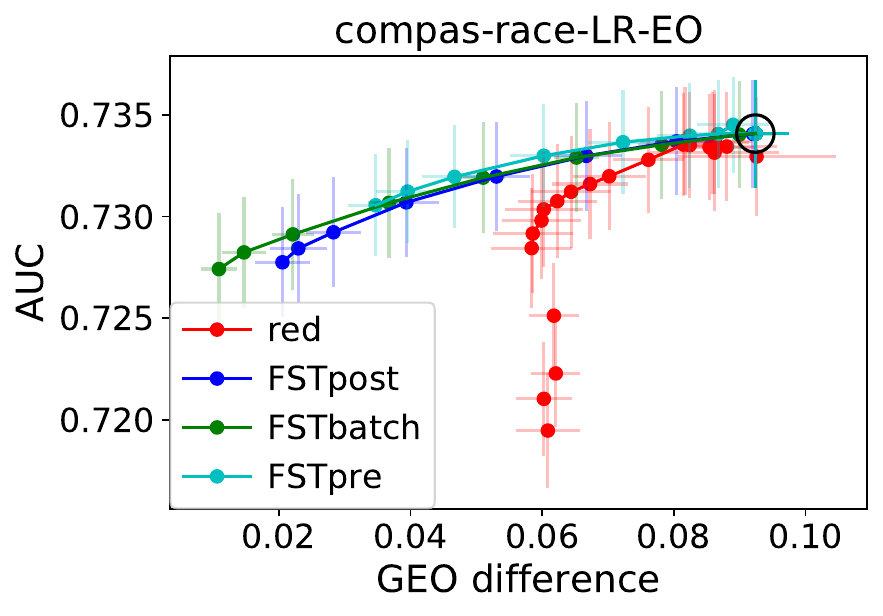}
  \label{fig:compas_2_LR_EO_acc_AUC}
  \end{subfigure}
  \begin{subfigure}[b]{0.32\columnwidth}
  \includegraphics[width=\columnwidth]{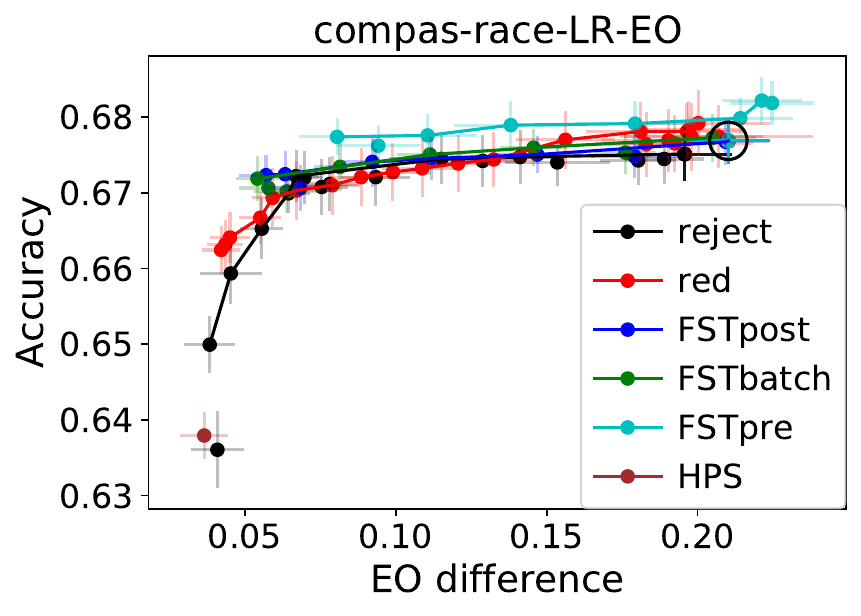}
  \label{fig:compas_2_LR_EO_acc_acc}
  \end{subfigure}
  \begin{subfigure}[b]{0.32\columnwidth}
  \includegraphics[width=\columnwidth]{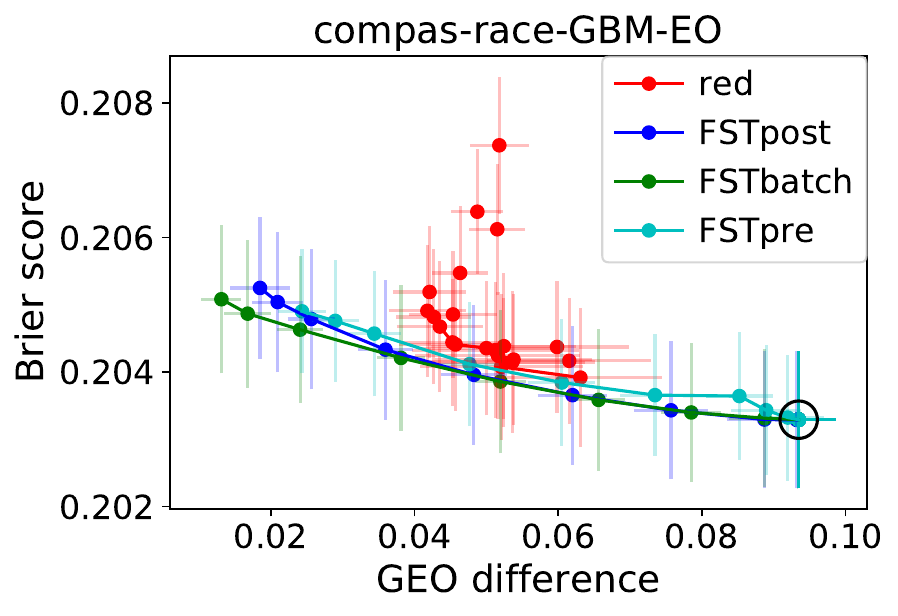}
  \label{fig:compas_2_GBM_EO_acc_Brier}
  \end{subfigure}
  \begin{subfigure}[b]{0.32\columnwidth}
  \includegraphics[width=\columnwidth]{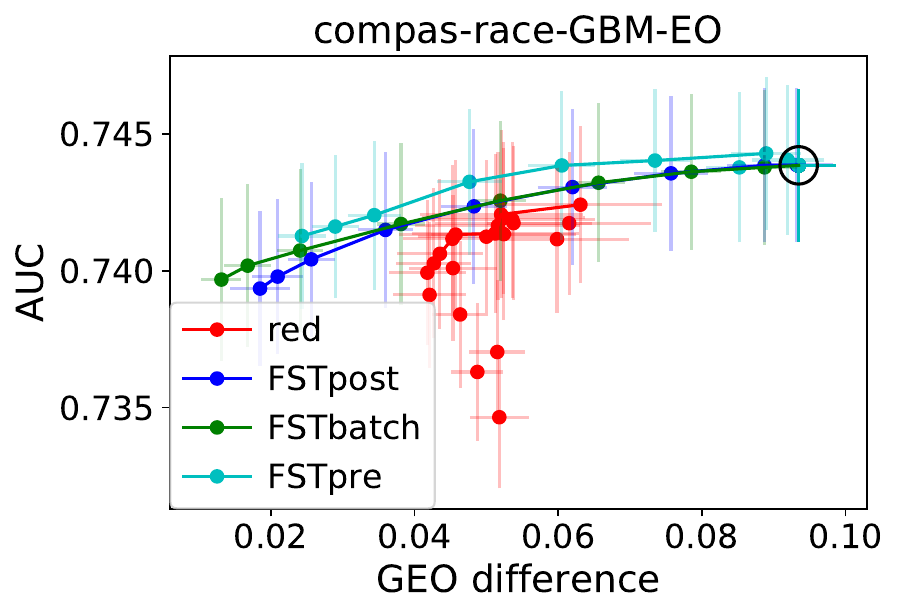}
  \label{fig:compas_2_GBM_EO_acc_AUC}
  \end{subfigure}
  \begin{subfigure}[b]{0.32\columnwidth}
  \includegraphics[width=\columnwidth]{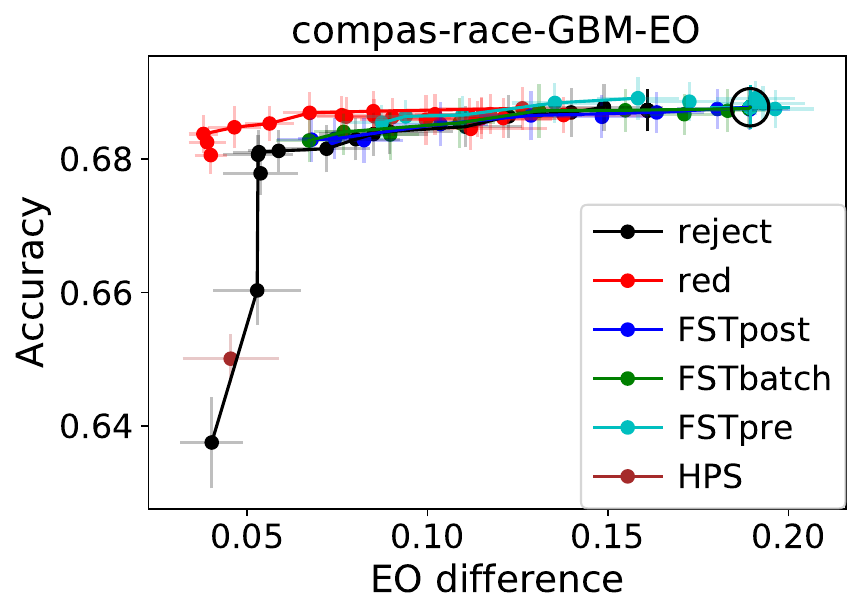}
  \label{fig:compas_2_GBM_EO_acc_acc}
  \end{subfigure}
  \caption{Trade-offs between fairness and classification performance on the COMPAS data set with race as the protected attribute and the protected attribute included in the features.}
  \label{fig:compas_2}
\end{figure}

\begin{figure}[t]
  \centering
  \begin{subfigure}[b]{0.32\columnwidth}
  \includegraphics[width=\columnwidth]{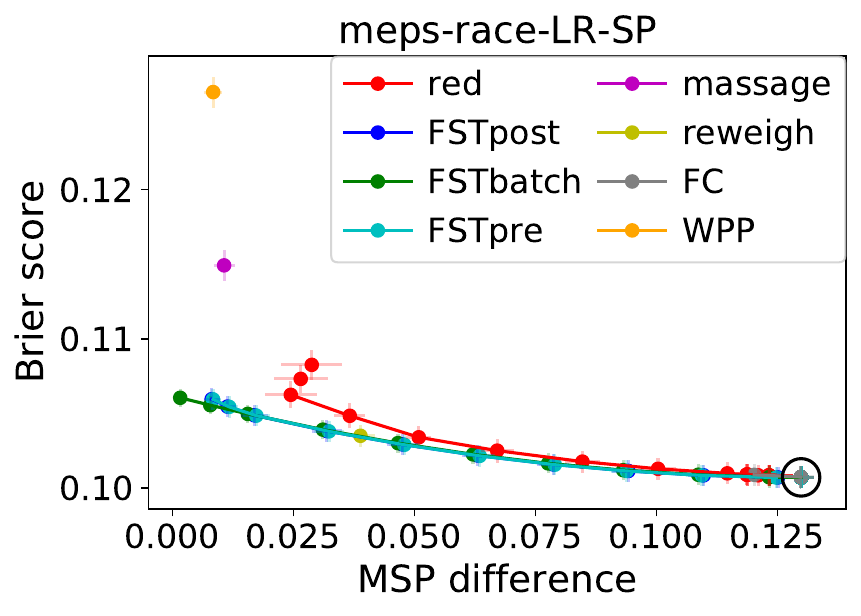}
  \label{fig:meps_1_LR_SP_acc_Brier}
  \end{subfigure}
  \begin{subfigure}[b]{0.32\columnwidth}
  \includegraphics[width=\columnwidth]{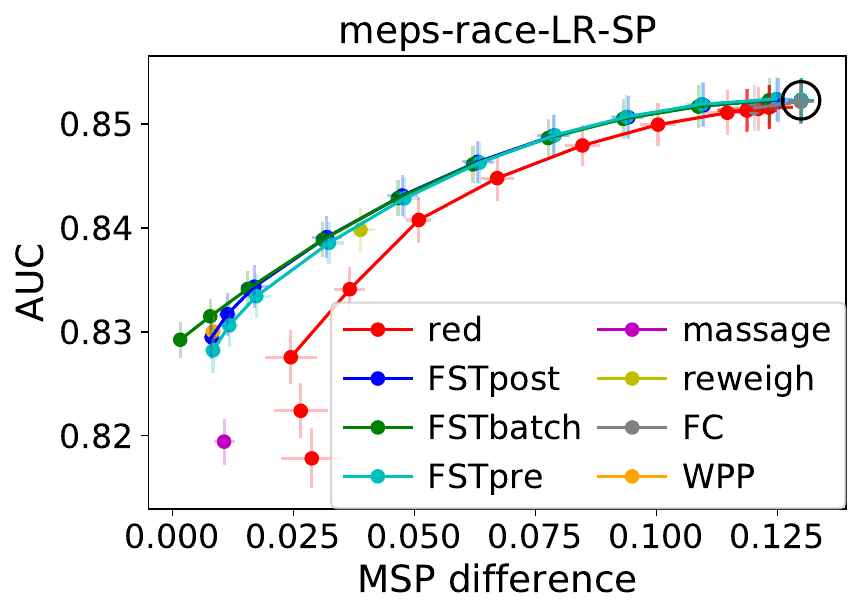}
  \label{fig:meps_1_LR_SP_acc_AUC}
  \end{subfigure}
  \begin{subfigure}[b]{0.32\columnwidth}
  \includegraphics[width=\columnwidth]{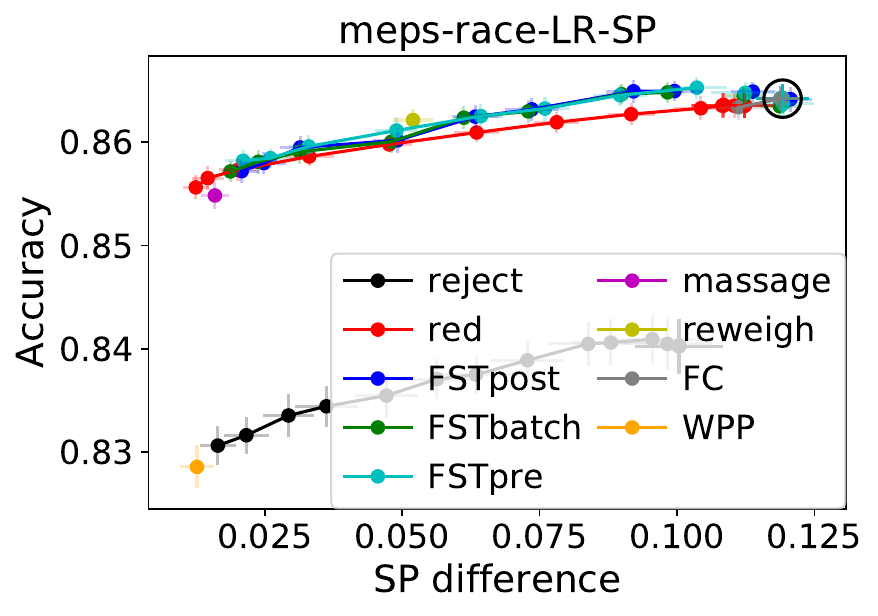}
  \label{fig:meps_1_LR_SP_acc_acc}
  \end{subfigure}
  \begin{subfigure}[b]{0.32\columnwidth}
  \includegraphics[width=\columnwidth]{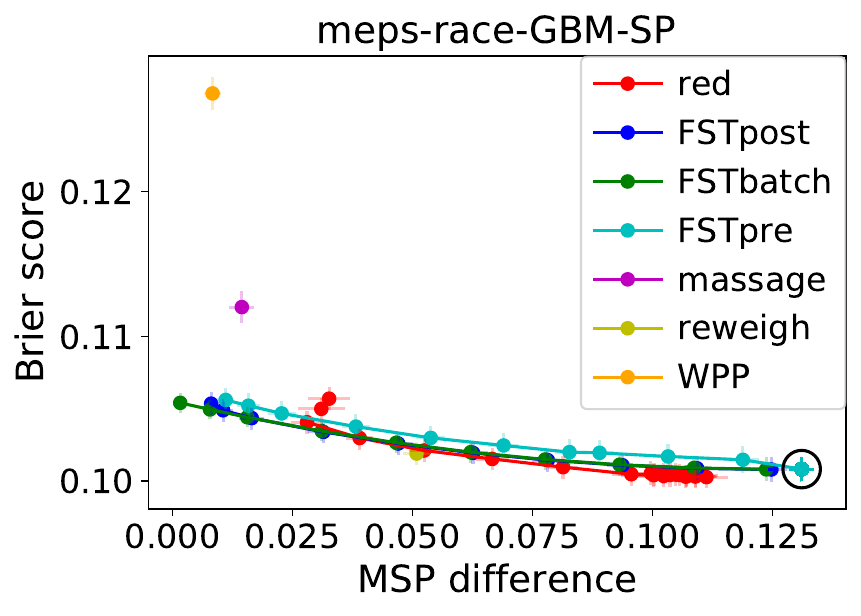}
  \label{fig:meps_1_GBM_SP_acc_Brier}
  \end{subfigure}
  \begin{subfigure}[b]{0.32\columnwidth}
  \includegraphics[width=\columnwidth]{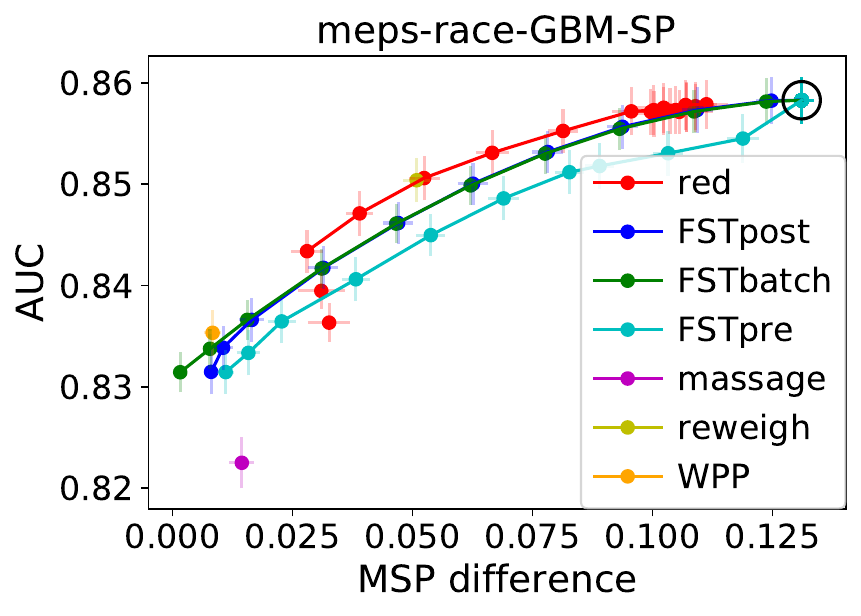}
  \label{fig:meps_1_GBM_SP_acc_AUC}
  \end{subfigure}
  \begin{subfigure}[b]{0.32\columnwidth}
  \includegraphics[width=\columnwidth]{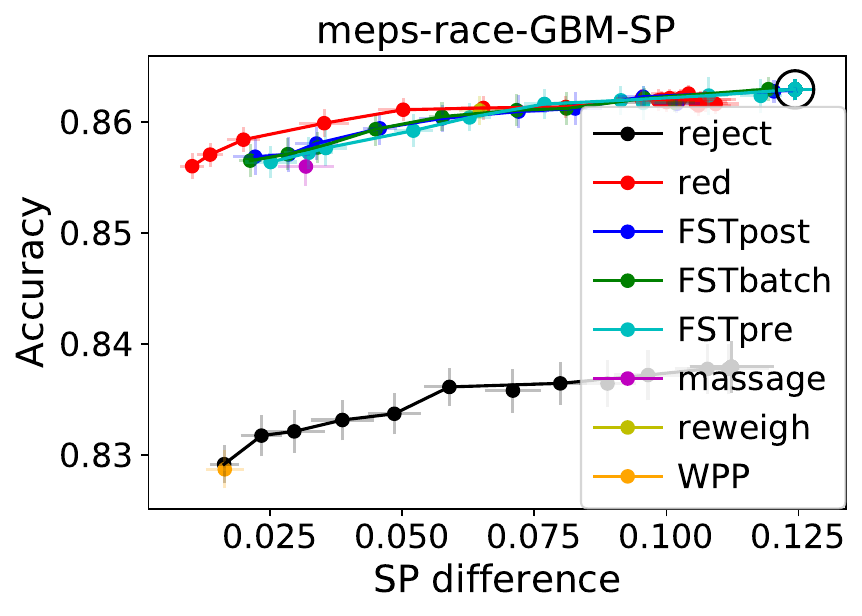}
  \label{fig:meps_1_GBM_SP_acc_acc}
  \end{subfigure}
  \begin{subfigure}[b]{0.32\columnwidth}
  \includegraphics[width=\columnwidth]{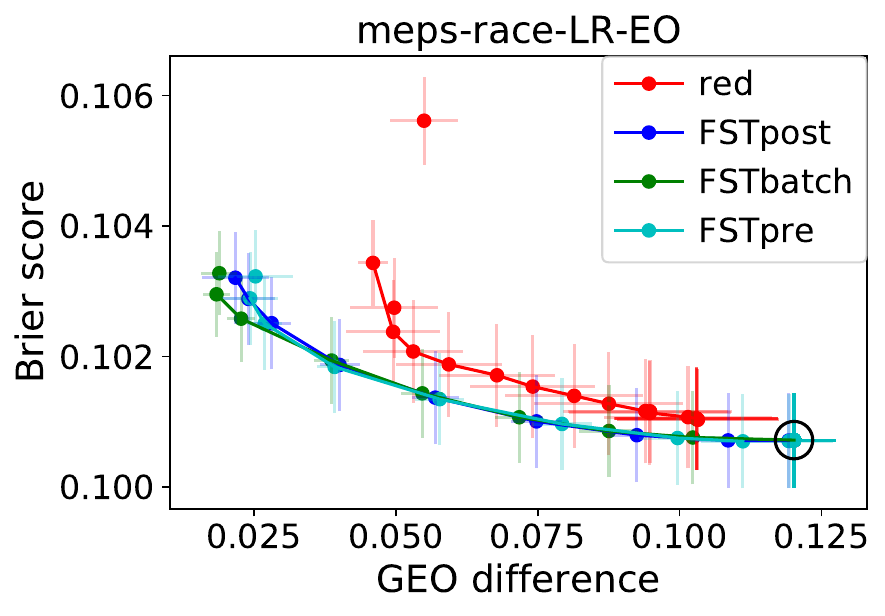}
  \label{fig:meps_1_LR_EO_acc_Brier}
  \end{subfigure}
  \begin{subfigure}[b]{0.32\columnwidth}
  \includegraphics[width=\columnwidth]{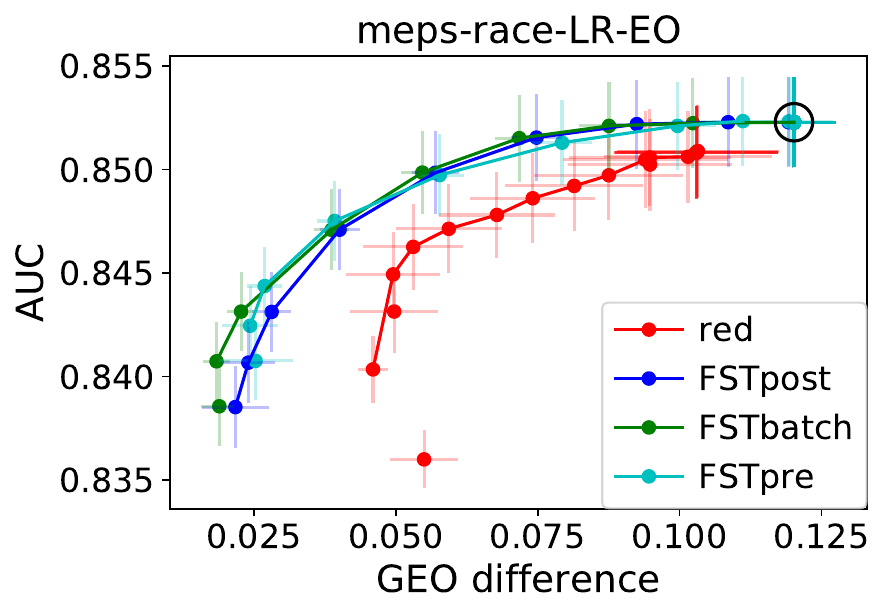}
  \label{fig:meps_1_LR_EO_acc_AUC}
  \end{subfigure}
  \begin{subfigure}[b]{0.32\columnwidth}
  \includegraphics[width=\columnwidth]{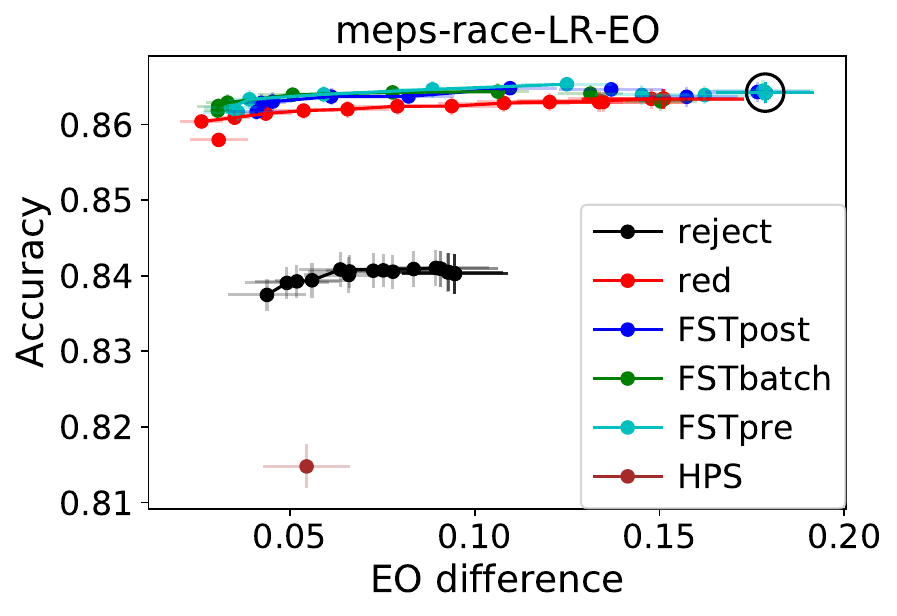}
  \label{fig:meps_1_LR_EO_acc_acc}
  \end{subfigure}
  \begin{subfigure}[b]{0.32\columnwidth}
  \includegraphics[width=\columnwidth]{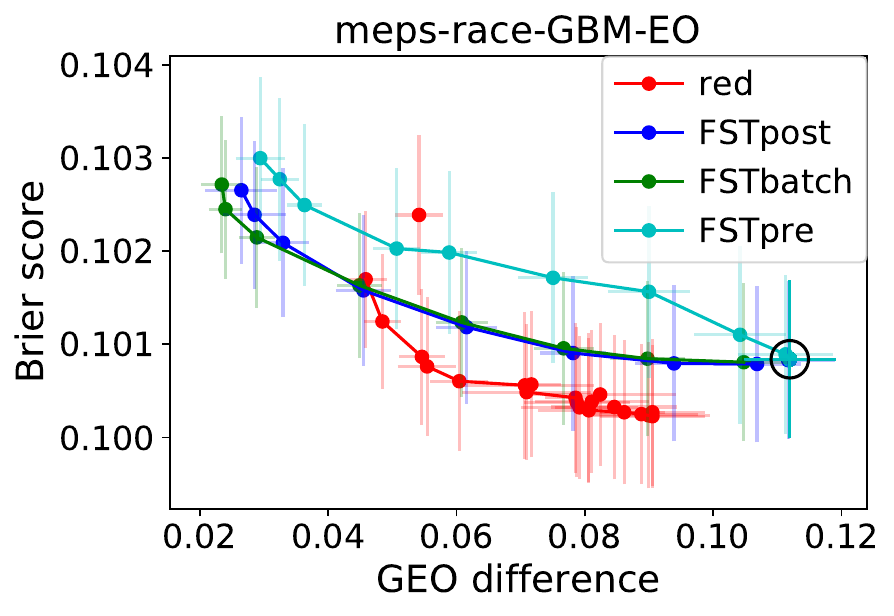}
  \label{fig:meps_1_GBM_EO_acc_Brier}
  \end{subfigure}
  \begin{subfigure}[b]{0.32\columnwidth}
  \includegraphics[width=\columnwidth]{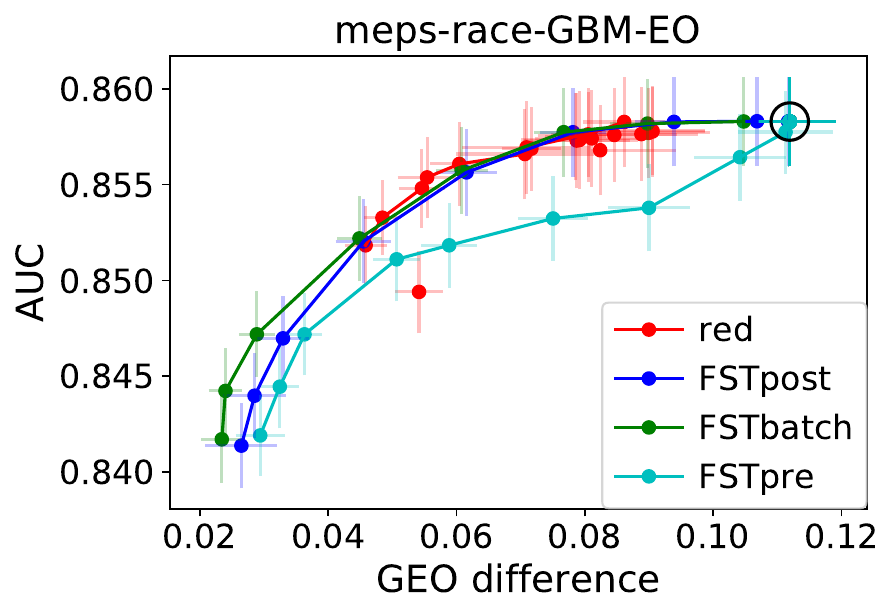}
  \label{fig:meps_1_GBM_EO_acc_AUC}
  \end{subfigure}
  \begin{subfigure}[b]{0.32\columnwidth}
  \includegraphics[width=\columnwidth]{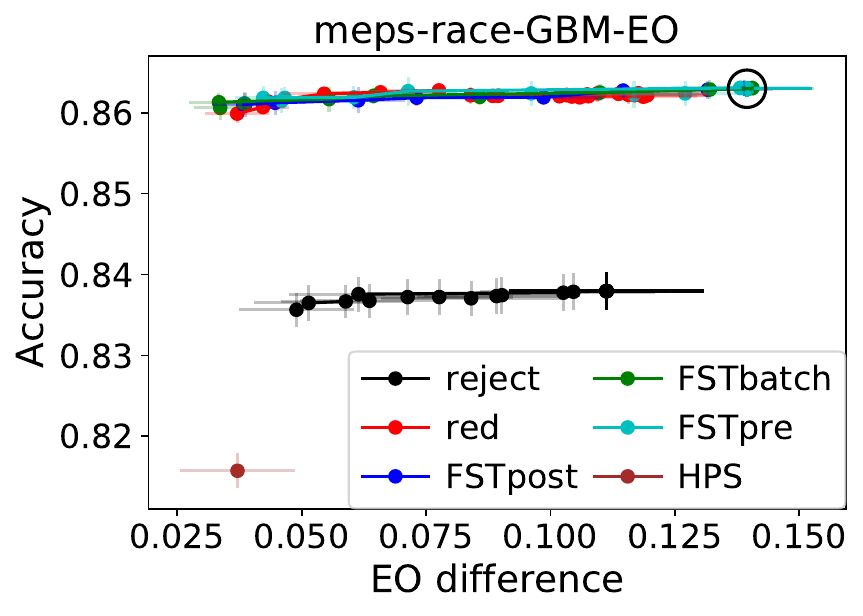}
  \label{fig:meps_1_GBM_EO_acc_acc}
  \end{subfigure}
  \caption{Trade-offs between fairness and classification performance on the MEPS data set with race as the protected attribute and the protected attribute included in the features.}
  \label{fig:meps_1}
\end{figure}

\subsection{Results with Exact Knowledge of Protected Attributes}
\label{sec:expt:Atest}

Figures~\ref{fig:adult_1}--\ref{fig:meps_1} show trade-offs between classification performance and fairness for the case where the features include the protected attribute $A$.
We defer results on the German credit data set to Appendix~\ref{sec:exptAdd} because its small size makes the results less conclusive. Appendix~\ref{sec:exptAdd} also presents separate comparisons with FERM using linear SVMs, and with OPP and DM using reduced feature sets, as mentioned above. 

Each of Figures~\ref{fig:adult_1}--\ref{fig:meps_1} corresponds to one data set-protected attribute combination. The left two columns show score-based measures: Brier score in the leftmost column and AUC in the middle column versus MSP or GEO differences on the x-axis. The rightmost column shows binary label-based measures, namely accuracy vs.~SP or EO differences. The rows correspond to combinations of base classifier (LR, GBM) and fairness measure targeted (SP, EO). Markers indicate mean values over the $10$ splits, error bars indicate standard errors in the means, and Pareto-optimal points have been connected with line segments to ease visualization.

Considering first the score-based plots (left and middle columns), FSTpost and FSTbatch achieve trade-offs that are at least as good as all other methods, with a few slight exceptions involving GBMs (e.g.,~MEPS in Figure~\ref{fig:meps_1}, AUC vs.~MSP difference in Figure~\ref{fig:adult_1}). In all cases, the advantage of FST lies in extending the Pareto frontiers farther to the left, attaining smaller MSP or GEO differences; this is especially apparent for GEO. FSTpre sometimes performs less well, e.g.,~with GBM on Adult (Figures~\ref{fig:adult_1} and \ref{fig:adult_2}) and MEPS (Figure~\ref{fig:meps_1}). This is likely due to the additional step of approximating the transformed score $r'(x)$ with the output of a classifier fit to the pre-processed data, which incurs loss.

Turning to the binary label-based plots (right column), the trade-offs for FSTpost and FSTbatch generally coincide with or are close to the trade-offs of the best method, and are even sometimes the best, despite not optimizing for binary metrics beyond tuning the binarization threshold for accuracy. Again FSTpre with GBM is worse on Adult, but FSTpre with LR is a top performer on COMPAS (Figures~\ref{fig:compas_1} and \ref{fig:compas_2}). The main disadvantage of FST is that its trade-off curves may not extend as far to the left as other methods, in particular on Adult. This is the converse of its advantage for score-based metrics. 

Among the existing methods, reductions is the strongest and also the most versatile, handling all cases that FST does. However, it is an in-processing method and far more computationally expensive, requiring an average of nearly $30$ calls to the base classification algorithm compared to one for FSTpost, FSTbatch and two for FSTpre. Reductions also returns a randomized classifier, which may not be desirable in some applications. The other in-processing method shown in Figures~\ref{fig:adult_1}--\ref{fig:meps_1} is FC, which applies only to the LR-SP rows (it is not compatible with GBM). It was not able to substantially reduce unfairness, particularly on COMPAS and MEPS and possibly due to the larger dimensionality of those data sets.

The post-processing methods of \citet{kamiran2012,hardt2016} are not designed to output scores and hence are omitted from the score-based plots. Reject option \citep{kamiran2012} performs close to the best in many cases, but not on COMPAS-gender (Figure~\ref{fig:compas_1}) and MEPS (Figure~\ref{fig:meps_1}) and at small unfairness values. HPS is limited to EO, does not have a parameter to vary the trade-off, and is less competitive. WPP and the pre-processing methods of \citet{kamiranC2012}, 
massaging and reweighing, likewise do not have a trade-off parameter and are limited to SP. As also observed by \citet{agarwal2018}, massaging is often dominated by other methods while reweighing lies on the Pareto frontier but with substantial disparity. WPP results in low disparity but its classification performance (Brier score, AUC, or accuracy) is sometimes less competitive.

\begin{figure}[t]
  \centering
  \begin{subfigure}[b]{0.32\columnwidth}
  \includegraphics[width=\columnwidth]{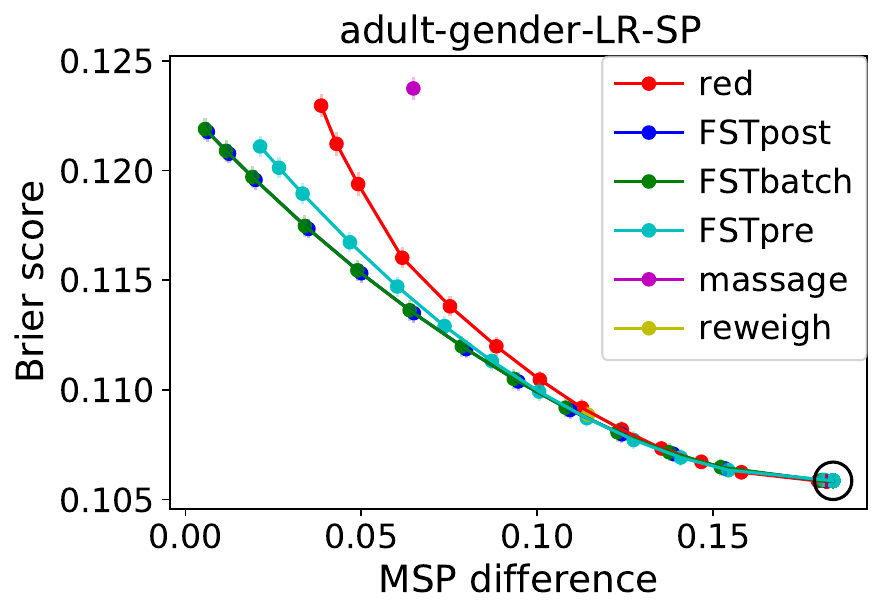}
  \label{fig:adult_1_LR_SP_acc_Brier_False}
  \end{subfigure}
  \begin{subfigure}[b]{0.32\columnwidth}
  \includegraphics[width=\columnwidth]{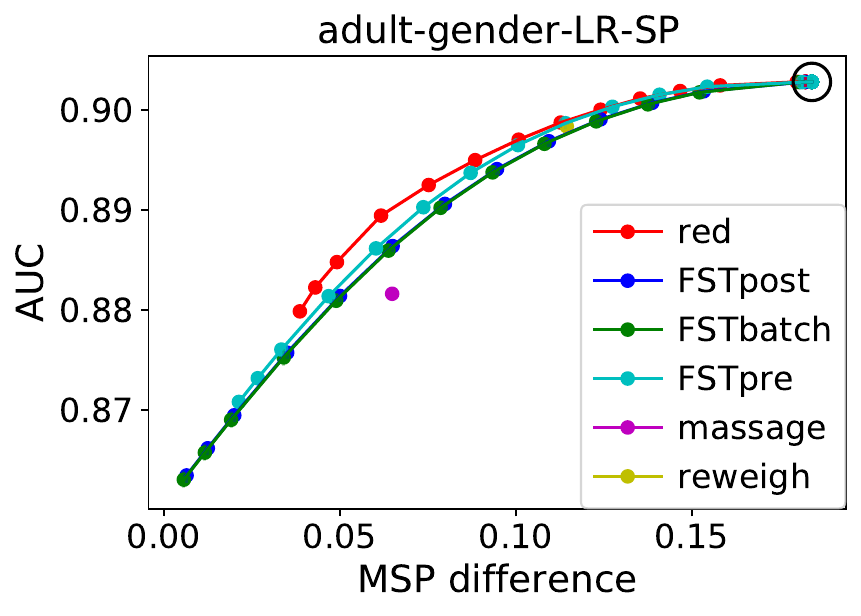}
  \label{fig:adult_1_LR_SP_acc_AUC_False}
  \end{subfigure}
  \begin{subfigure}[b]{0.32\columnwidth}
  \includegraphics[width=\columnwidth]{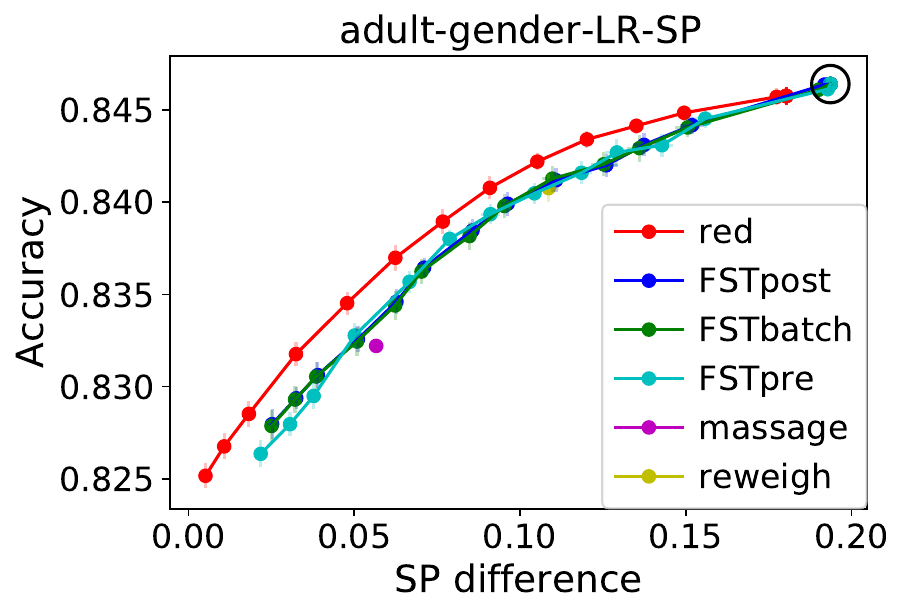}
  \label{fig:adult_1_LR_SP_acc_acc_False}
  \end{subfigure}
  \begin{subfigure}[b]{0.32\columnwidth}
  \includegraphics[width=\columnwidth]{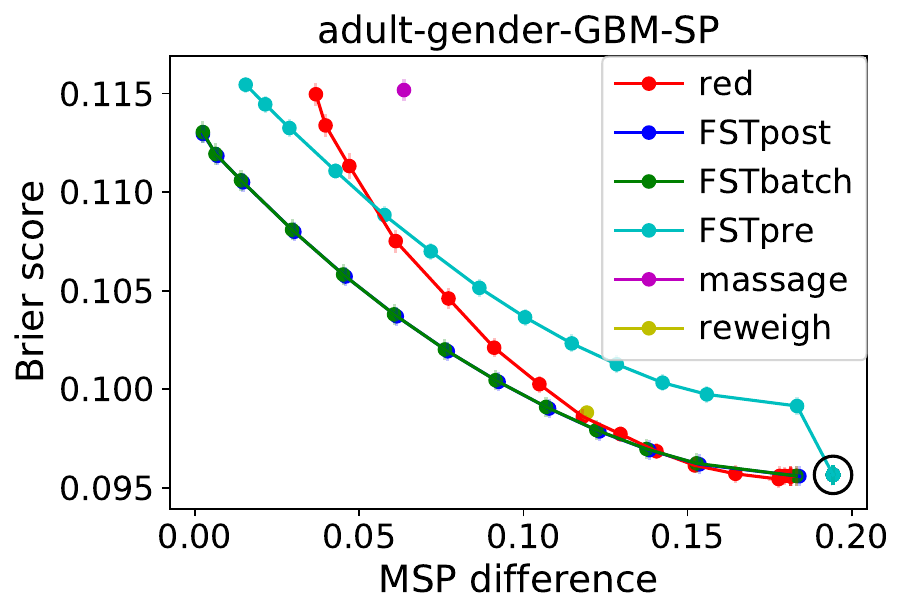}
  \label{fig:adult_1_GBM_SP_acc_Brier_False}
  \end{subfigure}
  \begin{subfigure}[b]{0.32\columnwidth}
  \includegraphics[width=\columnwidth]{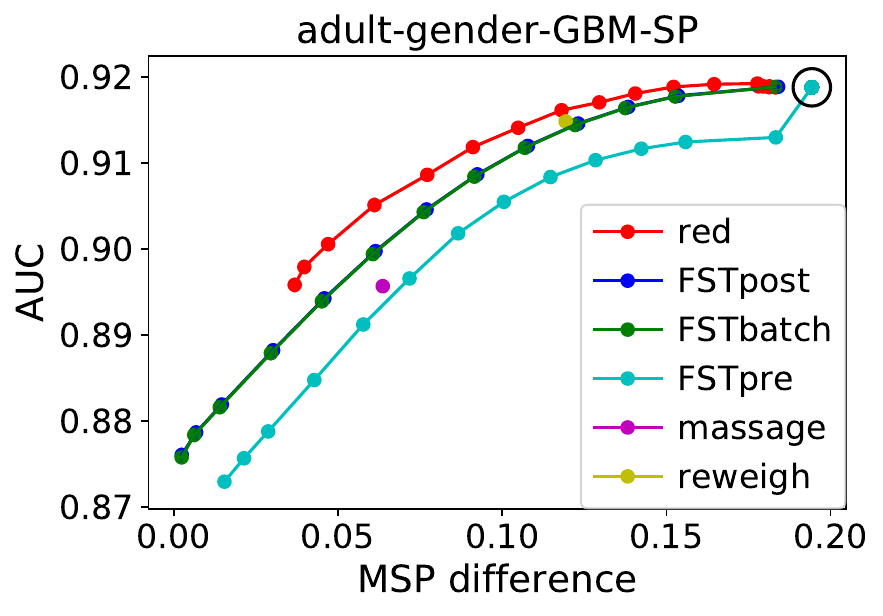}
  \label{fig:adult_1_GBM_SP_acc_AUC_False}
  \end{subfigure}
  \begin{subfigure}[b]{0.32\columnwidth}
  \includegraphics[width=\columnwidth]{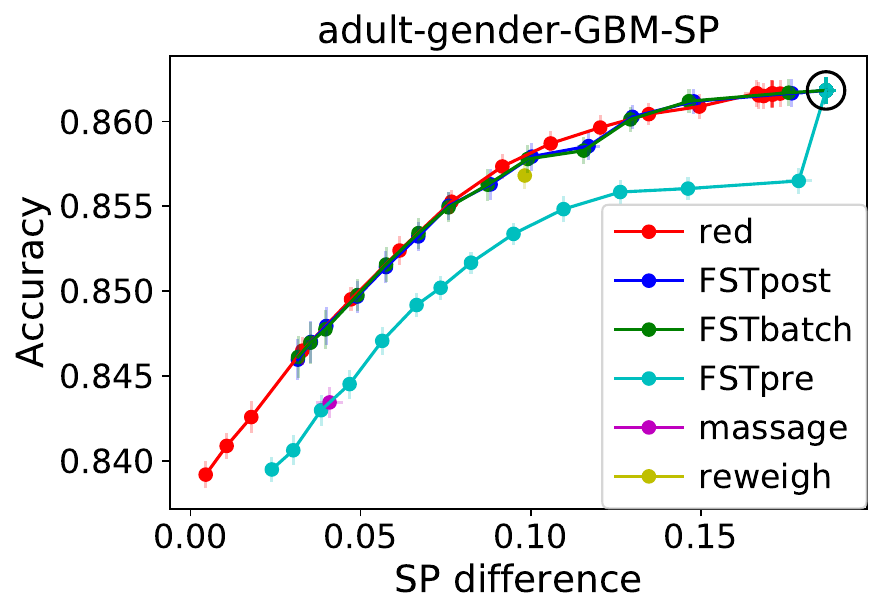}
  \label{fig:adult_1_GBM_SP_acc_acc_False}
  \end{subfigure}
  \begin{subfigure}[b]{0.32\columnwidth}
  \includegraphics[width=\columnwidth]{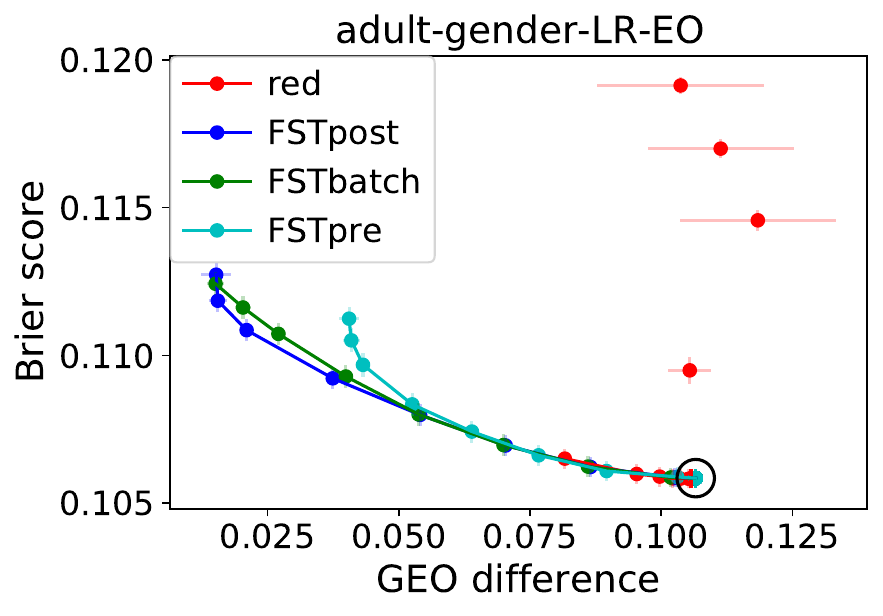}
  \label{fig:adult_1_LR_EO_acc_Brier_False}
  \end{subfigure}
  \begin{subfigure}[b]{0.32\columnwidth}
  \includegraphics[width=\columnwidth]{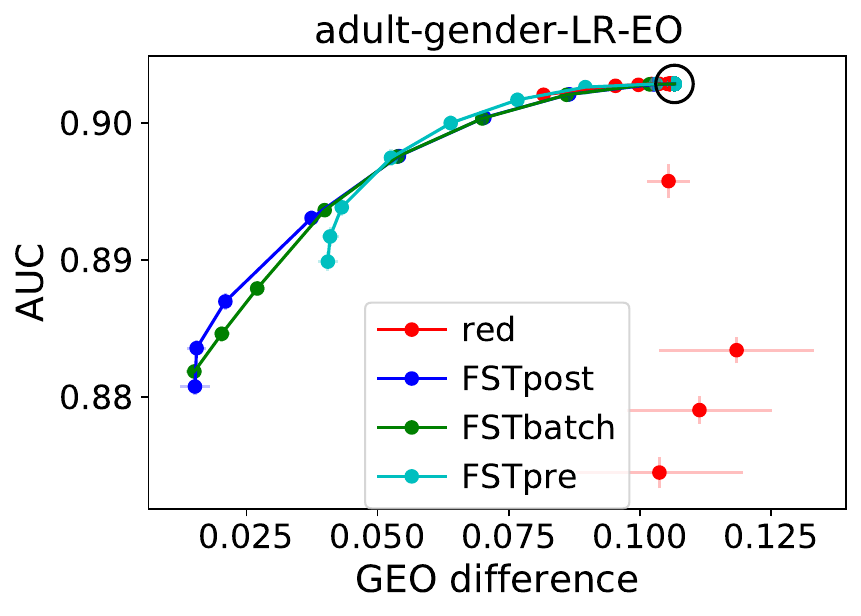}
  \label{fig:adult_1_LR_EO_acc_AUC_False}
  \end{subfigure}
  \begin{subfigure}[b]{0.32\columnwidth}
  \includegraphics[width=\columnwidth]{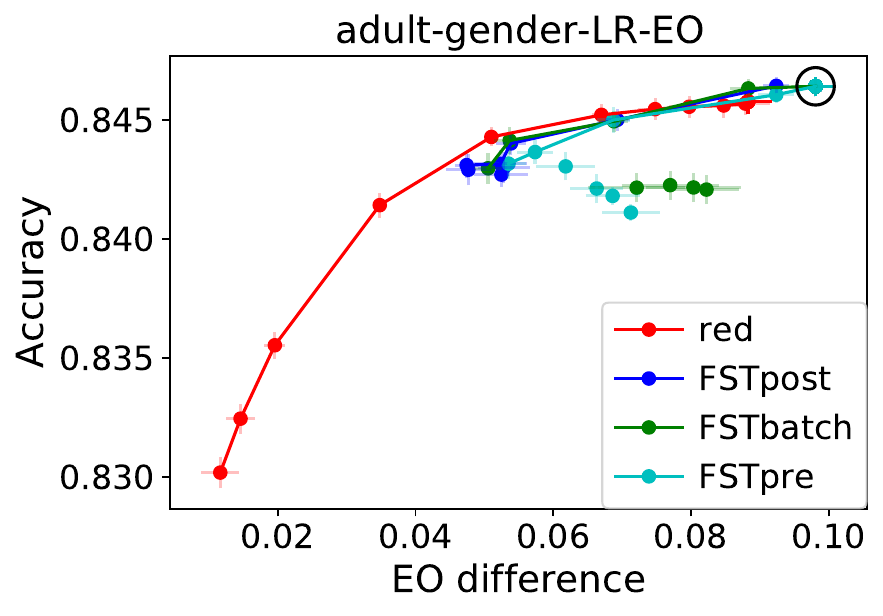}
  \label{fig:adult_1_LR_EO_acc_acc_False}
  \end{subfigure}
  \begin{subfigure}[b]{0.32\columnwidth}
  \includegraphics[width=\columnwidth]{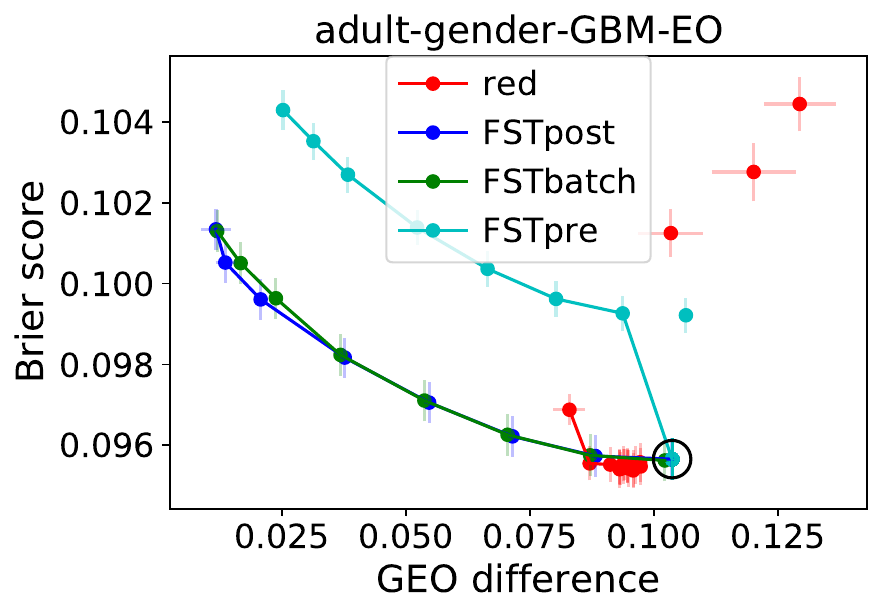}
  \label{fig:adult_1_GBM_EO_acc_Brier_False}
  \end{subfigure}
  \begin{subfigure}[b]{0.32\columnwidth}
  \includegraphics[width=\columnwidth]{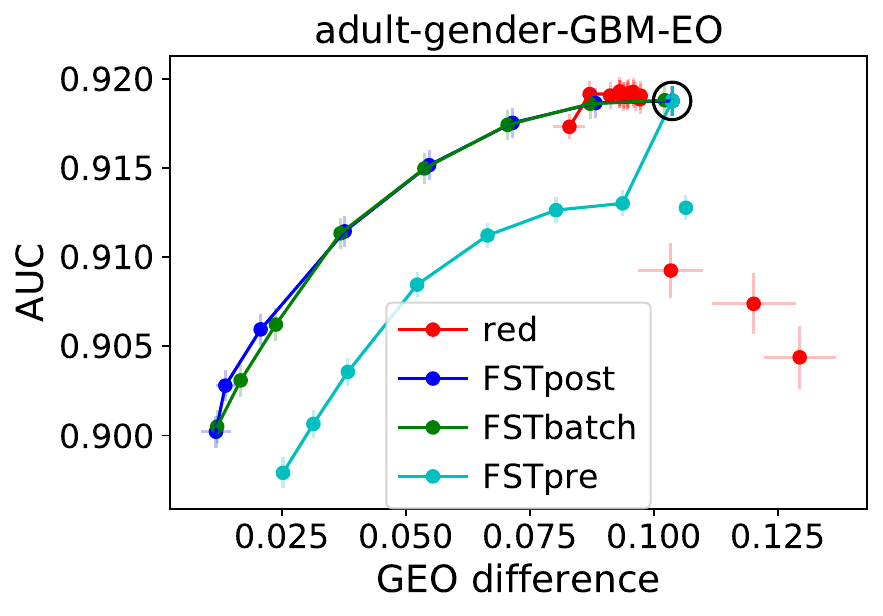}
  \label{fig:adult_1_GBM_EO_acc_AUC_False}
  \end{subfigure}
  \begin{subfigure}[b]{0.32\columnwidth}
  \includegraphics[width=\columnwidth]{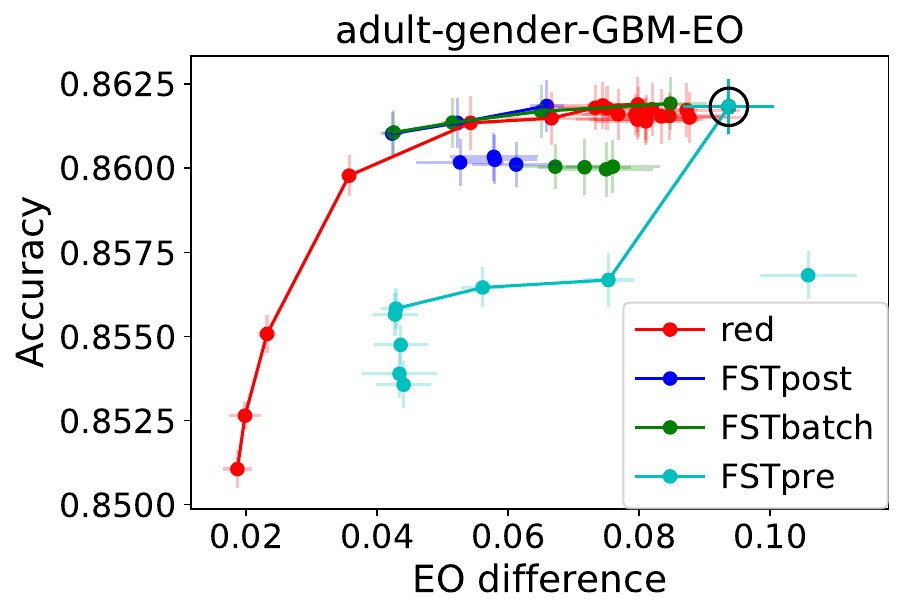}
  \label{fig:adult_1_GBM_EO_acc_acc_False}
  \end{subfigure}
  \caption{Trade-offs between fairness and classification performance on the Adult Income data set with gender as the protected attribute and the protected attribute excluded from the features.}
  \label{fig:adult_1_False}
\end{figure}

\begin{figure}[t]
  \centering
  \begin{subfigure}[b]{0.32\columnwidth}
  \includegraphics[width=\columnwidth]{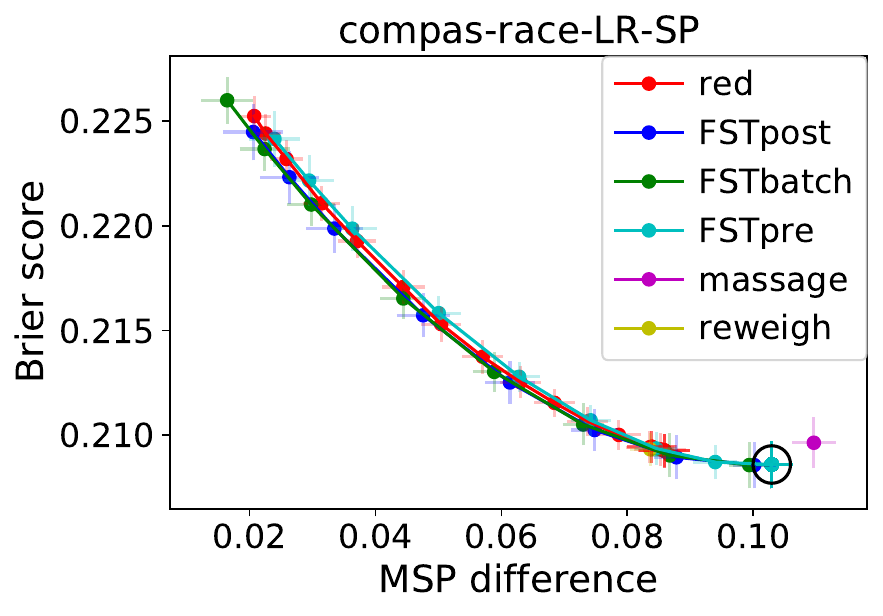}
  \label{fig:compas_2_LR_SP_acc_Brier_False}
  \end{subfigure}
  \begin{subfigure}[b]{0.32\columnwidth}
  \includegraphics[width=\columnwidth]{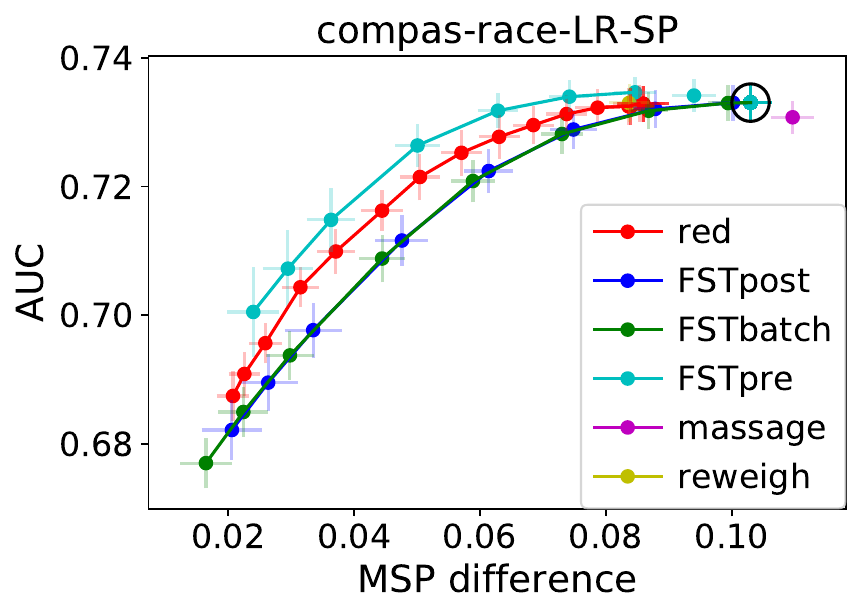}
  \label{fig:compas_2_LR_SP_acc_AUC_False}
  \end{subfigure}
  \begin{subfigure}[b]{0.32\columnwidth}
  \includegraphics[width=\columnwidth]{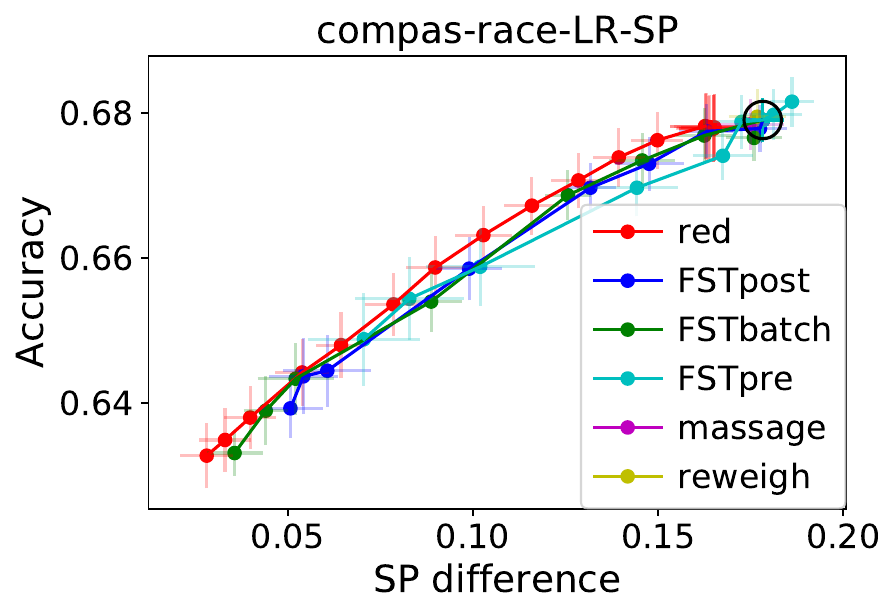}
  \label{fig:compas_2_LR_SP_acc_acc_False}
  \end{subfigure}
  \begin{subfigure}[b]{0.32\columnwidth}
  \includegraphics[width=\columnwidth]{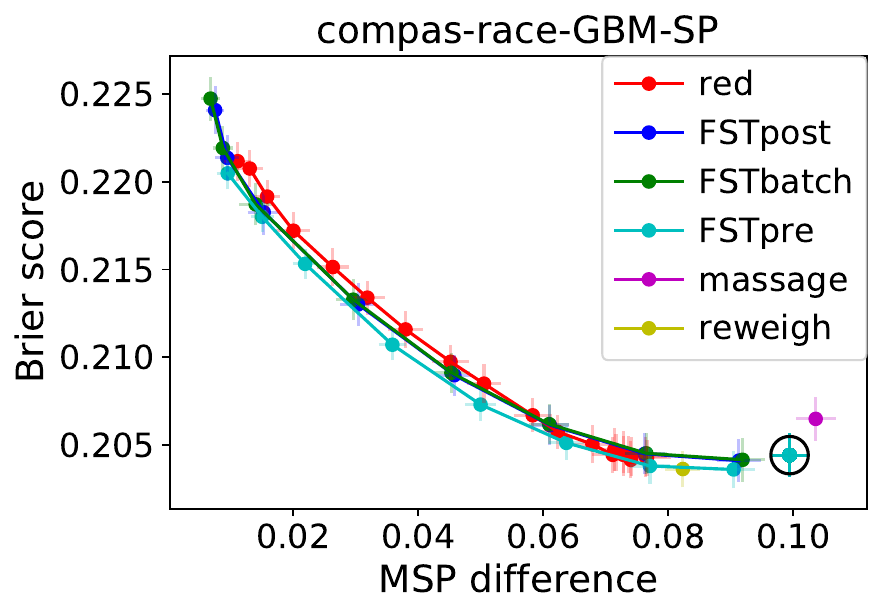}
  \label{fig:compas_2_GBM_SP_acc_Brier_False}
  \end{subfigure}
  \begin{subfigure}[b]{0.32\columnwidth}
  \includegraphics[width=\columnwidth]{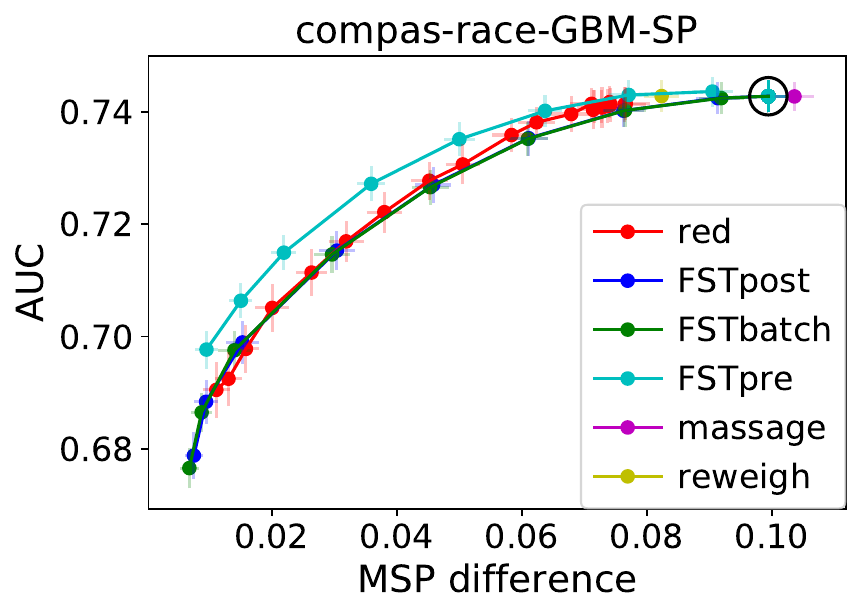}
  \label{fig:compas_2_GBM_SP_acc_AUC_False}
  \end{subfigure}
  \begin{subfigure}[b]{0.32\columnwidth}
  \includegraphics[width=\columnwidth]{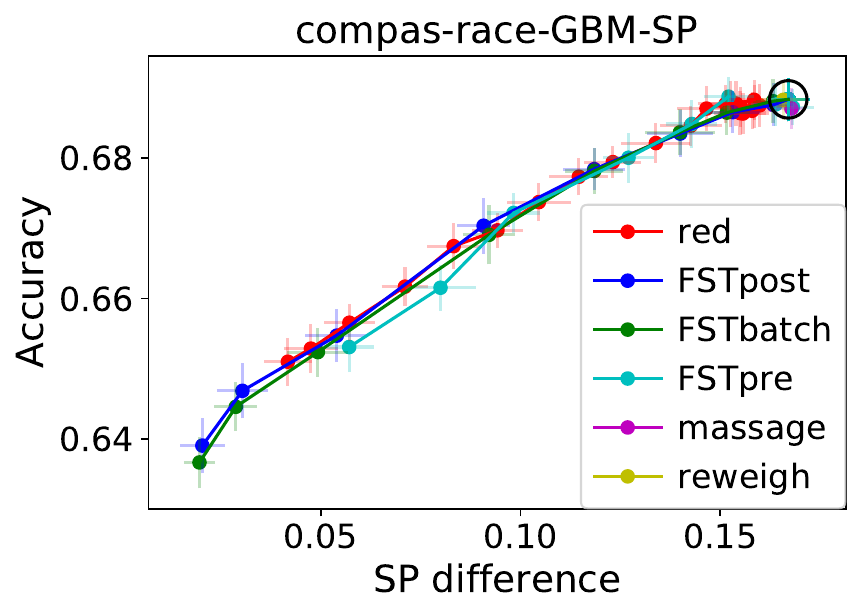}
  \label{fig:compas_2_GBM_SP_acc_acc_False}
  \end{subfigure}
  \begin{subfigure}[b]{0.32\columnwidth}
  \includegraphics[width=\columnwidth]{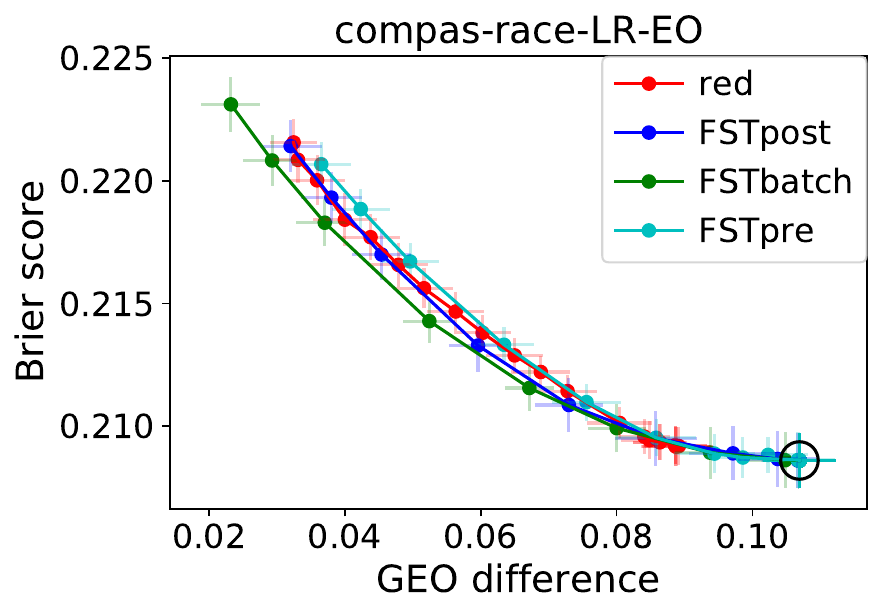}
  \label{fig:compas_2_LR_EO_acc_Brier_False}
  \end{subfigure}
  \begin{subfigure}[b]{0.32\columnwidth}
  \includegraphics[width=\columnwidth]{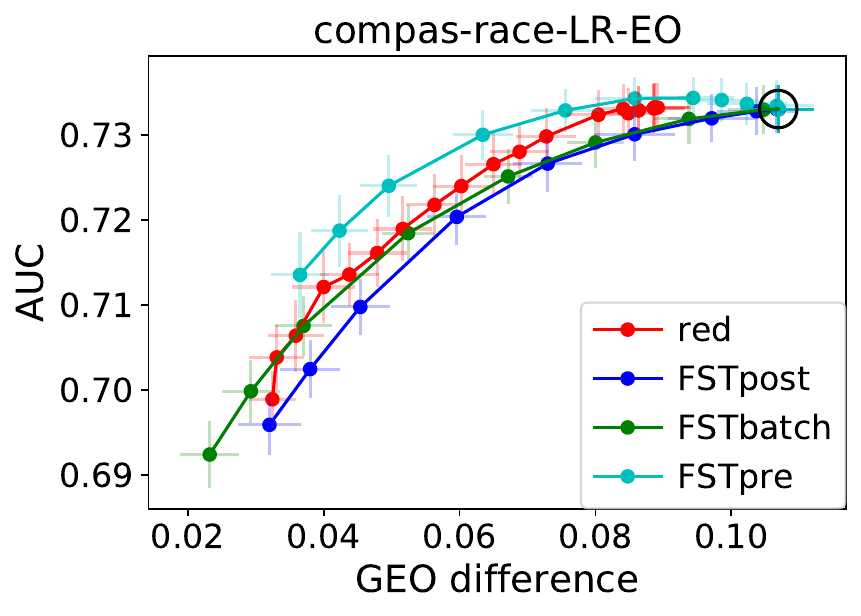}
  \label{fig:compas_2_LR_EO_acc_AUC_False}
  \end{subfigure}
  \begin{subfigure}[b]{0.32\columnwidth}
  \includegraphics[width=\columnwidth]{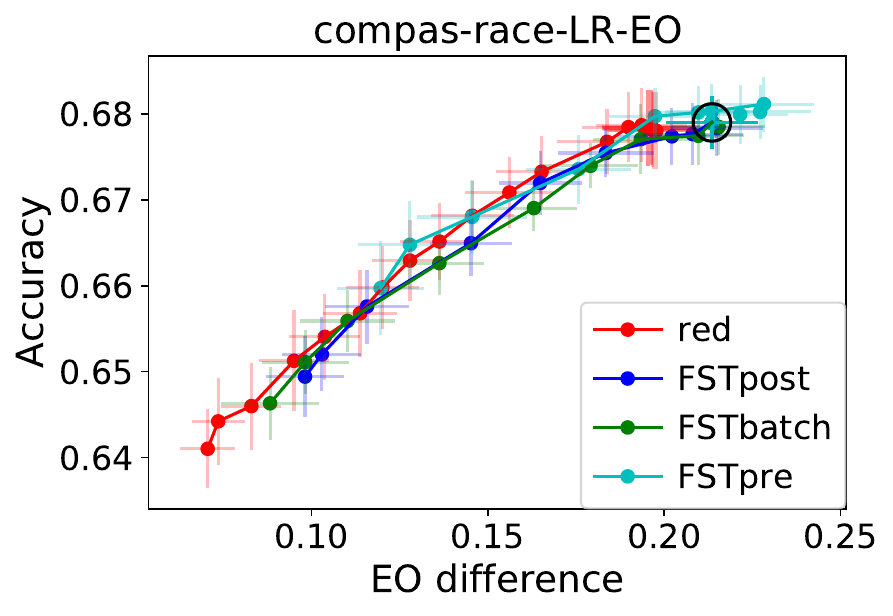}
  \label{fig:compas_2_LR_EO_acc_acc_False}
  \end{subfigure}
  \begin{subfigure}[b]{0.32\columnwidth}
  \includegraphics[width=\columnwidth]{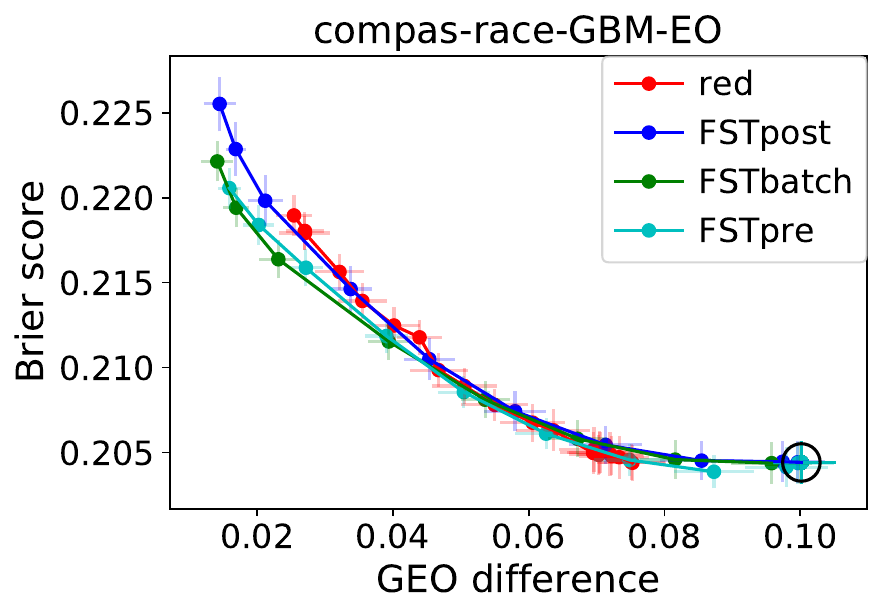}
  \label{fig:compas_2_GBM_EO_acc_Brier_False}
  \end{subfigure}
  \begin{subfigure}[b]{0.32\columnwidth}
  \includegraphics[width=\columnwidth]{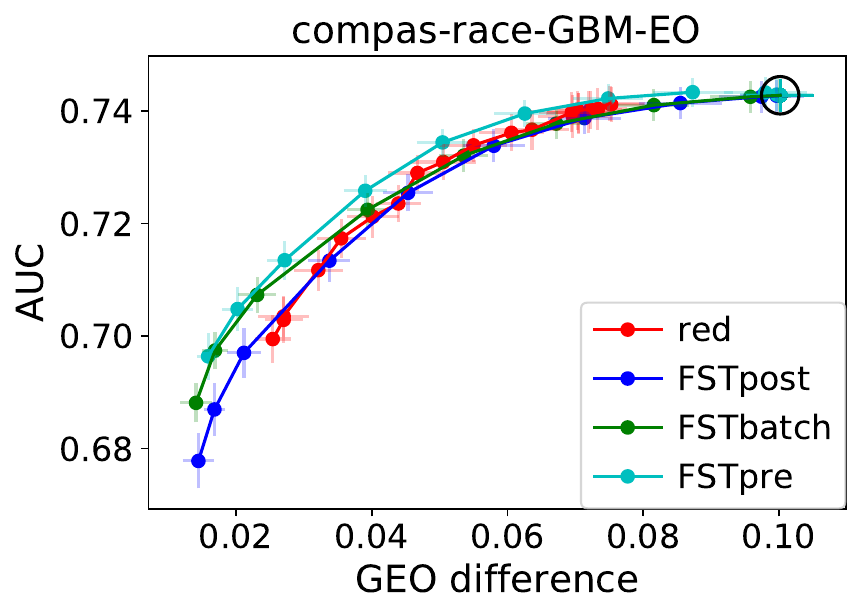}
  \label{fig:compas_2_GBM_EO_acc_AUC_False}
  \end{subfigure}
  \begin{subfigure}[b]{0.32\columnwidth}
  \includegraphics[width=\columnwidth]{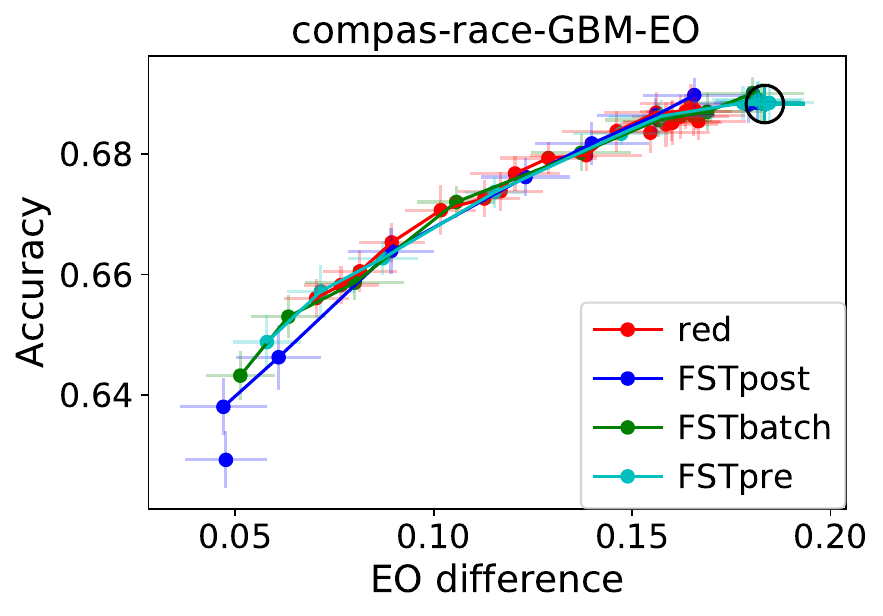}
  \label{fig:compas_2_GBM_EO_acc_acc_False}
  \end{subfigure}
  \caption{Trade-offs between fairness and classification performance on the COMPAS data set with race as the protected attribute and the protected attribute excluded from the features.}
  \label{fig:compas_2_False}
\end{figure}

\subsection{Results with Inexact Knowledge of Protected Attributes}
\label{sec:expt:noAtest}

We now present results for the case where $A$ is excluded from the features and is not available at test time. We compare a smaller set of methods that can handle this case. For FST, we use the training data to train a probabilistic classifier for $A$ based on $X$ (for MSP) or $X, Y$ (for GEO), as discussed in Section~\ref{sec:proc:origScore}. The same base classifier (LR or GBM) is used for this purpose. The classifier is used to approximate $p_{A\given X}(a\given x)$ in \eqref{eqn:dualMSP_rhat} or $p_{A\given X,Y}(a\given x,y)$ in \eqref{eqn:dualMEO_rhat}, which are in turn used to compute $\mu(x)$ in both the fit and transform steps in Section~\ref{sec:proc}.

The resulting trade-offs between classification performance and fairness are shown in Figures~\ref{fig:adult_1_False} and \ref{fig:compas_2_False}. Many of the patterns observed in Figures~\ref{fig:adult_1}--\ref{fig:meps_1} reappear in Figure~\ref{fig:adult_1_False} (Adult-gender): FSTpost and FSTbatch dominate the Brier score column; FSTpre achieves worse Brier scores, AUC, and accuracies with GBMs on Adult; FST achieves smaller score-based disparities while reductions achieves smaller binary prediction disparities (especially for GEO/EO); and reductions can obtain slightly better trade-offs with AUC and accuracy. All methods are more similar on COMPAS-race in Figure~\ref{fig:compas_2_False}. In particular, FSTpre no longer lags and may have a slight advantage in the AUC column.

\subsection{Results with More Than Two Protected Groups}
\label{sec:expt:adult_both}

The FST problem formulation also applies to non-binary protected groups. We evaluate this case using the Adult Income data set with both gender and race as protected attributes, giving rise to four protected groups (White males, Black males, White females, Black females). Here we do not compare FST with other methods as many of them do not handle more than two protected groups.

\begin{figure}[t]
  \centering
  \begin{subfigure}[b]{0.32\columnwidth}
  \includegraphics[width=\columnwidth]{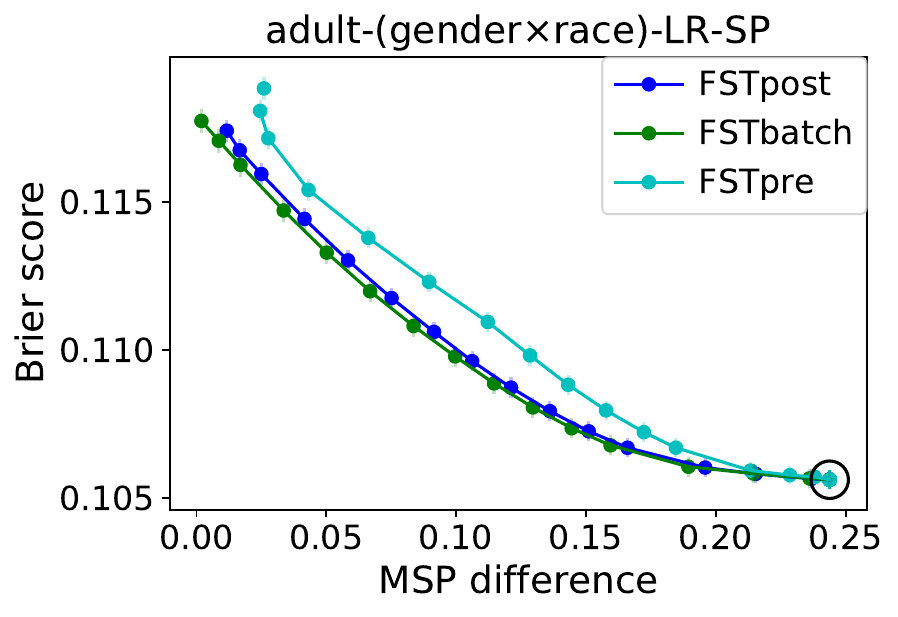}
  \label{fig:adult_both_LR_SP_acc_Brier}
  \end{subfigure}
  \begin{subfigure}[b]{0.32\columnwidth}
  \includegraphics[width=\columnwidth]{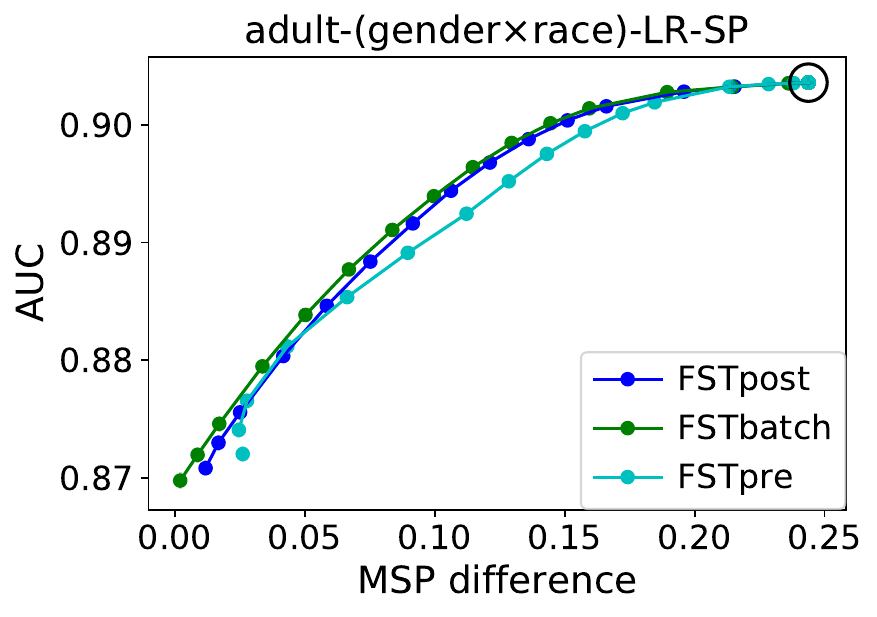}
  \label{fig:adult_both_LR_SP_acc_AUC}
  \end{subfigure}
  \begin{subfigure}[b]{0.32\columnwidth}
  \includegraphics[width=\columnwidth]{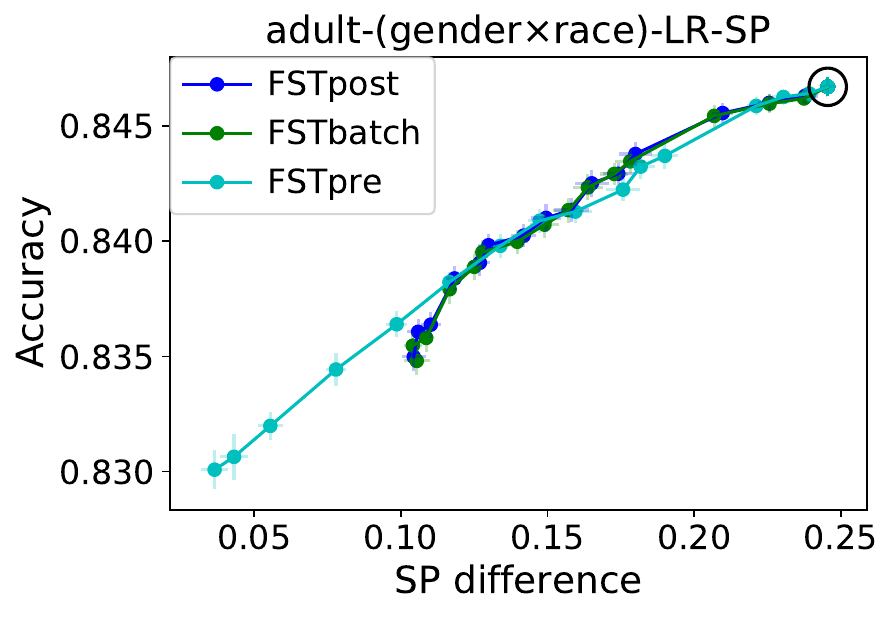}
  \label{fig:adult_both_LR_SP_acc_acc}
  \end{subfigure}
  \begin{subfigure}[b]{0.32\columnwidth}
  \includegraphics[width=\columnwidth]{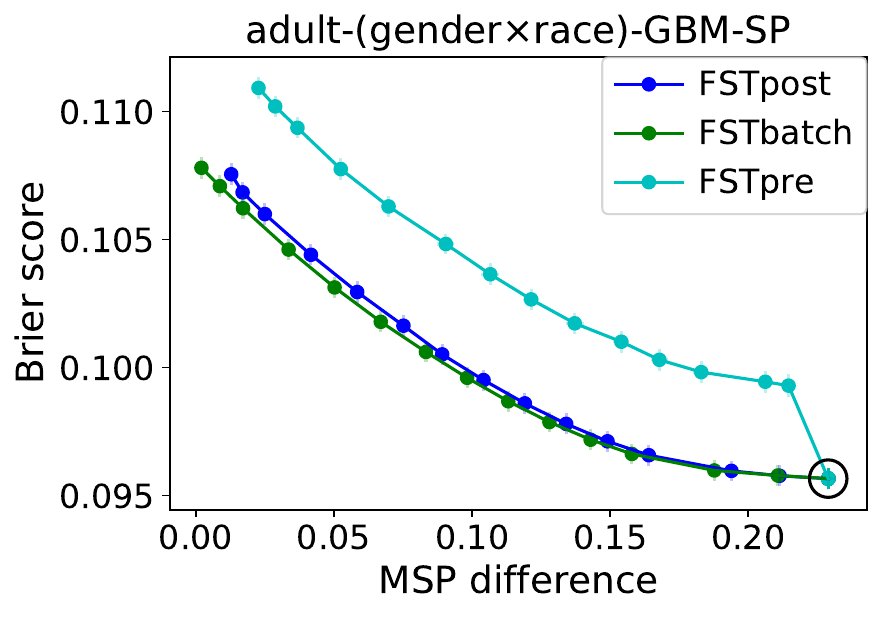}
  \label{fig:adult_both_GBM_SP_acc_Brier}
  \end{subfigure}
  \begin{subfigure}[b]{0.32\columnwidth}
  \includegraphics[width=\columnwidth]{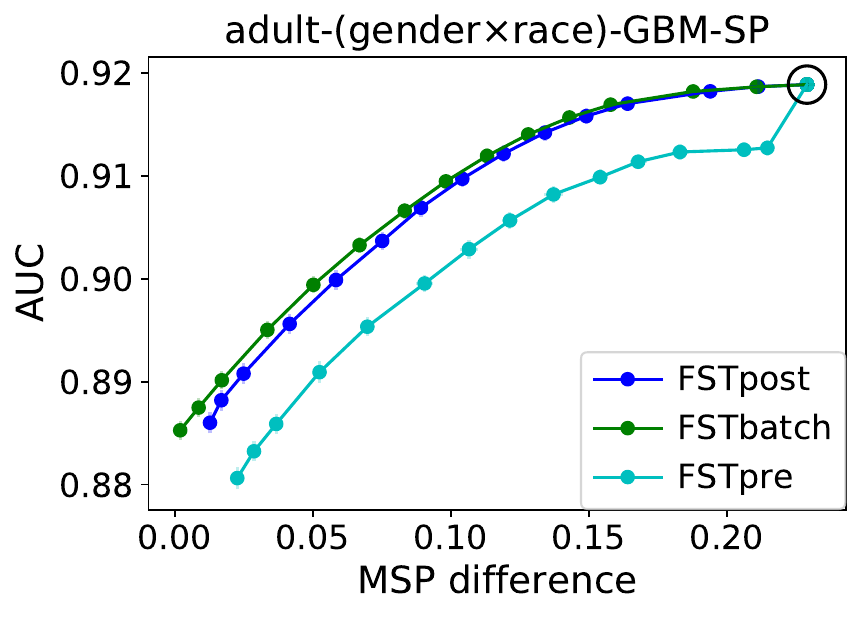}
  \label{fig:adult_both_GBM_SP_acc_AUC}
  \end{subfigure}
  \begin{subfigure}[b]{0.32\columnwidth}
  \includegraphics[width=\columnwidth]{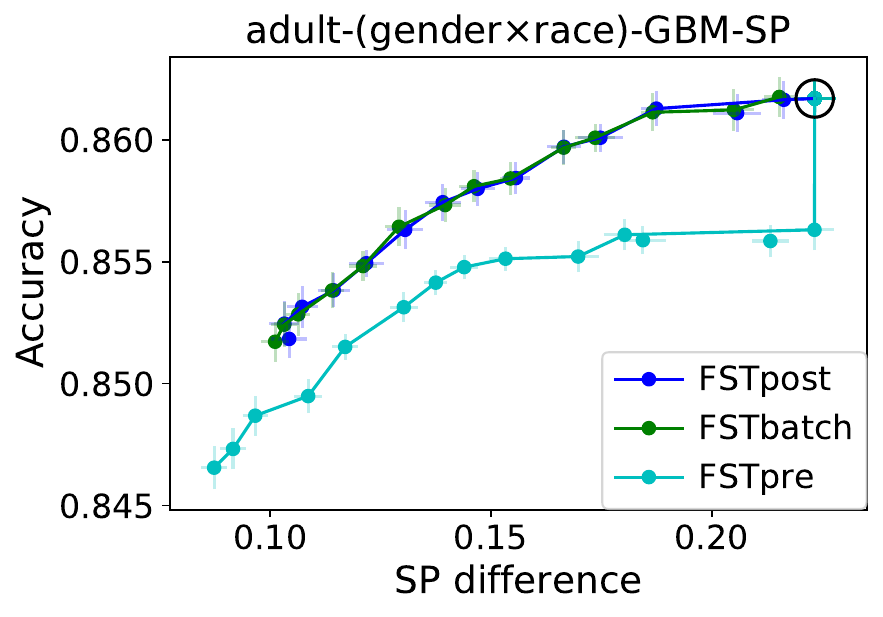}
  \label{fig:adult_both_GBM_SP_acc_acc}
  \end{subfigure}
  \begin{subfigure}[b]{0.32\columnwidth}
  \includegraphics[width=\columnwidth]{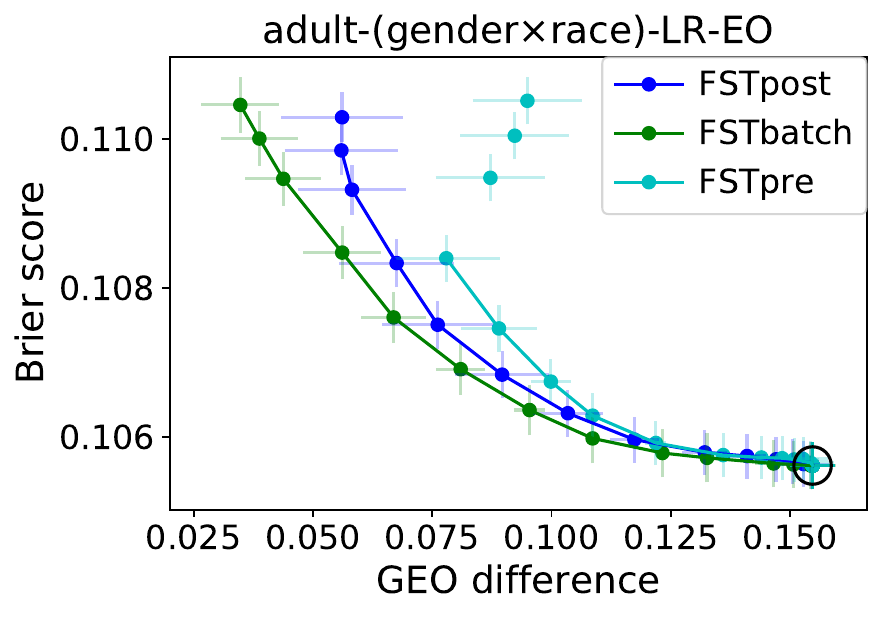}
  \label{fig:adult_both_LR_EO_acc_Brier}
  \end{subfigure}
  \begin{subfigure}[b]{0.32\columnwidth}
  \includegraphics[width=\columnwidth]{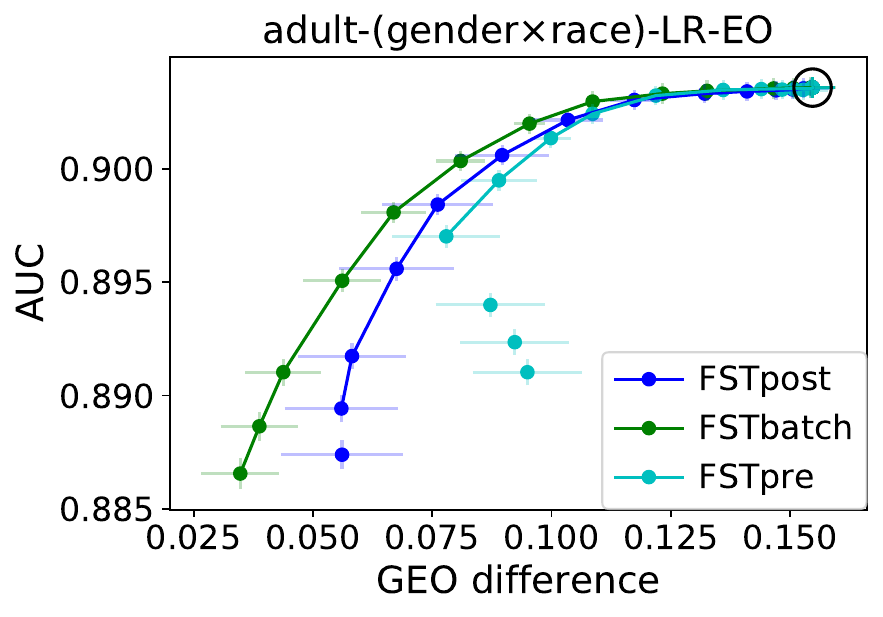}
  \label{fig:adult_both_LR_EO_acc_AUC}
  \end{subfigure}
  \begin{subfigure}[b]{0.32\columnwidth}
  \includegraphics[width=\columnwidth]{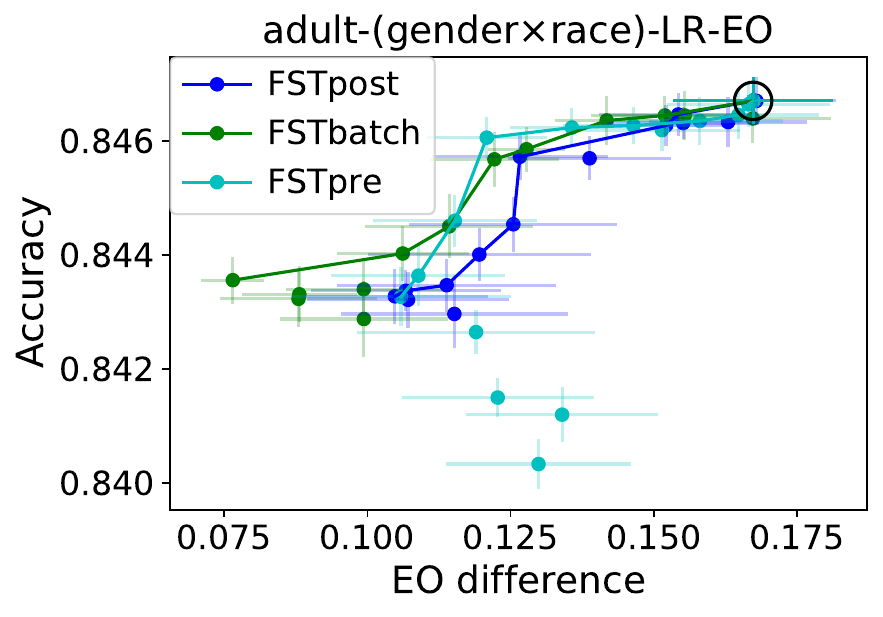}
  \label{fig:adult_both_LR_EO_acc_acc}
  \end{subfigure}
  \begin{subfigure}[b]{0.32\columnwidth}
  \includegraphics[width=\columnwidth]{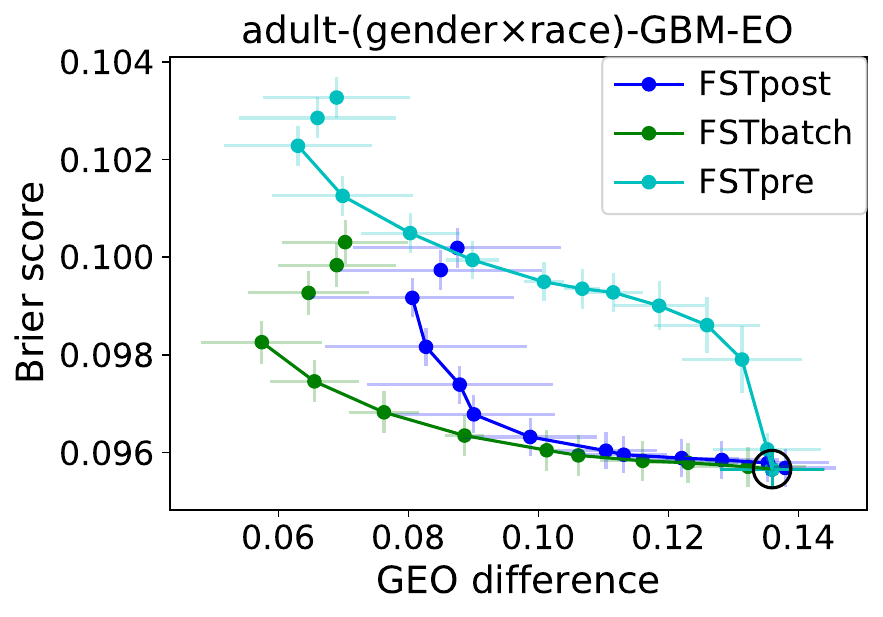}
  \label{fig:adult_both_GBM_EO_acc_Brier}
  \end{subfigure}
  \begin{subfigure}[b]{0.32\columnwidth}
  \includegraphics[width=\columnwidth]{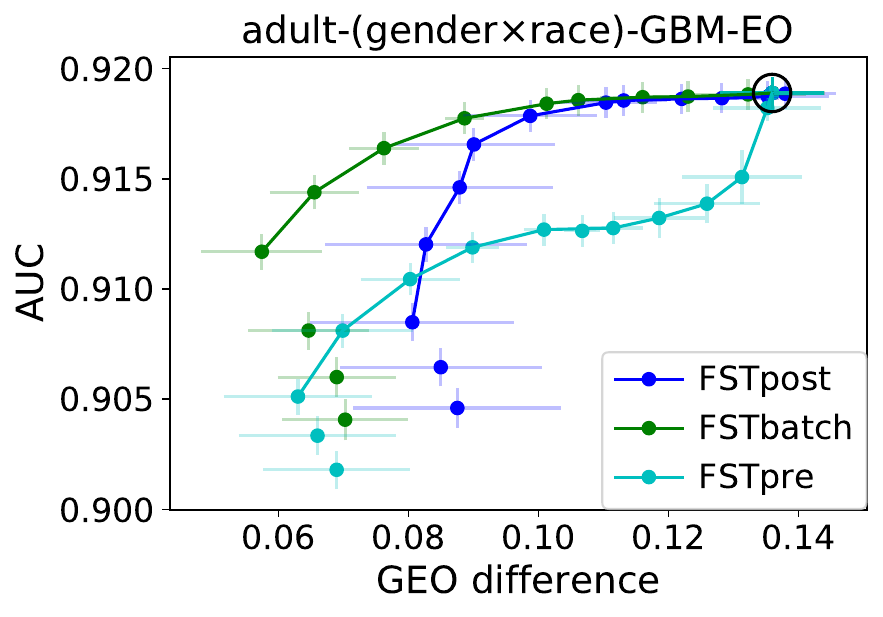}
  \label{fig:adult_both_GBM_EO_acc_AUC}
  \end{subfigure}
  \begin{subfigure}[b]{0.32\columnwidth}
  \includegraphics[width=\columnwidth]{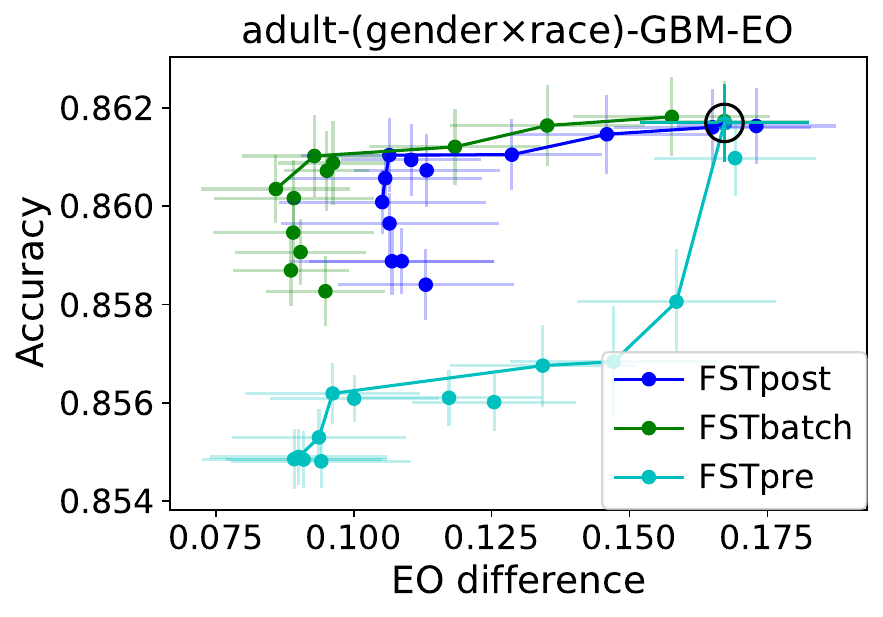}
  \label{fig:adult_both_GBM_EO_acc_acc}
  \end{subfigure}
  \caption{Trade-offs between fairness and classification performance on the Adult Income data set with both gender and race as protected attributes and the protected attributes included in the features.}
  \label{fig:adult_both}
\end{figure}

The results are shown in Figure~\ref{fig:adult_both} in the same style as Figures~\ref{fig:adult_1}--\ref{fig:compas_2_False}. With four protected groups, MSP and SP difference are computed as the largest difference in means between any two of the groups. Similarly, GEO and EO difference are computed as the largest (generalized) FPR or TPR difference between any two groups.

Figure~\ref{fig:adult_both} shows similar behavior to Figures~\ref{fig:adult_1} and \ref{fig:adult_2} in particular. First, the pre-processing extension FSTpre results in worse Brier score, AUC, and accuracy values when applied to GBMs. Second, in the right-most column, the binary label-based measures of SP and EO difference are not reduced as much as the score-based MSP and GEO difference. In general, the SP and EO difference values are higher in Figure~\ref{fig:adult_both} than in Figures~\ref{fig:adult_1} and \ref{fig:adult_2}, due to having four groups (six possible pairs) instead of two. FSTpre does achieve significantly lower SP difference  with LR than the post-processing versions (top right panel).

One difference compared to Figures~\ref{fig:adult_1} and \ref{fig:adult_2} is that there is clearer separation between FSTpost, which is fit on training data, and FSTbatch, which is fit on test data. Specifically, for EO (bottom two rows), FSTbatch attains better trade-offs than FSTpost. A possible explanation is the greater difficulty of fairness generalization with effectively eight groups (four protected groups and two labels), which FSTbatch is able to sidestep to a degree.

\section{Conclusion}
\label{sec:concl}

This paper studied the problem of fair probabilistic classification, and specifically the transformation of predicted probabilities (scores) to satisfy fairness constraints with a linearity property \eqref{eqn:fairnessLinIneq} while minimizing cross-entropy \eqref{eqn:crossentropy} with respect to the input scores. We introduced a flexible solution method called $\mathsf{FairScoreTransformer}$ (FST), whose output can be used directly as post-processing and can also be adapted to pre-process training data. \mname{} takes advantage of a closed-form expression for the optimal transformed scores, a low-dimensional convex optimization for the Lagrange multiplier parameters, and an ADMM decomposition of this convex optimization to offer a computationally efficient solution.
Theoretically, we showed in Section~\ref{sec:consistency} that FST has asymptotic and finite-sample optimality and fairness consistency properties. 
Via a comprehensive set of experiments (Section \ref{sec:expt} and Appendix~\ref{sec:exptAdd}), we numerically demonstrated that FST is either as competitive or outperforms several existing fairness intervention mechanisms over a range of settings and data sets. 

We note some limitations. First, FST inherently depends on well-calibrated classifiers  that approximate $p_{Y\given X}$ and, if necessary, $p_{A\given X}$ or $p_{A\given X,Y}$. This assumption of good calibration was made precise in Assumptions~\ref{ass:rhatTV}--\ref{ass:rhatKL}.
A poorly calibrated model 
(e.g., due lack of samples) may lead to transformed scores that do not achieve the target fairness criteria. Second, thresholding the transformed scores may have an adverse impact on fairness guarantees, as seen in the right-hand columns throughout Figures~\ref{fig:adult_1}--\ref{fig:compas_2_False}. Third, the pre-processing extension of FST depends on the classifier trained on the original data and how well it approximates $p_{Y\given X}$. The quality of this approximation limits subsequent classifiers trained on the re-weighted data.
Finally, like most pre- and post-processing methods, the score transformation found by the  FST is vulnerable to distribution shifts between training and deployment.

Future directions include: (1) characterizing the convergence rate of the ADMM iterations; (2) exploring alternative optimization algorithms for the empirical dual problem \eqref{eqn:dualSpecF}; (3) adapting \mname{} to non-binary outcomes $Y$; (4) adapting FST to fairness criteria that are not based on conditional means of scores \citep[e.g., calibration across groups as in][]{pleiss2017}; (5) extending to other modalities such as text and images.


\acks{F.P. Calmon would like to acknowledge support for this project from the National Science Foundation (NSF grant CIF-CAREER 1845852). All the authors thank the anonymous reviewers for their insightful comments during the review process, especially one reviewer whose attention to the proofs led to correction of flaws.}


\newpage

\appendix

\section{Proofs}
\label{sec:proofs}

This appendix contains all proofs deferred from the main paper, organized by section and by theorem.

\subsection{Proofs for Section~\ref{sec:opt}}

Here we provide proofs of Propositions~\ref{prop:rstar} and \ref{prop:dualspecialize} and derivations for Table~\ref{tab:menon}.

\subsubsection{Proof of Proposition \ref{prop:rstar}}
\label{sec:proof:rstar}
\begin{proof}
We manipulate the conditional mean scores as follows:
\begin{align*}
    \mathbb{E}\bigl[ r'(X) \given \cE_{lj} \bigr] &= \frac{\mathbb{E}\bigl[ r'(X) \ones((A,X,Y) \in \cE_{lj}) \bigr]}{\Pr(\cE_{lj})}\\
    &= \frac{\mathbb{E}\bigl[ \mathbb{E}\bigl[ r'(X) \ones((A,X,Y) \in \cE_{lj}) \given X \bigr] \bigr]}{\Pr(\cE_{lj})}\\
    &= \frac{\mathbb{E}\bigl[ r'(X) \Pr(\cE_{lj} \given X) \bigr]}{\Pr(\cE_{lj})},
\end{align*}
where in the second line we have iterated expectations and then moved $r'(X)$ outside of the conditional expectation given $X$.  Defining $\mu(X)$ according to \eqref{eqn:mu},  the Lagrangian \eqref{eqn:Lagrangian1} becomes 
\begin{equation}\label{eqn:Lagrangian2}
    L(r', \lambda) = \mathbb{E} \bigl[ \hat{r}(X) \log r'(X) + (1-\hat{r}(X)) \log(1 - r'(X)) - \mu(X) r'(X) \bigr] + \sum_{l=1}^L c_l \lambda_l.
\end{equation}

It can be seen from \eqref{eqn:Lagrangian2} that the maximization with respect to the primal variable $r'(X)$ can be done independently for each $X = x$.  Noting that $L(r',\lambda)$ is a concave function of $r'$ (sum of logarithmic and linear terms), a necessary and sufficient condition of optimality is that the partial derivatives with respect to each $r'(x)$ are equal to zero:
\begin{equation}\label{eqn:LagrangianOptCond}
\frac{\hat{r}(x)}{r'(x)} - \frac{1 - \hat{r}(x)}{1 - r'(x)} - \mu(x) = 0 \quad \forall x \in \cX.
\end{equation}
This condition can be rearranged into the quadratic equation 
\[
\mu(x) r'(x)^2 - (1 + \mu(x)) r'(x) + \hat{r}(x) = 0,
\]
whose solution is 
\begin{equation}
    \nonumber
    r^*\bigl(\mu(x); \hat{r}(x)\bigr) = \begin{cases}
    \dfrac{1 + \mu(x) - \sqrt{(1 + \mu(x))^2 - 4 \hat{r}(x) \mu(x)}}{2\mu(x)}, & \mu(x) \neq 0\\
    \hat{r}(x), & \mu(x) = 0,
    \end{cases}
\end{equation}
after eliminating the root outside of the interval $[0,1]$. 

Lastly, it can be seen that the substitution of $r^*$ into the expectation in \eqref{eqn:Lagrangian2} yields $\mathbb{E}[g(\mu(X); \hat{r}(X)]]$ where 
\begin{equation*}
g\bigl(\mu(x); \hat{r}(x)\bigr) \defined -H_b\Bigl(\hat{r}(x),r^*\left(\mu(x);\hat{r}(x) \right) \Bigr)  - \mu(x) r^*\bigl(\mu(x); \hat{r}(x)\bigr).
\end{equation*}

\end{proof}

\subsubsection{Proof of Proposition \ref{prop:dualspecialize}}
\label{sec:proof:dualspecialize}

\begin{proof}
We first specify the exact correspondences between \eqref{eqn:MSP}, \eqref{eqn:MEO} and \eqref{eqn:fairnessLinIneq}. The MSP constraint \eqref{eqn:MSP} can be obtained from \eqref{eqn:fairnessLinIneq} by setting $J = 2$, $l = (a,\pm)$ for $a\in\cA$ where $+$ corresponds to the $\leq \epsilon$ constraint and $-$ to the $\geq -\epsilon$ constraint, $L = 2\abs{\cA}$, $\cE_{(a,\pm),1} = \{A = a\}$, $\cE_{(a,\pm),2} = \Omega$ (the entire sample space), $c_l = \epsilon$, and $b_{(a,\pm),j} = \mp (-1)^j$.  
For the GEO constraint \eqref{eqn:MEO}, set $J = 2$, $l = (a,y,\pm)$ for $a\in\cA$, $y\in\{0,1\}$ and the same $\pm$ correspondences, $L = 4\abs{\cA}$, $\cE_{(a,y,\pm),1} = \{A = a, Y = y\}$, $\cE_{(a,y,\pm),2} = \{Y = y\}$, $c_l = \epsilon$, and $b_{(a,y,\pm),j} = \mp (-1)^j$.

\textit{Mean score parity constraints.}
\label{sec:opt:MSP}
For MSP \eqref{eqn:MSP}, let $\lambda_a^+$ and $\lambda_a^-$ respectively denote the Lagrange multipliers for the $\leq \epsilon$ and $\geq -\epsilon$ constraints for each $a \in \cA$.  With the correspondences identified above, 
the modifier $\mu(X,\lambda )$ {}becomes 
\begin{equation}\label{eqn:muMSP1}
\mu(X,\lambda) = \sum_{a\in\cA} \bigl(\lambda_a^+ - \lambda_a^-\bigr) \left( \frac{p_{A\given X}(a\given X)}{p_A(a)} - \frac{\Pr(\Omega\given X)}{\Pr(\Omega)} \right).
\end{equation}
For $\epsilon > 0$, at most one of the constraints can be active for each $a$ in \eqref{eqn:MSP}, and hence at optimality at most one of $\lambda_a^+$, $\lambda_a^-$ can be non-zero.  We can therefore interpret $\lambda_a^+$, $\lambda_a^-$ as the positive and negative parts of a real-valued Lagrange multiplier $\lambda_a = \lambda_a^+ - \lambda_a^-$, as done in linear programming \citep{bt1997}. Equation \eqref{eqn:muMSP1} can be rewritten as 
\begin{equation}\label{eqn:muMSP2}
\mu(X,\lambda) = \sum_{a\in\cA} \lambda_a \frac{p_{A\given X}(a\given X)}{p_A(a)} - \sum_{a\in\cA} \lambda_a.
\end{equation}
If $A$ is included in the features $X$, then $p_{A\given X}(a\given X) = \ones(a = A)$, where $A$ is the component of $X$ that is given, and \eqref{eqn:muMSP2} further simplifies to 
\[
\mu(X,\lambda) = \frac{\lambda_A}{p_A(A)} - \sum_{a\in\cA} \lambda_a.
\]
Interestingly, the only difference between the cases of including or excluding $A$ is that in the latter, \eqref{eqn:muMSP2} asks for $A$ to be inferred from the available features $X$, whereas in the former, $A$ can be used directly. 

In the objective function of \eqref{eqn:dualGeneral} we have 
\begin{equation}\label{eqn:1norm}
\sum_{l=1}^L c_l \lambda_l = \epsilon \sum_{a\in\cA} \bigl(\lambda_a^+ + \lambda_a^-\bigr) = \epsilon \norm{\lambda}_1
\end{equation}
upon recognizing that $(\lambda_a^+ + \lambda_a^-) = \abs{\lambda_a}$. Combining this with \eqref{eqn:muMSP2}, the dual problem for MSP is 
\begin{equation}
    \begin{split}
        \min_{\lambda} \quad &\mathbb{E}\left[ g\bigl(\mu(X); \hat{r}(X)\bigr) \right] + \epsilon \norm{\lambda}_1\\
        \st \quad &\mu(X,\lambda) = \sum_{a\in\cA} \lambda_a \frac{p_{A\given X}(a\given X)}{p_A(a)} - \sum_{a\in\cA} \lambda_a \nonumber.
    \end{split}
\end{equation}

\textit{Generalized equalized odds constraints.}
\label{sec:opt:MEO}
For GEO \eqref{eqn:MEO}, we similarly define Lagrange multipliers $\lambda_{a,y}^+$ and $\lambda_{a,y}^-$ for the $\leq \epsilon$ and $\geq -\epsilon$ constraints.  The modifier $\mu(X)$ is given by 
\begin{align}
    \mu(X,\lambda) &= \sum_{a\in\cA} \sum_{y\in\{0,1\}} \bigl(\lambda_{a,y}^+ - \lambda_{a,y}^-\bigr) \left( \frac{p_{A,Y\given X}(a,y\given X)}{p_{A,Y}(a,y)} - \frac{p_{Y\given X}(y\given X)}{p_{Y}(y)} \right)\nonumber\\
    &= \sum_{y\in\{0,1\}} \frac{p_{Y\given X}(y\given X)}{p_{Y}(y)} \sum_{a\in\cA} \lambda_{a,y} \left( \frac{p_{A\given X,Y}(a\given X,y)}{p_{A\given Y}(a\given y)} - 1\right), \label{eqn:muMEO}
\end{align}
where we have similarly identified $\lambda_{a,y} = \lambda_{a,y}^+ - \lambda_{a,y}^-$ and factored the joint distribution of $A,Y$.  If $A$ is included in $X$, \eqref{eqn:muMEO} simplifies to 
\[
\mu(X,\lambda) = \sum_{y\in\{0,1\}} \frac{p_{Y\given X}(y\given X)}{p_{Y}(y)} \left( \frac{\lambda_{A,y}}{p_{A\given Y}(A\given y)} - \sum_{a\in\cA} \lambda_{a,y} \right). 
\]
Again, the difference between the two cases lies in whether $A$ must be inferred, this time from $X$ and $Y$.  We also have an analogue to \eqref{eqn:1norm} where the summation and $\ell_1$ norm now run over all $(a,y)$.  The dual problem for GEO is therefore 
\begin{equation}
    \begin{split}
        \min_{\lambda} \quad &\mathbb{E}\left[ g\bigl(\mu(X,\lambda); \hat{r}(X)\bigr) \right] + \epsilon \norm{\lambda}_1\\
        \st \quad &\mu(X,\lambda) = \sum_{y\in\{0,1\}} \frac{p_{Y\given X}(y\given X)}{p_{Y}(y)} \sum_{a\in\cA} \lambda_{a,y} \left( \frac{p_{A\given X,Y}(a\given X,y)}{p_{A\given Y}(a\given y)} - 1\right) \nonumber.
    \end{split}
\end{equation}
\end{proof}

\subsubsection{Derivations for Table~\ref{tab:menon}}
\label{sec:proof:menon}

As stated in Section~\ref{sec:opt:compare}, for the right-hand column of Table~\ref{tab:menon}, we assume that the optimal transformed score $r^*(\mu(X); r(X))$ is thresholded at the cost-sensitive threshold $c$ to obtain a binary prediction, $\hat{Y}(X) = \ones(r^*(\mu(X); r(X)) > c)$. By virtue of the monotonicity of $r^*(\mu; r)$ in $r$ (Lemma~\ref{clm:monotonic} and Figure~\ref{fig:rStar_r}), this is equivalent to thresholding $r(X)$ at a transformed threshold, which can be determined by setting $r^* = c$ and inverting \eqref{eqn:r*} (see Appendix~\ref{sec:proof:rstar} for the quadratic equation that leads to Equation~\ref{eqn:r*}). The result is 
\begin{equation}\label{eqn:Yhat}
    \hat{Y}(X) = \ones\left(r(X) - c(1-c) \mu(X) > c\right),
\end{equation}
i.e., an additive modification to the threshold that is proportional to $\mu(X)$.

We now discuss each row in Table~\ref{tab:menon} in turn. For the case of SP, \citet[Proposition~4]{menon2018} show that the classifier that minimizes \eqref{eqn:menonCostSens} is given by 
\begin{equation}\label{eqn:menonSP}
    \hat{Y}^*(X) = \ones\left(r(X) - \lambda(\bar{\eta}(X) - 1/2) > c\right),
\end{equation}
which is of the form $\ones(h(X) > c)$ with $h(X)$ as given in the corresponding entry of Table~\ref{tab:menon}. Two notes: (1) the $1/2$ in \eqref{eqn:menonSP} comes from the equivalence of their MD criterion to a second cost-sensitive risk with weight $\bar{c} = 1/2$ \citep[Lemma~2]{menon2018}; (2) the case where the threshold is met with equality is ignored for simplicity. On the other hand, for the thresholded optimal fair score \eqref{eqn:Yhat} and the case of SP, the constraint in \eqref{eqn:dualMSP_rhat} gives 
\[
\mu(X) = \lambda_0 \left(\frac{1-\bar{\eta}(X)}{p_A(0)} - 1\right) + \lambda_1 \left(\frac{\bar{\eta}(X)}{p_A(1)} - 1\right),
\]
and hence 
\begin{equation}\label{eqn:YhatSP}
    \hat{Y}(X) = \ones\left(r(X) - c(1-c) \left(\lambda_0 \left(\frac{1-\bar{\eta}(X)}{p_A(0)} - 1\right) + \lambda_1 \left(\frac{\bar{\eta}(X)}{p_A(1)} - 1\right)\right) > c\right).
\end{equation}
This corresponds to the rightmost entry in the SP, $A$ not known row.

For the case of SP and $A$ known, \eqref{eqn:menonSP} simplifies to \citep[Cor.~5]{menon2018}
\begin{equation}
    \hat{Y}^*(X) = \begin{cases}
    \ones\left(r(X) + \lambda/2 > c\right), & A = 0,\\
    \ones\left(r(X) - \lambda/2 > c\right), & A = 1,
    \end{cases}
    \nonumber
\end{equation}
while \eqref{eqn:YhatSP} becomes 
\begin{equation}
    \hat{Y}(X) = \begin{cases}
    \ones\left(r(X) + c(1-c) p_A(1) \left(\frac{\lambda_1}{p_A(1)} - \frac{\lambda_0}{p_A(0)}\right) > c\right), & A = 0,\\
    \ones\left(r(X) - c(1-c) p_A(0) \left(\frac{\lambda_1}{p_A(1)} - \frac{\lambda_0}{p_A(0)}\right) > c\right), & A = 1.
    \end{cases}
    \nonumber
\end{equation}
The above two equations yield the SP, $A$ known row in Table~\ref{tab:menon}.

For the case of EOpp, \citet[Proposition~6]{menon2018} specify the optimal classifier as follows:
\begin{equation}\label{eqn:menonEOpp}
    \hat{Y}^*(X) = \ones\left(\left(1 - \frac{\lambda}{p_Y(1)} (\bar{\eta}(X) - 1/2) \right) r(X) > c\right),
\end{equation}
where now $\bar{\eta}(X) = p_{A\given X,Y}(1\given X,1)$. For the thresholded transformed score in \eqref{eqn:Yhat}, an expression for $\mu(X)$ in the case of EOpp is needed. This is given by the constraint in \eqref{eqn:dualMEO_rhat} restricted to $y=1$:
\[
    \mu(X) = \frac{r(X)}{p_{Y}(1)} \left(\lambda_{0} \left( \frac{1-\bar{\eta}(X)}{p_{A\given Y}(0\given 1)} - 1\right) + \lambda_{1} \left( \frac{\bar{\eta}(X)}{p_{A\given Y}(1\given 1)} - 1\right) \right),
\]
using the definitions of $r(X)$ and $\bar{\eta}(X)$ and dropping the second subscript $y=1$ from $\lambda_{01}$, $\lambda_{11}$. Substituting into \eqref{eqn:Yhat} yields 
\begin{equation}\label{eqn:YhatEOpp}
    \hat{Y}(X) = \ones\left(\left(1 - \frac{c(1-c)}{p_{Y}(1)} \left(\lambda_{0} \left( \frac{1-\bar{\eta}(X)}{p_{A\given Y}(0\given 1)} - 1\right) + \lambda_{1} \left( \frac{\bar{\eta}(X)}{p_{A\given Y}(1\given 1)} - 1\right) \right) \right) r(X) > c\right).
\end{equation}
This establishes the third row in Table~\ref{tab:menon}. 

For the last case of EOpp and $A$ known, \eqref{eqn:menonEOpp} and \eqref{eqn:YhatEOpp} simplify respectively to 
\begin{equation}
    \hat{Y}^*(X) = \begin{cases}
    \ones\left(\left(1 + \frac{\lambda}{2 p_Y(1)} \right) r(X) > c\right), & A = 0,\\
    \ones\left(\left(1 - \frac{\lambda}{2 p_Y(1)} \right) r(X) > c\right), & A = 1,
    \end{cases}
    \nonumber
\end{equation}
\begin{equation}
    \hat{Y}(X) = \begin{cases}
    \ones\left(\left(1 + \frac{c(1-c)}{p_{Y}(1)} p_{A\given Y}(1\given 1) \left(\frac{\lambda_{1}}{p_{A\given Y}(1\given 1)} - \frac{\lambda_{0}}{p_{A\given Y}(0\given 1)} \right) \right) r(X) > c\right), & A = 0,\\
    \ones\left(\left(1 - \frac{c(1-c)}{p_{Y}(1)} p_{A\given Y}(0\given 1) \left(\frac{\lambda_{1}}{p_{A\given Y}(1\given 1)} - \frac{\lambda_{0}}{p_{A\given Y}(0\given 1)} \right) \right) r(X) > c\right), & A = 1.
    \end{cases}
    \nonumber
\end{equation}

\subsection{Proofs for Section~\ref{sec:consistency:implications}}

We prove two implications of the assumptions discussed in Section~\ref{sec:consistency:implications}.

\subsubsection{Proof of Lemma~\ref{lem:chernoffRel}}
\label{sec:consistency:chernoffRel}

\begin{proof}
We prove the lemma for a generic probability $p$, which can be either $p_A(a)$ or $p_{A,Y}(a,y)$, and its empirical estimate $\hat{p}$. First we consider the event 
\[
\frac{p}{\hat{p}} - 1 > \varepsilon \quad \Longleftrightarrow \quad \hat{p} < \frac{p}{1+\varepsilon}
\]
for $\varepsilon > 0$. A version of the Chernoff-Hoeffding theorem bounds the probability of this event as 
\begin{equation}\label{eqn:chernoffRel_1}
\Pr\left(\hat{p} < \frac{p}{1+\varepsilon}\right) \leq \exp\left(-m D_{\mathrm{KL}}\left(\frac{p}{1+\varepsilon} \KLsep[\middle] p\right) \right),
\end{equation}
recalling the definition of Bernoulli KL divergence $D_{\mathrm{KL}}(p\KLsep q)$ in \eqref{eqn:KL}. It can be shown by a second-order Taylor expansion that 
\[
D_{\mathrm{KL}}(p\KLsep q) \geq \frac{(p-q)^2}{2\max\{p,q\}}.
\]
Applying this to \eqref{eqn:chernoffRel_1} yields 
\begin{align*}
    \Pr\left(\hat{p} < \frac{p}{1+\varepsilon}\right) \leq \exp\left(- \frac{m p^2 (1/(1+\varepsilon) - 1)^2}{2p} \right) = \exp\left(-\frac{mp}{2} \left(\frac{\varepsilon}{1+\varepsilon}\right)^2\right).
\end{align*}
Setting the right-hand side equal to $\delta / 2$ and solving for $\varepsilon$,
\begin{align}
\frac{\varepsilon}{1+\varepsilon} &= \sqrt{\frac{2\log(2/\delta)}{mp}},\nonumber\\
\varepsilon &= \frac{\sqrt{2\log(2/\delta)}}{\sqrt{mp} - \sqrt{2\log(2/\delta)}},\label{eqn:chernoffRel_2}
\end{align}
which requires $mp > 2\log(2/\delta)$.

For the other direction 
\[
\frac{p}{\hat{p}} - 1 < -\varepsilon \quad \Longleftrightarrow \quad \hat{p} > \frac{p}{1-\varepsilon},
\]
a similar calculation results in 
\[
\Pr\left(\hat{p} > \frac{p}{1-\varepsilon}\right) \leq \exp\left(-\frac{mp\varepsilon^2}{2(1-\varepsilon)}\right) < \exp\left(-\frac{mp}{2} \left(\frac{\varepsilon}{1+\varepsilon}\right)^2\right).
\]
Hence by taking $\varepsilon$ as in \eqref{eqn:chernoffRel_2}, we have $\abs{(p/\hat{p}) - 1} \leq \varepsilon$ with probability at least $1-\delta$.
\end{proof}

\subsubsection{Proof of Lemma~\ref{lem:KL_L1}}
\label{sec:consistency:KL_L1}

\begin{proof}
Assumption~\ref{ass:rhatKL} means that for every $\epsilon > 0$, we have 
\[
\Pr\left(\EE{D_{\mathrm{KL}}(r(X) \KLsep \hat{r}(X)} \leq \epsilon \right) \to 1,
\]
where the probability is with respect to the random estimator $\hat{r}$. 

Suppose then that $\hat{r}$ satisfies $\EE{D_{\mathrm{KL}}(r(X) \KLsep \hat{r}(X)} \leq 2\epsilon$. It can be shown via a second-order Taylor expansion that $D_{\mathrm{KL}}(r \KLsep \hat{r}) \geq 2(r - \hat{r})^2$ for any $r, \hat{r} \in [0,1]$. Hence we also have $\EE{(r(X) - \hat{r}(X))^2} \leq \epsilon$. Furthermore, by the law of total expectations, 
\begin{align*}
    &\EE{\abs{r(X) - \hat{r}(X)}}\\ 
    &\qquad = \Pr\left(\abs{r(X) - \hat{r}(X)} \leq \sqrt{\epsilon} \right) \EE{\abs{r(X) - \hat{r}(X)} \given[\middle] \abs{r(X) - \hat{r}(X)} \leq \sqrt{\epsilon}}\\
    &\qquad \quad {} + \Pr\left(\abs{r(X) - \hat{r}(X)} > \sqrt{\epsilon} \right) \EE{\abs{r(X) - \hat{r}(X)} \given[\middle] \abs{r(X) - \hat{r}(X)} > \sqrt{\epsilon}}\\
    &\qquad \leq \sqrt{\epsilon} + \Pr\left(\abs{r(X) - \hat{r}(X)} > \sqrt{\epsilon} \right) \EE{\abs{r(X) - \hat{r}(X)} \given[\middle] \abs{r(X) - \hat{r}(X)} > \sqrt{\epsilon}}\\
    &\qquad \leq \sqrt{\epsilon} + \Pr\left(\abs{r(X) - \hat{r}(X)} > \sqrt{\epsilon} \right) \EE{\frac{\abs{r(X) - \hat{r}(X)}^2}{\sqrt{\epsilon}} \given[\middle] \abs{r(X) - \hat{r}(X)} > \sqrt{\epsilon}}\\
    &\qquad \leq \sqrt{\epsilon} + \frac{1}{\sqrt{\epsilon}} \EE{\abs{r(X) - \hat{r}(X)}^2}\\
    &\qquad \leq 2\sqrt{\epsilon},
\end{align*}
where we have used the conditions $\abs{r(X) - \hat{r}(X)} \leq \sqrt{\epsilon}$ and $\abs{r(X) - \hat{r}(X)} > \sqrt{\epsilon}$ to obtain the first and second inequalities above, respectively. The third inequality follows from a similar total expectations decomposition of $\EE{\abs{r(X) - \hat{r}(X)}^2}$ and the last inequality from $\EE{\abs{r(X) - \hat{r}(X)}^2} \leq \epsilon$. 

We have thus shown for any $\epsilon > 0$ that $\EE{D_{\mathrm{KL}}(r(X) \KLsep \hat{r}(X)} \leq 2\epsilon$ implies $\EE{\abs{r(X) - \hat{r}(X)}} \leq 2\sqrt{\epsilon}$. Hence 
\[
\Pr\left(\EE{\abs{r(X) - \hat{r}(X)}} \leq 2\sqrt{\epsilon}\right) \geq \Pr\left(\EE{D_{\mathrm{KL}}(r(X) \KLsep \hat{r}(X)} \leq 2\epsilon \right) \to 1,
\]
as required for Assumption~\ref{ass:rhatL1}.
\end{proof}

\subsection{Proofs for Asymptotic Dual Optimality}
\label{sec:consistency:dualProofs}

This section completes the proof of Theorem~\ref{thm:dualConsistency} (asymptotic dual optimality), as was outlined in Section~\ref{sec:consistency:dual}.

\subsubsection{Proof of Lemma~\ref{lem:Lambda0}}
\label{sec:consistency:Lambda0}

\begin{proof}
We prove the lemma only for the sub-level set of the population dual, $\{\lambda: J(\lambda) \leq J(0)\}$. The argument for the empirical dual $\hat{J}(\lambda)$ is entirely analogous. The inclusion in the $\ell_1$ ball $\Lambda_0$ is proven by showing that the first term $\EE{g(\mu(X); r(X))}$ in $J(\lambda)$ is bounded from below by a constant. The expectation $\EE{g(\mu(X); r(X))}$ is in fact the dual objective function corresponding to a primal problem in which $\epsilon = 0$, i.e.,~perfect fairness is required (zero MSP or GEO difference). By weak duality, $\EE{g(\mu(X); r(X))}$ is lower bounded by the objective value of any primal solution satisfying perfect fairness. The set of constant score functions $r'(X) = r'$ is a family of such solutions since their conditional means do not depend on $A$ or $Y$. The corresponding primal objective value is 
\[
-\EE{H_b(r(X), r')} = \log r' \EE{r(X)} + \log(1-r') \EE{1-r(X)} = m_Y \log r' + (1-m_Y) \log(1-r'),
\]
where $m_Y = \EE{Y} = \EE{r(X)}$ since $r(X) = p_{Y\given X}(1\given X)$. Maximizing this with respect to $r'$ yields 
\begin{equation}\label{eqn:E[g]LB}
\EE{g\bigl(\mu(X); r(X)\bigr)} \geq \max_{r'\in [0,1]} -\EE{H_b(r(X), r')} = -H_b(m_Y, m_Y) \geq -\log 2,
\end{equation}
where the last inequality is due to binary entropy being bounded by $\log 2$.

Now for $\lambda$ such that $J(\lambda) \leq J(0)$, we have
\[
\EE{g\bigl(\mu(X); r(X)\bigr)} + \epsilon \norm{\lambda}_1 \leq \EE{g\bigl(0; r(X)\bigr)} = \EE{-H_b(r(X), r(X))},
\]
using \eqref{eqn:g} and the fact that $r^*(0; r(x)) = r(x)$. Since binary entropy $H_b(r,r)$ is non-negative,
\[
\EE{g\bigl(\mu(X); r(X)\bigr)} + \epsilon \norm{\lambda}_1 \leq 0, \quad \lambda : J(\lambda) \leq J(0).
\]
Combining this with \eqref{eqn:E[g]LB} and dividing by $\epsilon$ (allowed by Assumption~\ref{ass:epsilon}) gives the result.
\end{proof}

\subsubsection{Auxiliary Lemmas}

Here we establish bounds on functions that are used to prove subsequent lemmas.

\begin{lemma}\label{lem:gLipschitz}
The function $g(\mu; r)$ is $1$-Lipschitz in $\mu$ for any fixed $r \in [0,1]$.
\end{lemma}
\begin{proof}
By the mean value theorem, 
\[
\abs*{g(\mu_2; r) - g(\mu_1; r)} = \abs*{\left.\frac{\partial g(\mu; r)}{\partial\mu}\right\rvert_{\bar{\mu}}} \abs{\mu_2 - \mu_1} 
\]
for any $\mu_1 \leq \mu_2$ and some $\bar{\mu} \in [\mu_1, \mu_2]$. From \eqref{eqn:df(tlambda)}, $\abs{\partial g(\mu;r) / \partial\mu} = r^*(\mu; r) \leq 1$ and the result follows.
\end{proof}

\begin{lemma}\label{lem:gUB}
Under Assumptions~\ref{ass:Atrue}, \ref{ass:pAY}, and \ref{ass:epsilon}, 
\[
\abs[\big]{g\bigl(\lambda^T \boldf(x); r(x)\bigr)} \leq \left(1 + \frac{1}{\epsilon} \left(\frac{1}{\eta} - 1\right) \right) \log 2 \defined \bar{G} \quad \forall \, \lambda \in \Lambda_0, \; x \in \cX.
\]
The same bound holds if $\boldf$ is replaced by $\hat{\boldf}$ and $r(x)$ by $\hat{r}(x)$.
\end{lemma}
\begin{proof}
By the triangle inequality and Lemma~\ref{lem:gLipschitz},
\begin{align*}
    \abs[\big]{g\bigl(\lambda^T \boldf(x); r(x)\bigr)} &\leq \abs[\big]{g\bigl(0; r(x)\bigr)} + \abs[\big]{g\bigl(\lambda^T \boldf(x); r(x)\bigr) - g\bigl(0; r(x)\bigr)}\\
    &\leq \abs[\big]{g\bigl(0; r(x)\bigr)} + \abs[\big]{\lambda^T \boldf(x)}.
\end{align*}
As in the proof of Lemma~\ref{lem:Lambda0}, $\abs{g(0; r(x))} = H_b(r(x),r(x)) \leq \log 2$, while 
$\abs{\lambda^T \boldf(x)} \leq \norm{\lambda}_1 \norm{\boldf}_\infty$ from H\"{o}lder's inequality. Thus 
\[
\abs[\big]{g\bigl(\lambda^T \boldf(x); r(x)\bigr)} \leq \log 2 + \norm{\lambda}_1 \norm{\boldf}_\infty,
\]
and combining this with Lemmas~\ref{lem:Lambda0} and \ref{lem:fUB} yields the result. The same proof holds if $\boldf$ is replaced by $\hat{\boldf}$ and $r(x)$ by $\hat{r}(x)$ since Lemma~\ref{lem:fUB} applies equally to $\hat{\boldf}$ and both $r(x), \hat{r}(x) \in [0,1]$.
\end{proof}

\begin{lemma}\label{lem:fUB}
Under Assumptions~\ref{ass:Atrue} and \ref{ass:pAY},
\[
\norm{\boldf(x)}_{\infty} \leq \frac{1}{\eta} - 1, \quad \norm{\hat{\boldf}(x)}_{\infty} \leq \frac{1}{\eta} - 1 \qquad \forall \, x \in \cX.
\]
\end{lemma}
\begin{proof}
For the MSP case,
expression \eqref{eqn:fMSP} implies that 
\[
\norm{\boldf(X)}_{\infty} \leq \max_{a\in\cA} \max \left\{\frac{1}{p_A(a)} - 1, 1 \right\} = \max \left\{ \max_{a\in\cA} \frac{1}{p_A(a)} - 1, 1\right\} \leq \frac{1}{\eta} - 1
\]
by Assumption~\ref{ass:pAY}. 
The same is true for $\hat{\boldf}$.

For the GEO case, we infer from \eqref{eqn:fMEO} that
\begin{align*}
\norm{\boldf(X)}_{\infty} &\leq \max_{a\in\cA,\, r\in[0,1]} \max \left\{ \frac{1-r(X)}{p_Y(0)} \left(\frac{1}{p_{A\given Y}(a\given 0)} - 1\right), \frac{r(X)}{p_Y(1)} \left(\frac{1}{p_{A\given Y}(a\given 1)} - 1\right), \right.\\
&\quad \qquad \qquad \qquad \quad \; \left. \frac{1-r(X)}{p_Y(0)}, \frac{r(X)}{p_Y(1)} \right\}\\
&\leq \max_{a\in\cA} \max \left\{ \frac{1}{p_Y(0)} \left(\frac{1}{p_{A\given Y}(a\given 0)} - 1\right), \frac{1}{p_Y(1)} \left(\frac{1}{p_{A\given Y}(a\given 1)} - 1\right), \frac{1}{p_Y(0)}, \frac{1}{p_Y(1)} \right\}\\
&\leq \max\left\{ \frac{1}{\eta} - \frac{1}{p_Y(0)}, \frac{1}{\eta} - \frac{1}{p_Y(1)} \right\}\\
&\leq \frac{1}{\eta} - 1,
\end{align*}
where the second line results from four separate maximizations over $r$, the third line from applying Assumption~\ref{ass:pAY} 
to the first two terms and using $\abs{\cA} \geq 2$ to drop the last two terms, and the last line from $p_Y(y) \leq 1$. Again the same is true for $\hat{\boldf}$.
\end{proof}

\subsubsection{Proof of Lemma~\ref{lem:dualConsistency1}}
\begin{proof}
The quantity of interest is the supremum over a set $\Lambda_0$ of the difference between an empirical average and expectation of the same function $g\bigl(\lambda^T \hat{\boldf}(A, \hat{r}(X)); \hat{r}(X)\bigr)$, which is a function of parameters $\lambda \in \Lambda_0$ and random variables $A$, $\hat{r}(X)$. This difference is analogous to the difference between empirical and expected risks in statistical learning theory, with $g$ playing the role of the loss function and $\lambda$ the model parameters. The supremum can therefore be bounded using learning theory tools for establishing uniform convergence.

We consider in particular the Rademacher complexity of the function class $\{g\bigl(\lambda^T \hat{\boldf}(A, \hat{r}(X)); \hat{r}(X)\bigr) : \lambda \in \Lambda_0\}$:
\begin{equation}\label{eqn:rademacher}
    R_n(\Lambda_0) = \EE{\sup_{\lambda\in\Lambda_0} \abs*{\frac{1}{n} \sum_{i=1}^n \sigma_i g\bigl(\lambda^T \hat{\boldf}(A_i, \hat{r}(X_i)); \hat{r}(X_i)\bigr)} },
\end{equation}
where $\sigma_i$, $i = 1,\dots,n$ are i.i.d.~Rademacher random variables and the expectation is with respect to $\{\sigma_i, A_i, X_i\}_{i=1}^n$. Using the standard learning theory arguments of McDiarmid's inequality and symmetrization \citep[see e.g.,][and references therein]{liang2016notes,duchinotes}, the supremum of the one-sided difference satisfies 
\begin{multline}\label{eqn:dualConsistency1_1}
    \Pr\left( \sup_{\lambda\in\Lambda_0} \frac{1}{n} \sum_{i=1}^n g\bigl(\lambda^T \hat{\boldf}(a_i, \hat{r}(x_i)); \hat{r}(x_i)\bigr) - \EE{g\bigl(\lambda^T \hat{\boldf}(A, \hat{r}(X)); \hat{r}(X)\bigr)} > 2R_n(\Lambda_0) + \varepsilon \right)\\ 
    \leq \exp\left(-\frac{n\varepsilon^2}{2 \bar{G}^2}, \right),
\end{multline}
where the only difference is that $g$ is bounded by $\bar{G}$ defined in Lemma~\ref{lem:gUB} instead of the unit interval $[0,1]$, and hence $\bar{G}^2$ appears in the exponent above. A similar bound holds for the difference in the other direction. Setting the right-hand side of \eqref{eqn:dualConsistency1_1} equal to $\delta / 2$ and applying the union bound, we therefore have 
\begin{equation}\label{eqn:dualConsistency1_2}
    \sup_{\lambda\in\Lambda_0} \abs*{\frac{1}{n} \sum_{i=1}^n g\bigl(\lambda^T \hat{\boldf}(a_i, \hat{r}(x_i)); \hat{r}(x_i)\bigr) - \EE{g\bigl(\lambda^T \hat{\boldf}(A, \hat{r}(X)); \hat{r}(X)\bigr)}} \leq 2R_n(\Lambda_0) + \bar{G} \sqrt{\frac{2 \log(2/\delta)}{n}}
\end{equation}
with probability at least $1 - \delta$.

The proof is completed by obtaining an upper bound on the Rademacher complexity $R_n(\Lambda_0)$. For this, we consider the \emph{empirical} Rademacher complexity obtained by conditioning on $A_i = a_i$, $\hat{r}(X_i) = \hat{r}_i$ in \eqref{eqn:rademacher}:
\begin{equation*}
    \hat{R}_n(\Lambda_0) = \EE{\sup_{\lambda\in\Lambda_0} \abs*{\frac{1}{n} \sum_{i=1}^n \sigma_i g\bigl(\lambda^T \hat{\boldf}_i; \hat{r}_i\bigr)} },
\end{equation*}
where $\hat{\boldf}_i = \hat{\boldf}(a_i, \hat{r}_i)$ is also fixed. The Rademacher complexity $R_n(\Lambda_0)$ is then the expectation of $\hat{R}_n(\Lambda_0)$ over $\{A_i, \hat{r}(X_i)\}$. First we subtract and add $g(0; \hat{r}_i)$, 
\begin{align*}
    \hat{R}_n(\Lambda_0) &= \EE{\sup_{\lambda\in\Lambda_0} \abs*{\frac{1}{n} \sum_{i=1}^n \sigma_i \left( g\bigl(\lambda^T \hat{\boldf}_i; \hat{r}_i\bigr) - g\bigl(0; \hat{r}_i\bigr) + g\bigl(0; \hat{r}_i\bigr) \right)} },
\end{align*}
and treat $g(0; \hat{r}_i)$ as a $\lambda$-independent translation of the function class. Recalling from the proof of Lemma~\ref{lem:Lambda0} that $\abs{g(0; \hat{r}_i)} = H_b(\hat{r}_i, \hat{r}_i) \leq \log 2$ and using \citet[Thm.~12.5]{bartlett2002rademacher},
\begin{equation}\label{eqn:empRademacher1}
    \hat{R}_n(\Lambda_0) \leq \EE{\sup_{\lambda\in\Lambda_0} \abs*{\frac{1}{n} \sum_{i=1}^n \sigma_i \left( g\bigl(\lambda^T \hat{\boldf}_i; \hat{r}_i\bigr) - g\bigl(0; \hat{r}_i\bigr) \right)}} + \frac{\log 2}{\sqrt{n}}.
\end{equation}
The first term in \eqref{eqn:empRademacher1} is the empirical Rademacher complexity of a composition of functions, $\hat{\boldf}_i \mapsto \lambda^T \hat{\boldf}_i$ and $\mu \mapsto g(\mu; \hat{r}_i) - g(0; \hat{r}_i)$. By Lemma~\ref{lem:gLipschitz}, the second function satisfies the conditions of the Lipschitz composition property of \citet{ledoux1991probability} \citep[see also][Thm.~12.4]{bartlett2002rademacher}, with Lipschitz constant $1$. Applying their lemma results in 
\begin{align*}
    \EE{\sup_{\lambda\in\Lambda_0} \abs*{\frac{1}{n} \sum_{i=1}^n \sigma_i \left( g\bigl(\lambda^T \hat{\boldf}_i; \hat{r}_i\bigr) - g\bigl(0; \hat{r}_i\bigr)\right)}} &\leq 2 \EE{\sup_{\lambda\in\Lambda_0} \abs*{\frac{1}{n} \sum_{i=1}^n \sigma_i \lambda^T \hat{\boldf}_i}}\\ 
    &\leq 2 \EE{\sup_{\lambda\in\mathcal{B}_1((\log 2)/\epsilon)} \abs*{\frac{1}{n} \sum_{i=1}^n \sigma_i \lambda^T \hat{\boldf}_i}},
\end{align*}
where the second inequality is due to $\Lambda_0$ being contained in the $\ell_1$ ball of radius $(\log 2) / \epsilon$, $\mathcal{B}_1((\log 2) / \epsilon)$. Since $\lambda^T \hat{\boldf}_i$ is now linear in $\lambda$, we may use the standard steps of restricting to the vertices of $\mathcal{B}_1((\log 2) / \epsilon)$ and applying Massart's finite class lemma to obtain 
\[
    \EE{\sup_{\lambda\in\Lambda_0} \abs*{\frac{1}{n} \sum_{i=1}^n \sigma_i \left( g\bigl(\lambda^T \hat{\boldf}_i; \hat{r}_i\bigr) - g\bigl(0; \hat{r}_i\bigr)\right)}} \leq 2 \sup_{\lambda\in\mathcal{B}_1((\log 2)/\epsilon)} \abs[\big]{\lambda^T \hat{\boldf}_i} \sqrt{\frac{2\log(2L)}{n}},
\]
with $L = \dim(\lambda)$. Applying H\"{o}lder's inequality $\abs[\big]{\lambda^T \hat{\boldf}_i} \leq \norm{\lambda}_1 \norm[\big]{\hat{\boldf}_i}_\infty$ and Lemma~\ref{lem:fUB}, 
\begin{equation}\label{eqn:empRademacher2}
    \EE{\sup_{\lambda\in\Lambda_0} \abs*{\frac{1}{n} \sum_{i=1}^n \sigma_i \left( g\bigl(\lambda^T \hat{\boldf}_i; \hat{r}_i\bigr) - g\bigl(0; \hat{r}_i\bigr)\right)}} \leq \frac{2\log 2}{\epsilon} \left(\frac{1}{\eta} - 1\right) \sqrt{\frac{2\log(2L)}{n}}.
\end{equation}
Substituting \eqref{eqn:empRademacher2} into \eqref{eqn:empRademacher1} and taking expectations,
\begin{equation}\label{eqn:rademacherUB}
    R_n(\Lambda_0) \leq \frac{2\log 2}{\epsilon} \left(\frac{1}{\eta} - 1\right) \sqrt{\frac{2\log(2L)}{n}} + \frac{\log 2}{\sqrt{n}} < 2\bar{G} \sqrt{\frac{2 \log(2L)}{n}},
\end{equation}
where the last inequality comes from $1 < 2\sqrt{2\log(2L)}$.

Combining \eqref{eqn:dualConsistency1_2} and \eqref{eqn:rademacherUB} yields the result.
\end{proof}

\subsubsection{Proof of Lemma~\ref{lem:dualConsistency3}}

\begin{proof}
We regard $\EE{g\bigl(\lambda^T \hat{\boldf}(A, \hat{r}(X)); \hat{r}(X)\bigr)}$ as the expectation of a function of $A$ and induced random variable $\hat{r}(X)$, and $\EE{g\bigl(\lambda^T \hat{\boldf}(A, r(X)); r(X)\bigr)}$ as the expectation of the same function of $A$ and induced random variable $r(X)$. Lemma~\ref{lem:gUB} asserts that $g(\lambda^T \hat{\boldf}(a, r(x)); r(x))$ is a bounded (and continuous) function for all $\lambda \in \Lambda_0$ and $x \in \cX$. Therefore the convergence in distribution stated in Assumption~\ref{ass:rhatTV} can be used to bound the difference in expectations.

More concretely, we have with probability at least $1-\delta$,
\begin{align*}
    &\sup_{\lambda\in\Lambda_0} \abs*{\EE{g\bigl(\lambda^T \hat{\boldf}(A, \hat{r}(X)); \hat{r}(X)\bigr)} - \EE{g\bigl(\lambda^T \hat{\boldf}(A, r(X)); r(X)\bigr)}}\\
    & = \sup_{\lambda\in\Lambda_0} \abs*{\sum_{a\in\cA} p_A(a) \left( \EE{g\bigl(\lambda^T \hat{\boldf}(a, \hat{r}(X)); \hat{r}(X)\bigr) \given A=a} - \EE{g\bigl(\lambda^T \hat{\boldf}(a, r(X)); r(X)\bigr) \given A=a} \right)}\\
    & \leq \sup_{\lambda\in\Lambda_0} \sum_{a\in\cA} p_A(a) \sup_{x\in\cX} \abs[\big]{g\bigl(\lambda^T \hat{\boldf}(a, r(x)); r(x)\bigr)} D_{\mathrm{TV}}\bigl(\hat{r}(X) \given A=a, r(X) \given A=a\bigr)\\
    & \leq \left(1 + \frac{1}{\epsilon} \left(\frac{1}{\eta} - 1\right) \right) (\log 2) \sum_{a\in\cA} p_A(a) D_{\mathrm{TV}}\bigl(\hat{r}(X) \given A=a, r(X)\given A=a\bigr)\\
    & \leq \left(1 + \frac{1}{\epsilon} \left(\frac{1}{\eta} - 1\right) \right) (\log 2) E_{\mathrm{TV}}(m, \delta).
\end{align*}
The first inequality bounds the difference in expectations by the product of the total variation distance and the supremum of the function $g$. We then apply Lemma~\ref{lem:gUB} to obtain the second inequality and invoke Assumption~\ref{ass:rhatTV}.
\end{proof}

\subsubsection{Proof of Lemma~\ref{lem:dualConsistency2}}

\begin{proof}
Let us first consider a single point $x \in \cX$. Using the Lipschitz property of $g$ in Lemma~\ref{lem:gLipschitz},
\[
\abs[\big]{g\bigl(\lambda^T \hat{\boldf}(a, r(x)); r(x)\bigr) - g\bigl(\lambda^T \boldf(a, r(x)); r(x)\bigr)} \leq \abs*{\lambda^T \bigl(\hat{\boldf}(a, r(x)) - \boldf(a, r(x))\bigr)}.
\]
H\"{o}lder's inequality then gives 
\begin{equation}\label{eqn:dualConsistency2_1}
\abs[\big]{g\bigl(\lambda^T \hat{\boldf}(a, r(x)); r(x)\bigr) - g\bigl(\lambda^T \boldf(a, r(x)); r(x)\bigr)} \leq \norm{\lambda}_1 \norm[\big]{\hat{\boldf}(a, r(x)) - \boldf(a, r(x))}_{\infty}.
\end{equation}
Now for the expectations over $\cX$, it follows from the triangle inequality, \eqref{eqn:dualConsistency2_1}, and Lemma~\ref{lem:Lambda0} that 
\begin{align}
&\sup_{\lambda\in\Lambda_0} \abs*{\EE{g\bigl(\lambda^T \hat{\boldf}(A, r(X)); r(X)\bigr)} - \EE{g\bigl(\lambda^T \boldf(A, r(X)); r(X)\bigr)}}\nonumber\\
&\qquad \leq \sup_{\lambda\in\Lambda_0} \EE{\abs*{g\bigl(\lambda^T \hat{\boldf}(A, r(X)); r(X)\bigr) - g\bigl(\lambda^T \boldf(A, r(X)); r(X)\bigr)}}\nonumber\\
&\qquad \leq \sup_{\lambda\in\Lambda_0} \norm{\lambda}_1 \EE{\norm[\big]{\hat{\boldf}(A, r(X)) - \boldf(A, r(X))}_{\infty}}\nonumber\\
&\qquad \leq \frac{\log 2}{\epsilon} \EE{\norm[\big]{\hat{\boldf}(A,r(X)) - \boldf(A,r(X))}_{\infty}}.\label{eqn:dualConsistency2_2}
\end{align}

Next we obtain bounds on the expected $\ell_\infty$ norm in \eqref{eqn:dualConsistency2_2}, considering the MSP and GEO cases separately. In the former case, from \eqref{eqn:fMSP} we obtain
\[
\norm[\big]{\hat{\boldf}(A,r(X)) - \boldf(A,r(X))}_{\infty} = \max_{a\in\cA} \ones(A=a) \abs*{\frac{1}{\hat{p}_A(a)} - \frac{1}{p_A(a)}} = \abs*{\frac{1}{\hat{p}_A(A)} - \frac{1}{p_A(A)}}.
\]
Hence 
\begin{align*}
\EE{\norm[\big]{\hat{\boldf}(A,r(X)) - \boldf(A,r(X))}_{\infty}} &= \sum_{a\in\cA} p_A(a) \abs*{\frac{1}{\hat{p}_A(a)} - \frac{1}{p_A(a)}}\\ 
&= \sum_{a\in\cA} \abs*{\frac{p_A(a)}{\hat{p}_A(a)} - 1}.
\end{align*}
Applying Lemma~\ref{lem:chernoffRel} to each term, together with a union bound over $a \in \cA$, gives 
\begin{equation}\label{eqn:dualConsistency2_MSP}
\EE{\norm[\big]{\hat{\boldf}(A,r(X)) - \boldf(A,r(X))}_{\infty}} \leq \sum_{a\in\cA} \frac{\sqrt{2 \log(2L/\delta)}}{\sqrt{m p_A(a)} - \sqrt{2 \log(2L/\delta)}}
\end{equation}
with probability at least $1-\delta$, noting that $\abs{\cA} = L$ here.

For the GEO case, we use \eqref{eqn:fMEO}, the definition $r(X) = p_{Y\given X}(1\given X)$, and the triangle inequality to obtain 
\begin{align*}
    &\norm[\big]{\hat{\boldf}(A,r(X)) - \boldf(A,r(X))}_{\infty}\\ 
    &= \max_{y\in\{0,1\}} p_{Y\given X}(y\given X) \max_{a\in\cA} \abs*{\ones(A=a) \left(\frac{1}{\hat{p}_{A,Y}(a,y)} - \frac{1}{p_{A,Y}(a,y)}\right) - \left(\frac{1}{\hat{p}_Y(y)} - \frac{1}{p_Y(y)}\right)}\\
    &\leq \max_{y\in\{0,1\}} p_{Y\given X}(y\given X) \left( \abs*{\frac{1}{\hat{p}_{A,Y}(A,y)} - \frac{1}{p_{A,Y}(A,y)}} + \abs*{\frac{1}{\hat{p}_Y(y)} - \frac{1}{p_Y(y)}} \right)\\
    &\leq \sum_{y\in\{0,1\}} p_{Y\given X}(y\given X) \left( \abs*{\frac{1}{\hat{p}_{A,Y}(A,y)} - \frac{1}{p_{A,Y}(A,y)}} + \abs*{\frac{1}{\hat{p}_Y(y)} - \frac{1}{p_Y(y)}} \right).
\end{align*}
Taking expectations with respect to $X$ (which includes $A$),
\begin{align*}
    \EE{\norm[\big]{\hat{\boldf}(A,r(X)) - \boldf(A,r(X))}_{\infty}} &\leq \sum_{y\in\{0,1\}} \sum_{a\in\cA} p_{A,Y}(a,y) \abs*{\frac{1}{\hat{p}_{A,Y}(a,y)} - \frac{1}{p_{A,Y}(a,y)}}\\
    &\qquad {} + \sum_{y\in\{0,1\}} p_Y(y) \abs*{\frac{1}{\hat{p}_Y(y)} - \frac{1}{p_Y(y)}}\\
    &= \sum_{y\in\{0,1\}} \sum_{a\in\cA} \abs*{\frac{p_{A,Y}(a,y)}{\hat{p}_{A,Y}(a,y)} - 1} + \sum_{y\in\{0,1\}} \abs*{\frac{p_Y(y)}{\hat{p}_Y(y)} - 1}.
\end{align*}
The final right-hand side is a sum of $2(\abs{\cA} + 1) = L + 2$ terms of the form in Lemma~\ref{lem:chernoffRel}. Applying this lemma and a union bound over the $L+2$ events, we have with probability at least $1-\delta$,
\begin{align}
    \EE{\norm[\big]{\hat{\boldf}(A,r(X)) - \boldf(A,r(X))}_{\infty}} \leq &\sum_{y\in\{0,1\}} \sum_{a\in\cA} \frac{\sqrt{2 \log(2(L+2)/\delta)}}{\sqrt{m p_{A,Y}(a,y)} - \sqrt{2 \log(2(L+2)/\delta)}}\nonumber\\
    &{} + \sum_{y\in\{0,1\}} \frac{\sqrt{2 \log(2(L+2)/\delta)}}{\sqrt{m p_{Y}(y)} - \sqrt{2 \log(2(L+2)/\delta)}}.\label{eqn:dualConsistency2_MEO}
\end{align}

The proof is completed by substituting \eqref{eqn:dualConsistency2_MSP} or \eqref{eqn:dualConsistency2_MEO} into \eqref{eqn:dualConsistency2_2}.
\end{proof}

\subsection{Proofs for Asymptotic Primal Feasibility}
\label{sec:consistency:primalFeasProofs}

This section completes the proof of Theorem~\ref{thm:primalFeas} (asymptotic primal feasibility), following the outline in Section~\ref{sec:consistency:primalFeas}.

\subsubsection{Proof of Lemma~\ref{lem:fairConsistency0}}

\begin{proof}
We follow the same steps and use the same learning theory tools as in the proof of Lemma~\ref{lem:dualConsistency1}. The function class is now 
\[
\left\{\hat{f}_l(A, \hat{r}(X)) r^*\bigl(\lambda^T \hat{\boldf}(A, \hat{r}(X)); \hat{r}(X)\bigr): \lambda \in \Lambda_0\right\}
\]
and we consider its Rademacher complexity, again denoted by $R_n(\Lambda_0)$. Below we note the differences with respect to the proof of Lemma~\ref{lem:dualConsistency1}.
\begin{enumerate}
    \item Using Lemma~\ref{lem:fUB} and the fact that $\abs{r^*(\cdot)} \leq 1$, the class of functions is uniformly bounded by $(1/\eta) - 1$ in absolute value. Thus $(1/\eta) - 1$ replaces $\bar{G}$ in the last term in \eqref{eqn:dualConsistency1_2}.
    \item Considering the empirical Rademacher complexity fixes the factor $\hat{f}_l(a_i, \hat{r}(x_i))$. We again use Lemma~\ref{lem:fUB} to bound this factor by $(1/\eta) - 1$. We then focus on determining the empirical Rademacher complexity of $\bigl\{r^*\bigl(\lambda^T \hat{\boldf}(A, \hat{r}(X)); \hat{r}(X)\bigr): \lambda \in \Lambda_0\bigr\}$, scaled by $(1/\eta) - 1$.
    \item The translating function that is added and subtracted is $r^*(0; \hat{r}_i)$ instead of $g(0; \hat{r}_i)$. Since $\abs{r^*(0; \hat{r}_i)} = \abs{\hat{r}_i} \leq 1$, the last $(\log 2) / \sqrt{n}$ term in \eqref{eqn:empRademacher1} is replaced by $1/\sqrt{n}$ (which will be scaled by $(1/\eta) - 1$).
    \item From \eqref{eqn:d2f(tlambda)}, it can be verified that $\abs{\partial r^*(\mu; \hat{r}) / \partial\mu} \leq 1$. Hence, like $g(\mu; \hat{r})$, $r^*(\mu; \hat{r})$ is $1$-Lipschitz in $\mu$. 
    \item After using the Lipschitz composition property, the resulting class of $\ell_1$-bounded linear functions is the same as in the proof of Lemma~\ref{lem:dualConsistency1}.
\end{enumerate}

In summary, we obtain 
\[
    R_n(\Lambda_0) \leq \left(\frac{1}{\eta} - 1\right) \left(\frac{2\log 2}{\epsilon} \left(\frac{1}{\eta} - 1\right) \sqrt{\frac{2\log(2L)}{n}} + \frac{1}{\sqrt{n}} \right)
\]
and with probability at least $1-\delta$,
\begin{align*}
    &\sup_{\lambda\in\Lambda_0} \abs*{\frac{1}{n} \sum_{i=1}^n \hat{f}_l(a_i, \hat{r}(x_i)) r^*\bigl(\lambda^T \hat{\boldf}(a_i, \hat{r}(x_i)); \hat{r}(x_i)\bigr) - \EE{\hat{f}_l(A, \hat{r}(X)) r^*\bigl(\lambda^T \hat{\boldf}(A, \hat{r}(X)); \hat{r}(X)\bigr)}}\\
    &\qquad \leq \left(\frac{1}{\eta} - 1\right) \left( \frac{4\log 2}{\epsilon} \left(\frac{1}{\eta} - 1\right) \sqrt{\frac{2\log(2L)}{n}} + \frac{2}{\sqrt{n}} + \sqrt{\frac{2\log(2/\delta)}{n}} \right).
\end{align*}
The proof is completed by dividing the probability $\delta$ over $l = 1,\dots,L$ and applying the union bound.
\end{proof}

\subsubsection{Proof of Lemma~\ref{lem:fairConsistency1}}

\begin{proof}
In the case of GEO, the index $l = (a,y)$ for $a \in \cA$ and $y \in \{0,1\}$. Using \eqref{eqn:fMEO}, the quantity of interest is given by 
\begin{align*}
    &\abs*{\EE{\hat{f}_{a,y}(A, \hat{r}(X)) r'(X)} - \EE{\hat{f}_{a,y}(A, r(X)) r'(X)}}\\ 
    &\qquad\qquad = \abs*{\EE{(r(X) - \hat{r}(X)) \left(\frac{\ones(A=a)}{\hat{p}_{A,Y}(a,y)} - \frac{1}{\hat{p}_Y(y)} \right) r'(X)}}\\
    &\qquad\qquad \leq \EE{\abs[\big]{r(X) - \hat{r}(X)} \abs*{\frac{\ones(A=a)}{\hat{p}_{A,Y}(a,y)} - \frac{1}{\hat{p}_Y(y)}} r'(X)}.
\end{align*}
Similar to the proof of Lemma~\ref{lem:fUB}, we have 
\[
\abs*{\frac{\ones(A=a)}{\hat{p}_{A,Y}(a,y)} - \frac{1}{\hat{p}_Y(y)}} \leq \max_{a\in\cA, y\in\{0,1\}} \max\left\{\frac{1}{\hat{p}_{A,Y}(a,y)} - \frac{1}{\hat{p}_Y(y)}, \frac{1}{\hat{p}_Y(y)} \right\} \leq \frac{1}{\eta} - 1,
\]
and $\abs{r'(X)} \leq 1$. Hence we obtain the further bounds 
\begin{align*}
\abs*{\EE{\hat{f}_{a,y}(A, \hat{r}(X)) r'(X)} - \EE{\hat{f}_{a,y}(A, r(X)) r'(X)}} &\leq \left(\frac{1}{\eta} - 1\right) \EE{\abs[\big]{r(X) - \hat{r}(X)}}\\
&\leq \left(\frac{1}{\eta} - 1\right) E_{L_1}(m, \delta),
\end{align*}
where the last bound holds with probability $1-\delta$ by Assumption~\ref{ass:rhatL1}.
\end{proof}

\subsubsection{Proof of Lemma~\ref{lem:fairConsistency2}}

\begin{proof}
As in Lemma~\ref{lem:fairConsistency1}, the index $l = (a,y)$ for $a \in \cA$ and $y \in \{0,1\}$. We prove the lemma for $y = 0$; the case $y = 1$ is similar. 

Using \eqref{eqn:fMEO}, the quantity of interest is given by 
\begin{multline*}
    \abs*{\EE{\hat{f}_{a,0}(A, r(X)) r'(X)} - \EE{f_{a,0}(A, r(X)) r'(X)}}\\
    = \abs*{\EE{(1-r(X)) \left(\frac{\ones(A=a)}{\hat{p}_{A,Y}(a,0)} - \frac{\ones(A=a)}{p_{A,Y}(a,0)} - \frac{1}{\hat{p}_Y(0)} + \frac{1}{p_Y(0)}\right) r'(X)}}.
\end{multline*}
Recalling that $1 - r(x) = p_{Y\given X}(0\given x)$ and using Bayes' rule $p_{Y\given X}(0\given x) p_X(x) = p_Y(0) p_{X\given Y}(x\given 0)$, this can be rewritten as 
\begin{align}
    &\abs*{\EE{\hat{f}_{a,0}(A, r(X)) r'(X)} - \EE{f_{a,0}(A, r(X)) r'(X)}}\\ 
    & = \abs*{\EE{p_Y(0) \left(\frac{\ones(A=a)}{\hat{p}_{A,Y}(a,0)} - \frac{\ones(A=a)}{p_{A,Y}(a,0)} - \frac{1}{\hat{p}_Y(0)} + \frac{1}{p_Y(0)}\right) r'(X) \given[\middle] Y=0}}\nonumber\\
    & \leq \abs*{\EE{p_Y(0) \left(\frac{\ones(A=a)}{\hat{p}_{A,Y}(a,0)} - \frac{\ones(A=a)}{p_{A,Y}(a,0)}\right) r'(X) \given[\middle] Y=0}} + \abs*{\EE{\left(1 - \frac{p_Y(0)}{\hat{p}_Y(0)}\right) r'(X) \given[\middle] Y=0}}.\label{eqn:fairConsistency2_1}
\end{align}
The second term in \eqref{eqn:fairConsistency2_1} is bounded similarly as in the proof of Theorem~\ref{thm:primalFeas} for the MSP case:
\begin{align}
    \abs*{\EE{\left(1 - \frac{p_Y(0)}{\hat{p}_Y(0)}\right) r'(X) \given[\middle] Y=0}} &= \abs*{1 - \frac{p_Y(0)}{\hat{p}_Y(0)}} \abs*{\EE{r'(X) \given Y=0}}\nonumber\\
    &\leq \abs*{1 - \frac{p_Y(0)}{\hat{p}_Y(0)}}\nonumber\\
    &\leq \frac{\sqrt{2 \log(2/\delta)}}{\sqrt{m p_Y(0)} - \sqrt{2 \log(2/\delta)}},\label{eqn:fairConsistency2_2}
\end{align}
using
$\abs{r'(X)} \leq 1$ and Lemma~\ref{lem:chernoffRel} (with probability $1-\delta$). The first term in \eqref{eqn:fairConsistency2_1} is also similar: 
\begin{align}
    &\abs*{\EE{p_Y(0) \left(\frac{\ones(A=a)}{\hat{p}_{A,Y}(a,0)} - \frac{\ones(A=a)}{p_{A,Y}(a,0)}\right) r'(X) \given[\middle] Y=0}}\nonumber\\
    &\qquad = \abs*{\EE{p_Y(0) p_{A\given Y}(a\given 0) \left(\frac{1}{\hat{p}_{A,Y}(a,0)} - \frac{1}{p_{A,Y}(a,0)}\right) r'(X) \given[\middle] A=a, Y=0}}\nonumber\\
    &\qquad = \abs*{\frac{p_{A,Y}(a,0)}{\hat{p}_{A,Y}(a,0)} - 1} \abs*{\EE{r'(X) \given A=a, Y=0}}\nonumber\\
    &\qquad \leq \abs*{\frac{p_{A,Y}(a,0)}{\hat{p}_{A,Y}(a,0)} - 1}\nonumber\\
    &\qquad \leq \frac{\sqrt{2 \log(2/\delta)}}{\sqrt{m p_{A,Y}(a,0)} - \sqrt{2 \log(2/\delta)}} ,\label{eqn:fairConsistency2_3}
\end{align}
again using 
$\abs{r'(X)} \leq 1$ and Lemma~\ref{lem:chernoffRel}. Considering all $a \in \cA$ and $y \in \{0,1\}$, there are $2\abs{\cA} + 2 = L+2$ deviations to control in \eqref{eqn:fairConsistency2_2}, \eqref{eqn:fairConsistency2_3}, and hence we divide the probability $\delta$ by $L+2$ and apply the union bound. The result then follows from substituting \eqref{eqn:fairConsistency2_2}, \eqref{eqn:fairConsistency2_3} into \eqref{eqn:fairConsistency2_1}.
\end{proof}

\subsection{Proofs for Asymptotic Primal Optimality}
\label{sec:consistency:primalOptProofs}

We now complete the proof of Theorem~\ref{thm:primalOpt} as outlined in Section~\ref{sec:consistency:primalOpt}.

\subsubsection{Proof of Lemma~\ref{lem:primalConsistency1}}

\begin{proof}
Let $\hat{\mu}(x) = \hat{\lambda}^T \hat{\boldf}(a, \hat{r}(x))$. Using the definition of binary cross-entropy in \eqref{eqn:crossentropy}, the difference in question can be written as 
\begin{equation}\label{eqn:primalConsistency1_1}
\abs*{\EE{r(X) \log\frac{r^*\bigl(\hat{\mu}(X); r(X)\bigr)}{r^*\bigl(\hat{\mu}(X); \hat{r}(X)\bigr)} + (1-r(X)) \log\frac{1 - r^*\bigl(\hat{\mu}(X); r(X)\bigr)}{1 - r^*\bigl(\hat{\mu}(X); \hat{r}(X)\bigr)}} }.
\end{equation}
For $\hat{\mu}(X) = 0$, $r^*$ is an identity mapping and \eqref{eqn:primalConsistency1_1} reduces to 
\[
\abs*{\EE{r(X) \log\frac{r(X)}{\hat{r}(X)} + (1-r(X)) \log\frac{1-r(X)}{1-\hat{r}(X)}}} = \EE{D_{\mathrm{KL}}(r(X) \KLsep \hat{r}(X))}.
\]
Hence Assumption~\ref{ass:rhatKL} is a necessary condition for \eqref{eqn:primalConsistency1_1} to converge to zero in probability. We will show that Assumption~\ref{ass:rhatKL} is sufficient as well. 

Toward this end, we apply the triangle inequality to bound \eqref{eqn:primalConsistency1_1} as follows:
\begin{align*}
    &\abs*{\EE{r(X) \log\frac{r^*\bigl(\hat{\mu}(X); r(X)\bigr)}{r^*\bigl(\hat{\mu}(X); \hat{r}(X)\bigr)} + (1-r(X)) \log\frac{1 - r^*\bigl(\hat{\mu}(X); r(X)\bigr)}{1 - r^*\bigl(\hat{\mu}(X); \hat{r}(X)\bigr)}} }\\
    &\qquad \leq \EE{\abs*{ r(X) \log\frac{r^*\bigl(\hat{\mu}(X); r(X)\bigr)}{r^*\bigl(\hat{\mu}(X); \hat{r}(X)\bigr)} + (1-r(X)) \log\frac{1 - r^*\bigl(\hat{\mu}(X); r(X)\bigr)}{1 - r^*\bigl(\hat{\mu}(X); \hat{r}(X)\bigr)} }}\\
    &\qquad \leq \EE{r(X) \abs*{\log\frac{r^*\bigl(\hat{\mu}(X); r(X)\bigr)}{r^*\bigl(\hat{\mu}(X); \hat{r}(X)\bigr)}} + (1-r(X)) \abs*{\log\frac{1 - r^*\bigl(\hat{\mu}(X); r(X)\bigr)}{1 - r^*\bigl(\hat{\mu}(X); \hat{r}(X)\bigr)}} }\\
    &\qquad \leq \EE{r(X) \sup_{\mu} \, \abs*{\log\frac{r^*\bigl(\mu; r(X)\bigr)}{r^*\bigl(\mu; \hat{r}(X)\bigr)}} + (1-r(X)) \sup_{\mu} \, \abs*{\log\frac{1 - r^*\bigl(\mu; r(X)\bigr)}{1 - r^*\bigl(\mu; \hat{r}(X)\bigr)}} }.
\end{align*}
The last inequality results from replacing $\hat{\mu}(X)$ with the supremum over $\mu$ for a given $r(x)$, $\hat{r}(x)$. Applying Lemma~\ref{lem:primalConsistency3} and defining 
\[
\bar{D}(r \KLsep \hat{r}) = r \abs*{\log\frac{1 - \sqrt{1 - r}}{1 - \sqrt{1 - \hat{r}}}} + (1-r) \abs*{\log\frac{1 - \sqrt{r}}{1 - \sqrt{\hat{r}}}}
\]
results in  
\begin{equation}\label{eqn:primalConsistency1_2}
    \abs*{\EE{r(X) \log\frac{r^*\bigl(\hat{\mu}(X); r(X)\bigr)}{r^*\bigl(\hat{\mu}(X); \hat{r}(X)\bigr)} + (1-r(X)) \log\frac{1 - r^*\bigl(\hat{\mu}(X); r(X)\bigr)}{1 - r^*\bigl(\hat{\mu}(X); \hat{r}(X)\bigr)}} }
    \leq \EE{\bar{D}(r(X) \KLsep \hat{r}(X))}.
\end{equation}
We have thus eliminated $\hat{\mu}(X)$ from the expectation.

We proceed to bound the right-hand side of \eqref{eqn:primalConsistency1_2}. Using the law of total expectations, for any $\epsilon > 0$,
\begin{align*}
    &\EE{\bar{D}(r(X) \KLsep \hat{r}(X))}\\
    &\qquad = \Pr\bigl(D_{\mathrm{KL}}(r(X) \KLsep \hat{r}(X)) \leq \epsilon\bigr) \EE{\bar{D}(r(X) \KLsep \hat{r}(X)) \given[\middle] D_{\mathrm{KL}}(r(X) \KLsep \hat{r}(X)) \leq \epsilon}\\
    &\qquad \quad {} + \Pr\bigl(D_{\mathrm{KL}}(r(X) \KLsep \hat{r}(X)) > \epsilon\bigr) \EE{\bar{D}(r(X) \KLsep \hat{r}(X)) \given[\middle] D_{\mathrm{KL}}(r(X) \KLsep \hat{r}(X)) > \epsilon}\\
    &\qquad \leq \sup_{r, \hat{r}}\left\{ \bar{D}(r \KLsep \hat{r}) : D_{\mathrm{KL}}(r \KLsep \hat{r}) \leq \epsilon \right\}\\
    &\qquad \quad {} + \Pr\bigl(D_{\mathrm{KL}}(r(X) \KLsep \hat{r}(X)) > \epsilon\bigr) \EE{\bar{D}(r(X) \KLsep \hat{r}(X)) \given[\middle] D_{\mathrm{KL}}(r(X) \KLsep \hat{r}(X)) > \epsilon},
\end{align*}
where we have bounded the first expectation by the conditional supremum. Applying Lemma~\ref{lem:primalConsistency5} to the second expectation, 
\begin{align}
    &\EE{\bar{D}(r(X) \KLsep \hat{r}(X))} \nonumber\\
    &\qquad \leq \sup_{r, \hat{r}}\left\{ \bar{D}(r \KLsep \hat{r}) : D_{\mathrm{KL}}(r \KLsep \hat{r}) \leq \epsilon \right\} + 2 \Pr\bigl(D_{\mathrm{KL}}(r(X) \KLsep \hat{r}(X)) > \epsilon\bigr)\nonumber\\
    &\qquad \quad {} + \Pr\bigl(D_{\mathrm{KL}}(r(X) \KLsep \hat{r}(X)) > \epsilon\bigr) \EE{D_{\mathrm{KL}}(r(X) \KLsep \hat{r}(X)) \given[\middle] D_{\mathrm{KL}}(r(X) \KLsep \hat{r}(X)) > \epsilon}\nonumber\\
    &\qquad \leq \sup_{r, \hat{r}}\left\{ \bar{D}(r \KLsep \hat{r}) : D_{\mathrm{KL}}(r \KLsep \hat{r}) \leq \epsilon \right\} + 2 \Pr\bigl(D_{\mathrm{KL}}(r(X) \KLsep \hat{r}(X)) > \epsilon\bigr)\nonumber\\
    &\qquad \quad {} + \EE{D_{\mathrm{KL}}(r(X) \KLsep \hat{r}(X))}\nonumber\\
    &\qquad \leq \sup_{r, \hat{r}}\left\{ \bar{D}(r \KLsep \hat{r}) : D_{\mathrm{KL}}(r \KLsep \hat{r}) \leq \epsilon \right\} + \left(1 + \frac{2}{\epsilon}\right) \EE{D_{\mathrm{KL}}(r(X) \KLsep \hat{r}(X))}.\label{eqn:primalConsistency1_3}
\end{align}
The second inequality above is implied by a similar application of total expectation to $\EE{D_{\mathrm{KL}}(r(X) \KLsep \hat{r}(X))}$, and the third inequality is due to Markov's inequality. By Assumption~\ref{ass:rhatKL}, the last term in \eqref{eqn:primalConsistency1_3} converges to zero in probability for any $\epsilon > 0$. 

We now take $\epsilon \to 0$ and argue that the first right-hand side term in \eqref{eqn:primalConsistency1_3} also converges to zero. This is because the condition $D_{\mathrm{KL}}(r \KLsep \hat{r}) \leq \epsilon$ excludes the cases $\hat{r} = 0$ unless $r = 0$, and $\hat{r} = 1$ unless $r = 1$, which would cause $\bar{D}(r \KLsep \hat{r})$ to diverge. $\bar{D}(r \KLsep \hat{r})$ is therefore bounded and continuous on the set $\{(r, \hat{r}) : D_{\mathrm{KL}}(r \KLsep \hat{r}) \leq \epsilon\}$. As $\epsilon \to 0$, this set shrinks toward the line $r = \hat{r}$ where $\bar{D}(r \KLsep \hat{r}) = 0$. The lemma is thus proven by combining \eqref{eqn:primalConsistency1_2}, \eqref{eqn:primalConsistency1_3} and taking $\epsilon \to 0$.
\end{proof}

\subsubsection{Auxiliary Lemmas for Lemma~\ref{lem:primalConsistency1}}

\begin{lemma}\label{lem:primalConsistency3}
For $r^*(\mu; r)$ defined in \eqref{eqn:r*},
\begin{align*}
\sup_{\mu} \, \abs*{\log\frac{r^*(\mu; r)}{r^*(\mu; \hat{r})}} &= \abs*{\log\frac{1 - \sqrt{1 - r}}{1 - \sqrt{1 - \hat{r}}}},\\
\sup_{\mu} \, \abs*{\log\frac{1 - r^*(\mu; r)}{1 - r^*(\mu; \hat{r})}} &= \abs*{\log\frac{1 - \sqrt{r}}{1 - \sqrt{\hat{r}}}}.
\end{align*}
\end{lemma}
\begin{proof}
We prove the first identity. From \eqref{eqn:r*} we have 
\begin{align*}
    \frac{\partial}{\partial\mu} \log r^*(\mu; r) &= \frac{1}{1 + \mu - \sqrt{(1+\mu)^2 - 4r\mu}} \left(1 - \frac{1+\mu-2r}{\sqrt{(1+\mu)^2 - 4r\mu}} \right) - \frac{1}{\mu}\\
    &= \frac{\sqrt{(1+\mu)^2 - 4r\mu} - (1+\mu) + 2r}{1 + \mu - \sqrt{(1+\mu)^2 - 4r\mu}} \frac{1}{\sqrt{(1+\mu)^2 - 4r\mu}} - \frac{1}{\mu}\\
    &= \frac{-4r\mu + 2r\left(1 + \mu + \sqrt{(1+\mu)^2 - 4r\mu}\right)}{4r\mu \sqrt{(1+\mu)^2 - 4r\mu}} - \frac{1}{\mu}\\
    &= \frac{1-\mu}{2\mu \sqrt{(1+\mu)^2 - 4r\mu}} - \frac{1}{2\mu},
\end{align*}
where the third equality comes from multiplying numerator and denominator by $1 + \mu + \sqrt{(1+\mu)^2 - 4r\mu}$ and using the identity $(a-b)(a+b) = a^2 - b^2$. Hence
\begin{align}
    \frac{\partial}{\partial\mu} \log\frac{r^*(\mu; r)}{r^*(\mu; \hat{r})} &= \frac{1-\mu}{2\mu} \left(\frac{1}{\sqrt{(1+\mu)^2 - 4r\mu}} - \frac{1}{\sqrt{(1+\mu)^2 - 4\hat{r}\mu}} \right)\nonumber\\
    &= \frac{1-\mu}{2\mu} \left( \frac{\sqrt{(1+\mu)^2 - 4\hat{r}\mu} - \sqrt{(1+\mu)^2 - 4r\mu}}{\sqrt{(1+\mu)^2 - 4r\mu} \sqrt{(1+\mu)^2 - 4\hat{r}\mu}} \right)\nonumber\\
    &= \frac{(1-\mu) \bigl(r^*(\mu; r) - r^*(\mu; \hat{r})\bigr)}{\sqrt{(1+\mu)^2 - 4r\mu} \sqrt{(1+\mu)^2 - 4\hat{r}\mu}},\label{eqn:primalConsistency3_1}
\end{align}
using the definition of $r^*(\mu; r)$ \eqref{eqn:r*} in the last line. 

We now consider three cases: (1) $r = \hat{r}$, (2) $r > \hat{r}$, and (3) $r < \hat{r}$. (1) If $r = \hat{r}$, then $\log(r^*(\mu; r)/r^*(\mu;\hat{r})) = 0$ for all $\mu$ and the identity is true. (2) For $r > \hat{r}$, Lemma~\ref{clm:monotonic} implies that $r^*(\mu; r) > r^*(\mu; \hat{r})$ also. It follows that $\log(r^*(\mu; r)/r^*(\mu;\hat{r}))$ is positive for all $\mu$. Furthermore from \eqref{eqn:primalConsistency3_1}, $\log(r^*(\mu; r)/r^*(\mu;\hat{r}))$ increases with $\mu$ for $\mu < 1$ and decreases for $\mu > 1$. The maximum therefore occurs at $\mu = 1$, and the substitution of $r^*(1; r) = 1 - \sqrt{1 - r}$ yields the desired identity. (3) For $r < \hat{r}$, the same arguments show that $\log(r^*(\mu; r)/r^*(\mu;\hat{r}))$ is negative for all $\mu$, decreases with $\mu$ for $\mu < 1$, and increases for $\mu > 1$. The maximum absolute value occurs therefore at $\mu = 1$ as well.

The proof of the second identity in the lemma statement is analogous by symmetry and shows that the maximizing value is $\mu = -1$.
\end{proof}

\begin{lemma}\label{lem:primalConsistency4}
For $r, \hat{r} \in [0,1]$, 
\begin{align*}
    -0.6140 &\leq r \log \frac{1 - \sqrt{1 - r}}{1 - \sqrt{1 - \hat{r}}} \leq r \log \frac{r}{\hat{r}} + \log 2,\\
    -0.6140 &\leq (1-r) \log \frac{1 - \sqrt{r}}{1 - \sqrt{\hat{r}}} \leq (1-r) \log \frac{1-r}{1-\hat{r}} + \log 2.
\end{align*}
\end{lemma}
\begin{proof}
As with the proof of Lemma~\ref{lem:primalConsistency3}, we prove only the first line of inequalities. The second line follows by symmetry.

To obtain the lower bound, we observe that the quantity of interest is minimized for any $r \in [0,1]$ by taking $\hat{r} = 1$. Numerical minimization of the resulting quantity $r \log(1 - \sqrt{1 - r})$ over $r \in [0,1]$ then yields the lower bound.

To obtain the upper bound, we maximize the quantity 
\begin{equation}\label{eqn:primalConsistency4_1}
r \log \frac{1 - \sqrt{1 - r}}{1 - \sqrt{1 - \hat{r}}} - r \log \frac{r}{\hat{r}} = r \log\left(\frac{1 - \sqrt{1 - r}}{r} \frac{\hat{r}}{1 - \sqrt{1 - \hat{r}}}\right).
\end{equation}
It can be verified that 
\begin{equation}\label{eqn:primalConsistency4_2}
\frac{d}{d\hat{r}} \frac{\hat{r}}{1 - \sqrt{1 - \hat{r}}} = -\frac{1}{2\sqrt{1 - \hat{r}}} < 0,
\end{equation}
which implies that \eqref{eqn:primalConsistency4_1} is monotonically decreasing in $\hat{r}$ for any $r \in [0,1]$ and is maximized by taking $\hat{r} \to 0$. By l'H\^{o}pital's rule, 
\begin{equation}\label{eqn:primalConsistency4_3}
\lim_{\hat{r} \to 0} \frac{\hat{r}}{1 - \sqrt{1 - \hat{r}}} = 2,
\end{equation}
and it remains to maximize 
\begin{equation}\label{eqn:primalConsistency4_4}
r \log \frac{2(1 - \sqrt{1 - r})}{r}.
\end{equation}
The calculation in \eqref{eqn:primalConsistency4_2} also implies that $\log(2(1-\sqrt{1-r}) / r)$ is monotonically \emph{increasing} in $r$, and \eqref{eqn:primalConsistency4_3} implies that 
\[
\lim_{r \to 0} \log\frac{2(1 - \sqrt{1 - r})}{r} = 0,
\]
so that $\log(2(1-\sqrt{1-r}) / r) \geq 0$ for $r \in [0,1]$. It follows that the quantity in \eqref{eqn:primalConsistency4_4} is monotonically increasing in $r$, as the product of two non-negative and monotonically increasing functions, and is therefore maximized at $r = 1$, yielding $\log 2$. This proves the upper bound.
\end{proof}

\begin{lemma}\label{lem:primalConsistency5}
For $r, \hat{r} \in [0,1]$, 
\begin{align*}
    \bar{D}(r \KLsep \hat{r}) \equiv r \abs*{\log \frac{1 - \sqrt{1 - r}}{1 - \sqrt{1 - \hat{r}}}} + (1-r) \abs*{\log \frac{1 - \sqrt{r}}{1 - \sqrt{\hat{r}}}} &\leq D_{\mathrm{KL}}(r \KLsep \hat{r}) + 2.
\end{align*}
\end{lemma}
\begin{proof}
Again it suffices to bound the first term of $\bar{D}(r \KLsep \hat{r})$ because the second term is analogous. From Lemma~\ref{lem:primalConsistency4} we have 
\begin{align*}
    r \abs*{\log \frac{1 - \sqrt{1 - r}}{1 - \sqrt{1 - \hat{r}}}} &= \max\left\{ r \log \frac{1 - \sqrt{1 - r}}{1 - \sqrt{1 - \hat{r}}}, -r \log \frac{1 - \sqrt{1 - r}}{1 - \sqrt{1 - \hat{r}}} \right\}\\
    &\leq \max\left\{r \log\frac{r}{\hat{r}} + \log 2, 0.6140\right\},
\end{align*}
where we note that $r \log(r/\hat{r}) + \log 2$ is always positive since its minimum value is $\log 2-1/e = 0.3253$ at $(r,\hat{r}) = (1/e, 1)$. We may further and more simply bound the above by 
\[
\max\left\{r \log\frac{r}{\hat{r}} + \log 2, 0.6140\right\} \leq r \log\frac{r}{\hat{r}} + \frac{1}{e} + 0.6140 < \log\frac{r}{\hat{r}} + 1,
\]
from which the result follows.
\end{proof}

\subsubsection{Proof of Lemma~\ref{lem:primalConsistency2}}

\begin{proof}
Let $\hat{\mu} = \hat{\lambda}^T \hat{\boldf}(a, \hat{r}(x))$ and $\mu = \hat{\lambda}^T \boldf(a, r(x))$. By the mean value theorem, 
\[
\abs*{H_b\left(r(x), r^*(\hat{\mu}; r(x))\right) - H_b\left(r(x), r^*(\mu; r(x))\right)} = \abs*{\left. \frac{\partial H_b\bigl(r, r^*(\mu; r)\bigr)}{\partial\mu} \right\rvert_{\mu=\hat{\lambda}^T \bar{\boldf}} }  \abs{\hat{\mu} - \mu},
\]
where $\bar{\boldf}$ is a convex combination of $\hat{\boldf}(a, \hat{r}(x))$ and $\boldf(a, r(x))$. By differentiating \eqref{eqn:g} with respect to $\mu$ and combining with \eqref{eqn:df(tlambda)}, we find that
\[
\abs*{\frac{\partial H_b\bigl(r, r^*(\mu; r)\bigr)}{\partial\mu}} = \abs{\mu} \abs*{\frac{\partial r^*(\mu; r)}{\partial\mu}},
\]
and it can be verified using \eqref{eqn:d2f(tlambda)} that $\abs{\partial r^*(\mu; r) / \partial\mu} \leq 1$. Hence 
\begin{align}
    \abs*{H_b\left(r(x), r^*(\hat{\mu}; r(x))\right) - H_b\left(r(x), r^*(\mu; r(x))\right)} &\leq \abs[\big]{\hat{\lambda}^T \bar{\boldf}} \abs*{\hat{\lambda}^T \bigl(\hat{\boldf}(a, \hat{r}(x)) - \boldf(a, r(x))\bigr)}\nonumber\\
    &\leq \norm[\big]{\hat{\lambda}}_1^2 \norm[\big]{\bar{\boldf}}_{\infty} \norm[\big]{\hat{\boldf}(a, \hat{r}(x)) - \boldf(a, r(x))}_{\infty}\nonumber\\
    &\leq \left(\frac{\log 2}{\epsilon}\right)^2 \norm[\big]{\bar{\boldf}}_{\infty} \norm[\big]{\hat{\boldf}(a, \hat{r}(x)) - \boldf(a, r(x))}_{\infty},\label{eqn:primalConsistency2_1}
\end{align}
where the second line results from two applications of H\"{o}lder's inequality, and the third line from Lemma~\ref{lem:Lambda0} given $\hat{\lambda} \in \Lambda_0$. Since Lemma~\ref{lem:fUB} applies to both $\hat{\boldf}(a, \hat{r}(x))$ and $\boldf(a, r(x))$, we have 
\begin{equation}\label{eqn:primalConsistency2_2}
\norm[\big]{\bar{\boldf}}_{\infty} \leq \frac{1}{\eta} - 1
\end{equation}
for their convex combination as well. Using the triangle inequality, 
\begin{equation}\label{eqn:primalConsistency2_3}
\norm[\big]{\hat{\boldf}(a, \hat{r}(x)) - \boldf(a, r(x))}_{\infty} \leq \norm[\big]{\hat{\boldf}(a, \hat{r}(x)) - \boldf(a, \hat{r}(x))}_{\infty} + \norm[\big]{\boldf(a, \hat{r}(x)) - \boldf(a, r(x))}_{\infty}.
\end{equation}
In the case of MSP, the second right-hand side term is zero because $\boldf$ does not depend on $r$. For GEO, \eqref{eqn:fMEO} implies that 
\begin{align}
&\norm[\big]{\boldf(a, \hat{r}(x)) - \boldf(a, r(x))}_{\infty}\nonumber\\ 
&\qquad \qquad \leq \abs{\hat{r}(x) - r(x)} \max_{a\in\cA, y\in\{0,1\}} \max \left\{ \frac{1}{p_Y(y)} \left(\frac{1}{p_{A\given Y}(a\given y)} - 1\right), \frac{1}{p_Y(y)} \right\}\nonumber\\
&\qquad \qquad \leq \abs{\hat{r}(x) - r(x)} \left(\frac{1}{\eta} - 1\right),\label{eqn:primalConsistency2_4}
\end{align}
where the last inequality was derived in the proof of Lemma~\ref{lem:fUB}.

We combine \eqref{eqn:primalConsistency2_1}, \eqref{eqn:primalConsistency2_2}, \eqref{eqn:primalConsistency2_3}, and \eqref{eqn:primalConsistency2_4} to obtain 
\begin{multline*}
    \abs*{H_b\left(r(x), r^*(\hat{\mu}; r(x))\right) - H_b\left(r(x), r^*(\mu; r(x))\right)}\\ \leq \left(\frac{\log 2}{\epsilon}\right)^2 \left(\frac{1}{\eta} - 1\right) \left( \norm[\big]{\hat{\boldf}(a, \hat{r}(x)) - \boldf(a, \hat{r}(x))}_{\infty} + \left(\frac{1}{\eta} - 1\right) \abs{\hat{r}(x) - r(x)} \right),
\end{multline*}
where the $\abs{\hat{r}(x) - r(x)}$ term is absent in the MSP case. Taking expectations over $x \in \cX$,
\begin{align*}
    &\abs*{\EE{H_b\left(r(X), r^*\bigl(\hat{\lambda}^T \hat{\boldf}(A, \hat{r}(X)); r(X)\bigr)\right)} - \EE{H_b\left(r(X), r^*\bigl(\hat{\lambda}^T \boldf(A, r(X)); r(X)\bigr)\right)}}\\ 
    &\quad \leq \EE{\abs*{H_b\left(r(X), r^*\bigl(\hat{\lambda}^T \hat{\boldf}(A, \hat{r}(X)); r(X)\bigr)\right) - H_b\left(r(X), r^*\bigl(\hat{\lambda}^T \boldf(A, r(X)); r(X)\bigr)\right)}}\\
    &\quad \leq \left(\frac{\log 2}{\epsilon}\right)^2 \left(\frac{1}{\eta} - 1\right) \left( \EE{\norm[\big]{\hat{\boldf}(A, \hat{r}(X)) - \boldf(A, \hat{r}(X))}_{\infty}} + \left(\frac{1}{\eta} - 1\right) \EE{\abs{\hat{r}(X) - r(X)}} \right).
\end{align*}
The proof of Lemma~\ref{lem:dualConsistency2} shows that $\EE{\norm[\big]{\hat{\boldf}(A, \hat{r}(X)) - \boldf(A, \hat{r}(X))}_{\infty}}$ 
converges to zero in probability as $\hat{p}_A$, $\hat{p}_{A,Y}$, $\hat{p}_Y$ converge to the true probabilities. Assumption~\ref{ass:rhatL1convProb} ensures that $\EE{\abs{\hat{r}(X) - r(X)}}$ converges to zero in probability as well, completing the proof.
\end{proof}

\subsubsection{Proof of Lemma~\ref{lem:primalConsistency6}}

\begin{proof}
We apply the mean value theorem in the same way as in the proof of Lemma~\ref{lem:primalConsistency2}, where now $\hat{\mu} = \hat{\lambda}^T \boldf(a, r(x))$, $\mu = \lambda^{*T} \boldf(a, r(x))$, and the intermediate value is $\bar{\mu} = \bar{\lambda}^T \boldf(a, r(x))$ for some convex combination $\bar{\lambda}$ of $\hat{\lambda}$ and $\lambda^*$. Following the same steps as in the earlier proof, we obtain 
\begin{align*}
    \abs*{H_b\left(r(x), r^*(\hat{\mu}; r(x))\right) - H_b\left(r(x), r^*(\mu; r(x))\right)} &\leq \abs[\big]{\bar{\lambda}^T \boldf} \abs*{\bigl(\hat{\lambda} - \lambda^*\bigr)^T \boldf(a, r(x))}\\
    &\leq \norm[\big]{\bar{\lambda}}_1 \norm{\boldf}_{\infty} \abs*{\bigl(\hat{\lambda} - \lambda^*\bigr)^T \boldf(a, r(x))}\\
    &\leq \left(\frac{\log 2}{\epsilon}\right) \left(\frac{1}{\eta} - 1\right) \abs*{\bigl(\hat{\lambda} - \lambda^*\bigr)^T \boldf(a, r(x))},
\end{align*}
again using H\"{o}lder's inequality in the second line, and Lemmas~\ref{lem:Lambda0} and \ref{lem:fUB} in the third line. Taking expectations over $x \in \cX$ then yields 
\begin{align*}
    &\abs*{\EE{H_b\left(r(X), r^*\bigl(\hat{\lambda}^T \boldf(A, r(X)); r(X)\bigr)\right)} - \EE{H_b\left(r(X), r^*\bigl(\lambda^{*T} \boldf(A, r(X)); r(X)\bigr)\right)}}\\ 
    &\quad \leq \EE{\abs*{H_b\left(r(X), r^*\bigl(\hat{\lambda}^T \boldf(A, r(X)); r(X)\bigr)\right) - H_b\left(r(X), r^*\bigl(\lambda^{*T} \boldf(A, r(X)); r(X)\bigr)\right)}}\\
    &\quad \leq \left(\frac{\log 2}{\epsilon}\right) \left(\frac{1}{\eta} - 1\right) \EE{\abs*{\bigl(\hat{\lambda} - \lambda^*\bigr)^T \boldf(A, r(X))}}.
\end{align*}
The proof is completed by Lemma~\ref{lem:primalConsistency7} below.
\end{proof}

\begin{lemma}\label{lem:primalConsistency7}
Under Assumptions~\ref{ass:Atrue}, \ref{ass:pAY}, \ref{ass:rhatTV}, \ref{ass:epsilon}, \ref{ass:strongConvex},
\[
\EE{\abs*{\bigl(\hat{\lambda} - \lambda^*\bigr)^T \boldf(A, r(X))}} \overset{p}{\to} 0.
\]
\end{lemma}
\begin{proof}
We prove that the quantity in question converges to zero in the $L_2$ norm, 
\begin{equation}\label{eqn:primalConsistency7_1}
\EE{\abs*{\bigl(\hat{\lambda} - \lambda^*\bigr)^T \boldf(A, r(X))}^2} \overset{p}{\to} 0,
\end{equation}
As shown in the proof of Lemma~\ref{lem:KL_L1}, this implies convergence in the $L_1$ norm as in the lemma statement.

To establish \eqref{eqn:primalConsistency7_1}, we use Theorem~\ref{thm:dualConsistency}, which implies that for any $\varepsilon > 0$, we have $J(\hat{\lambda}) \leq J(\lambda^*) + \varepsilon$ with probability converging to $1$ as $n, m \to \infty$. Assume then that $J(\hat{\lambda}) \leq J(\lambda^*) + \varepsilon$. Define $\lambda_{\alpha} = \alpha \lambda^* + (1-\alpha) \hat{\lambda}$ for $\alpha \in [0,1]$, and 
\begin{equation}\label{eqn:G}
G(\lambda) = \EE{g\bigl(\lambda^T \boldf(A, r(X)); r(X)\bigr)}
\end{equation}
to be the first term in the population dual objective function \eqref{eqn:dualMSP_rhat}, \eqref{eqn:dualMEO_rhat}. From \citet[Proposition~A.23b]{bertsekas1999}, we have the following second-order expansion:
\begin{align}
G\bigl(\hat{\lambda}\bigr) &= G(\lambda_\alpha) + \bigl(\hat{\lambda} - \lambda_\alpha\bigr)^T \nabla G(\lambda_\alpha) + \frac{1}{2} \bigl(\hat{\lambda} - \lambda_\alpha\bigr)^T \nabla^2 G(\bar{\lambda}) \bigl(\hat{\lambda} - \lambda_\alpha\bigr)\nonumber\\
&= G(\lambda_\alpha) + \alpha \bigl(\hat{\lambda} - \lambda^*\bigr)^T \nabla G(\lambda_\alpha) + \frac{1}{2} \alpha^2 \bigl(\hat{\lambda} - \lambda^*\bigr)^T \nabla^2 G(\bar{\lambda}) \bigl(\hat{\lambda} - \lambda^*\bigr),\label{eqn:primalConsistency7_2}
\end{align}
where $\bar{\lambda}$ is some point on the line segment between $\hat{\lambda}$ and $\lambda_\alpha$. By differentiating \eqref{eqn:G} and using \eqref{eqn:df(tlambda)}, we find 
\begin{align}
    \nabla G(\lambda) &= \EE{-r^*\bigl(\lambda^T \boldf(A, r(X)); r(X)\bigr) \boldf(A, r(X))},\nonumber\\
    \nabla^2 G(\lambda) &= \EE{s\bigl(\lambda^T \boldf(A, r(X)); r(X)\bigr) \boldf(A, r(X)) \boldf^T(A, r(X))},\label{eqn:primalConsistency7_3}
\end{align}
where $s(\mu; r) = -\partial r^*(\mu; r) / \partial\mu$. Substituting \eqref{eqn:primalConsistency7_3} into \eqref{eqn:primalConsistency7_2} yields 
\[
G\bigl(\hat{\lambda}\bigr) = G(\lambda_\alpha) + \alpha \bigl(\hat{\lambda} - \lambda^*\bigr)^T \nabla G(\lambda_\alpha) + \frac{1}{2} \alpha^2 \EE{s\bigl(\bar{\lambda}^T \boldf(A, r(X)); r(X)\bigr) \abs*{\bigl(\hat{\lambda} - \lambda^*\bigr)^T \boldf(A, r(X))}^2}.
\]
Since $\bar{\lambda}$ also lies on the line segment between $\hat{\lambda}$ and $\lambda^*$, the application of Assumption~\ref{ass:strongConvex} yields 
\begin{equation}\label{eqn:primalConsistency7_4}
G\bigl(\hat{\lambda}\bigr) \geq G(\lambda_\alpha) + \alpha \bigl(\hat{\lambda} - \lambda^*\bigr)^T \nabla G(\lambda_\alpha) + \frac{1}{2} \alpha^2 \tau \EE{\abs*{\bigl(\hat{\lambda} - \lambda^*\bigr)^T \boldf(A, r(X))}^2}.
\end{equation}

By repeating the above steps for $\lambda^*$ in place of $\hat{\lambda}$, we also obtain 
\begin{equation}\label{eqn:primalConsistency7_5}
G\bigl(\lambda^*\bigr) \geq G(\lambda_\alpha) + (1-\alpha) \bigl(\lambda^* - \hat{\lambda}\bigr)^T \nabla G(\lambda_\alpha) + \frac{1}{2} (1-\alpha)^2 \tau \EE{\abs*{\bigl(\hat{\lambda} - \lambda^*\bigr)^T \boldf(A, r(X))}^2}.
\end{equation}
Multiplying \eqref{eqn:primalConsistency7_4} by $1-\alpha$, \eqref{eqn:primalConsistency7_5} by $\alpha$, and summing,
\begin{equation}\label{eqn:primalConsistency7_6}
(1-\alpha) G\bigl(\hat{\lambda}\bigr) + \alpha G\bigl(\lambda^*\bigr) \geq G(\lambda_\alpha) + \frac{1}{2} \alpha (1-\alpha) \tau \EE{\abs*{\bigl(\hat{\lambda} - \lambda^*\bigr)^T \boldf(A, r(X))}^2}.
\end{equation}
Since the $\ell_1$ norm is convex, we also have 
\begin{equation}\label{eqn:primalConsistency7_7}
    (1-\alpha) \epsilon \norm[\big]{\hat{\lambda}}_1 + \alpha \epsilon \norm[\big]{\lambda^*}_1 \geq \epsilon \norm[\big]{\lambda_\alpha}_1.
\end{equation}
Adding \eqref{eqn:primalConsistency7_6} and \eqref{eqn:primalConsistency7_7}, 
\begin{align*}
    (1-\alpha) J\bigl(\hat{\lambda}\bigr) + \alpha J\bigl(\lambda^*\bigr) &\geq J(\lambda_\alpha) + \frac{1}{2} \alpha (1-\alpha) \tau \EE{\abs*{\bigl(\hat{\lambda} - \lambda^*\bigr)^T \boldf(A, r(X))}^2}\\
    &\geq J\bigl(\lambda^*\bigr) + \frac{1}{2} \alpha (1-\alpha) \tau \EE{\abs*{\bigl(\hat{\lambda} - \lambda^*\bigr)^T \boldf(A, r(X))}^2},
\end{align*}
where the last inequality is due to the optimality of $\lambda^*$. Using the assumption $J(\hat{\lambda}) \leq J(\lambda^*) + \varepsilon$, we arrive at 
\[
\frac{1}{2} \alpha \tau \EE{\abs*{\bigl(\hat{\lambda} - \lambda^*\bigr)^T \boldf(A, r(X))}^2} \leq \varepsilon,
\]
and since $\alpha \in [0,1]$ was arbitrary, we take $\alpha \to 1$ to yield 
\begin{equation}\label{eqn:primalConsistency7_8}
\EE{\abs*{\bigl(\hat{\lambda} - \lambda^*\bigr)^T \boldf(A, r(X))}^2} \leq \frac{2\varepsilon}{\tau}.
\end{equation}

We have thus shown that the near-optimality condition $J(\hat{\lambda}) \leq J(\lambda^*) + \varepsilon$ implies \eqref{eqn:primalConsistency7_8} for any $\varepsilon > 0$. Since the former holds with probability converging to $1$, this proves \eqref{eqn:primalConsistency7_1}.

\end{proof}

\section{ADMM Algorithm Details}

In this appendix, we present a closed-form solution for step \eqref{eq:iter-mu} in the ADMM algorithm of Section~\ref{sec:proc:ADMM}. We then describe alternative ADMM algorithms in Appendix~\ref{sec:proc:ADMMalt}.

\subsection{Closed-Form Solution for \eqref{eq:iter-mu}}
\label{sec:proc:muCubic}

To simplify notation, define $\trho = \rho n$ and drop the index $i$ and the hat from $\hat{r}$. Then the first-order optimality condition for \eqref{eq:iter-mu} can be written as 
\begin{equation}\label{eqn:muCubic1}
n \frac{\partial \mathsf{obj}(\mu)}{\partial \mu} = \trho(\mu - b) - r^*(\mu; r) = 0.
\end{equation}
Figure~\ref{fig:rStar_mu} shows that $r^*(\mu; r)$ decreases monotonically with $\mu$, and this can be proven by showing that expression~\eqref{eqn:d2f(tlambda)} for $dr^*(\mu; r) / d\mu$ does not change sign. It follows that the quantity in \eqref{eqn:muCubic1} strictly increases with $\mu$ and the equation has a unique solution.

To solve for the root of \eqref{eqn:muCubic1}, we use \eqref{eqn:r*} and rearrange to isolate the square root on one side:
\[
1 + \mu - 2 \trho \mu (\mu - b) = \sqrt{(1 + \mu)^2 - 4 r \mu}.
\]
Upon squaring both sides, it is seen that the zeroth-order terms in $\mu$ cancel to give
\begin{equation}\label{eqn:muCubic2}
\trho^2 \mu^2 (\mu - b)^2 - \trho \mu (1 + \mu) (\mu - b) = -r \mu.
\end{equation}
One solution to \eqref{eqn:muCubic2} is $\mu = 0$ but it satisfies the original condition \eqref{eqn:muCubic1} only if $r = -\trho b$. Assuming this is not the case, we divide both sides of \eqref{eqn:muCubic2} by $\mu$ and expand to yield the following cubic equation:
\begin{equation}\label{eqn:muCubic3}
    \mu^3 \underbrace{- (2b + (1/\trho))}_{a_2} \mu^2 + \underbrace{(b^2 + (b - 1) / \trho)}_{a_1} \mu + \underbrace{b/\trho + r/\trho^2}_{a_0} = 0.
\end{equation}

Toward obtaining a closed-form expression for the roots of \eqref{eqn:muCubic3}, define the coefficients of the corresponding \emph{depressed cubic} equation as 
\begin{align*}
    p &= 3a_1 - a_2^2,\\
    q &= a_2^3 - \frac{9}{2} a_1 a_2 + \frac{27}{2} a_0.
\end{align*}
We find empirically that cubic equation~\eqref{eqn:muCubic3} always has three real roots and that the desired root corresponding to the solution of \eqref{eqn:muCubic1} is given by the same one of these roots. We do not however have proofs of these facts. In the case of three real roots, they are given by the trigonometric formula 
\begin{equation}\label{eqn:muCubic4}
    \mu^* = \frac{1}{3} \left(2 \sqrt{-p} \cos{\left(\frac{1}{3} \left(\arccos\left(\frac{q}{p \sqrt{-p}}\right) - 2k\pi\right) \right)}  - a_2\right), \quad k = 0, 1, 2,
\end{equation}
and the desired root appears to always correspond to $k = 1$. In any case, the correct root can be identified by first noting that since $r^*(\mu; r) \in [0, 1]$, the solution to \eqref{eqn:muCubic1} must lie in the interval $[b, b + (1/\trho)]$. We may then compute all three roots in \eqref{eqn:muCubic4} and choose the one lying in $[b, b + (1/\trho)]$.

\subsection{Alternative ADMM Algorithms}
\label{sec:proc:ADMMalt}

This section presents alternative ADMM decompositions for the dual problems corresponding to MSP \eqref{eqn:dualMSP_rhat} and GEO \eqref{eqn:dualMEO_rhat}. To simplify notation, we suppress the hat symbols on $\hat{r}(x)$ and $\hat{\boldf}(x)$.

\subsubsection{Mean Score Parity}

Define auxiliary variables $\tlambda_a$ as follows:
\begin{equation}\label{eqn:tlambda_a}
\tlambda_a = \frac{\lambda_a}{p_A(a)} - \sum_{a'\in\cA} \lambda_{a'}, \quad a \in \cA,
\end{equation}
with $\tlambda = (\tlambda_a)_{a\in\cA}$. Then the empirical version of \eqref{eqn:dualMSP_rhat} can be written as 
\begin{equation}\label{eqn:dualMSP2}
    \begin{split}
        \min_{\lambda,\tlambda} \quad &\frac{1}{n} \sum_{i=1}^n g\bigl(\mu_i; r_i\bigr) + \epsilon \norm{\lambda}_1\\
        \st \quad &\mu_i = \sum_{a\in\cA} p_{A\given X}(a\given x_i) \tlambda_a,
    \end{split}
\end{equation}
where $\mu_i = \mu(x_i)$, $r_i = r(x_i)$, and we regard $\lambda$ and $\tlambda$ as two sets of optimization variables that are linearly related through \eqref{eqn:tlambda_a}. Let $\bB \in \mathbb{R}^{n\times d}$ be a matrix with entries $\bB_{i,a} = p_{A\given X}(a\given x_i)$ and rows $\bb_i^T$ so that we may write $\mu = \bB \tlambda$, $\mu_i = \bb_i^T \tlambda$. The objective function in \eqref{eqn:dualMSP2} is therefore separable between $\lambda$ and $\tlambda$. With $\ones$ denoting a vector of ones and $\bP_A$ the $d\times d$ diagonal matrix with diagonal entries $p_A(a)$, a scaled ADMM algorithm for \eqref{eqn:dualMSP2} consists of the following three steps in each iteration $k = 0,1,\dots$:
\begin{align}
    \tlambda^{k+1} &= \argmin_{\tlambda} \; \frac{1}{n} \sum_{i=1}^n g\bigl(\bb_i^T \tlambda; r_i\bigr) + \frac{\rho}{2} \norm*{\tlambda - \left(\bP_{A}^{-1} - \ones \ones^T\right) \lambda^k + u^k}_2^2 \label{eqn:ADMMtlambdaMSPS}\\
    \lambda^{k+1} &= \argmin_{\lambda} \; \epsilon \norm{\lambda}_1 + \frac{\rho}{2} \norm*{\left(\bP_{A}^{-1} - \ones \ones^T\right) \lambda - \tlambda^{k+1} - u^k}_2^2 \label{eqn:ADMMlambdaMSPS}\\
    u^{k+1} &= u^k + \tlambda^{k+1} - \left(\bP_{A}^{-1} - \ones \ones^T\right) \lambda^{k+1}.
\end{align}

The optimization in \eqref{eqn:ADMMlambdaMSPS} is an $\ell_1$-penalized quadratic minimization and can be handled by many convex solvers. The optimization in \eqref{eqn:ADMMtlambdaMSPS} can be solved using Newton's method.  Below we give the gradient and Hessian of the first term in \eqref{eqn:ADMMtlambdaMSPS}; the second Euclidean norm term is standard. First, using the definition of $g(\mu; r)$ in \eqref{eqn:g}, we find that 
\begin{align}
    \frac{dg(\mu; r)}{d\mu} &= -r^*(\mu; r)\label{eqn:df(tlambda)}\\
    \frac{d^2g(\mu; r)}{d\mu^2} &= -\frac{dr^*(\mu; r)}{d\mu} = \begin{cases}
    \dfrac{1}{2\mu^2} \left(1 - \dfrac{1 + (1-2r) \mu}{\sqrt{(1 + \mu)^2 - 4 r \mu}} \right), & \mu \neq 0\\
    r (1 - r), & \mu = 0.
    \end{cases}\label{eqn:d2f(tlambda)}
\end{align}
The simple form in \eqref{eqn:df(tlambda)} is due to $r^*(\mu; r)$ satisfying the optimality condition \eqref{eqn:LagrangianOptCond} and the ensuing cancellation of terms.  It is also related to \citet[Proposition~6.1.1]{bertsekas1999}. The gradient and Hessian of the first term in \eqref{eqn:ADMMtlambdaMSPS} are then given by 
\begin{align}
    \nabla\left(\frac{1}{n} \sum_{i=1}^n g\bigl(\bb_i^T \tlambda; r_i\bigr)\right) &= -\frac{1}{n} \bB^T \br^*\label{eqn:gradtlambda}\\
    \nabla^2\left(\frac{1}{n} \sum_{i=1}^n g\bigl(\bb_i^T \tlambda; r_i\bigr)\right) &= -\frac{1}{n} \bB^T \bH \bB,\label{eqn:hesstlambda}
\end{align}
where $\br^*$ is the $n$-dimensional vector with components $r^*(\mu_i; r_i)$ and $\bH$ is the $n\times n$ diagonal matrix with entries $dr^*(\mu_i; r_i)/d\mu_i$.  In the case where the features $X$ include the protected attribute $A$, $p_{A\given X}(a\given x_i) = \ones(a = a_i)$, $\bB$ is a sparse matrix with a single one in each row, and the Hessian in \eqref{eqn:hesstlambda} is diagonal. This implies that optimization \eqref{eqn:ADMMtlambdaMSPS} is separable over the components of $\tlambda$. 

\subsubsection{Generalized Equalized Odds}

In analogy with \eqref{eqn:tlambda_a} we define 
\begin{equation}\label{eqn:tlambda_ay}
\tlambda_{a,y} = \frac{\lambda_{a,y}}{p_{A\given Y}(a\given y)} - \sum_{a'\in\cA} \lambda_{a',y}, \quad a \in \cA, \; y \in \{0,1\}.
\end{equation}
Again let $\bB$ be a $n\times d$ matrix, recalling that $d = 2\abs{\cA}$ in the GEO case, with columns indexed by $(a, y)$ and entries
\begin{equation}\label{eqn:B}
\bB_{i,(a,y)} = \begin{cases}
\dfrac{(1-r(x_i)) p_{A\given X,Y}(a\given x_i, 0)}{p_Y(0)}, & y = 0\\
\dfrac{r(x_i) p_{A\given X,Y}(a\given x_i, 1)}{p_Y(1)}, & y = 1.
\end{cases}
\end{equation}
It can then be seen from the constraint in \eqref{eqn:dualMEO_rhat} that $\mu_i = \bb_i^T \tlambda$ as before and the empirical version of \eqref{eqn:dualMEO_rhat}, 
\begin{equation}\label{eqn:dualMEO2}
\min_{\lambda,\tlambda} \quad \frac{1}{n} \sum_{i=1}^n g\bigl(\bb_i^T \tlambda; r_i\bigr) + \epsilon \norm{\lambda}_1,
\end{equation}
is separable between $\lambda$ and $\tlambda$ subject to the linear relation \eqref{eqn:tlambda_ay}.  With $\bP_{A\given y}$ for $y = 0,1$ denoting the $\abs{\cA}\times \abs{\cA}$ diagonal matrix with diagonal entries $p_{A\given Y}(a\given y)$, the three steps in each ADMM iteration for \eqref{eqn:dualMEO2} are as follows:
\begin{align}
    \tlambda^{k+1} &= \argmin_{\tlambda} \; \frac{1}{n} \sum_{i=1}^n g\bigl(\bb_i^T \tlambda; r_i\bigr) + \frac{\rho}{2} \sum_{y=0}^1 \norm*{\tlambda_{\cdot,y} - \left(\bP_{A\given y}^{-1} - \ones \ones^T\right) \lambda_{\cdot,y}^k + u_{\cdot,y}^k}_2^2 \label{eqn:ADMMtlambdaMEO}\\
    \lambda_{\cdot,y}^{k+1} &= \argmin_{\lambda} \; \epsilon \norm{\lambda}_1 + \frac{\rho}{2} \norm*{\left(\bP_{A\given y}^{-1} - \ones \ones^T\right) \lambda - \tlambda_{\cdot,y}^{k+1} - u_{\cdot,y}^k}_2^2, \quad y = 0, 1 \label{eqn:ADMMlambdaMEO}\\
    u_{\cdot,y}^{k+1} &= u_{\cdot,y}^k + \tlambda_{\cdot,y}^{k+1} - \left(\bP_{A\given y}^{-1} - \ones \ones^T\right) \lambda_{\cdot,y}^{k+1}, \quad y = 0,1,
\end{align}
where $\tlambda_{\cdot,y}$, $\lambda_{\cdot,y}$, and $u_{\cdot,y}$ are $\abs{\cA}$-dimensional subvectors of $\tlambda$, $\lambda$ and $u$ consisting only of components with $y=0$ or $y=1$.
The optimization in \eqref{eqn:ADMMtlambdaMEO} is of the same form as \eqref{eqn:ADMMtlambdaMSPS} and can also be solved using Newton's method. The same expressions \eqref{eqn:gradtlambda}, \eqref{eqn:hesstlambda} hold for the gradient and Hessian of the first term in \eqref{eqn:ADMMtlambdaMEO}, where $\bB$ is now given by \eqref{eqn:B}. The optimization of $\lambda$ in \eqref{eqn:ADMMlambdaMEO} is separable over $y = 0, 1$ and is the same as step \eqref{eqn:ADMMlambdaMSPS} for MSP.

\section{Additional Experimental Results}
\label{sec:exptAdd}

This appendix presents results deferred from Section~\ref{sec:expt}, including results with log loss, those for the German credit risk data set, and comparisons with existing methods excluded from the main comparison due to their limitations.

\begin{figure}[t]
  \centering
  \begin{subfigure}[b]{0.32\columnwidth}
  \includegraphics[width=\columnwidth]{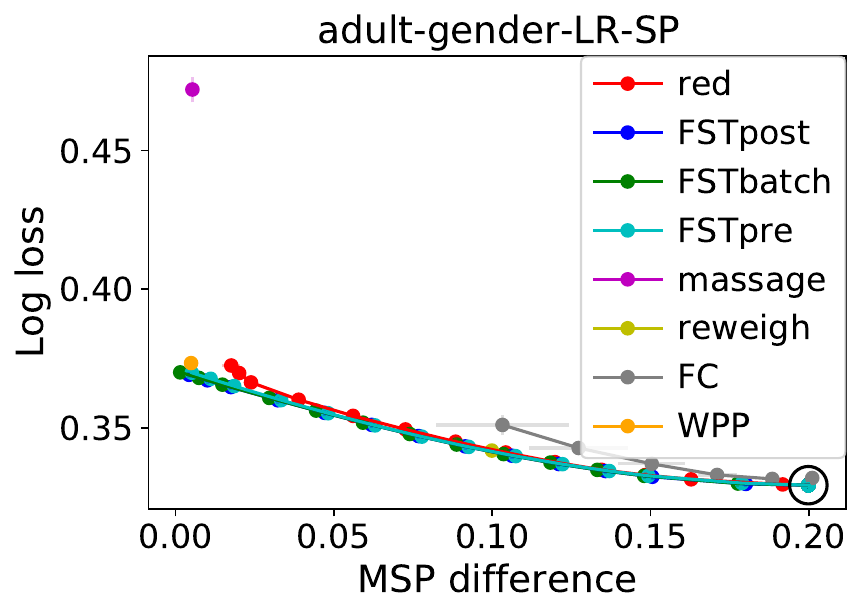}
  \label{fig:adult_1_LR_SP_acc_log}
  \end{subfigure}
  \begin{subfigure}[b]{0.32\columnwidth}
  \includegraphics[width=\columnwidth]{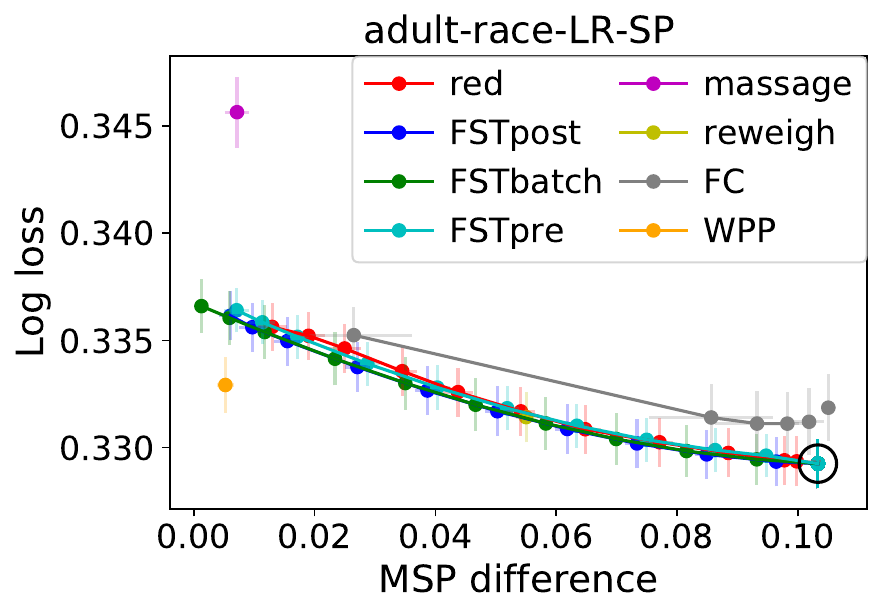}
  \label{fig:adult_2_LR_SP_acc_log}
  \end{subfigure}
  \begin{subfigure}[b]{0.32\columnwidth}
  \includegraphics[width=\columnwidth]{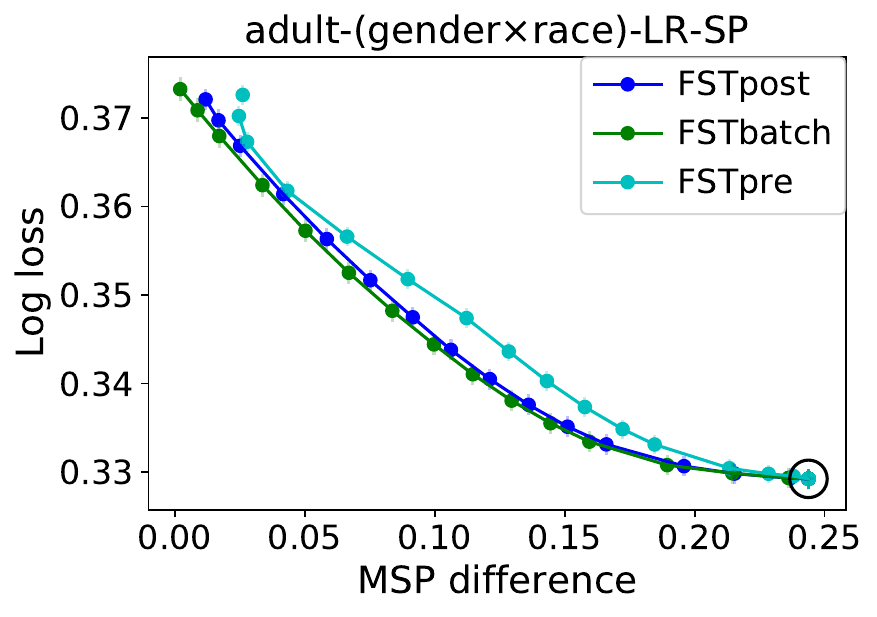}
  \label{fig:adult_both_LR_SP_acc_log}
  \end{subfigure}
  \begin{subfigure}[b]{0.32\columnwidth}
  \includegraphics[width=\columnwidth]{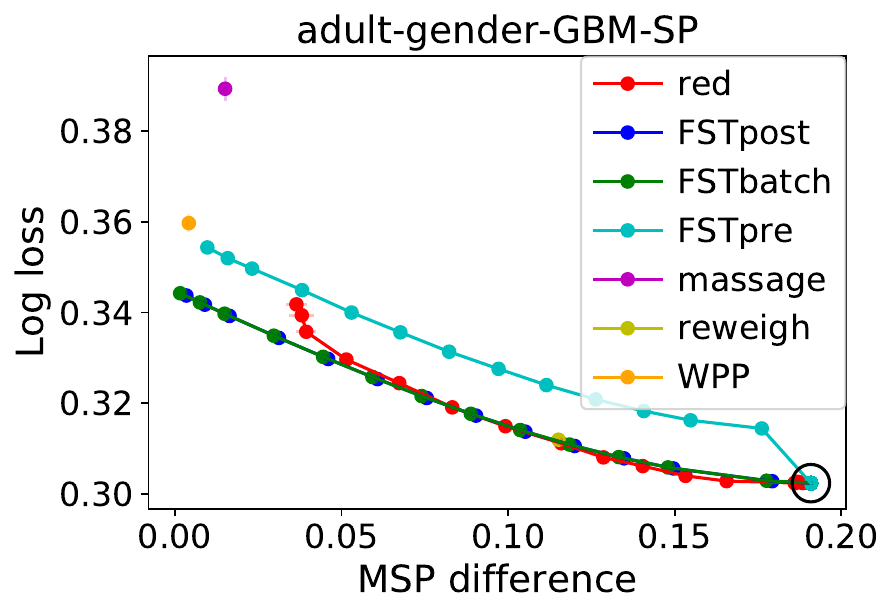}
  \label{fig:adult_1_GBM_SP_acc_log}
  \end{subfigure}
  \begin{subfigure}[b]{0.32\columnwidth}
  \includegraphics[width=\columnwidth]{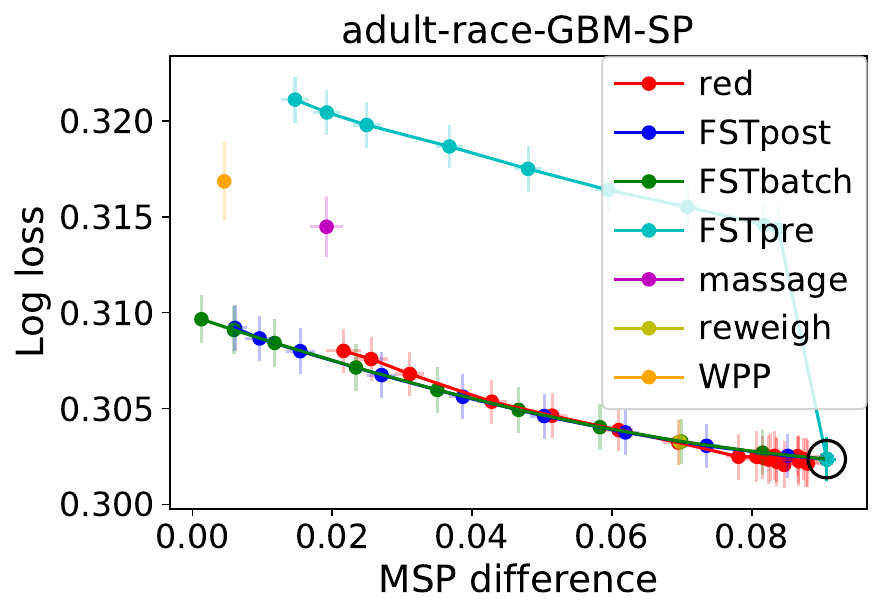}
  \label{fig:adult_2_GBM_SP_acc_log}
  \end{subfigure}
  \begin{subfigure}[b]{0.32\columnwidth}
  \includegraphics[width=\columnwidth]{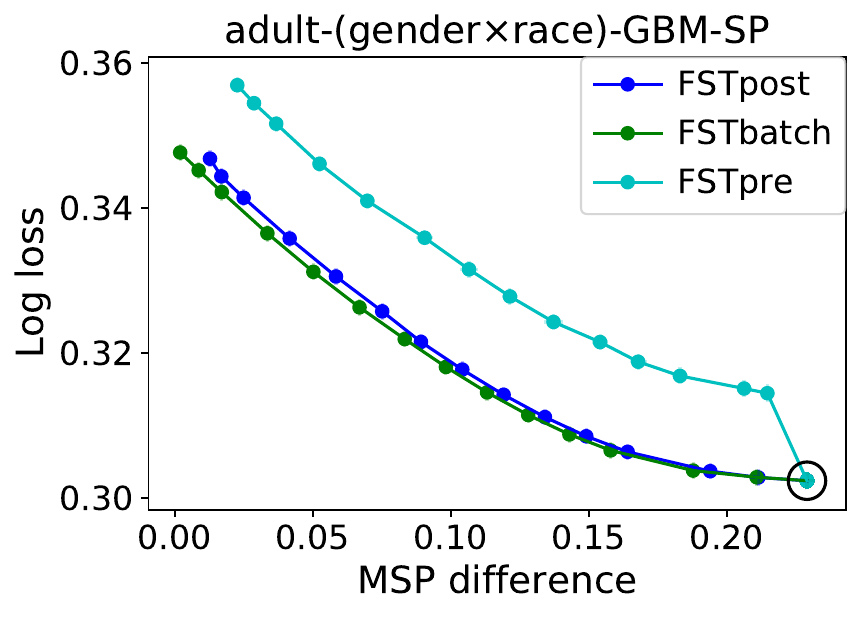}
  \label{fig:adult_both_GBM_SP_acc_log}
  \end{subfigure}
  \begin{subfigure}[b]{0.32\columnwidth}
  \includegraphics[width=\columnwidth]{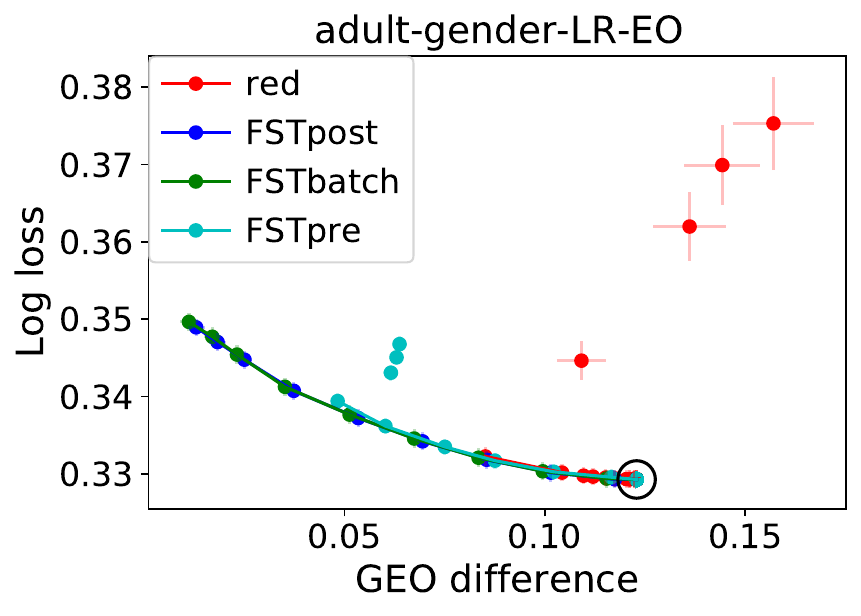}
  \label{fig:adult_1_LR_EO_acc_log}
  \end{subfigure}
  \begin{subfigure}[b]{0.32\columnwidth}
  \includegraphics[width=\columnwidth]{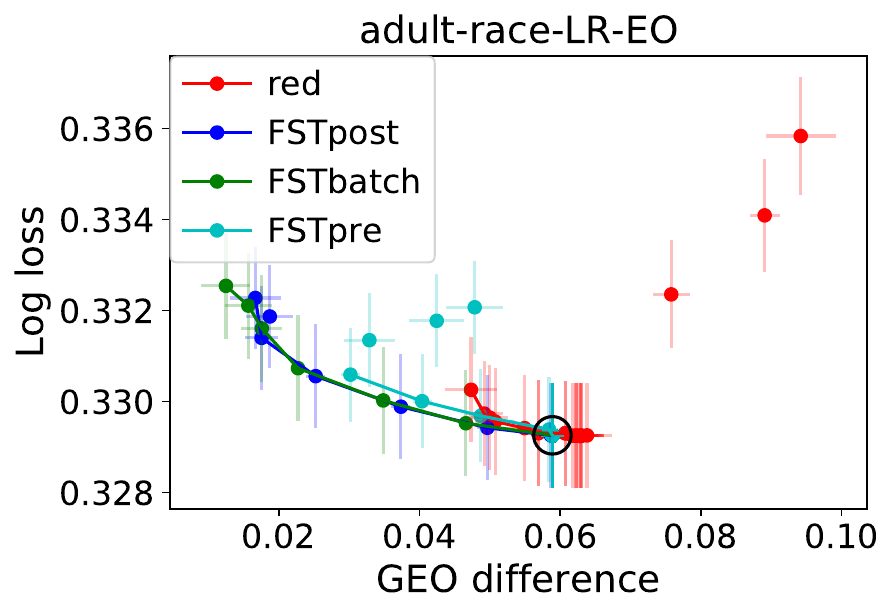}
  \label{fig:adult_2_LR_EO_acc_log}
  \end{subfigure}
  \begin{subfigure}[b]{0.32\columnwidth}
  \includegraphics[width=\columnwidth]{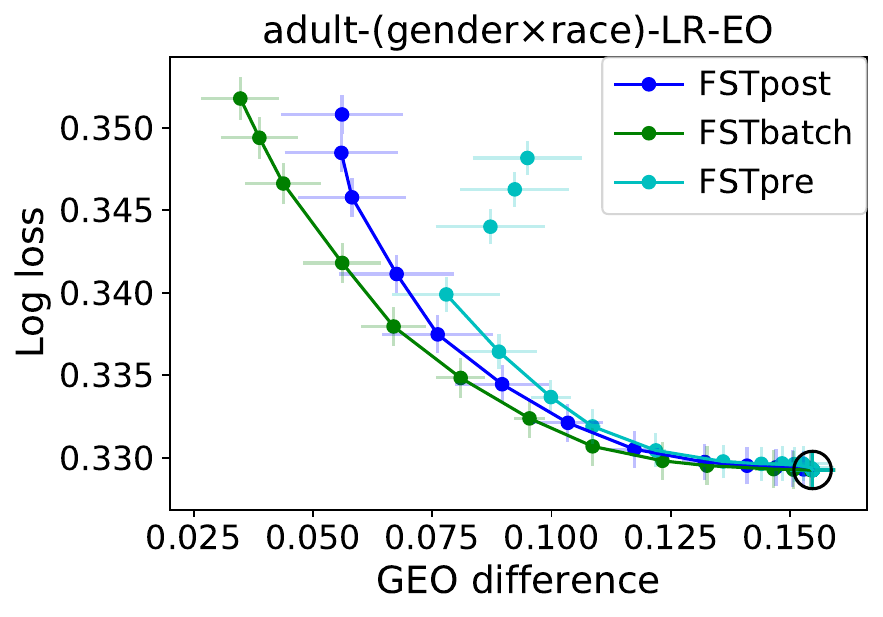}
  \label{fig:adult_both_LR_EO_acc_log}
  \end{subfigure}
  \begin{subfigure}[b]{0.32\columnwidth}
  \includegraphics[width=\columnwidth]{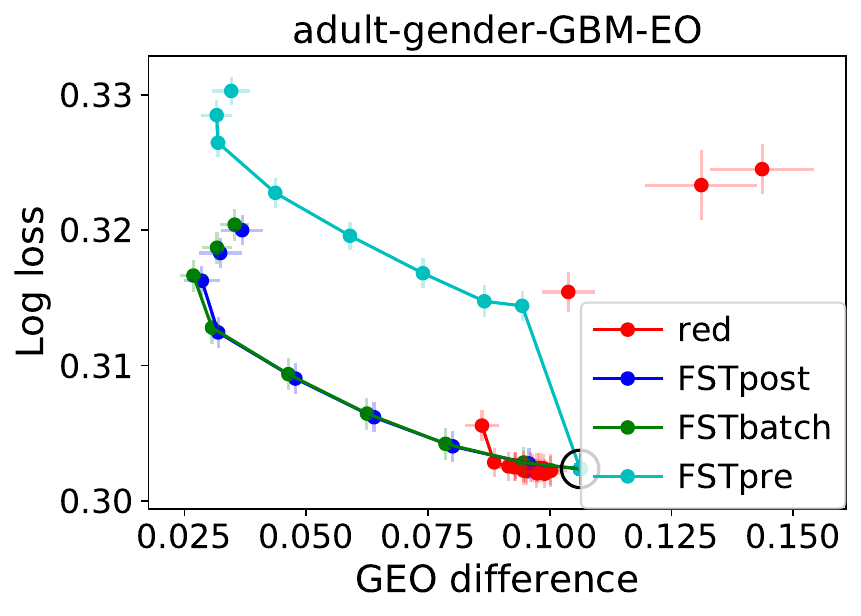}
  \label{fig:adult_1_GBM_EO_acc_log}
  \end{subfigure}
  \begin{subfigure}[b]{0.32\columnwidth}
  \includegraphics[width=\columnwidth]{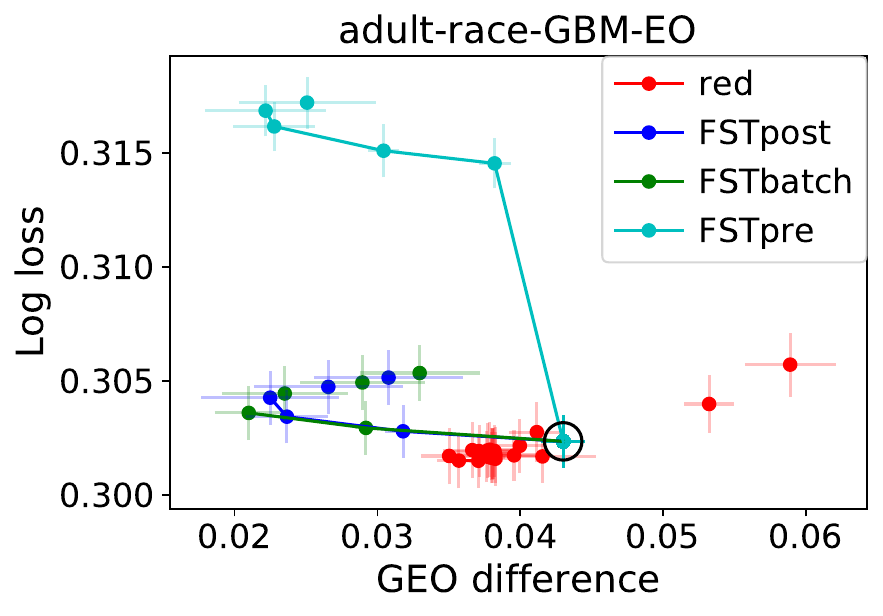}
  \label{fig:adult_2_GBM_EO_acc_log}
  \end{subfigure}
  \begin{subfigure}[b]{0.32\columnwidth}
  \includegraphics[width=\columnwidth]{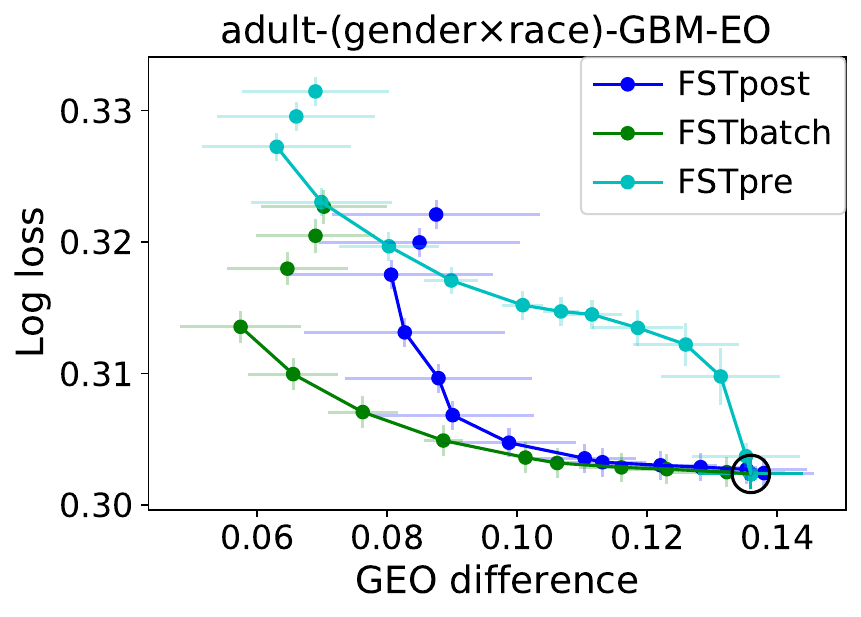}
  \label{fig:adult_both_GBM_EO_acc_log}
  \end{subfigure}
  \caption{Trade-offs between fairness and log loss on the Adult Income data set with the protected attributes included in the features.}
  \label{fig:adult_logloss}
\end{figure}

\subsection{Log Loss}
\label{sec:exptAdd:logloss}

\begin{figure}[t]
  \centering
  \begin{subfigure}[b]{0.32\columnwidth}
  \includegraphics[width=\columnwidth]{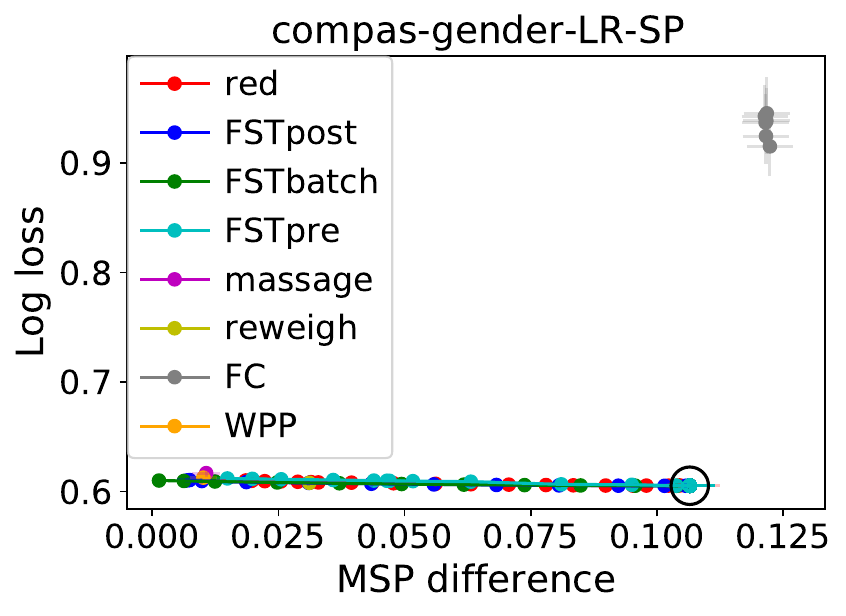}
  \label{fig:compas_1_LR_SP_acc_log}
  \end{subfigure}
  \begin{subfigure}[b]{0.32\columnwidth}
  \includegraphics[width=\columnwidth]{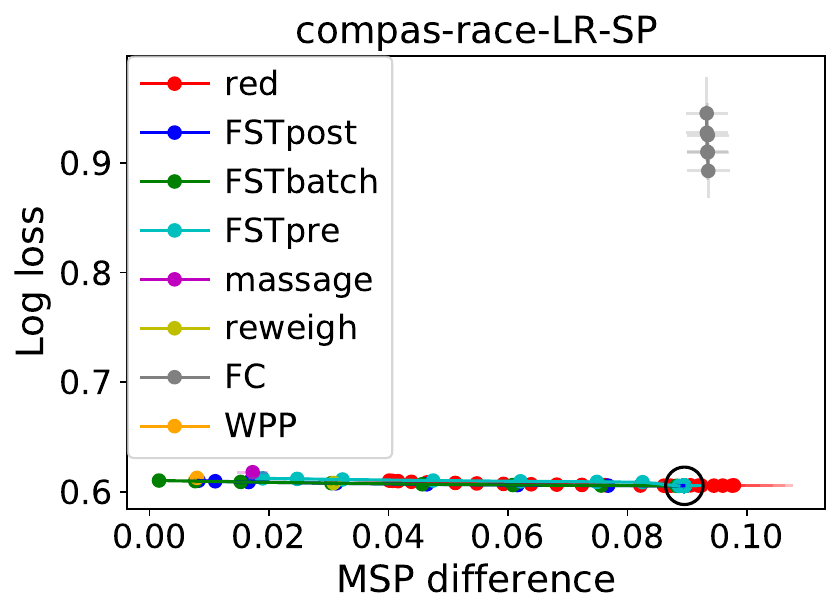}
  \label{fig:compas_2_LR_SP_acc_log}
  \end{subfigure}
  \begin{subfigure}[b]{0.32\columnwidth}
  \includegraphics[width=\columnwidth]{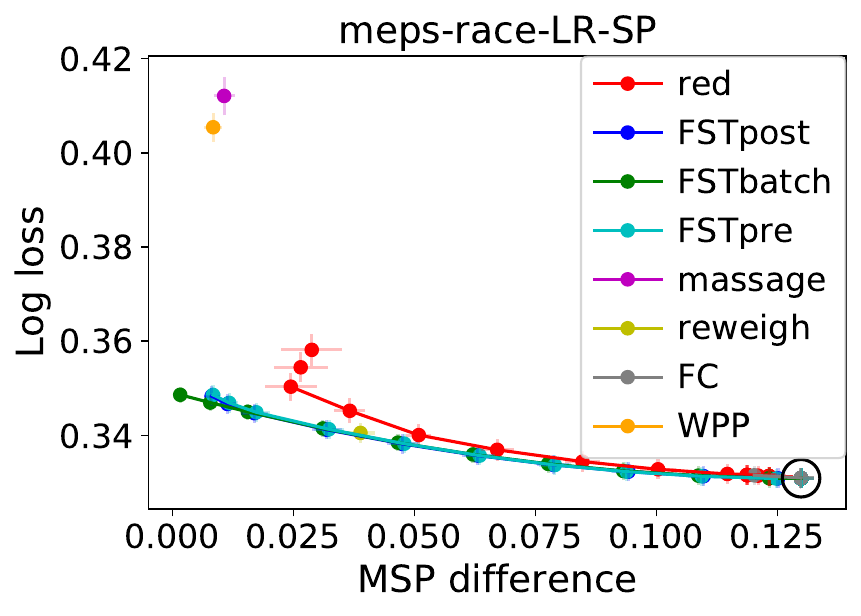}
  \label{fig:meps_1_LR_SP_acc_log}
  \end{subfigure}
  \begin{subfigure}[b]{0.32\columnwidth}
  \includegraphics[width=\columnwidth]{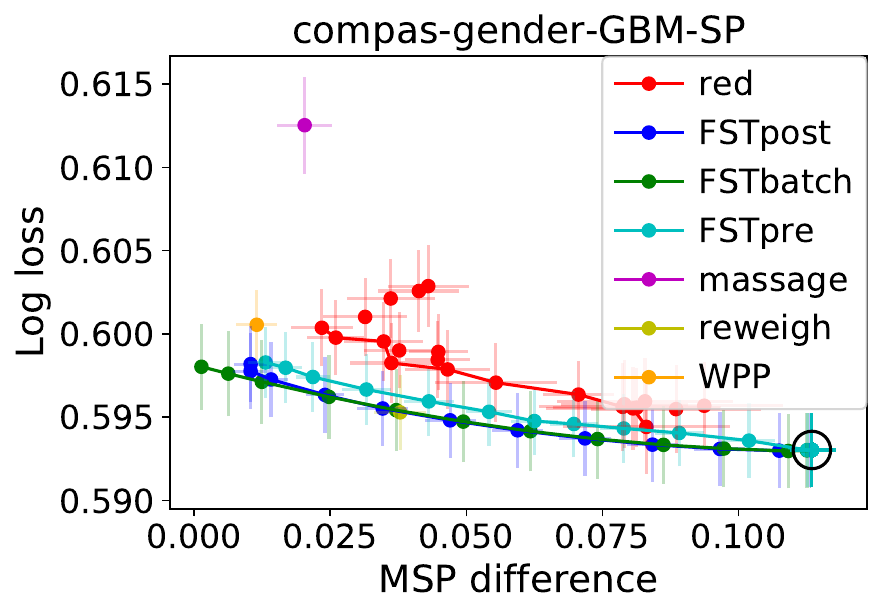}
  \label{fig:compas_1_GBM_SP_acc_log}
  \end{subfigure}
  \begin{subfigure}[b]{0.32\columnwidth}
  \includegraphics[width=\columnwidth]{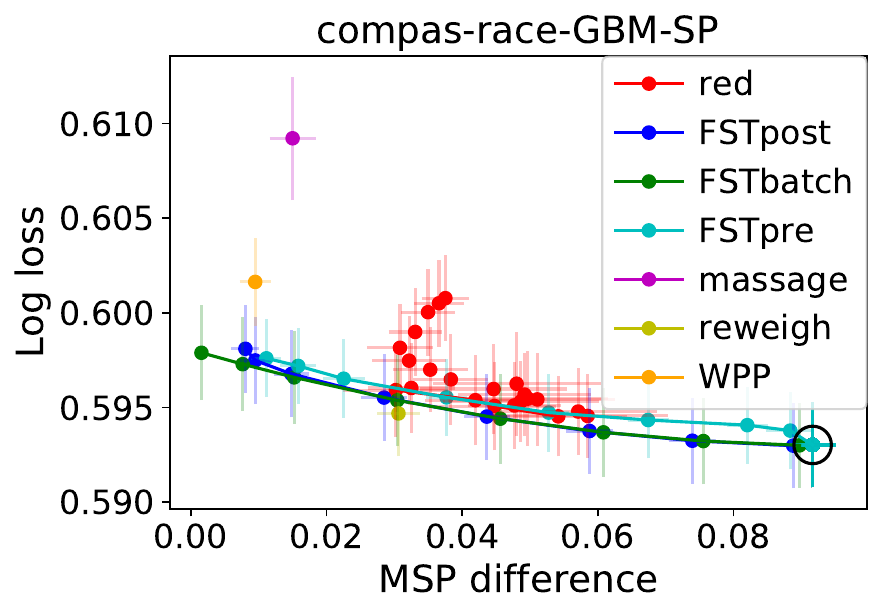}
  \label{fig:compas_2_GBM_SP_acc_log}
  \end{subfigure}
  \begin{subfigure}[b]{0.32\columnwidth}
  \includegraphics[width=\columnwidth]{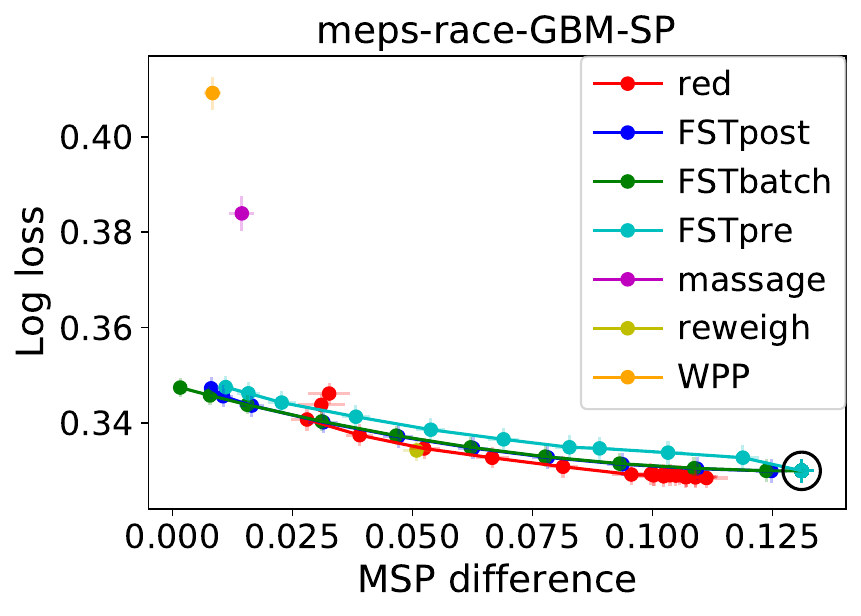}
  \label{fig:meps_1_GBM_SP_acc_log}
  \end{subfigure}
  \begin{subfigure}[b]{0.32\columnwidth}
  \includegraphics[width=\columnwidth]{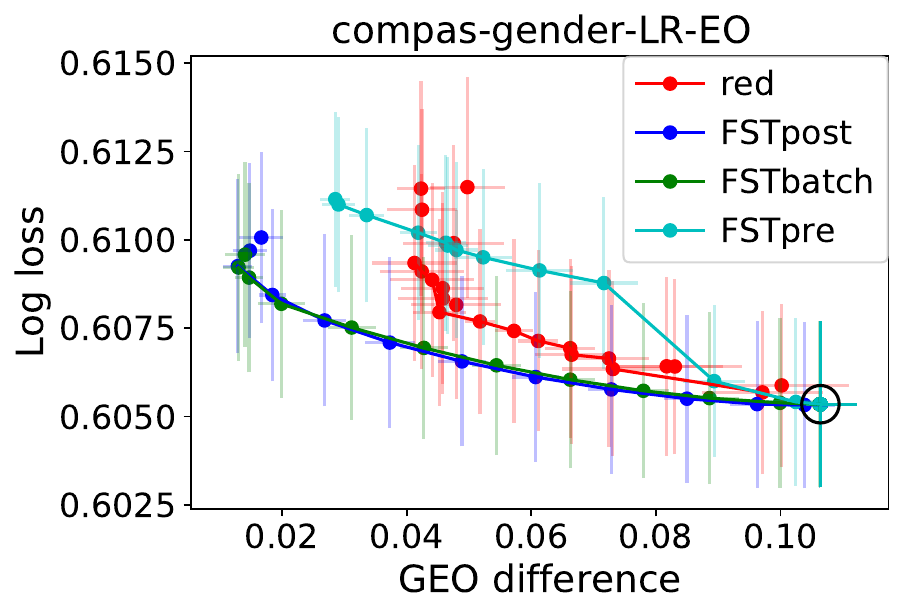}
  \label{fig:compas_1_LR_EO_acc_log}
  \end{subfigure}
  \begin{subfigure}[b]{0.32\columnwidth}
  \includegraphics[width=\columnwidth]{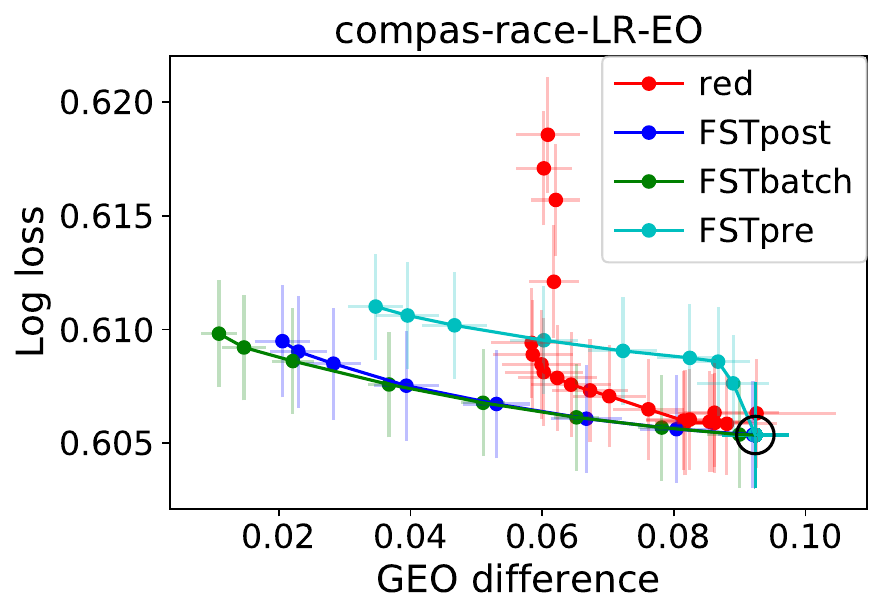}
  \label{fig:compas_2_LR_EO_acc_log}
  \end{subfigure}
  \begin{subfigure}[b]{0.32\columnwidth}
  \includegraphics[width=\columnwidth]{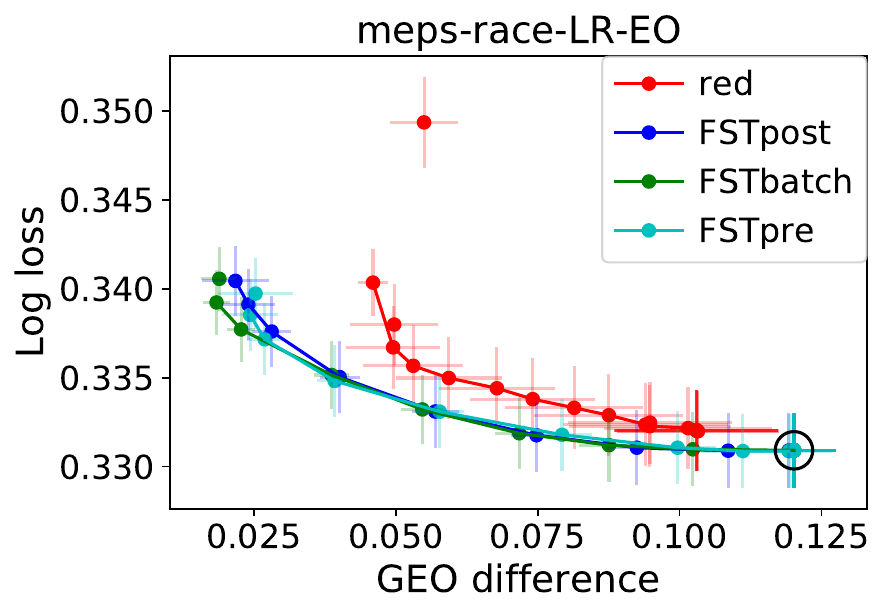}
  \label{fig:meps_1_LR_EO_acc_log}
  \end{subfigure}
  \begin{subfigure}[b]{0.32\columnwidth}
  \includegraphics[width=\columnwidth]{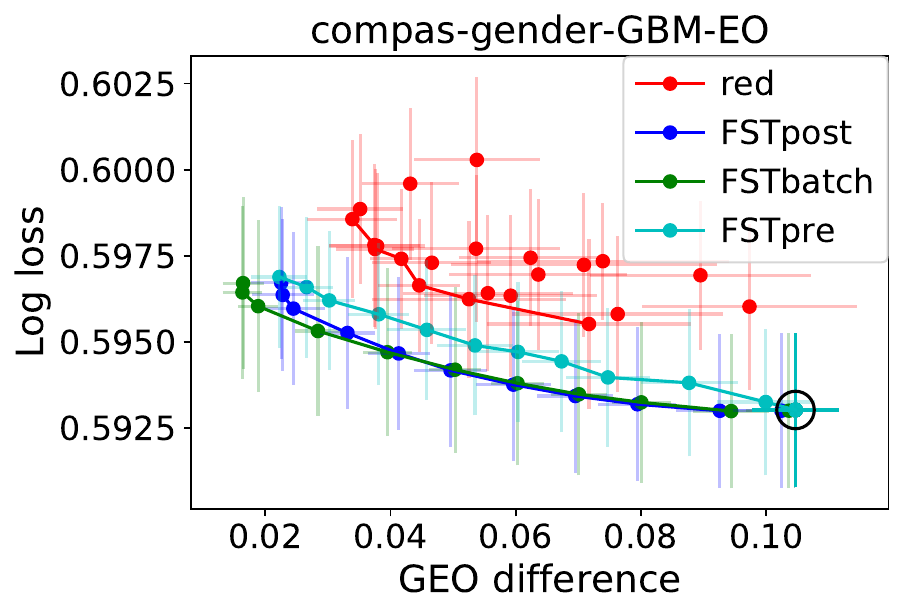}
  \label{fig:compas_1_GBM_EO_acc_log}
  \end{subfigure}
  \begin{subfigure}[b]{0.32\columnwidth}
  \includegraphics[width=\columnwidth]{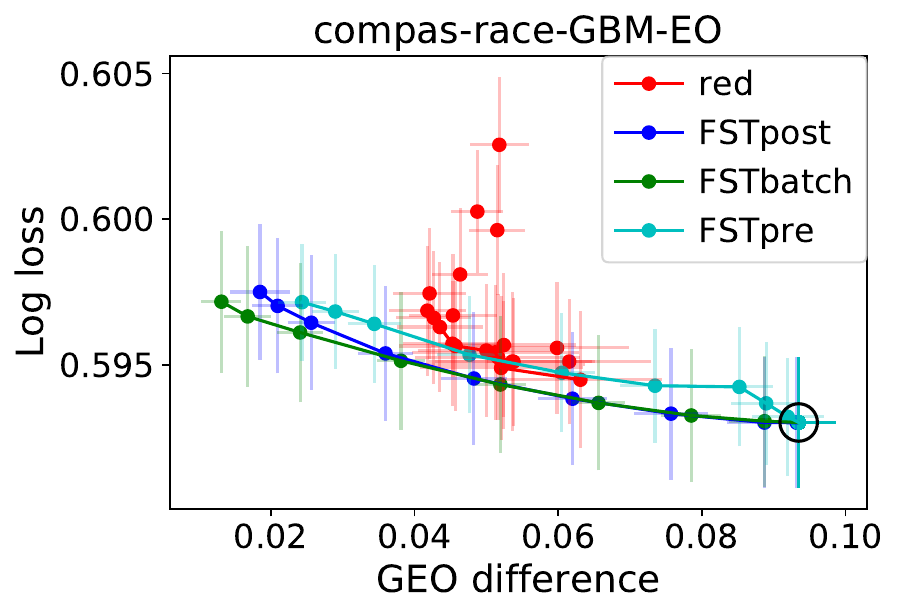}
  \label{fig:compas_2_GBM_EO_acc_log}
  \end{subfigure}
  \begin{subfigure}[b]{0.32\columnwidth}
  \includegraphics[width=\columnwidth]{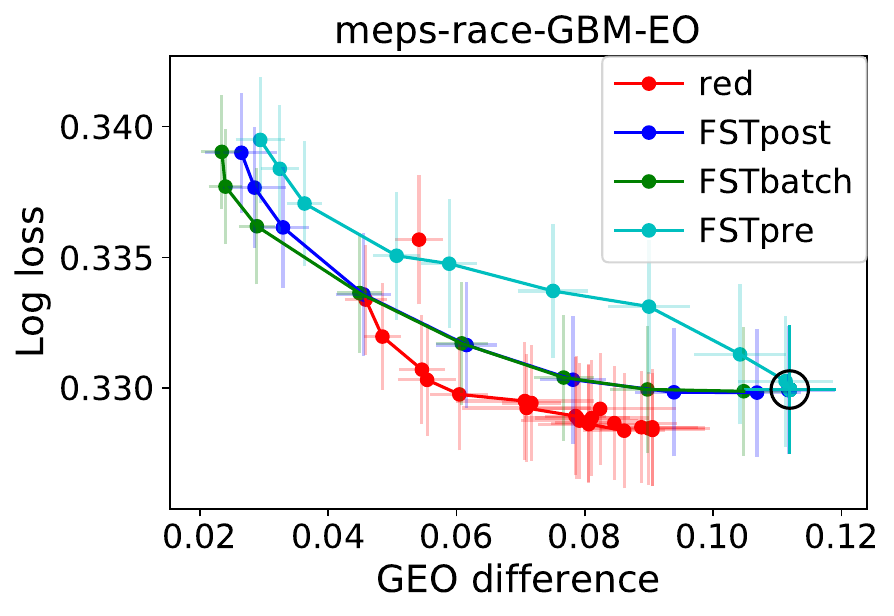}
  \label{fig:meps_1_GBM_EO_acc_log}
  \end{subfigure}
  \caption{Trade-offs between fairness and log loss on the COMPAS and MEPS data sets with the protected attributes included in the features.}
  \label{fig:compas_meps_logloss}
\end{figure}

Figures~\ref{fig:adult_logloss} and \ref{fig:compas_meps_logloss} show trade-offs between log loss and MSP or GEO fairness measures for the data set-protected attribute combinations considered in Figures~\ref{fig:adult_1}--\ref{fig:meps_1} and \ref{fig:adult_both} (i.e.,~with the protected attribute included in the features). The plots are quite similar to those for Brier score in Figures~\ref{fig:adult_1}--\ref{fig:meps_1} and \ref{fig:adult_both}.

\begin{figure}[t]
  \centering
  \begin{subfigure}[b]{0.32\columnwidth}
  \includegraphics[width=\columnwidth]{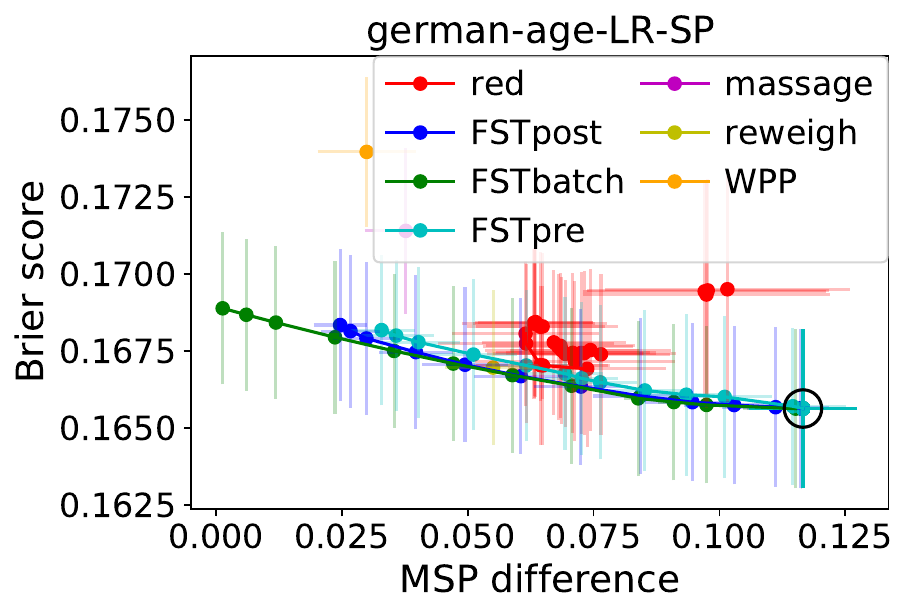}
  \label{fig:german_2_LR_SP_acc_Brier}
  \end{subfigure}
  \begin{subfigure}[b]{0.32\columnwidth}
  \includegraphics[width=\columnwidth]{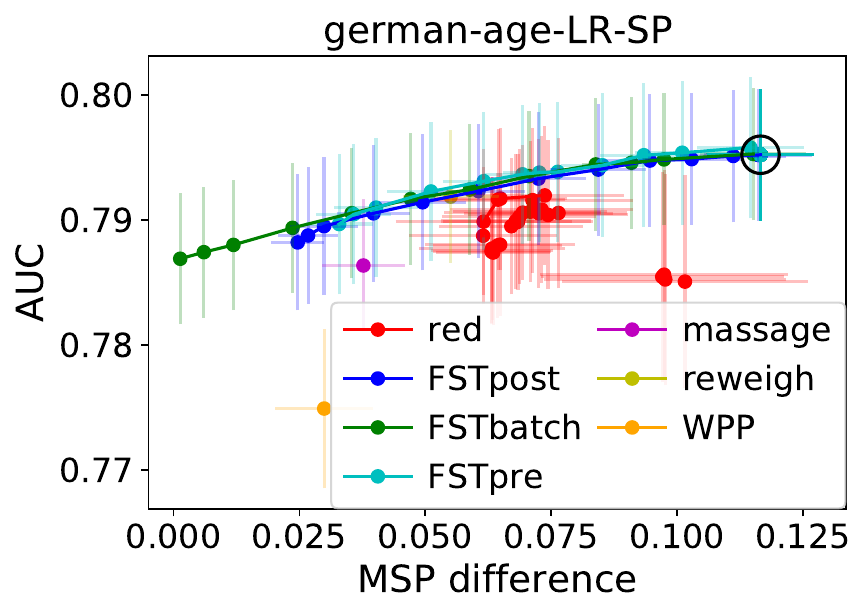}
  \label{fig:german_2_LR_SP_acc_AUC}
  \end{subfigure}
  \begin{subfigure}[b]{0.32\columnwidth}
  \includegraphics[width=\columnwidth]{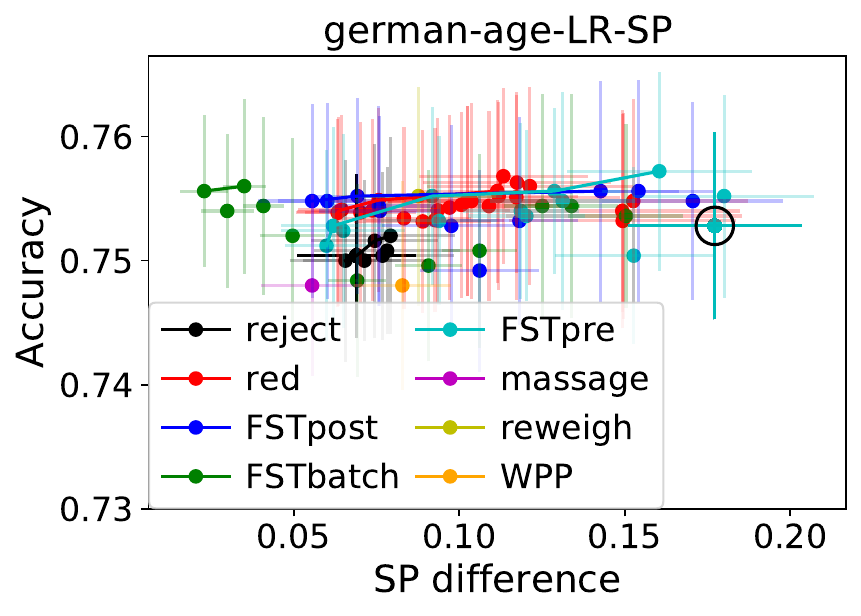}
  \label{fig:german_2_LR_SP_acc_acc}
  \end{subfigure}
  \begin{subfigure}[b]{0.32\columnwidth}
  \includegraphics[width=\columnwidth]{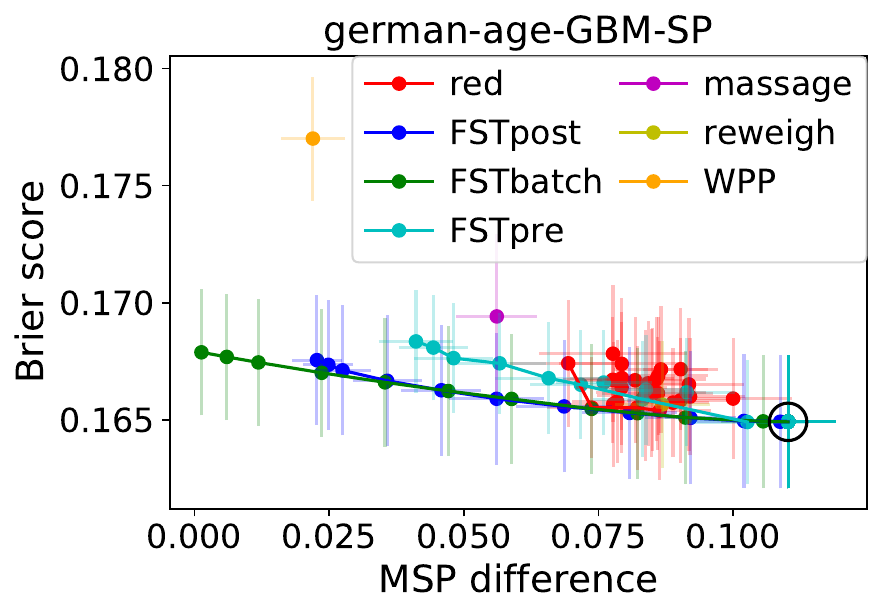}
  \label{fig:german_2_GBM_SP_acc_Brier}
  \end{subfigure}
  \begin{subfigure}[b]{0.32\columnwidth}
  \includegraphics[width=\columnwidth]{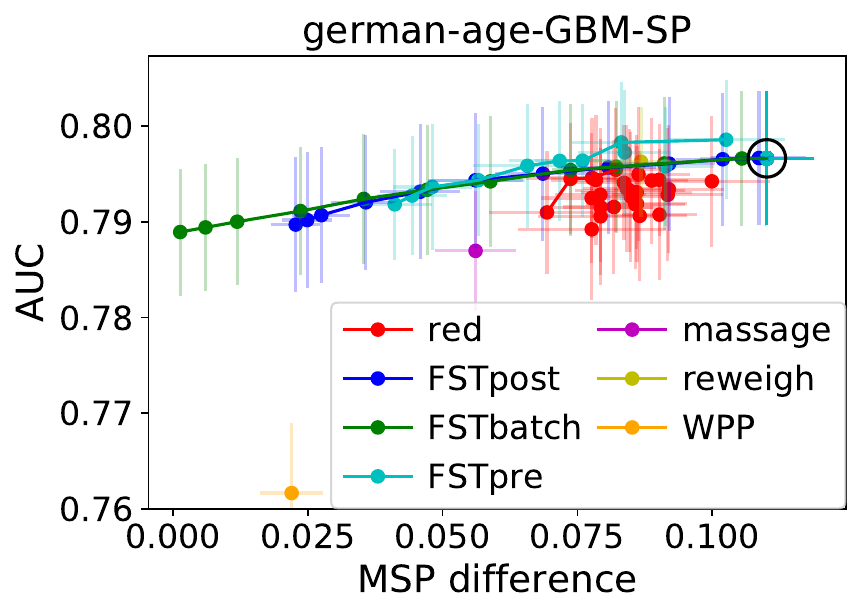}
  \label{fig:german_2_GBM_SP_acc_AUC}
  \end{subfigure}
  \begin{subfigure}[b]{0.32\columnwidth}
  \includegraphics[width=\columnwidth]{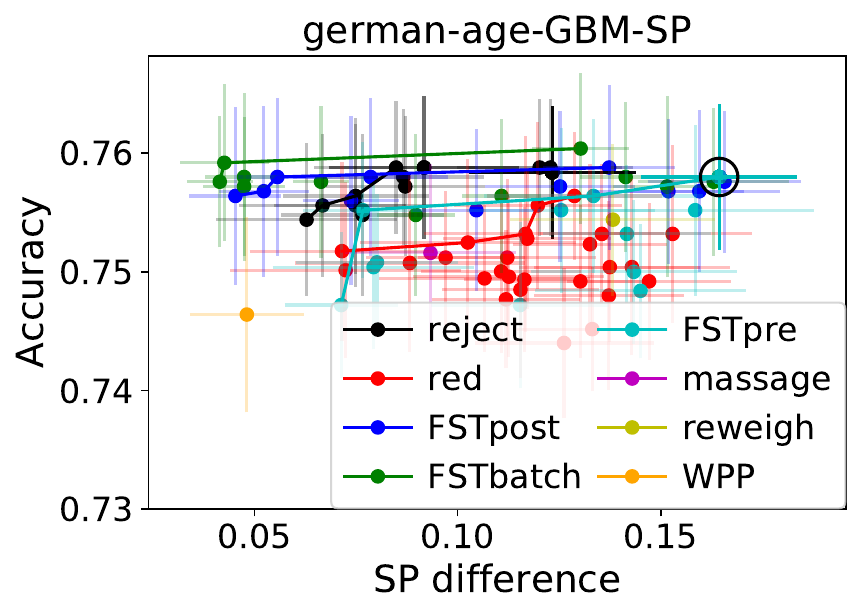}
  \label{fig:german_2_GBM_SP_acc_acc}
  \end{subfigure}
  \begin{subfigure}[b]{0.32\columnwidth}
  \includegraphics[width=\columnwidth]{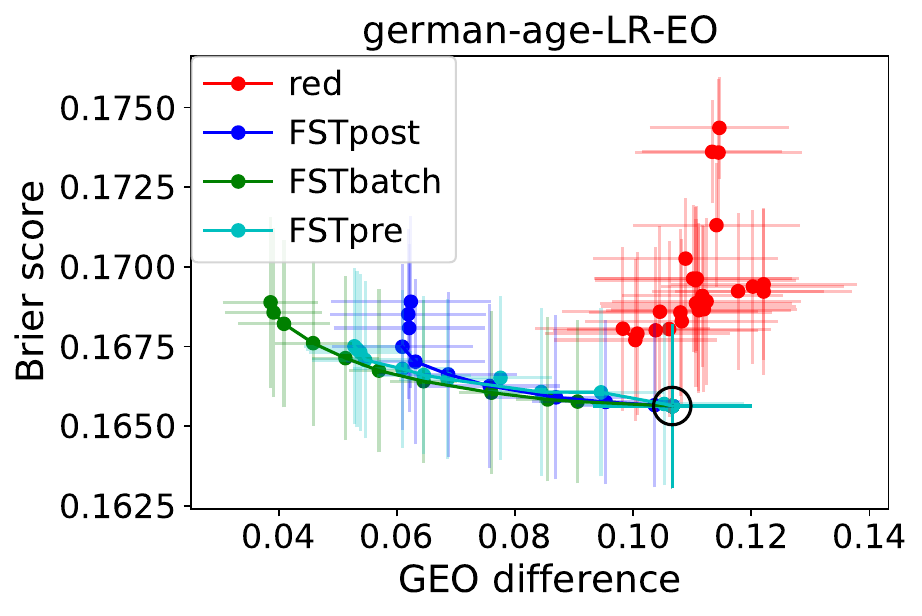}
  \label{fig:german_2_LR_EO_acc_Brier}
  \end{subfigure}
  \begin{subfigure}[b]{0.32\columnwidth}
  \includegraphics[width=\columnwidth]{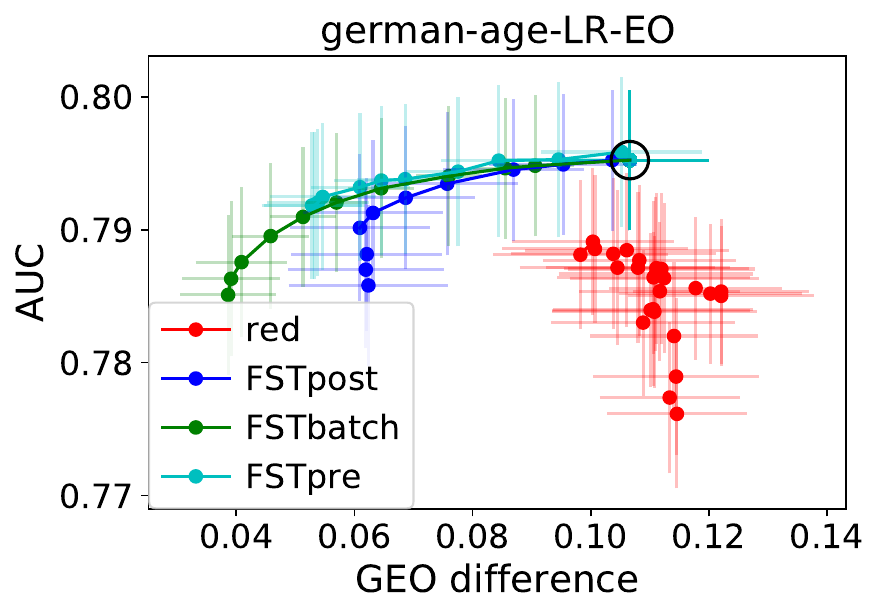}
  \label{fig:german_2_LR_EO_acc_AUC}
  \end{subfigure}
  \begin{subfigure}[b]{0.32\columnwidth}
  \includegraphics[width=\columnwidth]{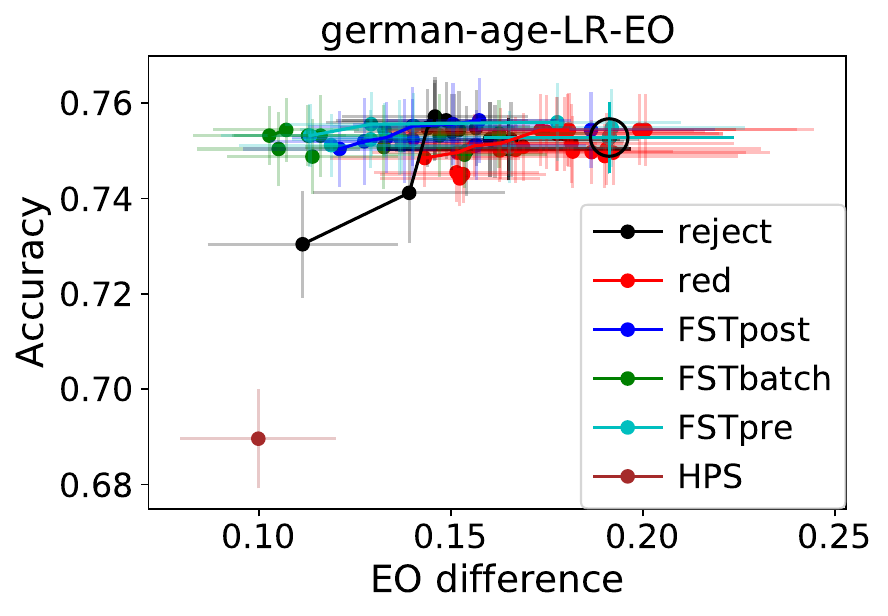}
  \label{fig:german_2_LR_EO_acc_acc}
  \end{subfigure}
  \begin{subfigure}[b]{0.32\columnwidth}
  \includegraphics[width=\columnwidth]{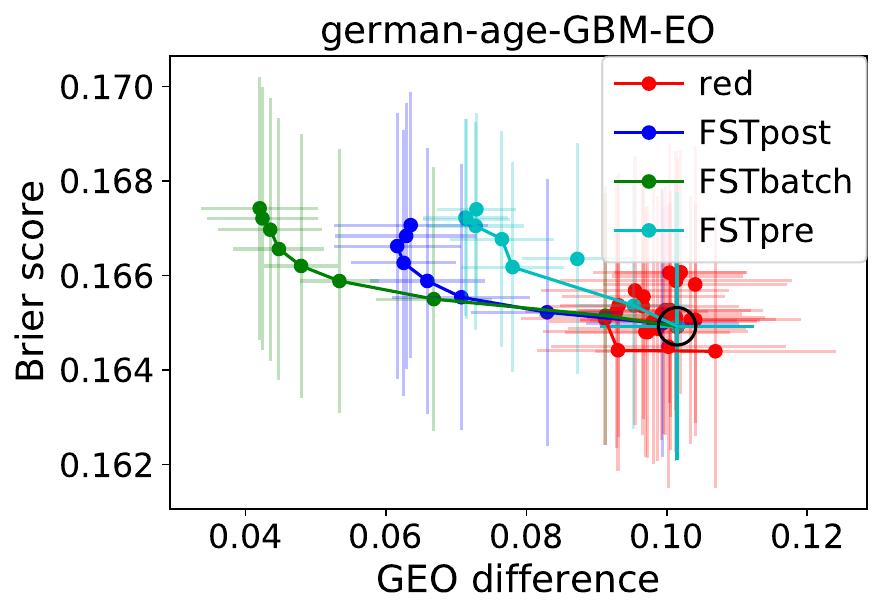}
  \label{fig:german_2_GBM_EO_acc_Brier}
  \end{subfigure}
  \begin{subfigure}[b]{0.32\columnwidth}
  \includegraphics[width=\columnwidth]{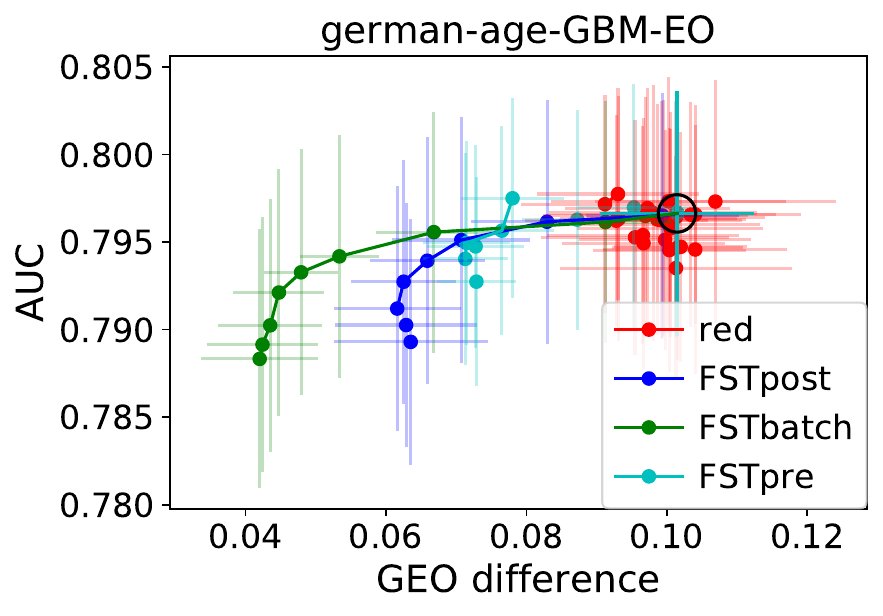}
  \label{fig:german_2_GBM_EO_acc_AUC}
  \end{subfigure}
  \begin{subfigure}[b]{0.32\columnwidth}
  \includegraphics[width=\columnwidth]{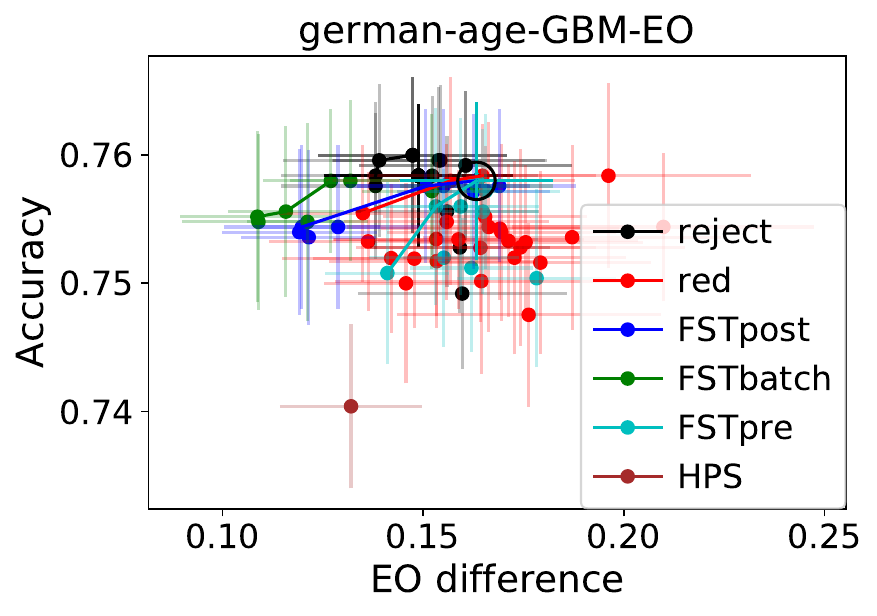}
  \label{fig:german_2_GBM_EO_acc_acc}
  \end{subfigure}
  \caption{Trade-offs between fairness and classification performance on the German credit data set with age as the protected attribute and the protected attribute included in the features.}
  \label{fig:german_2}
\end{figure}

\subsection{German Credit Risk Data Set}

\begin{figure}[t]
  \centering
  \begin{subfigure}[b]{0.32\columnwidth}
  \includegraphics[width=\columnwidth]{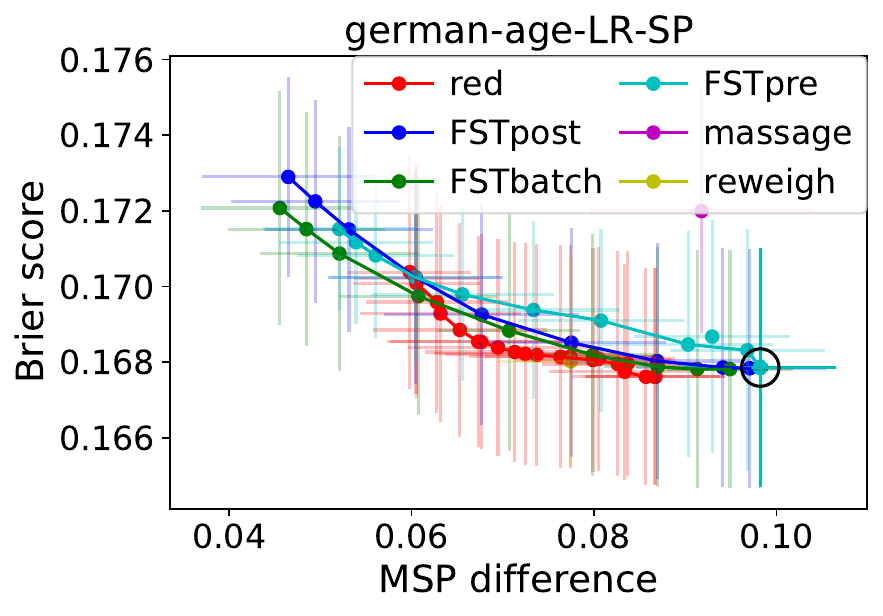}
  \label{fig:german_2_LR_SP_acc_Brier_False}
  \end{subfigure}
  \begin{subfigure}[b]{0.32\columnwidth}
  \includegraphics[width=\columnwidth]{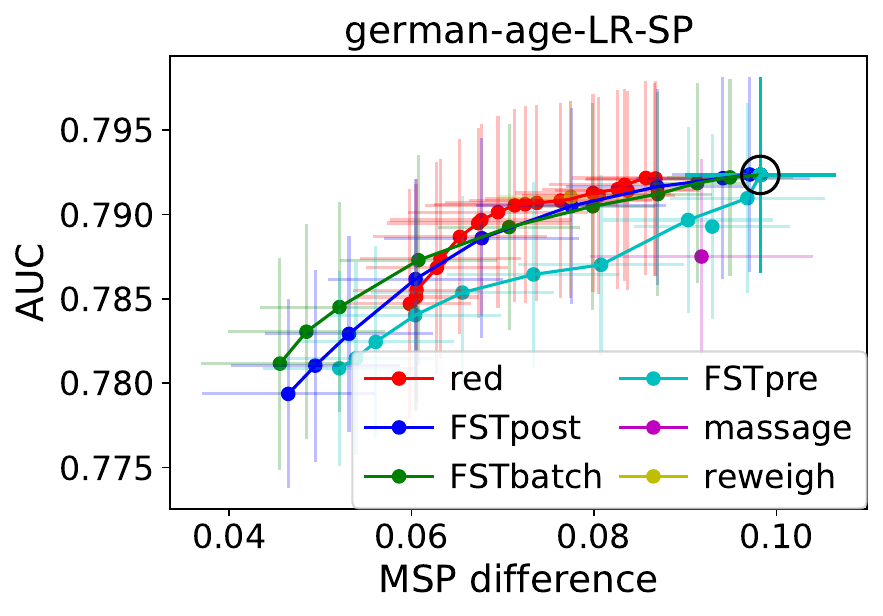}
  \label{fig:german_2_LR_SP_acc_AUC_False}
  \end{subfigure}
  \begin{subfigure}[b]{0.32\columnwidth}
  \includegraphics[width=\columnwidth]{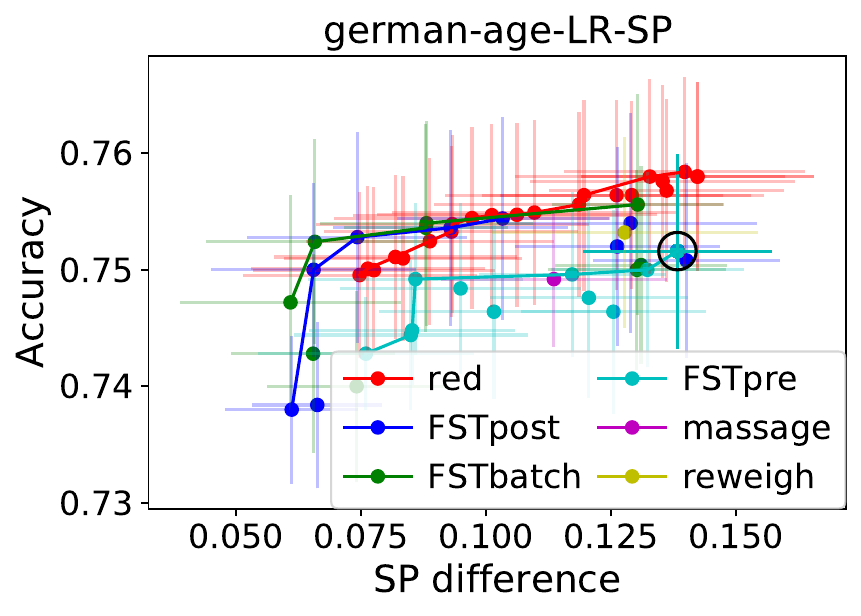}
  \label{fig:german_2_LR_SP_acc_acc_False}
  \end{subfigure}
  \begin{subfigure}[b]{0.32\columnwidth}
  \includegraphics[width=\columnwidth]{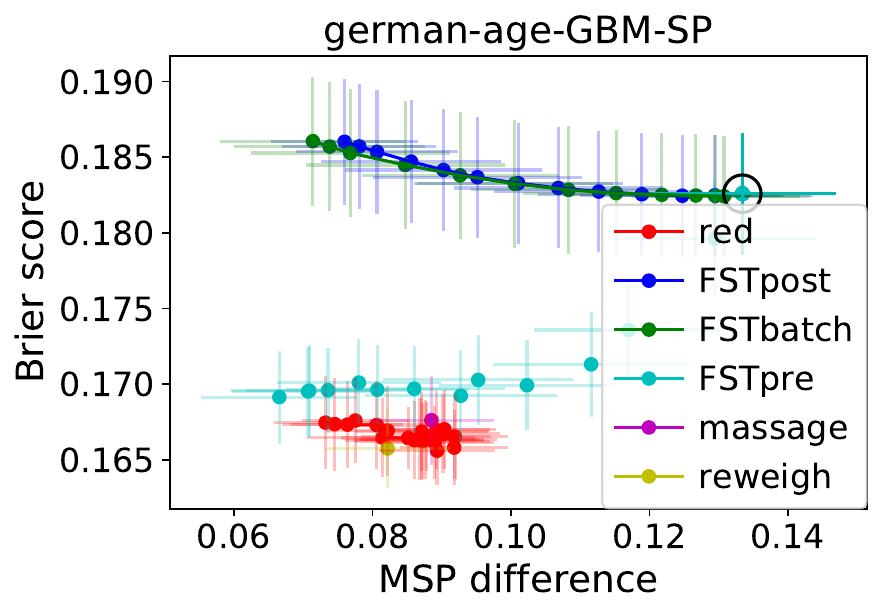}
  \label{fig:german_2_GBM_SP_acc_Brier_False}
  \end{subfigure}
  \begin{subfigure}[b]{0.32\columnwidth}
  \includegraphics[width=\columnwidth]{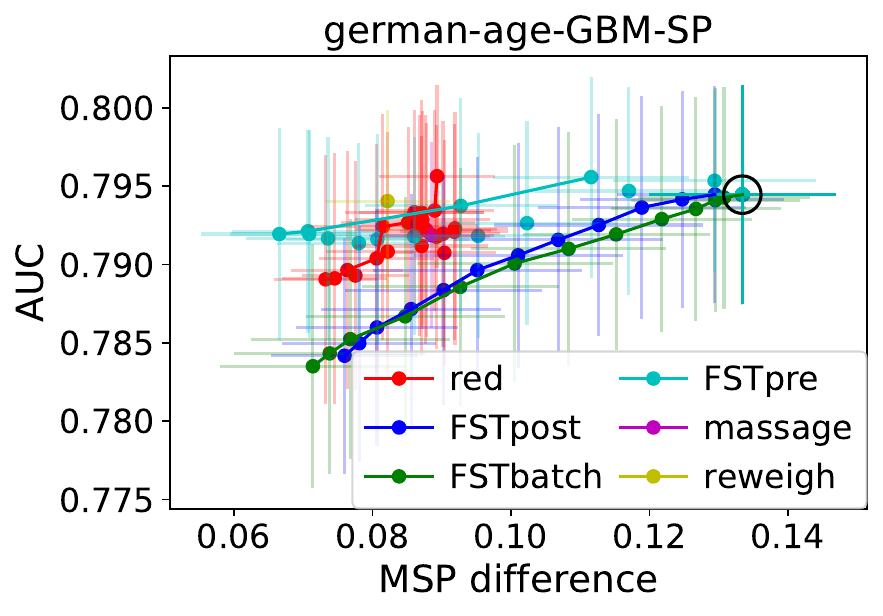}
  \label{fig:german_2_GBM_SP_acc_AUC_False}
  \end{subfigure}
  \begin{subfigure}[b]{0.32\columnwidth}
  \includegraphics[width=\columnwidth]{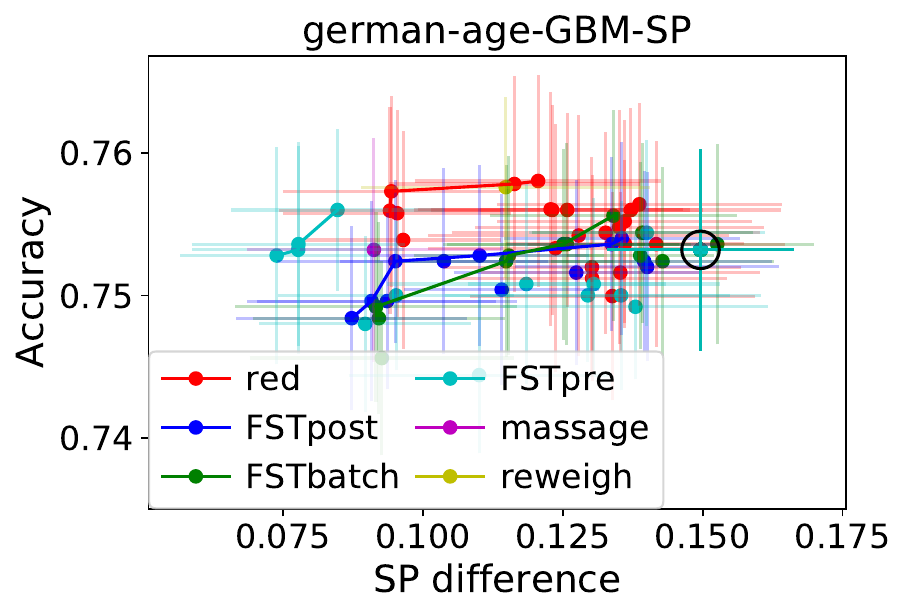}
  \label{fig:german_2_GBM_SP_acc_acc_False}
  \end{subfigure}
  \begin{subfigure}[b]{0.32\columnwidth}
  \includegraphics[width=\columnwidth]{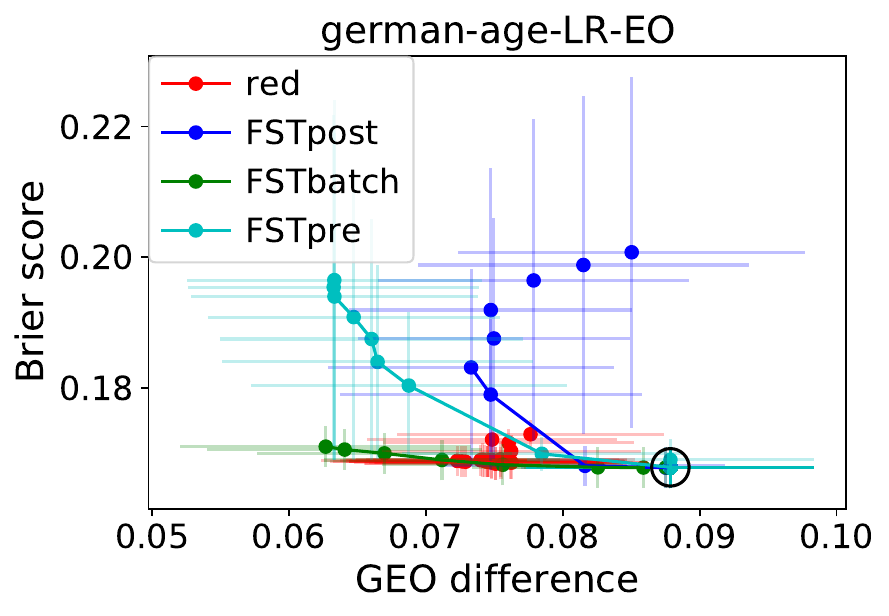}
  \label{fig:german_2_LR_EO_acc_Brier_False}
  \end{subfigure}
  \begin{subfigure}[b]{0.32\columnwidth}
  \includegraphics[width=\columnwidth]{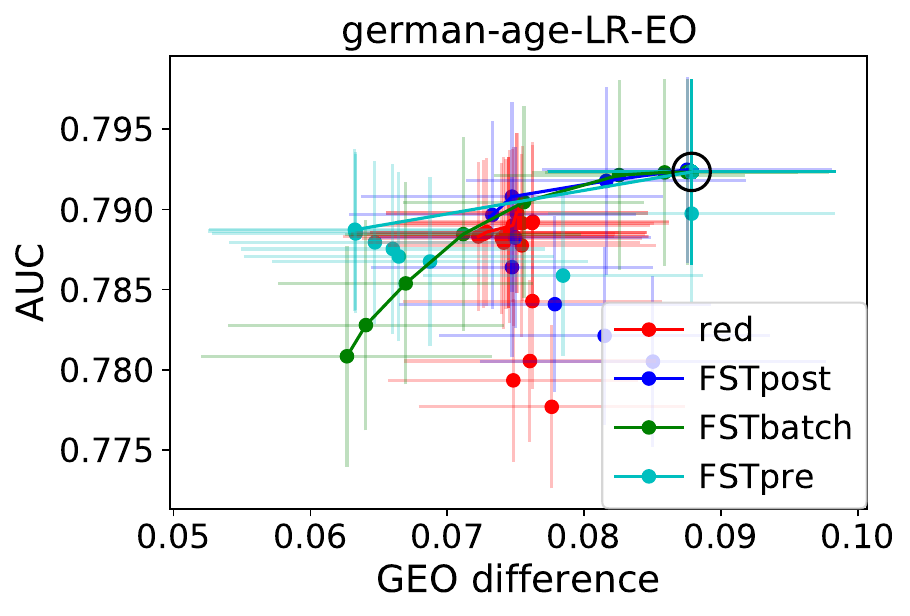}
  \label{fig:german_2_LR_EO_acc_AUC_False}
  \end{subfigure}
  \begin{subfigure}[b]{0.32\columnwidth}
  \includegraphics[width=\columnwidth]{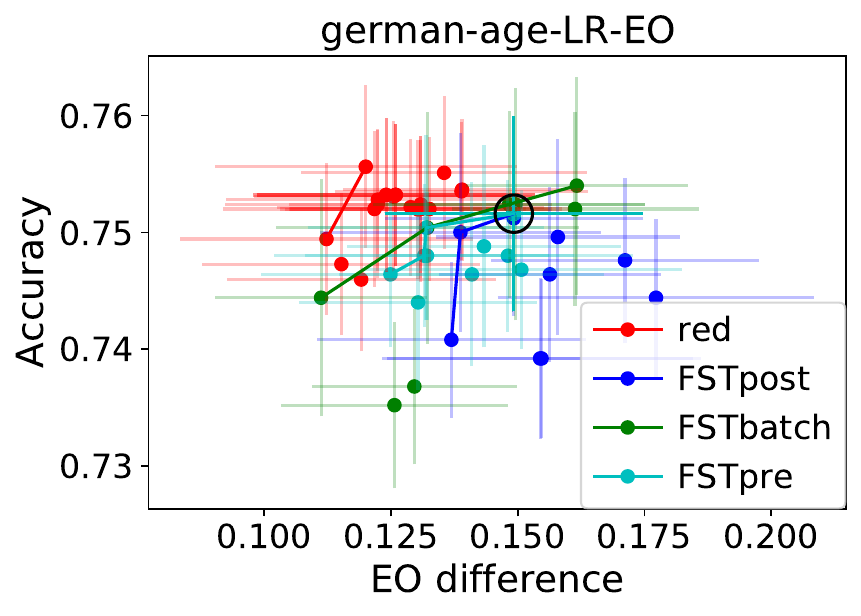}
  \label{fig:german_2_LR_EO_acc_acc_False}
  \end{subfigure}
  \begin{subfigure}[b]{0.32\columnwidth}
  \includegraphics[width=\columnwidth]{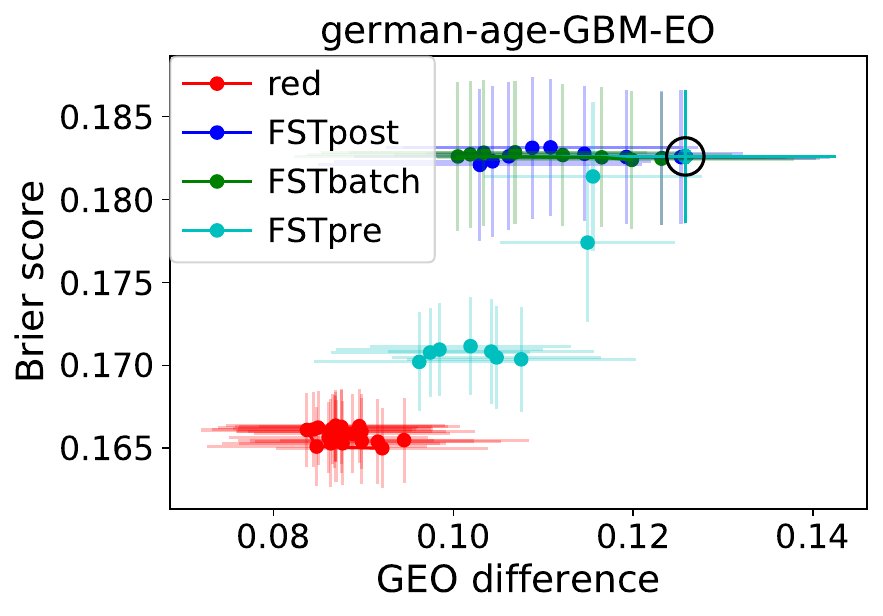}
  \label{fig:german_2_GBM_EO_acc_Brier_False}
  \end{subfigure}
  \begin{subfigure}[b]{0.32\columnwidth}
  \includegraphics[width=\columnwidth]{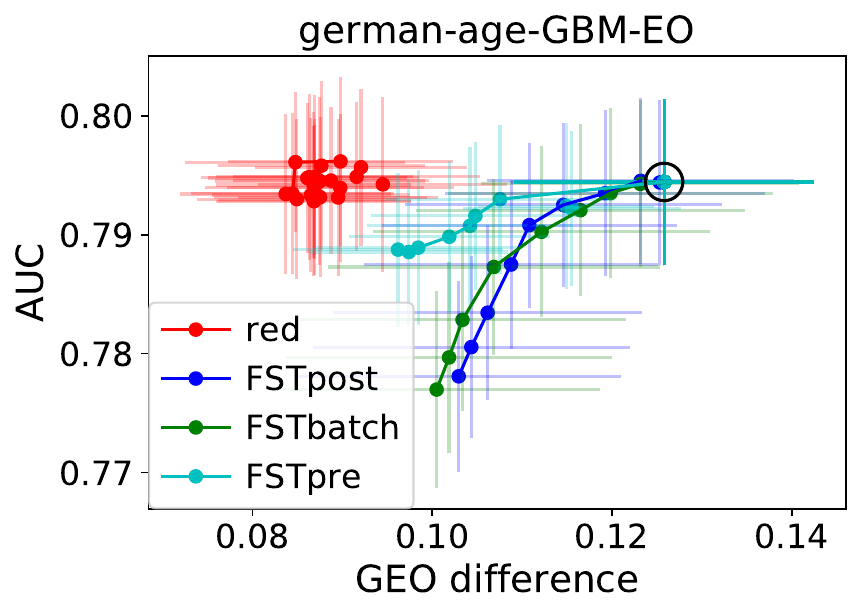}
  \label{fig:german_2_GBM_EO_acc_AUC_False}
  \end{subfigure}
  \begin{subfigure}[b]{0.32\columnwidth}
  \includegraphics[width=\columnwidth]{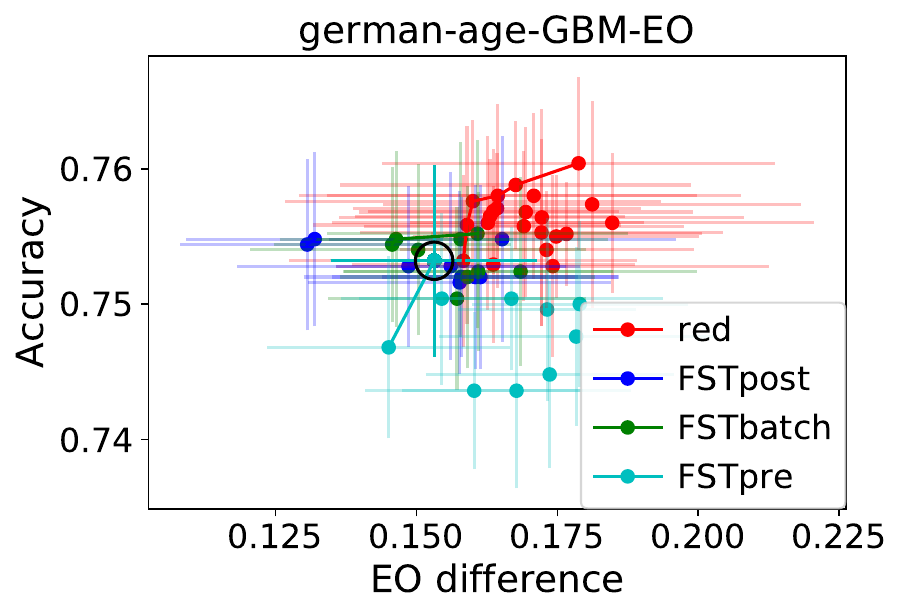}
  \label{fig:german_2_GBM_EO_acc_acc_False}
  \end{subfigure}
  \caption{Trade-offs between fairness and classification performance on the German credit data set with age as the protected attribute and the protected attribute excluded from the features.}
  \label{fig:german_2_False}
\end{figure}

Figures~\ref{fig:german_2} and \ref{fig:german_2_False} depict trade-offs between classification performance and fairness for the German credit data set, where the protected attribute of age is either included in or excluded from the features. These results are included for completeness as German is a standard data set in the fairness literature, but the small data set size and consequently large error bars make it difficult to draw conclusions.

\subsection{Individual Comparisons with Existing Methods}

\begin{figure*}[t]
  \centering
  \footnotesize
  \begin{subfigure}[b]{0.32\columnwidth}
  \includegraphics[width=\columnwidth]{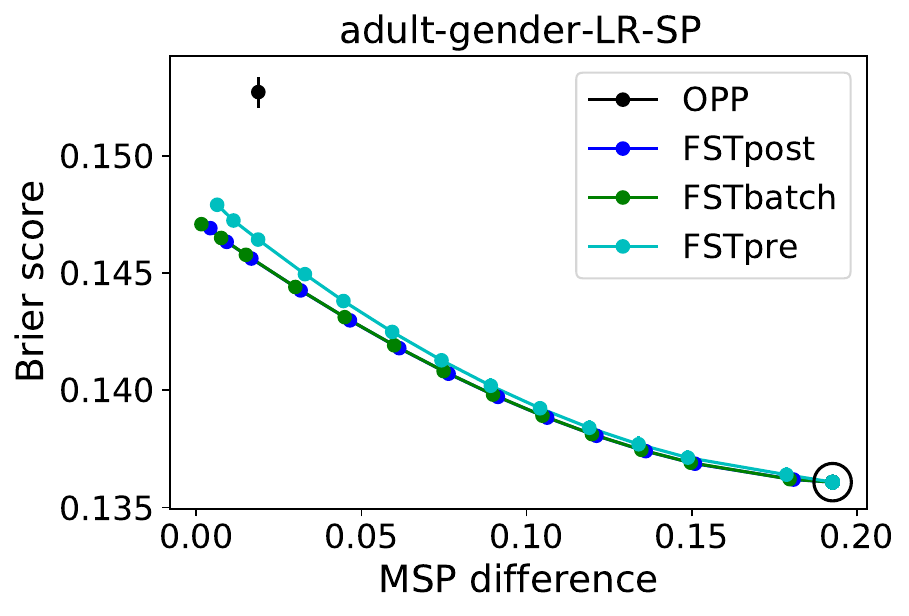}
  \label{fig:adult_1_LR_SP_acc_Brier_reduced}
  \end{subfigure}
  \begin{subfigure}[b]{0.32\columnwidth}
  \includegraphics[width=\columnwidth]{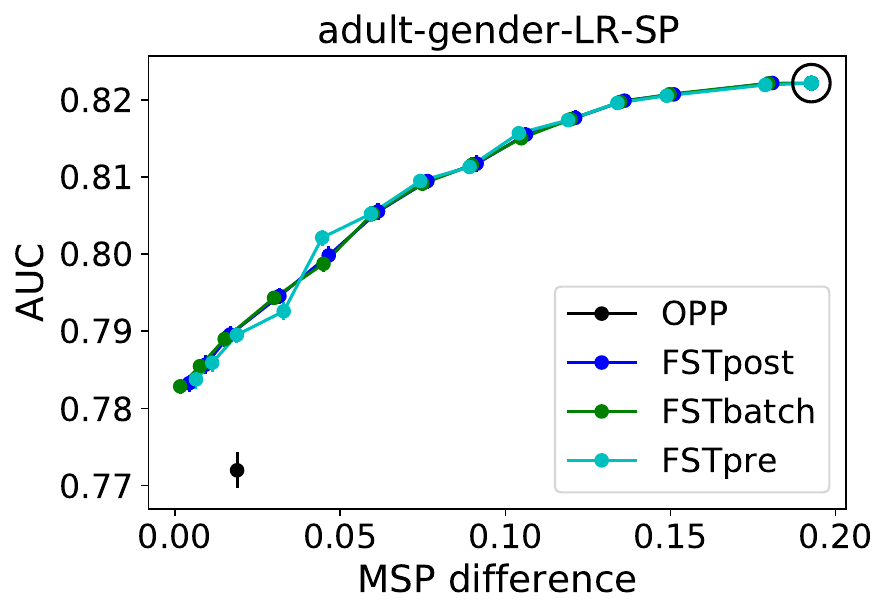}
  \label{fig:adult_1_LR_SP_acc_AUC_reduced}
  \end{subfigure}
  \begin{subfigure}[b]{0.32\columnwidth}
  \includegraphics[width=\columnwidth]{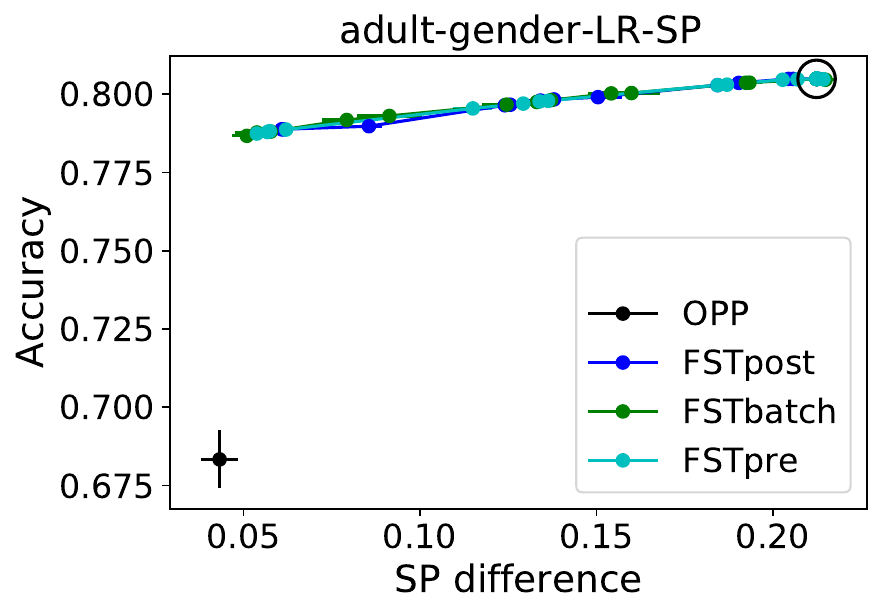}
  \label{fig:adult_1_LR_SP_acc_acc_reduced}
  \end{subfigure}
  \begin{subfigure}[b]{0.32\columnwidth}
  \includegraphics[width=\columnwidth]{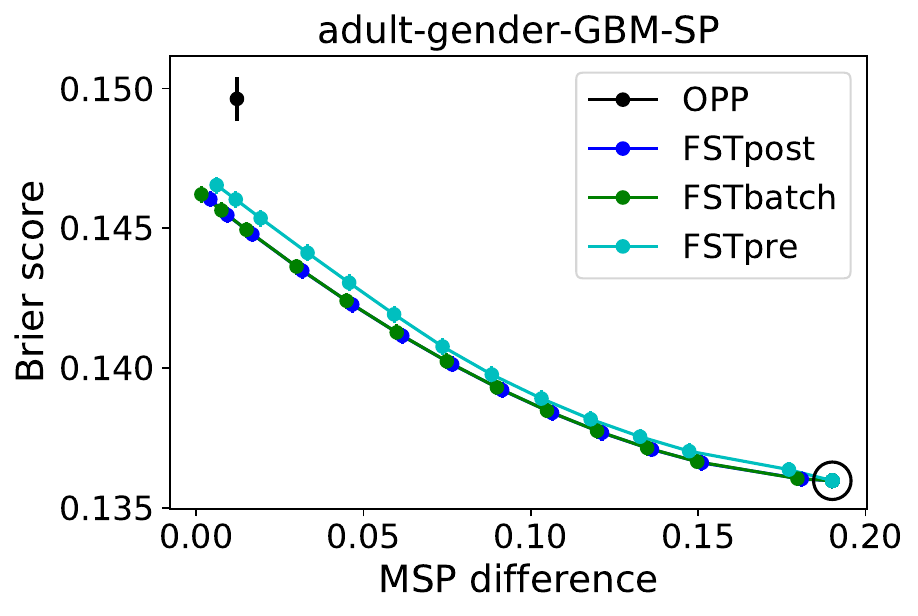}
  \label{fig:adult_1_GBM_SP_acc_Brier_reduced}
  \end{subfigure}
  \begin{subfigure}[b]{0.32\columnwidth}
  \includegraphics[width=\columnwidth]{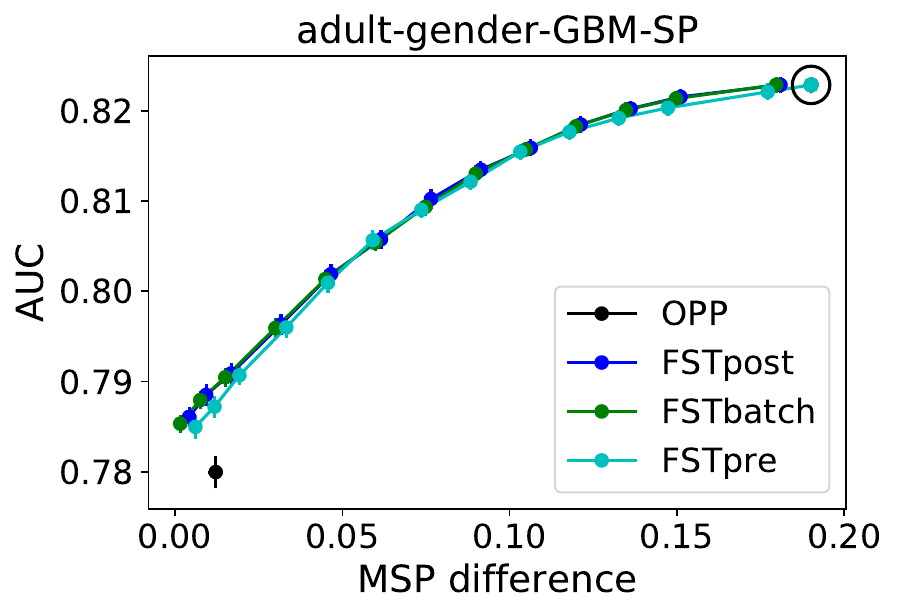}
  \label{fig:adult_1_GBM_SP_acc_AUC_reduced}
  \end{subfigure}
  \begin{subfigure}[b]{0.32\columnwidth}
  \includegraphics[width=\columnwidth]{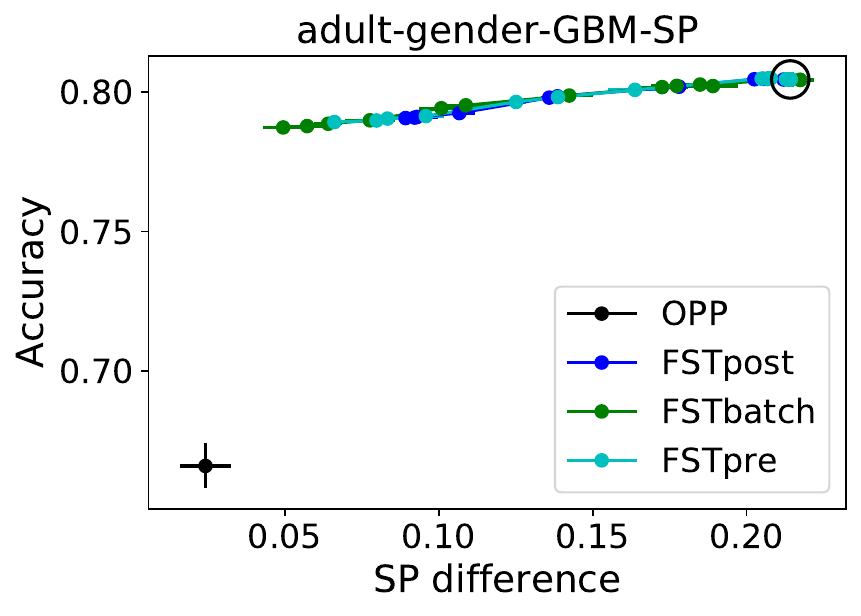}
  \label{fig:adult_1_GBM_SP_acc_acc_reduced}
  \end{subfigure}
  \caption{Trade-offs between statistical parity and classification performance measures for the Adult data set with a reduced set of features.}
  \label{fig:tradeoffsReduced}
\end{figure*}

As mentioned in Section~\ref{sec:expt}, we encountered computational difficulties in running the optimized pre-processing \citep[OPP,][]{calmon2017} and disparate mistreatment in-processing \citep[DM,][]{zafar2017fairness} methods. In the case of OPP, the method does not scale beyond feature dimensions of $\sim 5$. We have thus conducted separate experiments in which the set of features has been reduced.
Figure \ref{fig:tradeoffsReduced} shows the resulting trade-offs between statistical parity, which is what OPP addresses, and classification performance for the Adult data set. 
This limited comparison suggests that OPP is not competitive with FST. Unfortunately we were unable to obtain reasonable results for OPP on other data sets so do not show them here.

\begin{figure*}[t]
  \centering
  \footnotesize
  \begin{subfigure}[b]{0.32\columnwidth}
  \includegraphics[width=\columnwidth]{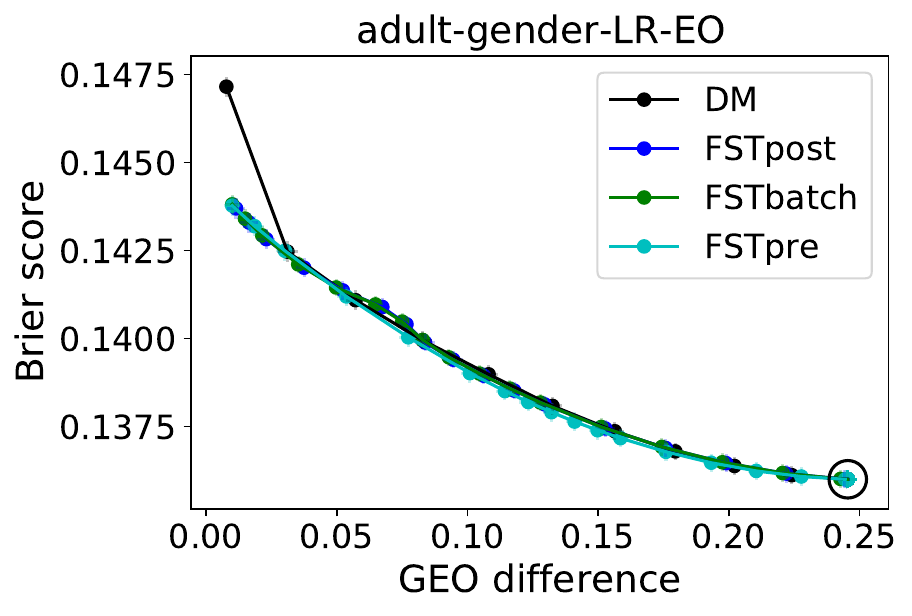}
  \label{fig:adult_1_LR_EO_acc_Brier_reduced}
  \end{subfigure}
  \begin{subfigure}[b]{0.32\columnwidth}
  \includegraphics[width=\columnwidth]{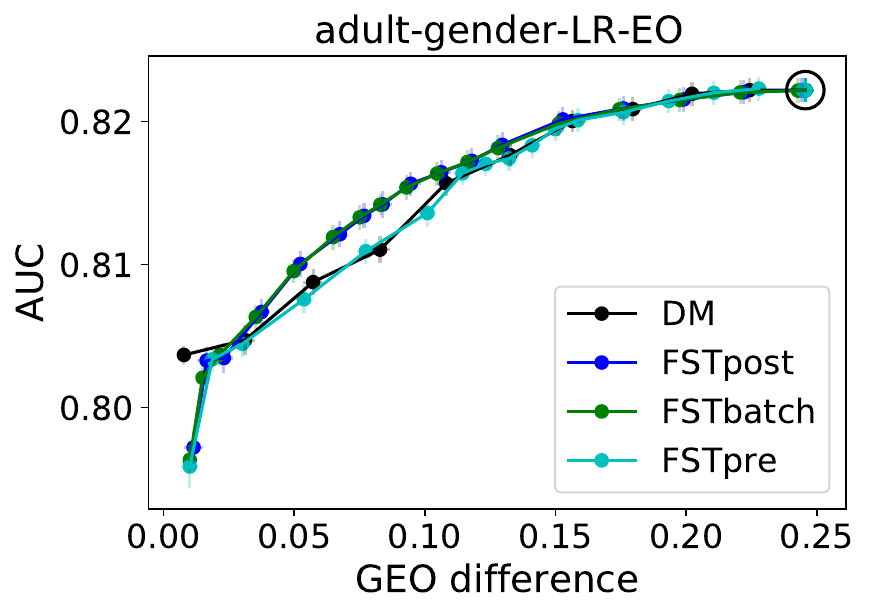}
  \label{fig:adult_1_LR_EO_acc_AUC_reduced}
  \end{subfigure}
  \begin{subfigure}[b]{0.32\columnwidth}
  \includegraphics[width=\columnwidth]{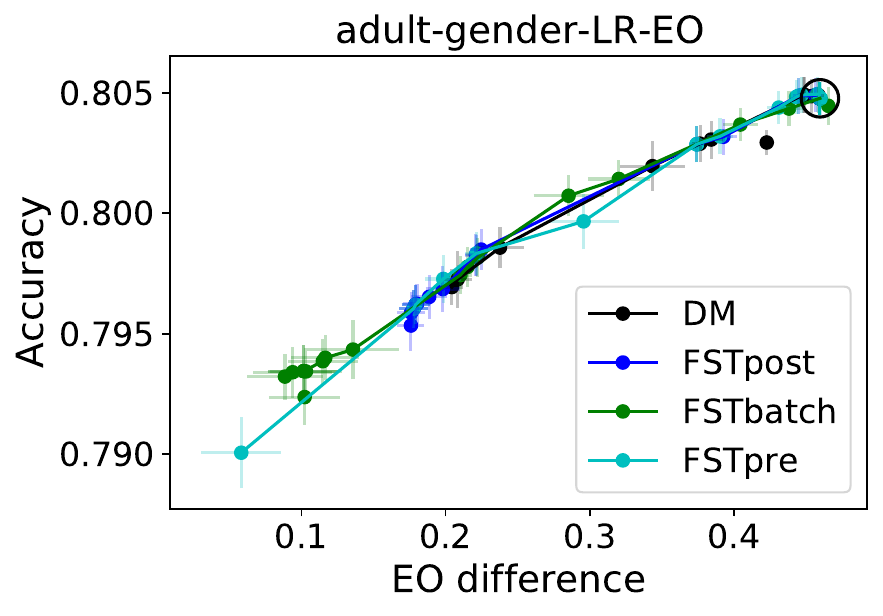}
  \label{fig:adult_1_LR_EO_acc_acc_reduced}
  \end{subfigure}
  \begin{subfigure}[b]{0.32\columnwidth}
  \includegraphics[width=\columnwidth]{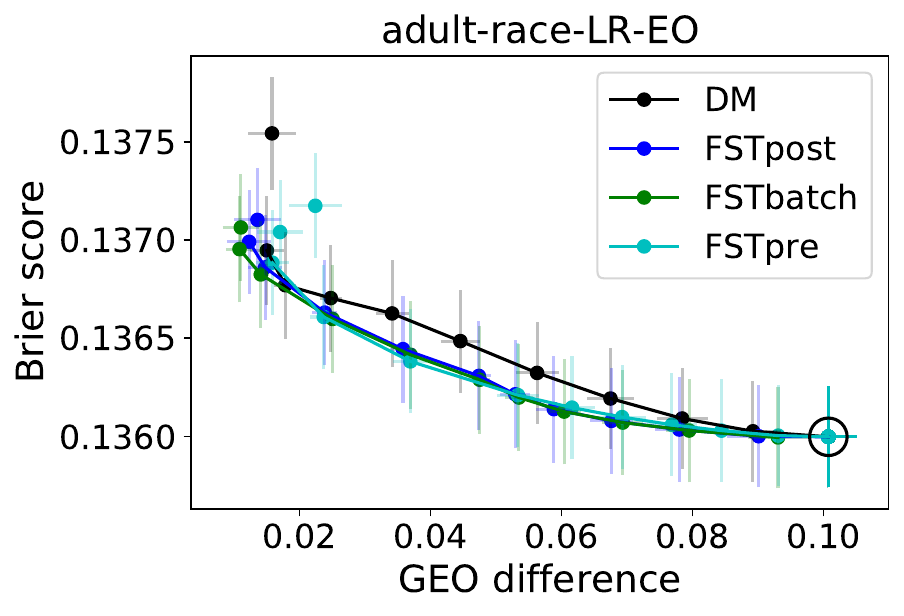}
  \label{fig:adult_2_LR_EO_acc_Brier_reduced}
  \end{subfigure}
  \begin{subfigure}[b]{0.32\columnwidth}
  \includegraphics[width=\columnwidth]{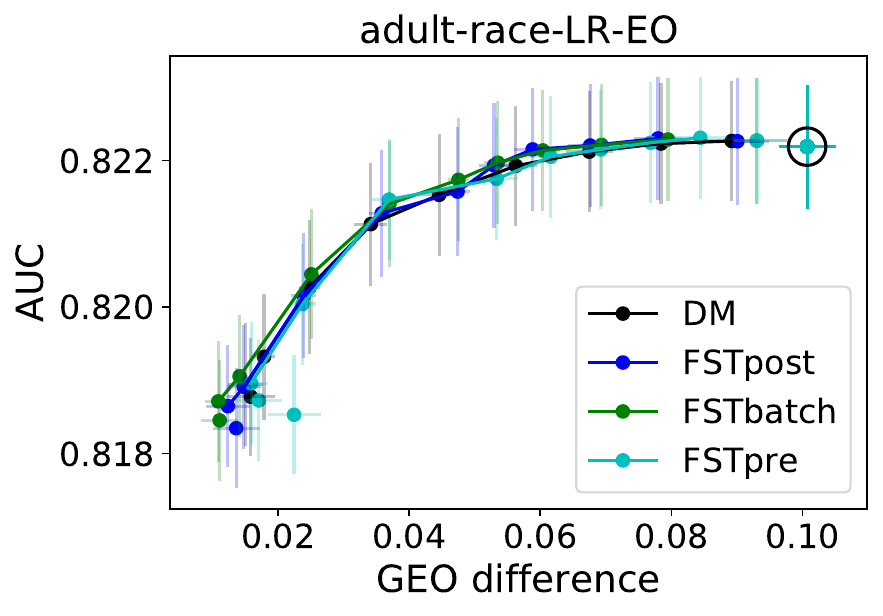}
  \label{fig:adult_2_LR_EO_acc_AUC_reduced}
  \end{subfigure}
  \begin{subfigure}[b]{0.32\columnwidth}
  \includegraphics[width=\columnwidth]{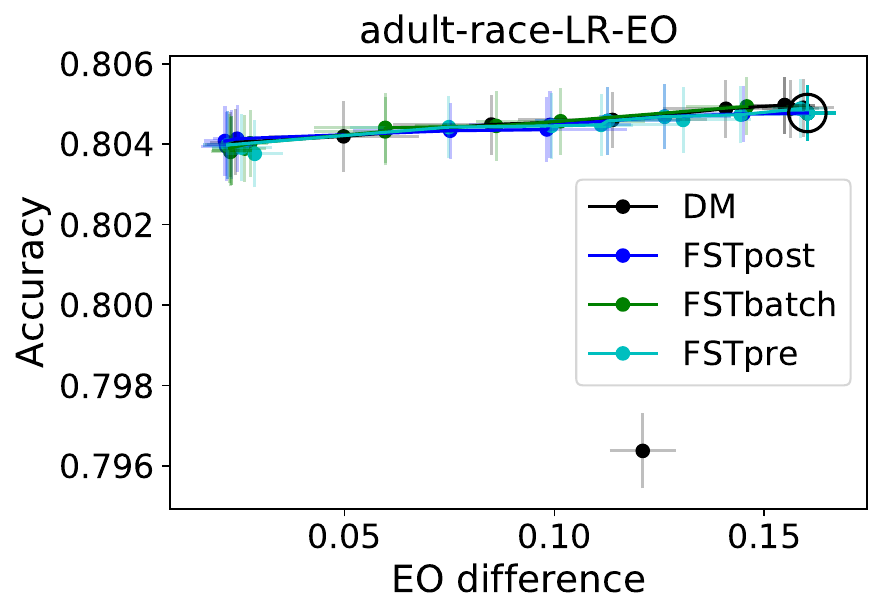}
  \label{fig:adult_2_LR_EO_acc_acc_reduced}
  \end{subfigure}
  \begin{subfigure}[b]{0.32\columnwidth}
  \includegraphics[width=\columnwidth]{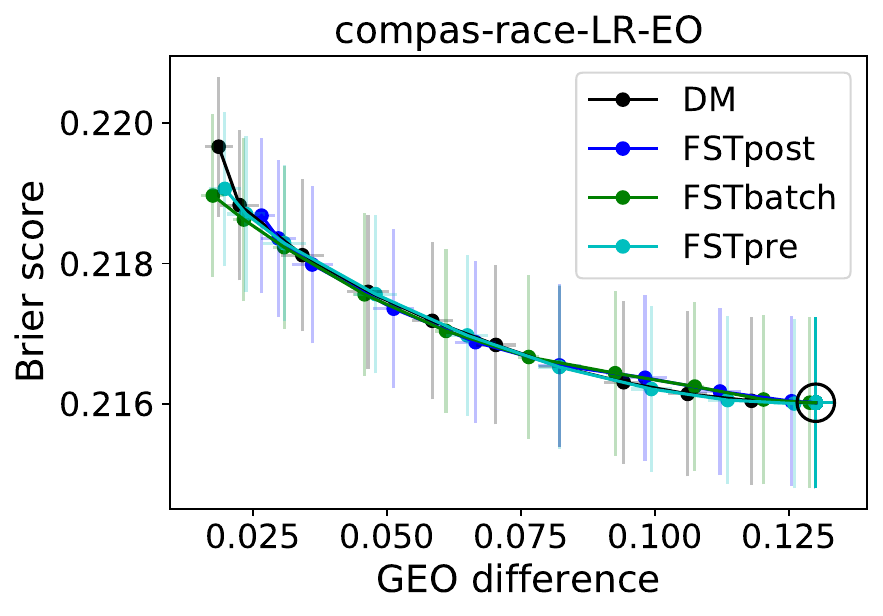}
  \label{fig:compas_2_LR_EO_acc_Brier_reduced}
  \end{subfigure}
  \begin{subfigure}[b]{0.32\columnwidth}
  \includegraphics[width=\columnwidth]{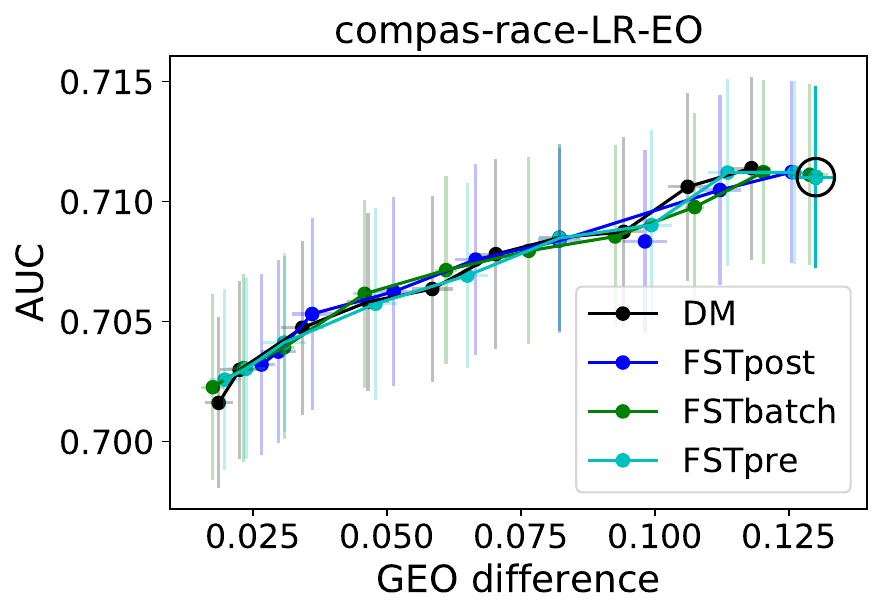}
  \label{fig:compas_2_LR_EO_acc_AUC_reduced}
  \end{subfigure}
  \begin{subfigure}[b]{0.32\columnwidth}
  \includegraphics[width=\columnwidth]{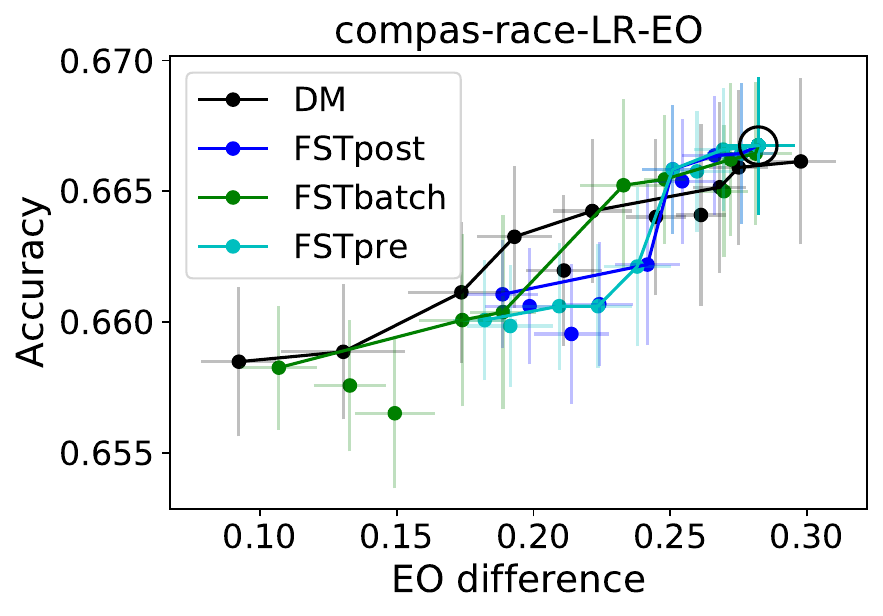}
  \label{fig:compas_2_LR_EO_acc_acc_reduced}
  \end{subfigure}
  \begin{subfigure}[b]{0.32\columnwidth}
  \includegraphics[width=\columnwidth]{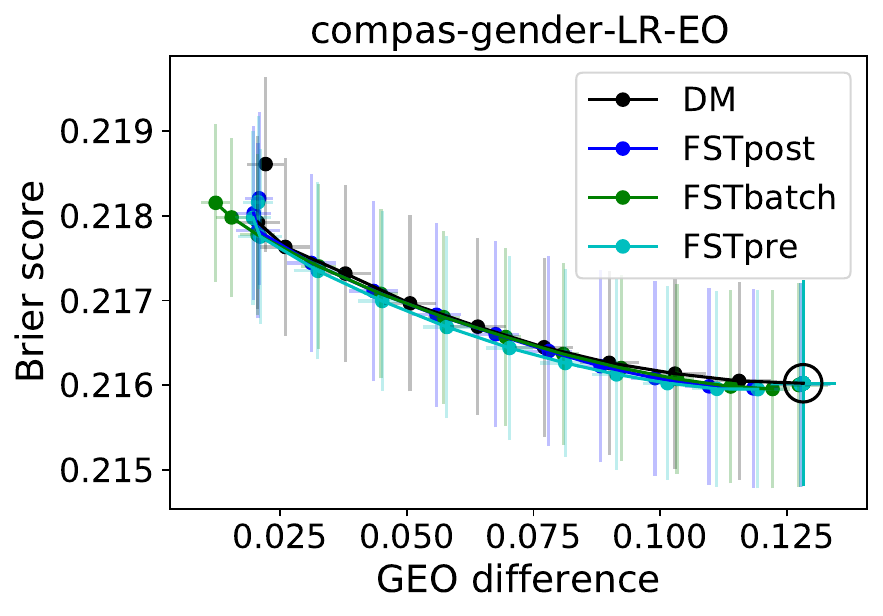}
  \label{fig:compas_1_LR_EO_acc_Brier_reduced}
  \end{subfigure}
  \begin{subfigure}[b]{0.32\columnwidth}
  \includegraphics[width=\columnwidth]{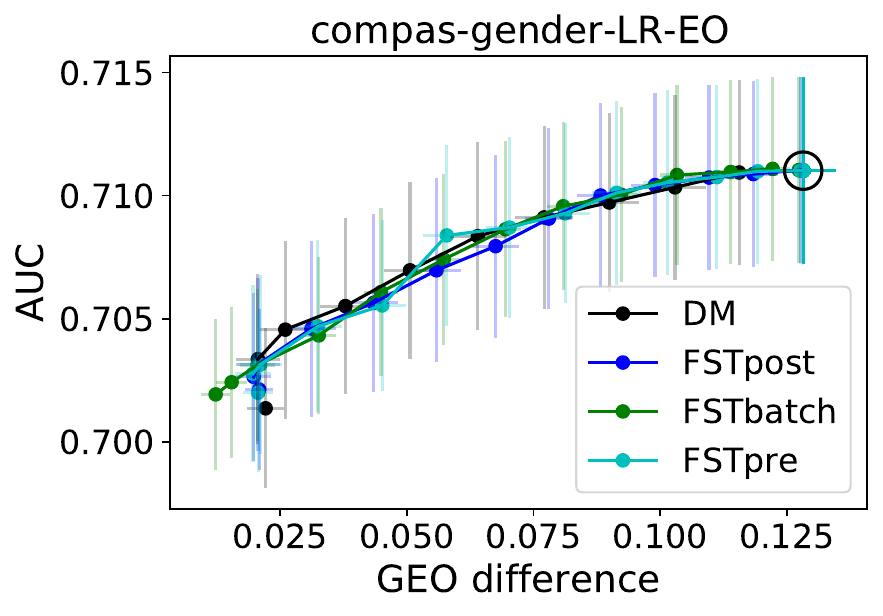}
  \label{fig:compas_1_LR_EO_acc_AUC_reduced}
  \end{subfigure}
  \begin{subfigure}[b]{0.32\columnwidth}
  \includegraphics[width=\columnwidth]{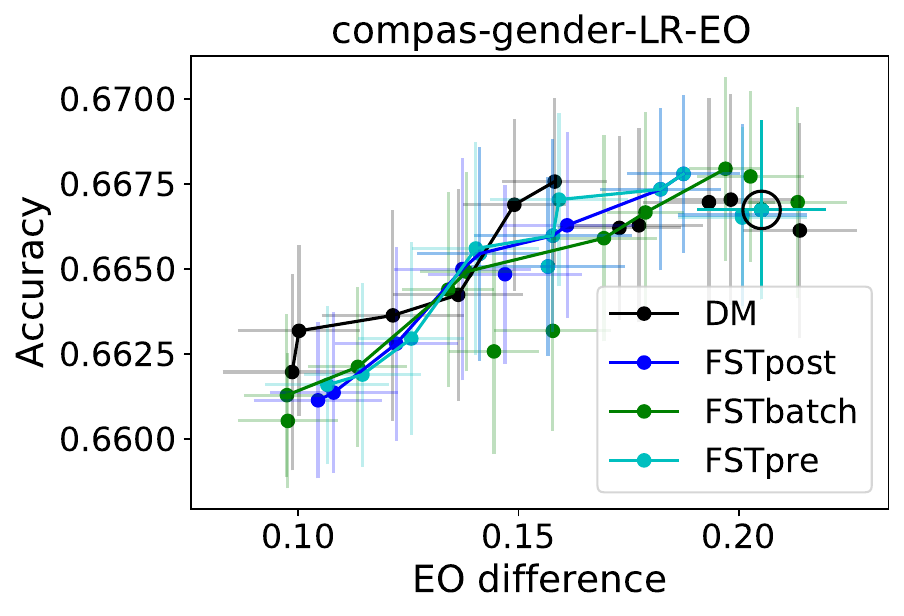}
  \label{fig:compas_1_LR_EO_acc_acc_reduced}
  \end{subfigure}
  \caption{Trade-offs between equalized odds and classification performance measures for the Adult and COMPAS data sets with a reduced set of features.} 
  \label{fig:tradeoffsReducedEO}
\end{figure*}

In the case of DM, when we ran the code\footnote{\url{https://github.com/mbilalzafar/fair-classification}} on data sets with a full set of features, the optimization either failed to converge or when it did converge, did not appreciably decrease the EO difference from that of an unconstrained logistic regression classifier. (The latter problem was also observed with FC, \citealp{zafar2017constraints}, in Figures~\ref{fig:compas_1}--\ref{fig:meps_1}.) We used a constraint type of 4 to impose both FNR and FPR constraints, in keeping with EO, and default values for the disciplined convex-concave programming (DCCP) parameters $\tau$ and $\mu$. For example on the Adult-gender combination, DM failed on 4 of the 10 training folds and converged on the others with little effect, while on the COMPAS-race combination, DM failed on 9 of 10 folds. We noticed that our version of the COMPAS data set has much higher dimension than the one used by \citet{zafar2017fairness}, due primarily to including a charge description feature and after one-hot encoding of categorical variables. Thus we opted to compare with DM using reduced feature sets, as with OPP. Figure~\ref{fig:tradeoffsReducedEO} shows the trade-offs obtained on the Adult and COMPAS data sets. On COMPAS, all methods are remarakably similar while on Adult, DM might be slightly worse. For example on Adult-gender (first row), DM does not reduce the EO difference below $0.2$ (right panel). We reiterate however our lack of success with DM on full-dimensional data sets.

\begin{figure*}[t]
  \centering
  \footnotesize
  \begin{subfigure}[b]{0.32\columnwidth}
  \includegraphics[width=\columnwidth]{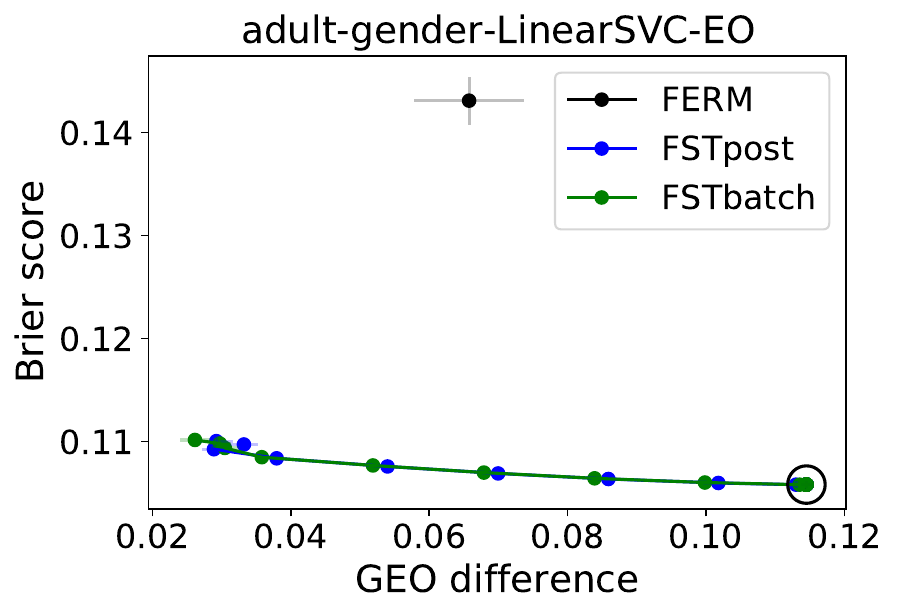}
  \label{fig:adult_1_LinearSVC_EO_acc_Brier}
  \end{subfigure}
  \begin{subfigure}[b]{0.32\columnwidth}
  \includegraphics[width=\columnwidth]{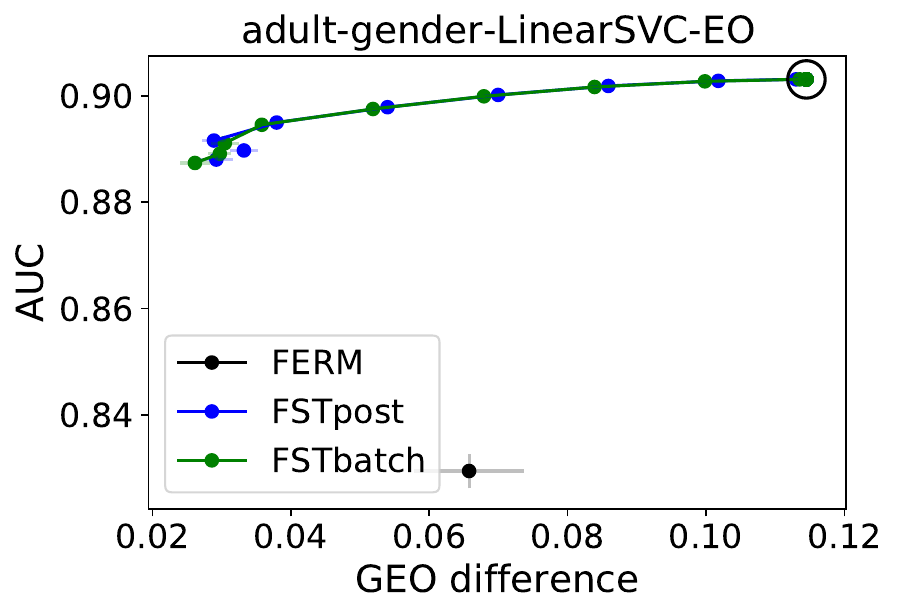}
  \label{fig:adult_1_LinearSVC_EO_acc_AUC}
  \end{subfigure}
  \begin{subfigure}[b]{0.32\columnwidth}
  \includegraphics[width=\columnwidth]{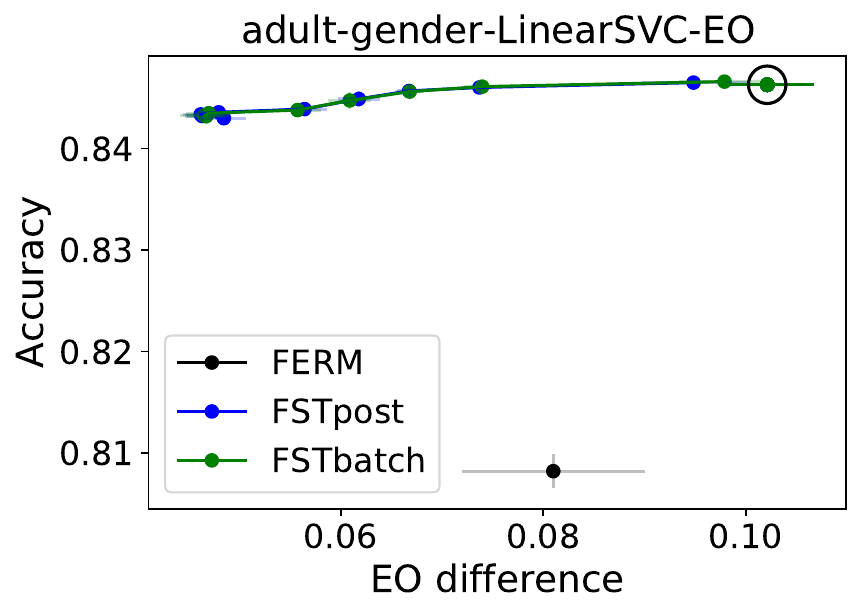}
  \label{fig:adult_1_LinearSVC_EO_acc_acc}
  \end{subfigure}
  \begin{subfigure}[b]{0.32\columnwidth}
  \includegraphics[width=\columnwidth]{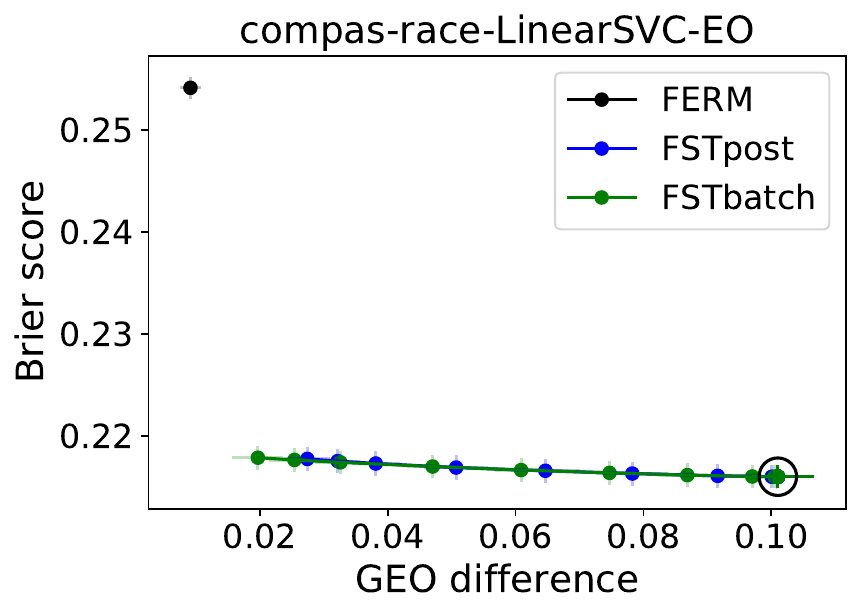}
  \label{fig:compas_2_LinearSVC_EO_acc_Brier}
  \end{subfigure}
  \begin{subfigure}[b]{0.33\columnwidth}
  \includegraphics[width=\columnwidth]{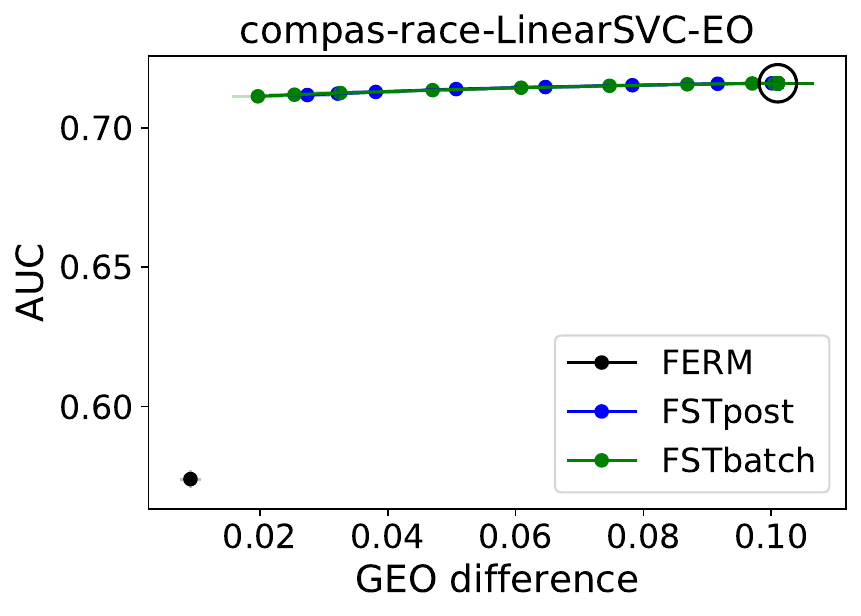}
  \label{fig:compas_2_LinearSVC_EO_acc_AUC}
  \end{subfigure}
  \begin{subfigure}[b]{0.32\columnwidth}
  \includegraphics[width=\columnwidth]{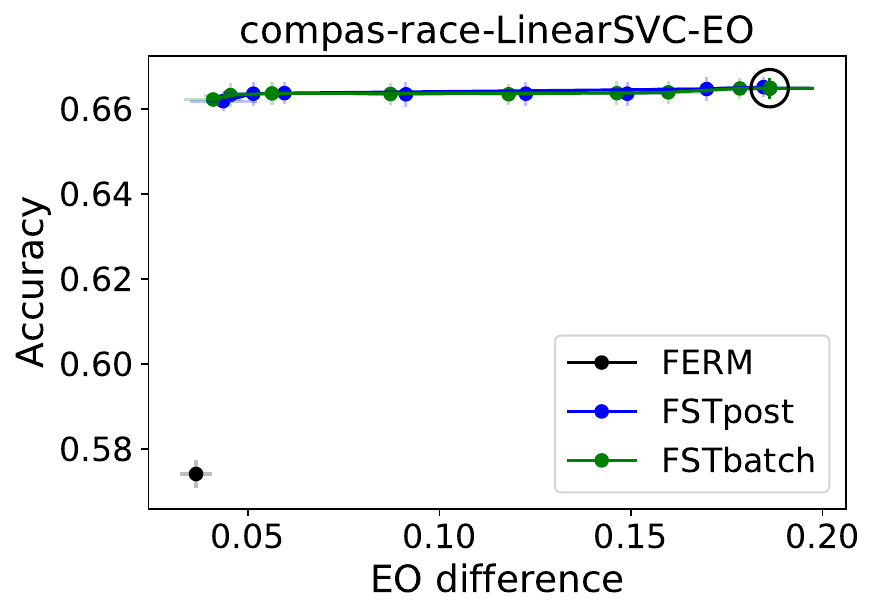}
  \label{fig:compas_2_LinearSVC_EO_acc_acc}
  \end{subfigure}
  \begin{subfigure}[b]{0.32\columnwidth}
  \includegraphics[width=\columnwidth]{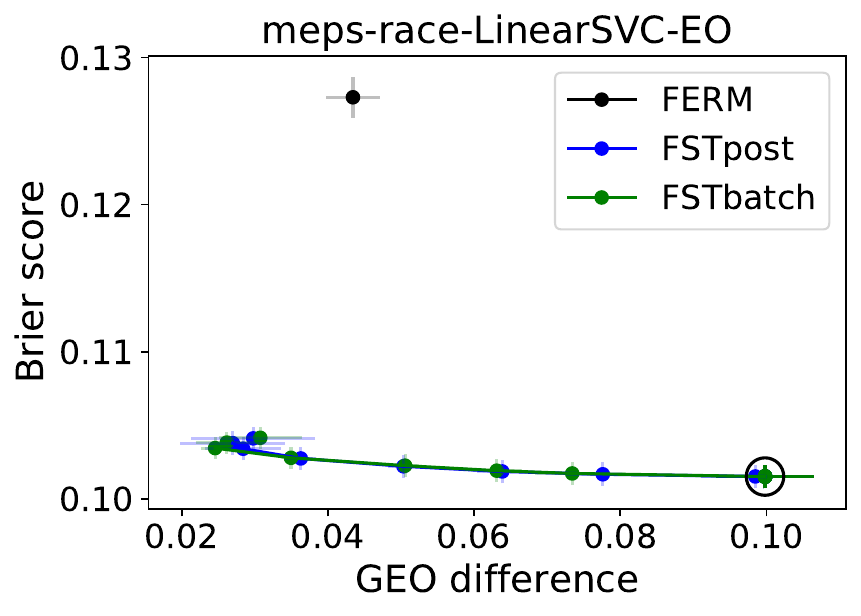}
  \label{fig:meps_1_LinearSVC_EO_acc_Brier}
  \end{subfigure}
  \begin{subfigure}[b]{0.32\columnwidth}
  \includegraphics[width=\columnwidth]{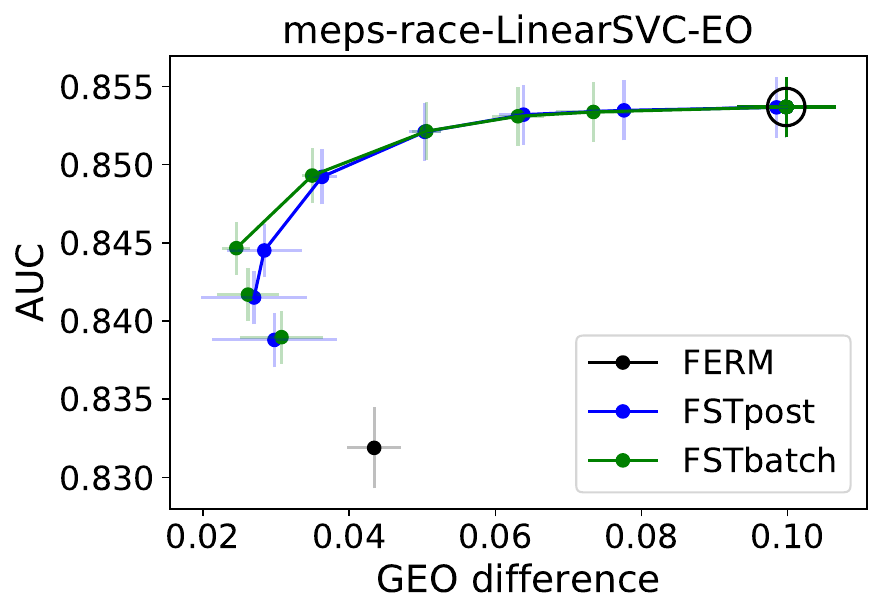}
  \label{fig:meps_1_LinearSVC_EO_acc_AUC}
  \end{subfigure}
  \begin{subfigure}[b]{0.32\columnwidth}
  \includegraphics[width=\columnwidth]{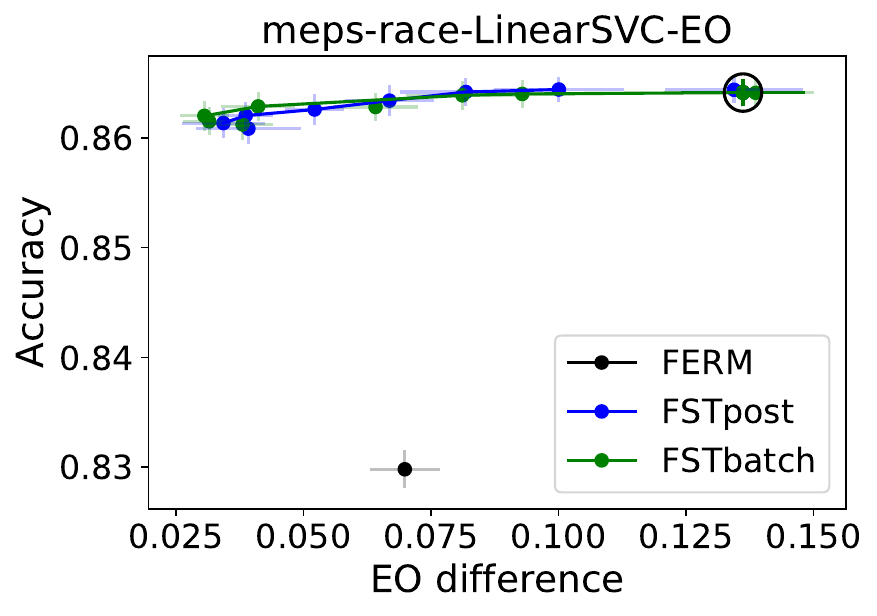}
  \label{fig:meps_1_LinearSVC_EO_acc_acc}
  \end{subfigure}
  \caption{Trade-offs between fairness and classification performance measures for FERM \citep{donini2018} and our proposed FST approaches.}
  \label{fig:comparisonsFERM}
\end{figure*}

We also compare FST to the Fair Empirical Risk Minimization (FERM) approach of \citet{donini2018}. We use the code\footnote{\url{https://github.com/jmikko/fair_ERM}} provided by the authors. FERM, although a general principle, has been specified only for binary classification problems with hinge loss as the loss function, and equal opportunity as the fairness constraint in \citet{donini2018}. The code provided by the authors implements linear and kernel support vector classifiers (SVC) with an equal opportunity constraint between two protected groups. During our experimentation, we observed that kernel SVC formulations of FERM were computationally impractical for the data sets we used (Adult, COMPAS, and MEPS). For example, experiments with the Adult data set using the RBF kernel SVC formulation did not finish even after waiting for $24$ hours, whereas the linear formulation took only minutes to complete.\footnote{Experiments were performed on a machine running Ubuntu OS with $32$ cores, and $64$ GB RAM.} We suspect that this is because the kernel SVC formulation is implemented using a generic convex optimization solver\footnote{\url{http://cvxopt.org/}} that does not incorporate any techniques for speedup specific to the problem. Hence we report results only for the linear SVC formulation. We also note that we use equalized odds as the fairness constraint in FST, which is stricter than the equal opportunity constraint used by FERM. These comparisons are illustrated in Figure \ref{fig:comparisonsFERM}. Clearly, our FST methods that post-process probability outputs from linear SVC (FSTpost, FSTbatch) outperform FERM substantially. We note however that the pre-processing variant of FST (FSTpre), which trains a second linear SVC model using sample weights described in Section~\ref{sec:proc:pre}, did not provide acceptable results. One possibility is that these sample weights, which are based on conditional probabilities, do not work well with the SVC problem formulation which is non-probabilistic.  Nevertheless, in general we see that among all the four in-processing approaches we compared, only the reductions approach \citep{agarwal2018} has performance competitive to ours.

\vskip 0.2in
\bibliography{arxiv}

\end{document}